\documentclass[a4paper,fleqn,5p,times]{elsarticle}

\usepackage[utf8]{inputenc}
\usepackage{adjustbox} 
\usepackage{amsmath}
\usepackage{amssymb}
\usepackage{booktabs}       
\usepackage{caption}
\usepackage{comment}
\usepackage[T1]{fontenc}    
\usepackage{float}
\usepackage{geometry}
\usepackage{graphicx}
\usepackage{hyperref}       
\usepackage[utf8]{inputenc} 
\graphicspath{ {./images/} }
\usepackage{lipsum}
\usepackage{longtable}
\usepackage{microtype}      

\usepackage{listings}

\usepackage{multirow}
\usepackage{multicol}
\usepackage{nicefrac}       
\usepackage{pifont}
\newcommand{\cmark}{\ding{51}} 
\newcommand{\xmark}{\ding{55}} 
\usepackage{subcaption}
\usepackage{tabularx}
\usepackage{tikz}
\usepackage{url}            

\usepackage{placeins}

\newcommand{\flame}{\textbf{F}}
\newcommand{\snowflake}{{\Large\textasteriskcentered}}
\usepackage{makecell}

\usepackage{natbib}
\setcitestyle{authoryear,open={(},close={)}}

\begin{document}

\let\WriteBookmarks\relax
\def\floatpagepagefraction{1}
\def\textpagefraction{.001}



\title{Quantitative Comparison of Fine-Tuning Techniques for Pretrained Latent Diffusion Models in the Generation of Unseen SAR Images}

\author[1,2]{Solène Debuysère}
\ead{solene.debuysere@onera.fr}

\author[1,2]{Nicolas Trouvé}
\ead{nicolas.trouve@onera.fr}

\author[1,2]{Nathan Letheule}
\ead{nathan.letheule@onera.fr}

\author[1,2]{Olivier Lévêque}
\ead{olivier.leveque@onera.fr}

\author[1,3]{Elise Colin}
\ead{elise.colin@onera.fr}

\cortext[cor1]{Corresponding author.}

\affiliation[1]{organization={Paris-Saclay University}, city={Gif-sur-Yvette (91190)}, country={France}}

\affiliation[2]{organization={ONERA - The French Aerospace Lab}, 
                department={The Electromagnetism and Radar Department (DEMR)}, 
                city={Palaiseau (91120)}, 
                country={France}}

\affiliation[3]{organization={ONERA - The French Aerospace Lab}, 
                department={The Information Processing and Systems Department (DTIS)}, 
                city={Palaiseau (91120)}, 
                country={France}}

\makeatletter
\def\printkeywords{}
\def\printnomenclature{}
\def\printendabstract{}
\def\printfirstpage{}
\makeatother

\begin{abstract}
We present a framework for adapting a large pretrained latent diffusion model to high-resolution Synthetic Aperture Radar (SAR) image generation. The approach enables controllable synthesis and the creation of rare or out-of-distribution scenes beyond the training set. Rather than training a task-specific small model from scratch, we adapt an open-source text-to-image foundation model to the SAR modality, using its semantic prior to align prompts with SAR imaging physics (side-looking geometry, slant-range projection, and coherent speckle with heavy-tailed statistics). Using a 100k-image SAR dataset, we compare full fine-tuning and parameter-efficient Low-Rank Adaptation (LoRA) across the UNet diffusion backbone, the Variational Autoencoder (VAE), and the text encoders. Evaluation combines (i) statistical distances to real SAR amplitude distributions, (ii) textural similarity via Gray-Level Co-occurrence Matrix (GLCM) descriptors, and (iii) semantic alignment using a SAR-specialized CLIP model. Our results show that a hybrid strategy—full UNet tuning with LoRA on the text encoders and a learned token embedding—best preserves SAR geometry and texture while maintaining prompt fidelity. The framework supports text-based control and multimodal conditioning (e.g., segmentation maps, TerraSAR-X, or optical guidance), opening new paths for large-scale SAR scene data augmentation and unseen scenario simulation in Earth observation.
\end{abstract}

\maketitle

\section{Introduction}

Synthetic Aperture Radar (SAR) has become a key modality in Earth observation, no longer restricted to governmental or defense institutions. It is supported by a growing diversity of platforms—from small satellites to airborne and drone systems operating across multiple frequency bands—and provides day-night, all-weather imaging for environmental monitoring, urban mapping, surveillance, and disaster assessment.

In this context, synthetic data are crucial for model testing, algorithm and sensor development, operational deployment, and data interpretation. Due to the variety of SAR applications, different simulation approaches are needed. In the literature, physics-based simulators typically require extensive information about objects' geometry and material, radar parameters, and other specific details, which are often difficult to obtain, for example, RaySAR from \cite{7730757}, SARCASTIC v2.0 from \cite{rs14112561}, and MOCEM \cite{5757200}. Most existing tools focus on the simulation of isolated targets or objects for detection or classification tasks. For large scene SAR simulation, a promising direction is the use of generative AI, enabling scalable labeled dataset augmentation with both controllable image synthesis and the creation of unseen, rare or challenging scenarios beyond existing datasets.

While Generative Adversarial Networks (GANs) have been explored mainly for isolated-target SAR image generation and augmentation (e.g., ships, vehicles, equipment models), they largely reproduce low-level statistics and offer limited multimodal control, constraining scene-level composition \cite{s20226673,isprs-annals-X-1-2024-83-2024,2018SPIE10752E..05L}. In contrast, Latent diffusion models (LDMs) presented in \cite{rombach2022highresolutionimagesynthesislatent} learn to invert a noise process in a compact latent space, enabling efficient generation with self- and cross-attention conditioning. As of foundation LDM, Stable Diffusion by \cite{rombach2022highresolutionimagesynthesislatent} uses a Variational Autoencoder (VAE) to map images to latent representations; text encoders provide embeddings that condition an attention-based UNet during denoising. Pretrained on large web-sourced text–image pairs of optical images, Stable Diffusion is a powerful generative model capable of synthesizing photorealistic scenes and layouts.

Adapting such foundation models to SAR remains challenging  because SAR imagery differs fundamentally from optical training datasets. SAR is acquired in a side-looking geometry with slant-range and azimuth coordinates (range–Doppler), producing layover, foreshortening, and shadow effects that depend on incidence angle and terrain. Coherent imaging creates speckle and heavy-tailed amplitude statistics, while the radiometric dynamic range is wide, with bright man-made backscatter coexisting with very low-return areas (e.g., calm water). These properties vary with polarization, incidence, resolution, and wavelength, complicating direct transfer and motivating domain adaptation.

Thus, we present an adaptable framework to fine-tune an open-source pretrained LDM—specifically Stable Diffusion XL (SDXL) by \cite{podell2023sdxlimprovinglatentdiffusion}—to the SAR modality. Our goal is to preserve the semantic prior and compositional abilities of the base text-to-image model while aligning generation with SAR imaging physics. To reduce acquisition-dependent variability, we curate a consistent high-resolution dataset (~40 cm, X-band) with similar incidence angles and processing, resulting in ~100,000 airborne SAR images acquired with ONERA’s SETHI sensor \cite{Baqu2019SethiR}.

To make synthetic SAR data useful for downstream tasks, the generative model must go beyond the training set by composing novel scene configurations on demand. We use the pretrained model’s semantic prior to describe and compose diverse environments (e.g., urban, forested, coastal) with spatial relations (e.g., “along,” “near,” “to the left of”). Our adaptation preserves the model’s language and relational understanding while aligning generation with SAR imagery, ensuring outputs are physically plausible SAR scenes preserving rather than synthesizing grayscale optical renderings with added speckle.

To find a non-trivial balance between overtraining—which would ensure a good understanding of SAR physics but would lead to a loss of model plasticity—and undertraining—which would result in a simple stylization of optical images—it is necessary to study the fine-tuning process carefully to identify the right trade-off. To address this, we investigate various fine-tuning approaches on the UNet backbone, VAE, and Text Encoders of Stable Diffusion XL—including adjustments to hyperparameters, weight updates, embedding learning (with a SAR-specific token), and loss regularization. These strategies aim to learn SAR-specific representations that capture speckle, texture, and reflectivity coherence.

Finally, an equally important challenge is how to evaluate the quality of generated SAR images. Traditional visual metrics in AI are ill-suited to this task, as they assume natural image statistics. In response, we propose a new evaluation framework, combining statistical distribution comparisons, texture analysis via Gray-Level Co-occurrence Matrices, and semantic alignment using a CLIP model fine-tuned on SAR-caption pairs. Using these metrics, we compare training configurations to make model behavior more explainable and to use the framework to other latent diffusion models (LDMs).

This adaptation offers numerous practical applications, including the generation of unseen, rare, and challenging scenes beyond existing datasets. It also enables controllable image synthesis with image-to-image generation while preserving spatial or statistical priors, improving tasks like adding structured spatial details and refining physics-based simulations. Here, we show that our method improves the realism of synthetic SAR imagery generated by ONERA’s EMPRISE simulator and of synthesis conditioned on TerraSAR-X imagery.

The paper first presents related work in generative modeling and fine-tuning methods, focusing on parameter-efficient adaptation. After describing the training dataset, the methodology section outlines the diffusion model architecture, fine-tuning strategies, and evaluation metrics. Experimental results, including quantitative analysis and visual examples of generated SAR images, are presented next. Our work on improving the realism of synthetic SAR images generated by ONERA’s EMPRISE simulator and enhancing TerraSAR-X satellite acquisitions is then discussed. It concludes with a discussion of our findings, limitations, and future directions.

\section{Related work}
Recent years have seen the emergence of large-scale generative models capable of synthesizing images from natural language prompts. These models, commonly referred to as foundation models, are typically pretrained on massive datasets and designed to capture high-level semantic alignment between visual and textual modalities. Their success has led to interest in adapting them to specialized domains, such as medical or optical remote sensing imaging. However, extending these models to unconventional modalities such as SAR, which differs both structurally and statistically from natural images, remains a largely underexplored challenge. In this section, we review the relevant efforts in basic vision language modeling and the fine-tuning strategies developed to adapt them effectively.

\subsection{Foundation Generative Vision-Language Models}

Vision-Language Models (VLMs) have demonstrated strong capabilities in learning joint embeddings and generative alignments across visual and textual modalities. Early models such as CoCa by \cite{yu2022cocacontrastivecaptionersimagetext} combine contrastive and generative objectives to jointly align and synthesize, while more recent architectures like CM3Leon and Chameleon by \cite{chameleonteam2025chameleonmixedmodalearlyfusionfoundation} implement early fusion designs, allowing unified multimodal generation through transformer-based architectures. These systems can generate both image and text outputs conditioned on joint multimodal inputs.

In parallel, a distinct family of models specifically focuses on text-to-image generation. Within this category, Stable Diffusion by \cite{rombach2022highresolutionimagesynthesislatent}, Flux by \cite{flux2024}, Imagen by \cite{saharia2022photorealistictexttoimagediffusionmodels}, and Parti by \cite{yu2023scalingautoregressivemultimodalmodels} represent several lines of research exploring different generation mechanisms: respectively, latent-space diffusion, pixel-space diffusion, and auto-regressive modeling. Among them, latent diffusion models such as Stable Diffusion have gained attention for their ability to generate high-resolution images with lower computational cost. This is achieved by operating in a compressed latent space, learned via a Variational Autoencoder (VAE), rather than directly in pixel space.
 
Stable Diffusion XL (SDXL) by \cite{podell2023sdxlimprovinglatentdiffusion}, in particular, is a flexible, open-source latent diffusion model consisting of three main components: (i) a VAE that compresses images into latent representations, (ii) a dual text encoder pipeline to transform prompts into embeddings, and (iii) a UNet backbone that performs conditional denoising in the latent domain. Numerous extensions have been proposed to control its generation process, including spatial guidance methods such as ControlNet by \cite{zhang2023addingconditionalcontroltexttoimage}, and global image-based conditioning methods like IP-Adapter by \cite{ye2023ipadaptertextcompatibleimage}. However, current research has mostly focused on optical image domains, and little is known about the model's ability to learn physically grounded or domain-specific concepts such as those present in SAR imagery.

\subsection{Fine-tuning approaches}
As the size of pretrained Vision-Language Models (VLMs) continues to grow, full fine-tuning—i.e., updating all model parameters—becomes increasingly impractical due to memory, compute, and data constraints. To address this, a family of Parameter-Efficient Fine-Tuning (PEFT) techniques has emerged, aiming to adapt large-scale models by updating only a small fraction of their weights parameters.

One of the most widely used PEFT strategies is Low-Rank Adaptation (LoRA) presented by \cite{hu2021loralowrankadaptationlarge}, which injects trainable low-rank matrices into the linear layers of the model. LoRA enables effective adaptation while keeping the majority of weights frozen, drastically reducing memory usage. Extensions such as QLoRA by \cite{dettmers2023qloraefficientfinetuningquantized} and DoRA by \cite{wang2024borabidimensionalweightdecomposedlowrank} further optimize efficiency by combining low-rank decomposition with quantization or weight reparameterization, especially for large language models (LLMs), and are increasingly being explored in vision and multimodal settings.

Other approaches focus on prompt-level conditioning rather than internal parameter modification. Prompt-based tuning methods such as CoOp by \cite{Zhou_2022} and VPT by \cite{jia2022visualprompttuning} learn input embeddings or prompts that guide the model without altering its architecture. These techniques are lightweight and adaptable, but may not represent entirely new visual domains when semantic gaps are large.

In contrast, DreamBooth presented by \cite{ruiz2023dreamboothfinetuningtexttoimage} enables explicit concept injection by associating new visual identities or styles with custom textual tokens. This method has been successfully applied to generate specific outputs from small datasets — for instance, \cite{Agrawal_2025} fine-tuned Stable Diffusion 3 on 300 samples of Jamini Roy-style paintings, achieving culturally accurate synthesis. Further control can be obtained using ControlNet and IPAdapter, which allow generation to be conditioned on structural priors such as edge maps, segmentation, or depth — especially effective for layout-sensitive domains.

Another promising method is textual inversion presented by \cite{gal2022imageworthwordpersonalizing}, which learns new visual concepts directly in the embedding space of the text encoder, without modifying the image generator. When combined with DreamBooth and LoRA, as in the paper from \cite{dai2025diffusionbasedsyntheticdatageneration}, it enables joint control over modality and identity — for example, generating paired visible-infrared images from shared prompts like "a [modality] photo of a [person] person".

Collectively, these techniques have made it feasible to adapt powerful diffusion models such as Stable Diffusion to novel visual concepts, ranging from new objects to artistic styles, while requiring relatively modest amounts of data and compute. However, most existing applications remain confined to optical image domain and focus on concept categories that are semantically close to those seen during pre-training (e.g., human faces, animals, or art styles). But these methods are not well-suited for learning entirely new domains, such as SAR imagery, which is structurally and statistically distinct. Indeed, LoRA-based method alone can't capture the spatial complexities of SAR data, including Rayleigh noise and imaging geometry. Prompt-based tuning methods like CoOp and VPT are limited when dealing with large semantic gaps, such as those between optical and SAR domains. DreamBooth or Textual inversion, though effective for small datasets, struggle with SAR's features, requiring larger, high-resolution data. 


\subsection{Generative Foundation models in Remote Sensing
}
In the field of remote sensing, most recent generative Vision-Language Models (VLMs) have focused primarily on optical imagery, with limited or no support for Synthetic Aperture Radar (SAR) data. Several foundation models have been trained from scratch on large-scale optical satellite image datasets, including RS5M and GeoRSCLIP by \cite{Zhang_2024}, DiffusionSat by \cite{khanna2024diffusionsatgenerativefoundationmodel}, MetaEarth by \cite{yu2024metaearthgenerativefoundationmodel}, CRS-Diff by \cite{tang2024crsdiffcontrollableremotesensing}, and HSIGene by \cite{pang2024hsigenefoundationmodelhyperspectral}, the latter targeting hyperspectral image generation. While most of these models are trained for representation learning, zero-shot classification, or retrieval in the optical domain, only a few are designed for image generation — and even fewer extend to radar-based modalities such as Synthetic Aperture Radar (SAR).

To our knowledge, Text2Earth by \cite{liu2025text2earthunlockingtextdrivenremote} is the first foundation model that incorporates both SAR and optical data for text-to-image generation. However, its SAR component relies on synthetic radar-like images produced via Pix2Pix translation from RGB inputs, rather than using real SAR measurements that include speckle noise, geometric distortions, and backscatter-specific statistical properties. As such, the model does not capture the full complexity of radar signal characteristics.

Other models, such as SARChat-InternVL2.5-8B presented by \cite{ma2025sarchatbench2mmultitaskvisionlanguagebenchmark}, have focused on improving multimodal understanding of SAR imagery through conversational tasks like description, counting, or spatial reasoning. However, this model is not able to do image generation, and its training data is limited to open-source object detection benchmarks. It does not address large-scale SAR image synthesis nor generalization across acquisition conditions.


In general, the development of generative models for SAR is still in its early stages, particularly for high-resolution data. While some efforts have explored training from scratch, such approaches are computationally prohibitive and require extensive domain-specific data. In contrast, adapting pretrained generative models - originally trained on optical images - to the SAR domain via fine-tuning offers a more scalable and practical alternative. Yet, this path remains largely underexplored. Our work positions itself in this gap, investigating how such pretrained models can be effectively adapted to synthesize realistic SAR imagery guided by textual prompts.

\section{Dataset Creation}
\label{sec:dataset-creation}

Unlike optical imaging, SAR systems actively transmit radar pulses toward the ground and record the backscattered signals. While optical image resolution is defined by the number of pixels per unit area, SAR images have two distinct resolutions: one in the range direction (perpendicular to the flight path) and one in the azimuth direction (along the flight path). SAR image formation involves two key steps: \textit{range compression}, which enhances resolution perpendicular to the sensor trajectory, and \textit{azimuth compression}, which improves resolution along the sensor motion. These processes rely on advanced signal processing techniques, such as the use of frequency-modulated pulses (chirps) and coherent integration of successive returns to synthesize a larger effective antenna aperture.\\

\begin{figure}
    \centering
    \begin{minipage}{0.22\textwidth}
        \centering
        \includegraphics[width=\textwidth]{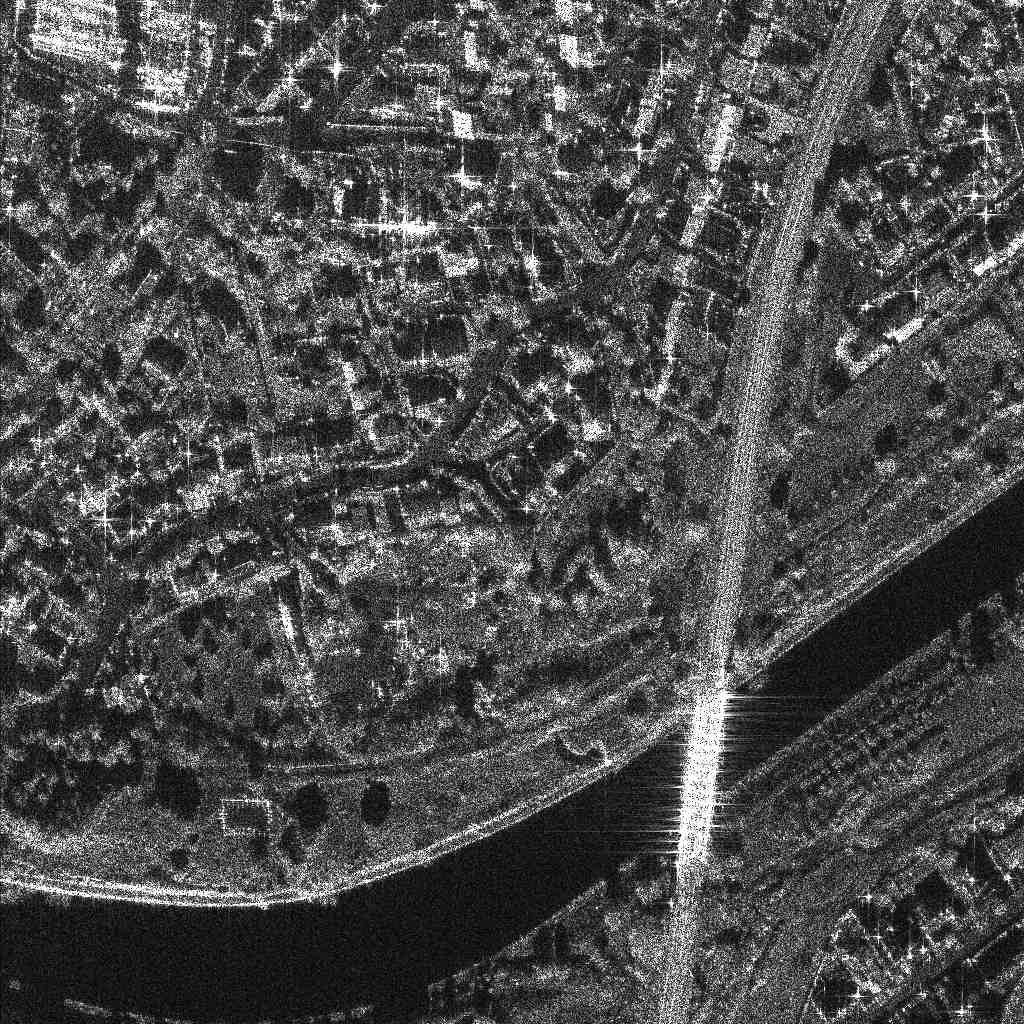}
        \\ (a) SAR Image
    \end{minipage}
    \hspace{0.02\textwidth}
    \begin{minipage}{0.22\textwidth}
        \centering
        \includegraphics[width=\textwidth]{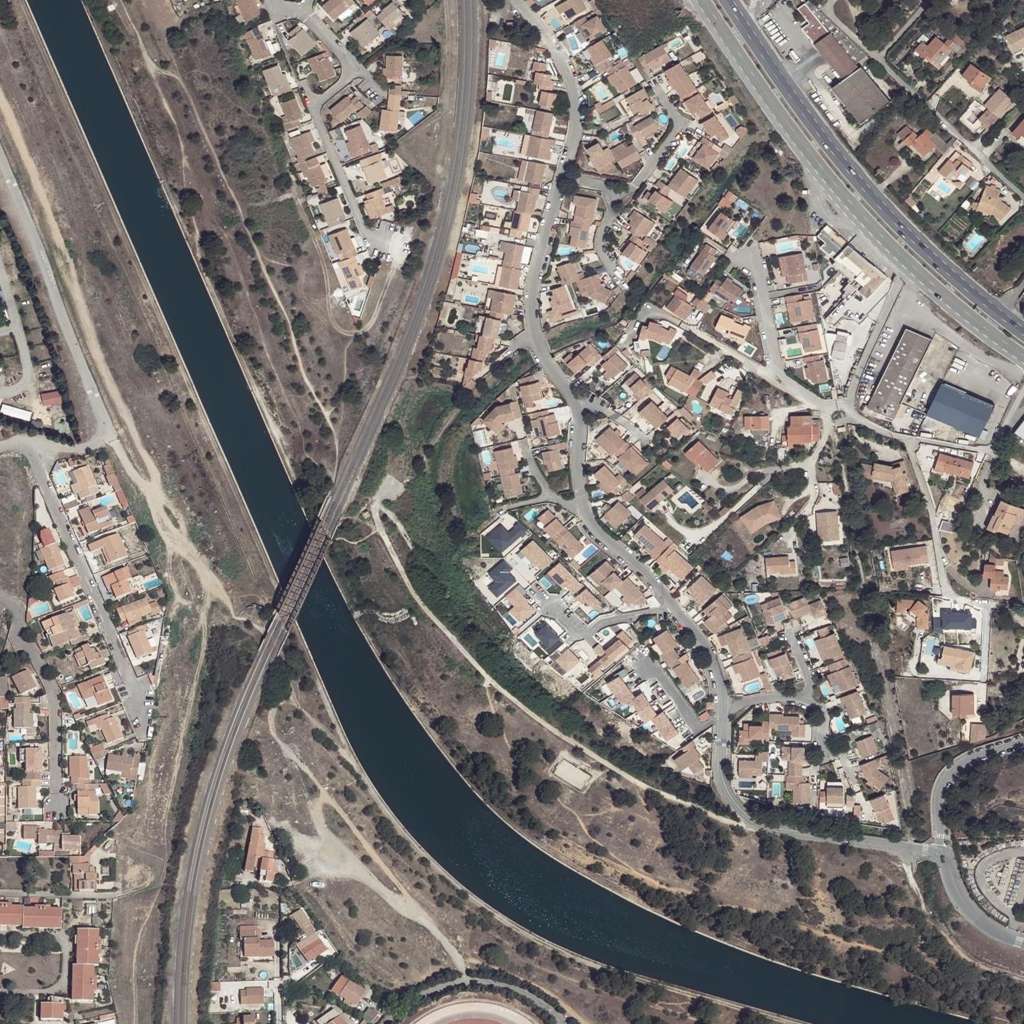}
        \\ (b) Optical Image
    \end{minipage}
    
    \vspace{0.02\textwidth}
    
    \begin{minipage}{0.22\textwidth}
        \centering
        \includegraphics[width=\textwidth]{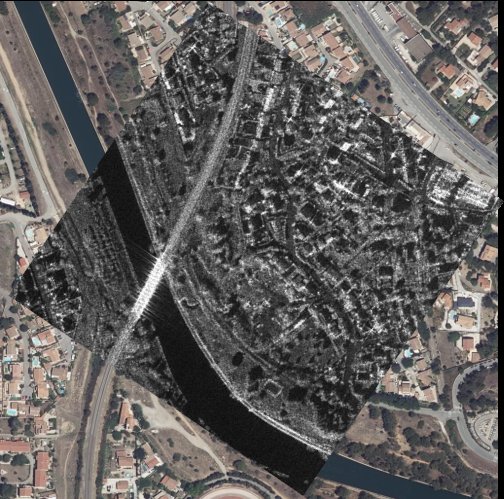}
        \\ (c) Optical vs SAR Images axis
    \end{minipage}
    \hspace{0.02\textwidth}
    \begin{minipage}{0.22\textwidth}
        \centering
        \includegraphics[width=\textwidth]{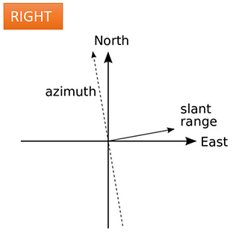}
        \\ (d) SAR Geometry Acquisition Plan
    \end{minipage}
    
    \caption{Pairs of optical (ground plan) and SAR (slant-range plan) images.}
    \label{fig:pairs}
\end{figure}

In our dataset, the resulting images are stored in what is known as the Single Look Complex (SLC) format, which means that images are acquired in the antenna reference frame, known as “slant range-azimuth” coordinate system, which is link to the radar's viewing geometry rather than geographic axes. As a result, SAR images are rotated with respect to true North. They are also geometrically distorted because the sampling in azimuth and slant-range directions does not correspond to equal ground distances. This leads to visual effects where structures like roads or rivers appear tilted or compressed when overlaid on optical images. For instance, in our example (see Figure (d) ~\ref{fig:pairs}), the SAR image appears diagonally inserted within the optical scene due to the acquisition in slant-range geometry during an ascending right-looking orbit. 

At ONERA's DEMR department, we conduct airborne campaigns using the SETHI radar system and process raw radar echoes into SLC-format SAR images (each approximately 40,000 × 7000 complex pixels). From this large archive, we built a training dataset by applying several post-processing steps to raw complex and amplitude images.

\paragraph{\textbf{Pre-processing Raw Data}} To ensure data quality, we first filtered the dataset by selecting images with sufficient metadata, choosing only those acquired in the X-band (8 to 12 GHz) and with HH or VV polarization, while excluding small or geographically overlapping scenes. The calibration factors were then applied to ensure radiometric accuracy. Then, we apply a correction matrix to complex images to refocus the spectrum in both directions to correct spectral misalignment caused by acquisition conditions. Finally, all images were downsampled in the frequency domain to a target resolution of 40 cm (in both azimuth and range directions).\\

\paragraph{\textbf{Training Dataset Creation}}
Our final dataset consists of refocused and resampled SAR images stored as complex-valued matrices. For training purposes, we work on amplitude images, whose pixel values follow a Rayleigh distribution — unbounded and highly skewed, with a small proportion (1–3\%) of very high-intensity scatterers. These bright pixels are critical as they correspond to strong reflectors such as buildings or metallic structures. For visualization and learning stability, we apply the following normalization, which is a common practice in visual interpretation of SAR images to stabilize their dynamic range and improve the quality of visualization:

{\small
\begin{equation}
A(r,y)_{\text{norm}} = \frac{A(r,y)}{\mu + 3 \cdot \sigma}
\end{equation}
}

where \(A(r,y)\) is the amplitude value at coordinates \((r, y)\), and \(\mu\) and \(\sigma\) are the mean and standard deviation of the amplitude image, respectively. Values are clipped between 0 and 1 to have 98\% of the values that fall in this interval. Thus, all pixels beyond the threshold are "flattened" at exactly 1. This creates an artificial saturation at pixel value 1. \\

However, it is important to note that this approach is a compromise. If the threshold is set too high to preserve the information of the strongest scatterers, it could cause all the weaker scatterers to be compressed into the same range of values, effectively losing the distinction between them. This would result in a loss of detail for the weaker scatterers, which are often important for accurate interpretation. Therefore, it’s essential to find a balance between preserving the information of the strongest reflectors (like buildings or metallic structures) while also maintaining enough precision to differentiate the weaker scatterers, which can be critical in some applications.

The normalized images are then cropped into standardized patches of size 1024 × 1024 pixels. For a subset of these patches, we created geo-aligned SAR–optical image pairs using optical imagery from the IGN database. Textual descriptions were automatically generated for the optical images using the foundation model CogVLM2~\cite{hong2024cogvlm2visuallanguagemodels}.

\begin{figure}[htb]
    \centering
    \begin{minipage}{0.48\textwidth}
        \centering
        \includegraphics[width=\textwidth]{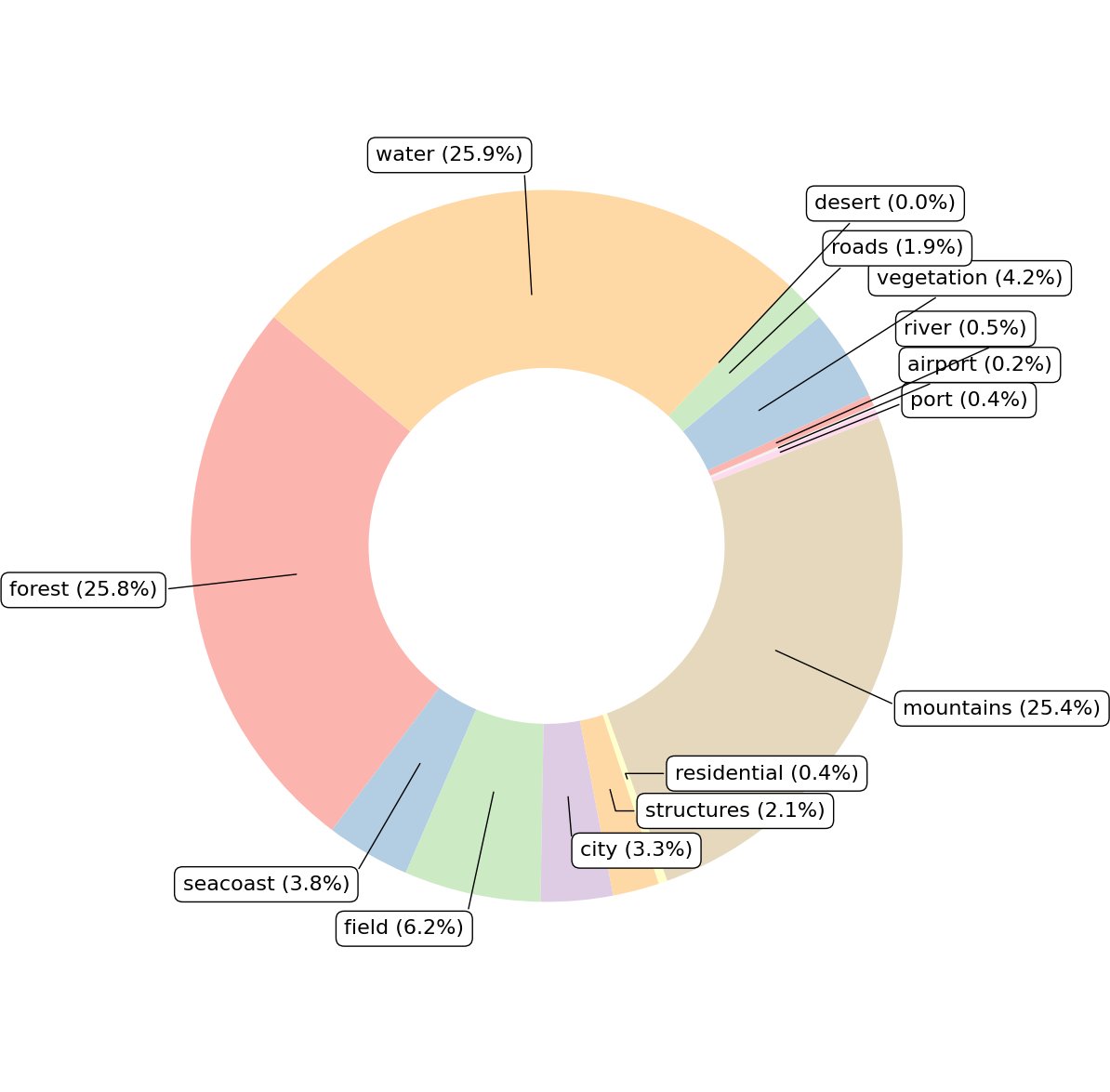}
    \end{minipage}
    \hspace{5pt}
    \begin{minipage}{0.27\textwidth}
        \centering
        \renewcommand{\arraystretch}{1.1}
        \setlength{\tabcolsep}{2pt}
        \scriptsize
        \begin{tabular}{lccc}
            \toprule
            \textbf{Category} & \textbf{Train} & \textbf{Validation} & \textbf{Test} \\
            \midrule
            Airport       & 0.17 & 0.16 & 0.23 \\
            City          & 3.27 & 3.13 & 3.33 \\
            Desert        & 0.01 & 0.01 & 0.01 \\
            Field         & 6.20 & 6.23 & 6.04 \\
            Forest        & 25.82 & 25.90 & 25.82 \\
            Mountains     & 25.36 & 25.44 & 25.46 \\
            Port          & 0.42 & 0.40 & 0.53 \\
            Residential   & 0.40 & 0.46 & 0.41 \\
            River         & 0.51 & 0.64 & 0.57 \\
            Roads         & 1.88 & 1.78 & 1.80 \\
            Seacoast      & 3.82 & 3.57 & 3.73 \\
            Structures    & 2.11 & 2.06 & 2.05 \\
            Vegetation    & 4.16 & 4.27 & 4.02 \\
            Water         & 25.89 & 25.95 & 26.00 \\
            \bottomrule
        \end{tabular}
    \end{minipage}
    \caption{(a) Training dataset labels repartition (b) Dataset repartition; train, validation and test}
    \label{fig:dataset}
\end{figure}

Because the statistical distribution of SAR amplitudes depends heavily on scene type, we categorized the dataset into semantic classes (e.g., forest, water, city) based on a manually constructed keyword dictionary applied to the generated captions. The class distribution is presented in Figure~\ref{fig:dataset}. We note that some mislabeling may occur due to inaccuracies in captioning or ambiguity in keyword matching. These class labels were aggregated for visualization and analysis.

Finally, we performed a cleaning stage to remove low-quality samples, such as blurred zones, blank images, and noisy or distorted data.

\section{Methodology}

\subsection{Stable Diffusion framework}
\label{sec:SDXL}
\paragraph{\textbf{Input–Output Representation}}
Stable Diffusion XL model operates in a latent space rather than directly in pixel space. Text–image pairs are processed independently into compact latent representations. First, a Variational Autoencoder (VAE) encodes an image $x$ into a latent representation $z$:

\begin{equation}
    z = E(x), \quad \tilde{x} = D(z)
\end{equation}

where $E$ and $D$ denote the encoder and decoder, respectively. The image is typically compressed by a factor of 8 along each spatial dimension. In parallel, the text prompt $y$, which describes the scene to be generated, is embedded into a semantic vector space via the model’s text encoders, yielding embeddings $\tau_{\theta}(y)$.\\

Stable Diffusion XL (SDXL) employs two separate text encoders. \textit{Text Encoder 1} (CLIP ViT-L) produces token-level embeddings of shape $[B, 77, 768]$ — one vector per token. These embeddings are used in the UNet's cross-attention layers for fine-grained conditioning. \textit{Text Encoder 2} (OpenCLIP ViT-bigG) also provides token-level embeddings, but in higher dimension $[B, 77, 1280]$, and additionally produces a global [CLS] token, which is passed through a learned linear projection to generate a global caption embedding, noted as \texttt{text\_embeds}. This projection complements the token-level conditioning and is injected globally into the UNet at each layer.

\begin{table}[htb]
\centering
\scriptsize
\adjustbox{max width=0.5\textwidth}{
\begin{tabular}{>{\raggedright}p{2.5cm}|>{\raggedright}p{2.5cm}|>{\raggedright\arraybackslash}p{2.5cm}}
\toprule
\textbf{Usage} & \textbf{Text Encoder 1 (CLIP ViT-L)} & \textbf{Text Encoder 2 (OpenCLIP ViT-bigG)} \\
\midrule
Token-wise embeddings & Yes $\rightarrow$ shape $[B, 77, 768]$ & Yes $\rightarrow$ shape $[B, 77, 1280]$ \\
\midrule
CLS token (global summary of captions) & Not used & Yes $\rightarrow$ shape $[B, 1280]$ (projected afterwards) \\
\midrule
Projection (\texttt{text\_projection}) & \textbf{No} & \textbf{Yes} (CLS $\rightarrow$ projection $\rightarrow$ \texttt{text\_embeds}) \\
\bottomrule
\end{tabular}
}
\caption{Comparison between Text Encoder 1 (CLIP ViT-L) and Text Encoder 2 (OpenCLIP ViT-bigG) in the SDXL architecture.}
\label{tab:sdxl_text_encoders}

\end{table}

The token - wise embeddings from both encoders are concatenated to form a $[B, 77, 2048]$ tensor, injected into each UNet layer via cross-attention. Moreover, the global projection vector of shape $[B, 1280]$ is passed through the \\ \texttt{added\_cond\_kwargs['text\_embeds']} layers.

\paragraph{\textbf{Training Process}} 

\begin{figure*}
    \centering
    \begin{adjustbox}{max width=0.88\textwidth}
    \includegraphics[width=\textwidth]{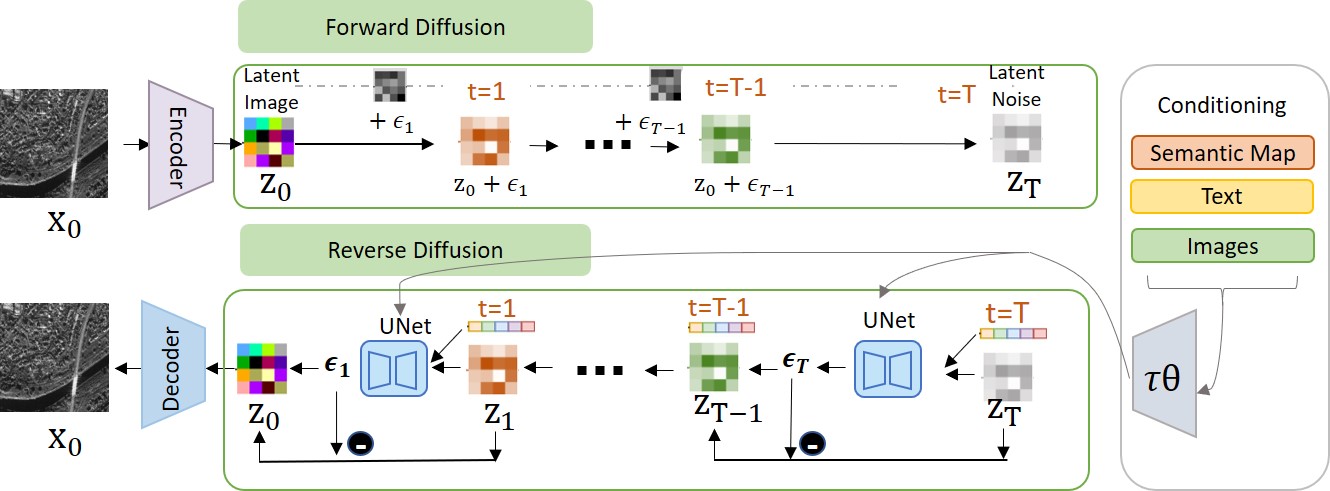}
    \end{adjustbox}
    \caption{ Forward and Reverse Process in Stable Diffusion XL}
    \label{fig:stable-diffusion-process}
    
\end{figure*}

As shown in the Figure~\ref{fig:stable-diffusion-process}, the model learns by noising training data (i.e. VAE Encoder output), through the successive addition of Gaussian noise (forward process). Then the model reverses this process (reverse process) to turn noise back into data by removing the noise added during each diffusion step. More specifically, the latent $z_t$ is the output of the VAE encoder from the input image $x_t$, and noise is added to simulate a step in the forward diffusion process, governed by a scheduler that defines the noise distribution at each $t$. More specifically, the forward diffusion process, which adds noise to an input data $z_0$ over $T$ timesteps by adding Gaussian noise, is defined as:

{\small
\begin{equation}
z_t = \sqrt{\bar{\alpha}_t} \, z_0 + \sqrt{1 - \bar{\alpha}_t} \, \epsilon, \quad \epsilon \sim \mathcal{N}(0, \mathbf{I})
\end{equation}
}

Here, $\bar{\alpha}_t = \prod_{s=1}^t \alpha_s$ denotes the cumulative noise attenuation schedule with $\alpha_t = 1 - \beta_t$. The UNet is trained to predict the noise component $\epsilon$ added at timestep $t$, conditioned on both time and prompt embeddings. This enables the model to learn how semantic information influences denoising across different degradation levels.

\begin{figure}
    \centering
    \begin{adjustbox}{max width=0.5\textwidth}
    \includegraphics[width=\textwidth]{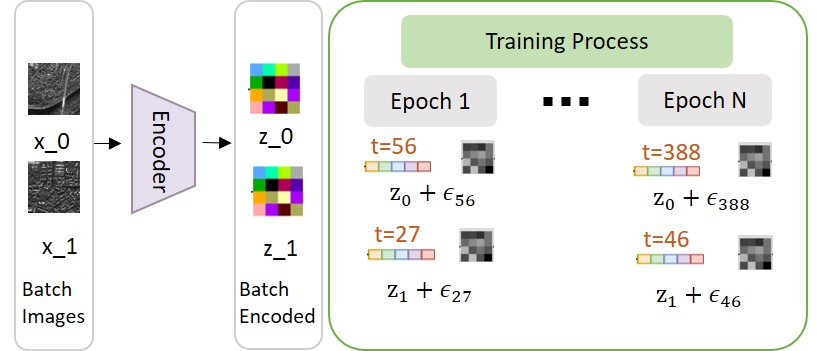}
    \end{adjustbox}
    \caption{Random timesteps sampling over a batch during training epochs}
    \label{fig:training}
    
\end{figure}

During training (Figure~\ref{fig:training}), the model receives a corrupted latent $z_t$ for a randomly sampled timestep $t \in [0, 1000]$ and is tasked with predicting the corresponding $\epsilon_t$. The loss is computed over multiple such noise levels per batch and epoch. Thus, to improve model performance without disrupting the pretrained knowledge, focusing on the final stages of the reverse diffusion process—corresponding to the earlier timesteps of the forward process during training—may offer a more effective refinement of the SAR image generation.

By randomly choosing timesteps over a batch during training, the model is exposed to a diverse set of degradation levels, which enables it to better capture the underlying patterns across different stages of the reverse diffusion process. Therefore, it is crucial to fine-tune the model over several epochs, ensuring that the model sees all the data multiple times. This also guarantees that different categories of data, such as fields, forests, cities, or seacoasts, are seen at various timesteps within the range $t \in [0,1000]$ during each training epoch.

During inference (generation), the process begins from a pure Gaussian noise sample $z_T$. The UNet iteratively denoises this latent through $T$ steps, each conditioned on the text prompt. Although the model was trained on a schedule of $T = 1000$ steps, it is common in practice to use only 25 to 50 steps for generation. At each step $t$, the model predicts $\epsilon_t$ and uses the reverse schedule to approximate $z_{t-1}$. After the final step, the latent $z_0$ is decoded by the VAE Decoder to produce the final image $x_0$.

\subsection{Training approaches and parameterization}

In this study, we compare the effects of two fine-tuning strategies on the main components of the SDXL architecture: the UNet backbone and the two text encoders (TE1 and TE2). The first approach involves \textit{full fine-tuning}, where all model weights are updated. While this allows maximum flexibility and capacity for domain adaptation, it is computationally expensive and increases the risk of overfitting, particularly in scenarios with limited training data.

The second approach uses \textit{Low-Rank Adaptation (LoRA)}, a parameter-efficient fine-tuning method. In this setting, the original model weights are kept frozen, and trainable low-rank matrices are injected into selected layers — typically in the attention and cross-attention modules. These additional parameters allow the model to learn task-specific adaptations with a significantly reduced memory and computational footprint.

Our goal is to evaluate whether LoRA is sufficient for adapting a pretrained latent diffusion model to SAR image generation, and under what conditions full fine-tuning is still required. We hypothesize the following:
\begin{itemize}
    \item \textbf{Full fine-tuning of the UNet} may be necessary to capture the low-level statistical and physical properties that characterize SAR images (e.g., speckle, radiometric contrast, geometry).
    \item \textbf{LoRA-based tuning of the text encoders} may help to preserve the model’s base language semantic knowledge, including spatial arrangements and object relationships, while adapting it to SAR imagery. 
\end{itemize}

In Section~\ref{sec:experiments}, we empirically evaluate these strategies across several model configurations. We use both semantic and statistical metrics to determine the most effective fine-tuning techniques for generating realistic and coherent SAR images from textual descriptions.

\subsection{Evaluation of generated SAR images}
\label{sec:evaluation-models-images}
\paragraph{\textbf{Evaluation Dataset}} 
To assess the realism of the generated SAR images, we use a test dataset composed of triplets: [captions, labels, real SAR images]. Since SAR amplitude pixel values distributions vary significantly depending on the type of scene, we perform a label-specific evaluation to account for this variability. The labels correspond to semantic scene categories --- \textit{forest, field, city, airport, seacoast, port, mountains, beach, industrial}, and \textit{residential} --- derived from a manually created keyword dictionary. For each label, the associated test captions are used as prompts to generate synthetic SAR images. In total, we generate 30 images per label across 11 categories, resulting in 330 generated images used for evaluation and comparison across model configurations.

\paragraph{\textbf{SAR Statistics analysis}} 
To compare the amplitude distributions between real and generated SAR images, we flatten each image into a one-dimensional array of pixel amplitudes. As described in Section~\ref{sec:dataset-creation}, a normalization factor is applied during pre-processing, and pixel values are clipped to the range [0, 1], introducing an artificial saturation peak at the upper bound.

To enable accurate statistical comparisons using the Kullback–Leibler (KL) divergence, we first exclude all pixel values corresponding to the saturated pixels (corresponding to 3\% maximum). We then compute the proportion of saturated pixels separately and renormalize the histograms over the remaining values so that the probability density integrates to 1.

The KL divergence is computed between the empirical amplitude distributions of real and generated images, separately for each semantic category. Given two discrete probability distributions, \( P \) (real SAR) and \( Q \) (generated SAR), estimated over amplitude bins \( i \), the KL divergence is given by:

{\small
\begin{equation}
D_{\mathrm{KL}}(P \parallel Q) = \sum_{i} P(i) \log \left( \frac{P(i)}{Q(i)} \right)
\label{eq:kl_divergence}
\end{equation}
}


\paragraph{\textbf{Prompt-Image Alignment}}
Beyond statistical similarity, we also evaluate how well the generated SAR images align semantically with their conditioning prompts. To this end, we fine-tuned a CLIP ViT-L/14 model on a separate dataset containing SAR image–caption pairs, using a batch size of 100, with the goal of embedding both modalities into a common latent space adapted to radar imagery.

To quantify alignment, we adopt two other evaluation strategies. First, we compute the \textit{ranking score}, which measures how well each image is matched to its correct caption among a set of other texts. For each batch of $N = 16$ image–text pairs, we extract normalized image embeddings $f_{\text{img}}(x_i)$ and text embeddings $f_{\text{text}}(t_j)$, and compute the cosine similarity matrix \( S \in \mathbb{R}^{N \times N} \):

{\small
\begin{equation}
S_{ij} = \frac{\langle f_\text{img}(x_i), f_\text{text}(t_j) \rangle}{\|f_\text{img}(x_i)\| \cdot \|f_\text{text}(t_j)\|}
\label{eq:similarity_score}
\end{equation}
}

We apply a softmax normalization across each row of the matrix to interpret the values as match probabilities. For each image $x_i$, we compute the rank $r_i$ of its corresponding ground-truth caption $t_i$ within the list of possible captions. We report the mean rank $r_\mu$, median, and variance ranks $r_\sigma$ over the evaluation set:

{\small
\begin{equation}
r_\mu = \frac{1}{N} \sum_{i=1}^{N} r_i \quad r_\sigma = \frac{1}{N} \sum_{i=1}^{N} (r_i - r_\nu)^2
\label{eq:rank}
\end{equation}
}

A lower mean rank indicates better semantic alignment between the generated image and its textual prompt.

In addition to ranking-based evaluation, we compute the \textit{cosine similarity} between each generated image and its prompt, using the same fine-tuned SAR-CLIP model. For each of the 11 semantic labels, we average these similarity scores across all generated samples to obtain a per-label, per-model alignment metric. The results are presented as heatmaps, where higher values indicate stronger text–image coherence. As a reference baseline, we compute the same similarity scores between real SAR images and their corresponding captions using the same CLIP model. These values are included at the bottom of each heatmap for visual comparison.

These evaluation approaches allows us to assess both \textit{relative ranking performance} (i.e., how uniquely matched each prompt is to its image) and \textit{absolute similarity} (i.e., how close the embedding vectors are).



\paragraph{\textbf{SAR Textural Indicators}}
\label{para:texture-analysis}
To evaluate the textural realism of the SAR images generated by our model, we compute textural indicators derived from the Gray-Level Co-occurrence Matrix (GLCM), following the classical method proposed by Haralick et al.~\cite{4309314}. This method is well-suited for SAR texture analysis, as it allows us to assess directional patterns and spatial relationships in the generated images.

We base this analysis on the same set of 330 labeled real and generated SAR images. From each image, we extract homogeneous patches of size 64 × 64 pixels, using segmentation masks inferred by the Segment Anything Model (SAM)~\cite{kirillov2023segment}. Specifically, we identify the largest mask in each image and apply a sliding kernel to extract patches that are fully contained within that region. For each patch, we compute the GLCM.

Given a patch of size $(N, N)$ with gray-level quantized amplitude values, the GLCM is defined as:

{\scriptsize
\begin{equation}
\text{GLCM}(l, k, \theta, d) = \frac{1}{N_d N_\theta} \sum_{x=1}^{M} \sum_{y=1}^{N} 
\begin{cases}
1, & \text{if } I(x,y) = l \text{ and } I(x+\Delta x, y+\Delta y) = k \\
0, & \text{otherwise}
\end{cases}
\label{eq:glcm}
\end{equation}
}

where $(\Delta x, \Delta y) = (\mathrm{round}(d\cos\theta), \mathrm{round}(d\sin\theta))$, and $d$ and $\theta$ denote the distance and orientation of the co-occurrence pair.

From the normalized GLCM, we compute four classical Haralick texture features:
\begin{itemize}
    \item \textbf{Correlation:} \scalebox{0.8}{$\sum_{l,k} \text{GLCM}(l,k)\frac{(l - \mu)(k - \mu_k)}{\sigma \sigma_k}$}
    \item \textbf{Homogeneity:} \scalebox{0.8}{$\sum_{l,k} \frac{\text{GLCM}(l,k)}{1 + (l-k)^2}$}
    \item \textbf{Contrast:} \scalebox{0.8}{$\sum_{l,k} \text{GLCM}(l,k)(l - k)^2$}
    \item \textbf{Entropy:} \scalebox{0.8}{$-\sum_{l,k} \text{GLCM}(l,k) \log(\text{GLCM}(l,k) + \epsilon)$}
\end{itemize}

These features characterize various aspects of image texture such as spatial regularity, smoothness, contrast, and randomness. We compute them across multiple orientations $\theta$ and distances $d$ to assess rotation-invariant properties. With the correlation, we can capture spatial dependencies (due to SAR image geometry acquisition), while contrast allows us to evaluate the dynamic range of SAR Rayleigh distributions, with high values for dark pixels (e.g., water) and low values for bright pixels (e.g., buildings). Moreover, the entropy captures the randomness in the texture, and the homogeneity quantifies the smoothness and uniformity of the texture. 

We then compare the distributions of these indicators between real and generated SAR images across semantic categories (\textit{forest, city, port}, etc.), analyzing both their mean values and their variation under rotation. This enables us to assess whether the model has captured label-specific structural patterns and preserved the geometric and statistical richness of SAR texture.

\section{Experiments and Results}
\label{sec:experiments}
In this section, we present a series of experiments conducted using Stable Diffusion XL to evaluate the impact of various fine-tuning configurations. Our goal is to assess the individual contribution of each architectural component (UNet, Text Encoder 1, and Text Encoder 2), and to investigate whether they can be adapted independently or require joint tuning for optimal performance.

All experiments are performed using our custom training dataset of 100,000 SAR image–caption pairs (1024 per 1024 pixels for each image). Training hyperparameters are kept fixed across all configurations to ensure consistency and fair comparison. Specifically, the learning rate is set to 5e-5 for the UNet and 4e-5 for both text encoders. These values were chosen based on empirical stability under the available computational budget (one NVIDIA H100 GPU), while fitting within memory constraints.

To promote reproducibility, we use a fixed random seed for all experiments. The training dataset is shuffled identically across configurations, and the same seed is used for image generation (1024 per 1024 pixels) during both training and evaluation. This ensures that differences in results arise solely from the fine-tuning strategy and not from data ordering or sampling variability.


\subsection{Importance of the noise offset}
As we know, SAR imagery has heavy-tailed amplitude distributions and high contrast between bright scatterers (e.g., buildings, ships) and low-reflectivity regions (e.g., calm water), we introduce a small noise offset during the forward diffusion process to add more stochasticity and help the model better capture the full dynamic range.

At each training step, the standard Gaussian noise $\varepsilon \sim \mathcal{N}(0, 1)$ is perturbed by an additional random term, defined as:

{\small
\begin{equation}
\varepsilon_{\text{offset}} = \varepsilon + \gamma \delta, \quad \delta \sim \mathcal{N}(0,1), \quad \gamma = 0.035
\end{equation}
}

This modification effectively shifts the noise distribution to $\mathcal{N}(0, 1 + \gamma^2)$, introducing a variability per sample and per channel without altering the spatial structure of the noise.

To evaluate its impact, we compare two training runs with identical LoRA settings on the UNet and both text encoders: \textit{rain-beach-6} (with noise offset) and \textit{umbrella-sand-8} (without noise offset). These configurations differ only in the inclusion of the noise offset.

\begin{figure}
    \centering
    \scriptsize
    \begin{adjustbox}{max width=0.48\textwidth}
    \begin{tabular}{p{2cm} p{0.3cm} p{0.3cm} p{0.18cm} p{0.18cm} p{0.18cm} p{0.18cm} 
                    >{\centering\arraybackslash}p{0.45cm} 
                    >{\centering\arraybackslash}p{0.45cm} >{\centering\arraybackslash}p{0.45cm} 
                    >{\centering\arraybackslash}p{0.45cm}}
        \toprule
        \multirow{2}{*}{\textbf{Train ID}} 
        & \multicolumn{2}{c}{\makecell{\textbf{UNet}\\\textbf{LoRA}}}
        & \multicolumn{2}{c}{\makecell{\textbf{TE1}\\\textbf{LoRA}}}
        & \multicolumn{2}{c}{\makecell{\textbf{TE2}\\\textbf{LoRA}}}
        & \multirow{2}{*}{\makecell{\textbf{Noise}\\\textbf{Offset}}}
        & \multicolumn{2}{c}{\makecell{\textbf{CLIP}\\\textbf{Rank}} $\downarrow$} 
        & \multirow{2}{*}{\textbf{KL} $\downarrow$} \\
        \cmidrule(lr){2-3} \cmidrule(lr){4-5} \cmidrule(lr){6-7} \cmidrule(lr){9-10}
        & \textbf{r} & \textbf{a}
        & \textbf{r} & \textbf{a}
        & \textbf{r} & \textbf{a}
        & & $\nu$ & $\sigma$ & \\
        \midrule
        rain-beach-6      & 256 & 128 & 8 & 4 & 8 & 4 & \cmark & 2.34 & 3.73 & 0.17 \\
        umbrella-sand-8   & 256 & 128 & 8 & 4 & 8 & 4 & \xmark & 2.54 & 4.40 & 1.16 \\
        \bottomrule
    \end{tabular}
    \end{adjustbox}
    \caption{Comparison of trainings - at epoch 8 - with and without noise offset and LoRA (r: rank, a: alpha).}
    \label{tab:noise_offset_metrics}
    
    \vspace{1em} 
    
    \begin{adjustbox}{max width=0.48\textwidth}
    \begin{tabular}{ccc}
        \includegraphics[width=0.22\textwidth]{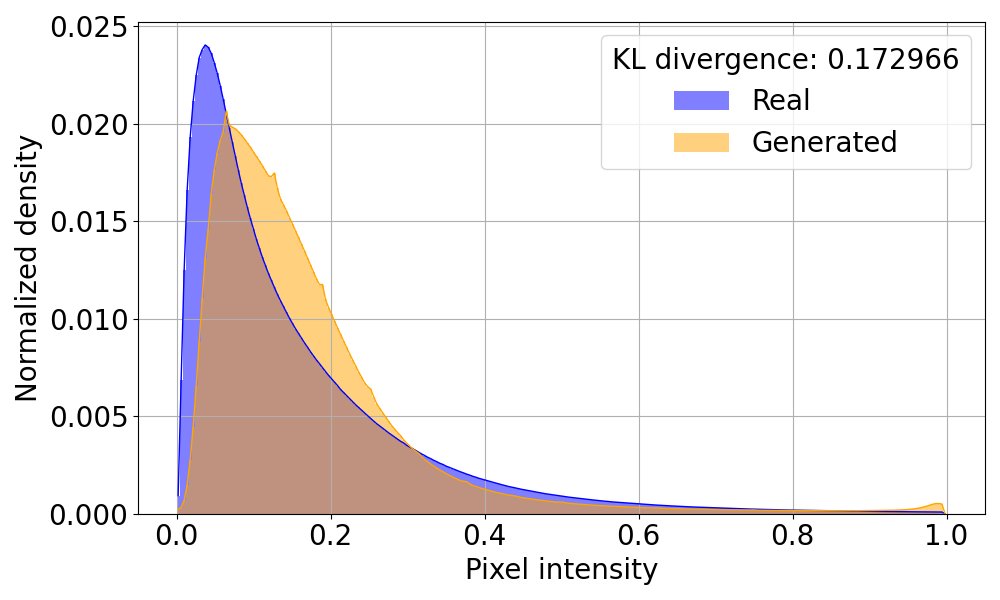} &
        \includegraphics[width=0.22\textwidth]{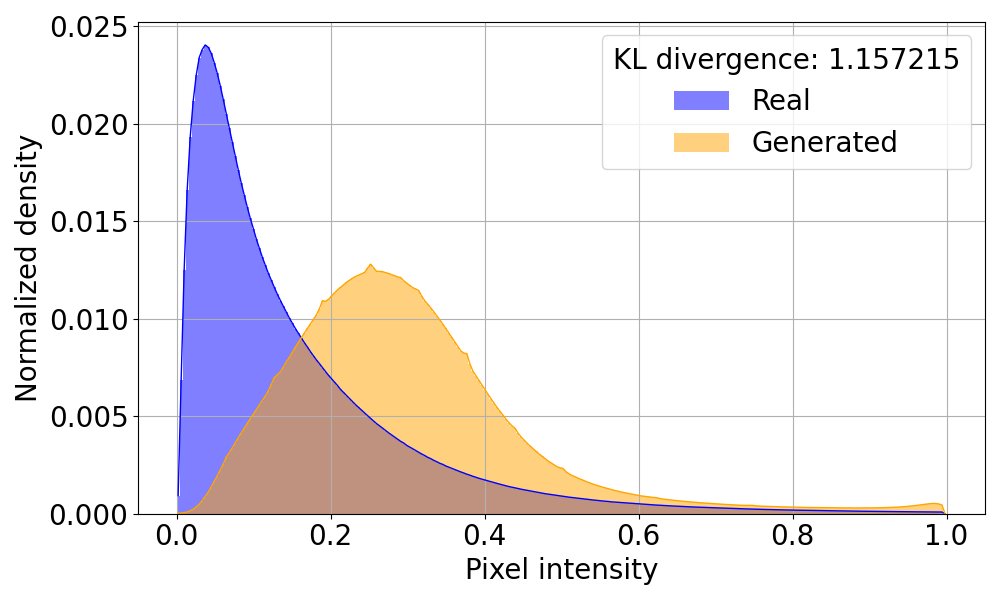} \\
        \textbf{rain-beach-6 (KL = 0.17)} & \textbf{umbrella-sand-8 (KL = 1.16)}
    \end{tabular}
    \end{adjustbox}
    \caption{Comparison of KL distances probability density distribution between 330 real and generated flattened images, with and without noise offset}
    \label{fig:histos_group1}
\end{figure}

\begin{figure}
    \centering
    \begin{adjustbox}{max width=0.5\textwidth}
    \begin{tabular}{ccc}
        \includegraphics[width=0.2\textwidth]{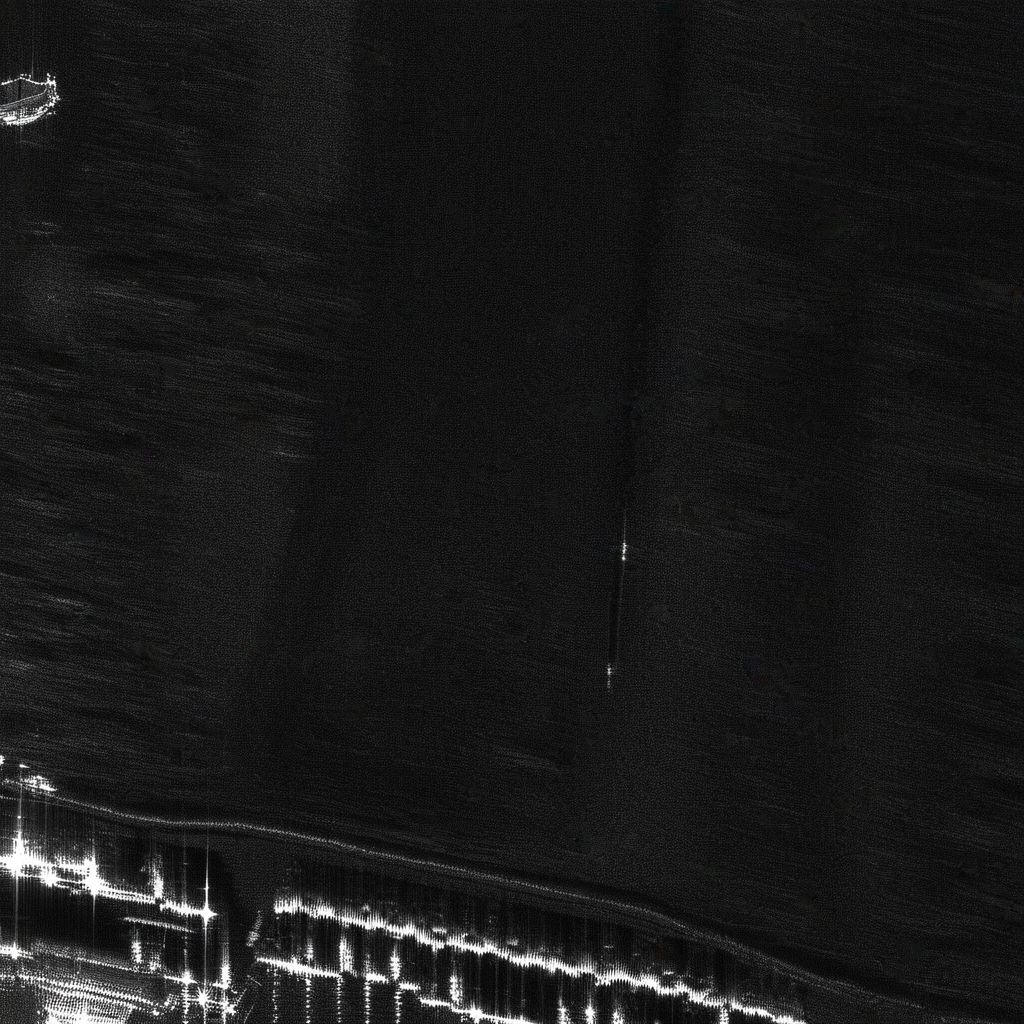} &
        \includegraphics[width=0.2\textwidth]{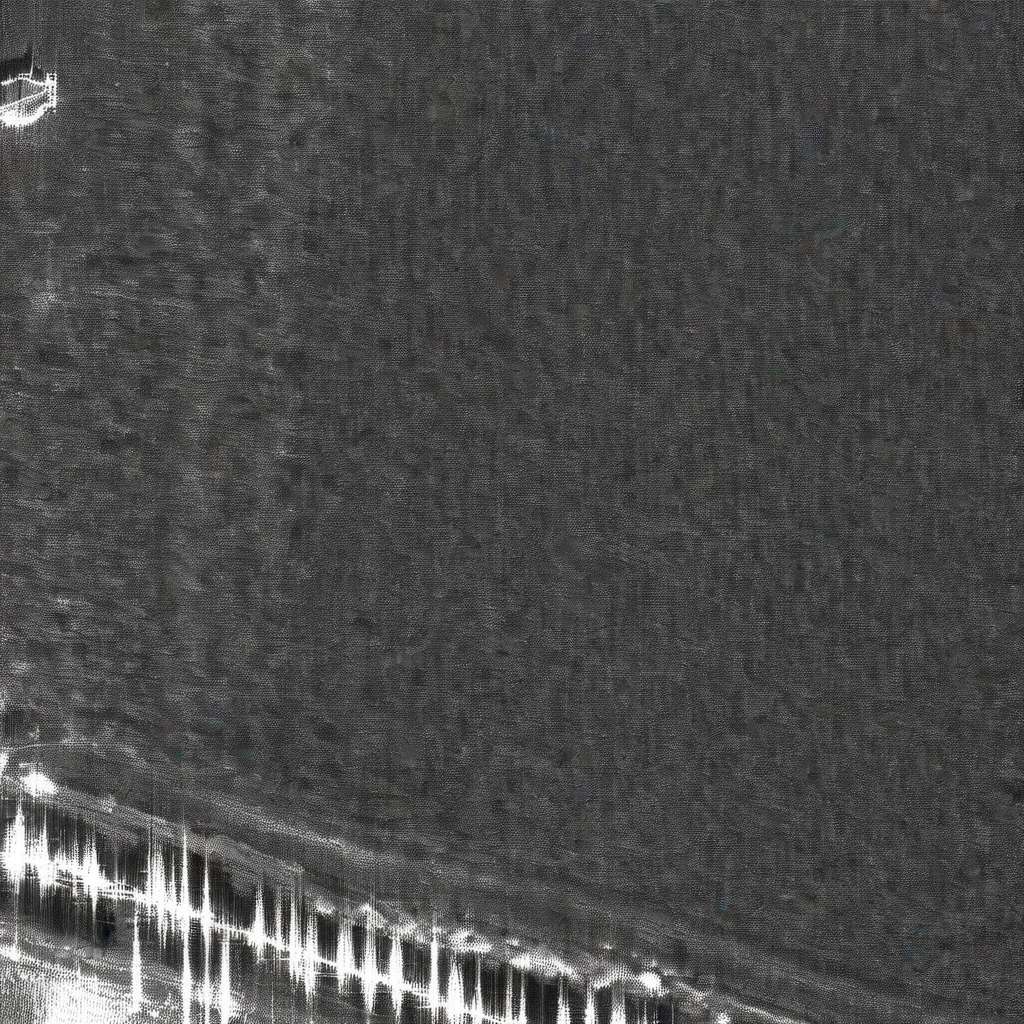} \\
        \textbf{rain-beach-6} & \textbf{umbrella-sand-8}
    \end{tabular}
    \end{adjustbox}
    \caption{Comparison of image generated (1024x1024px at 40cm) during training - at epoch 8 - with the same prompt: "A satellite view of a port with a boat in the water and a forest nearby." (see more examples in Appendix ~\ref{appendix:training-images})}
    \label{fig:images_rain-umbrella}
\end{figure}

As shown in Figure~\ref{fig:images_rain-umbrella}, without the noise offset, the generated images have lower contrast and a gray pixel distribution, which makes it difficult to capture important SAR-specific features, such as the contrast between land and sea. On the other hand, applying a small noise offset improves the dynamic range and enhances the physical realism of the textures, as seen in both the generated samples and the reduction in KL divergence (Table~\ref{tab:noise_offset_metrics}).We also notice that the noise offset affects the pixel distribution learning but does not alter the overall scene composition. Based on these results, we apply a noise offset by default in all subsequent training runs.

\subsection{Study on the UNet, TE1 and TE2}

The Text Encoders (TEs) and the UNet backbone have different roles, and understanding their contributions is essential for effective fine-tuning. To preserve the pre-existing language knowledge that is not specific to SAR, we primarily focus on the Text Encoders (TEs). Indeed, to generate complex environmental SAR scenes, such as urban, forest, and coastal areas, we use the TEs' understanding of language, spatial relationships, and object interactions. 

In Table~\ref{tab:training_lora_full}, we compare different fine-tuning strategies, for the UNet backbone, Text Encoder 1 (TE1), and Text Encoder 2 (TE2), to assess their relative importance and degree of independence. Each module is either fully fine-tuned (i.e., all weights are updated) or fine-tuned using Low-Rank Adaptation (LoRA) adapters.

By default, LoRA adapters are applied to the attention projection layers of the UNet: $[q_{\text{proj}}, k_{\text{proj}}, v_{\text{proj}}, \text{out}_{\text{proj}}]$. For the text encoders, LoRA is also applied to the same types of projection layers. Current libraries do not support LoRA on normalization layers; however, for the configuration identified as \textit{super-bowl-2}, we include additional convolutional LoRA adapters in layers $[\text{to}_q, \text{to}_k, \text{to}_v, \text{to}_{\text{out}}.0]$ as well as convolutional layers $[\text{conv}_1, \text{conv}_2]$ to analyze the contribution of convolutions to generative performance.

\begin{table}
    \centering
    \scriptsize
    \renewcommand{\arraystretch}{1.2}
    \setlength{\tabcolsep}{4pt}

    \begin{adjustbox}{max width=0.45\textwidth}
    \begin{tabular}{p{2cm} p{0.4cm} p{0.4cm} p{0.4cm} p{0.4cm} p{0.4cm} p{0.4cm} 
                    >{\centering\arraybackslash}p{0.5cm} >{\centering\arraybackslash}p{0.5cm} 
                    >{\centering\arraybackslash}p{0.65cm}}
        \toprule
        \multirow{2}{*}{\textbf{Train ID}} 
        & \multicolumn{2}{c}{\makecell{\textbf{UNet}\\\textbf{LoRA}}}
        & \multicolumn{2}{c}{\makecell{\textbf{TE1}\\\textbf{LoRA}}}
        & \multicolumn{2}{c}{\makecell{\textbf{TE2}\\\textbf{LoRA}}}
        & \multicolumn{2}{c}{\makecell{\textbf{CLIP}\\\textbf{Rank}} $\downarrow$} 
        & \multirow{2}{*}{\textbf{KL} $\downarrow$} \\
        \cmidrule(lr){2-3} \cmidrule(lr){4-5} \cmidrule(lr){6-7} \cmidrule(lr){8-9}
        & \textbf{r} & \textbf{a}
        & \textbf{r} & \textbf{a}
        & \textbf{r} & \textbf{a}
        & $\nu$ & $\sigma$
        & \\
        \midrule
        lake-mont-9       & \flame & \flame & \snowflake & \snowflake & \snowflake & \snowflake & 1.84 & 2.07 & 0.53 \\
        \textbf{soleil-up-7} & \flame & \flame & 8 & 4 & 8 & 4 & \textbf{1.61} & \textbf{1.17} & 0.42 \\
        mummy-pen-8       & \flame & \flame & \flame & \flame & \flame & \flame & 2.19 & 3.62 & 1.78 \\
        eau-vie-4         & 256 & 128 & \flame & \flame & \flame & \flame & 2.34 & 3.77 & 1.15 \\
        smile-road-5      & 256 & 128 & 8 & 4 & \flame & \flame & 2.77 & 5.49 & \textbf{0.032} \\
        rain-beach-6      & 256 & 128 & 8 & 4 & 8 & 4 & 2.34 & 3.73 & 1.17 \\
        king-kong-9       & 256 & 128 & \flame & \flame & 8 & 4 & 2.44 & 4.24 & 1.13 \\
        super-bowl-2      & 256 & 128 & 8 & 4 & 8 & 4 & 2.32 & 3.78 & 0.49 \\
        \bottomrule
    \end{tabular}
    \end{adjustbox}
    \caption{Comparison of training configurations - at epoch 8 - with UNet (\flame: all weights, \snowflake: weights freezed) and Text Encoders with LoRA (r: rank and a: alpha).}
    \label{tab:training_lora_full}
\end{table}

\begin{figure}[htb]
    \centering
    \begin{adjustbox}{max width=0.45\textwidth}
    \begin{tabular}{ccc}

        \includegraphics[width=0.22\textwidth]{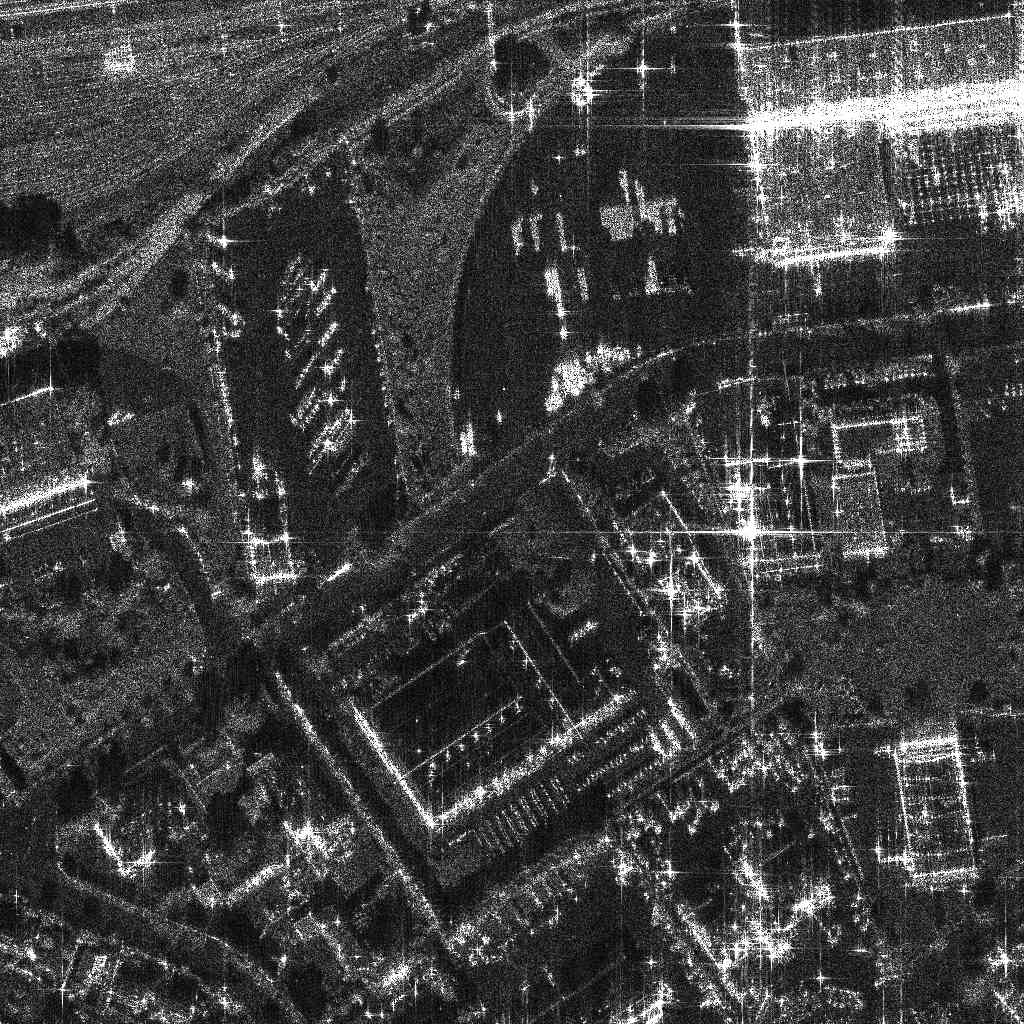} & \includegraphics[width=0.22\textwidth]{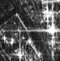} \\
        \textbf{Real SAR Image} & \textbf{Real image close up view} \\

        \includegraphics[width=0.22\textwidth]{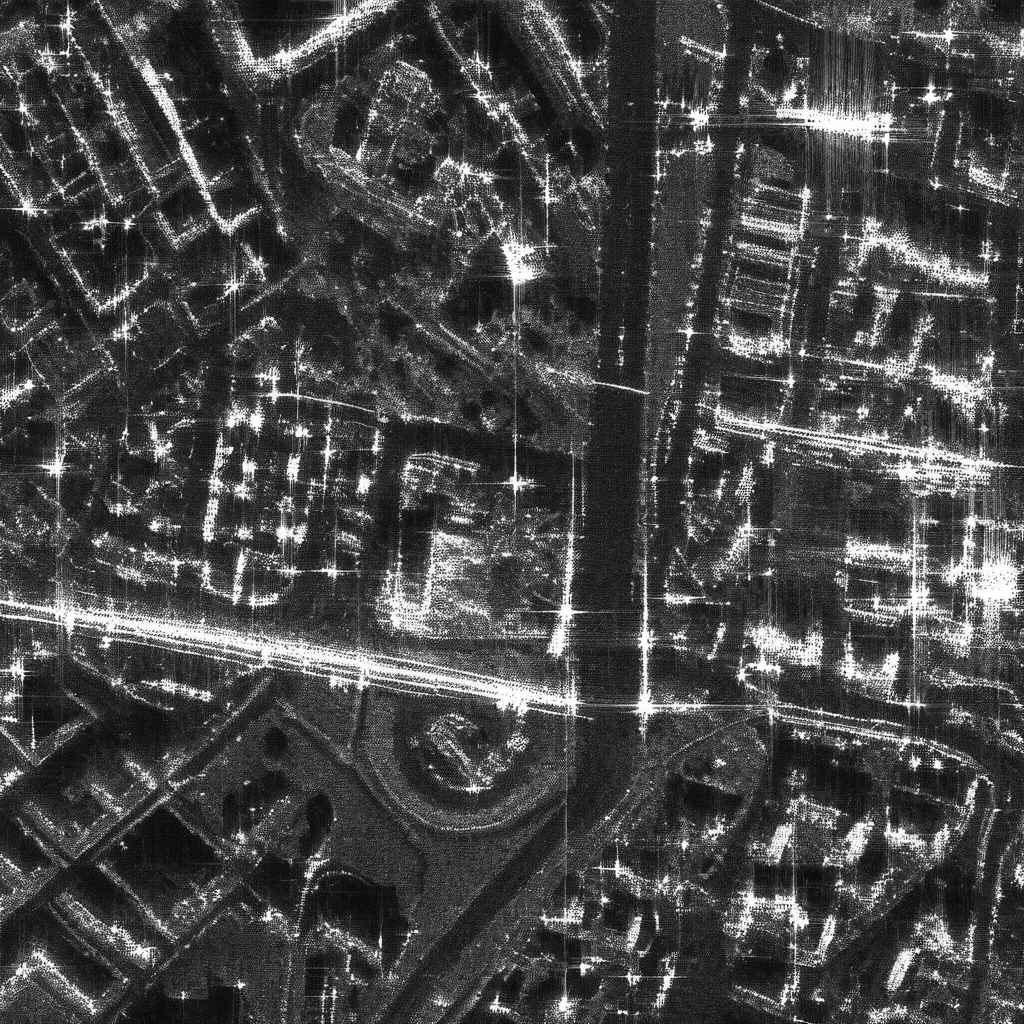} & \includegraphics[width=0.22\textwidth]{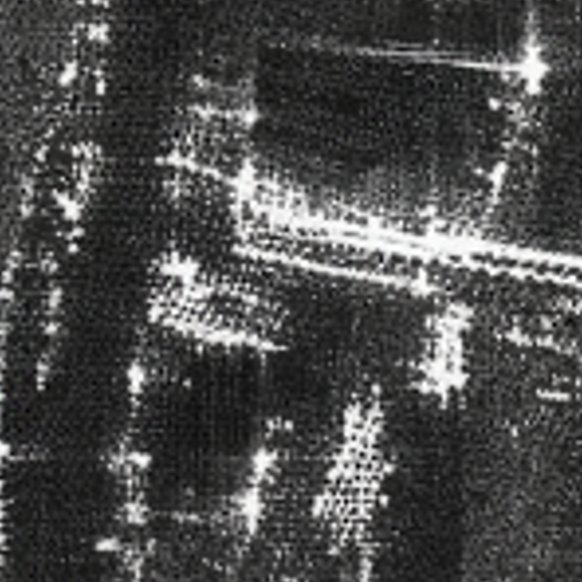} \\
        \textbf{soleil-up-7} & \textbf{soleil-up-7 close up view} \\
        
        \includegraphics[width=0.22\textwidth]{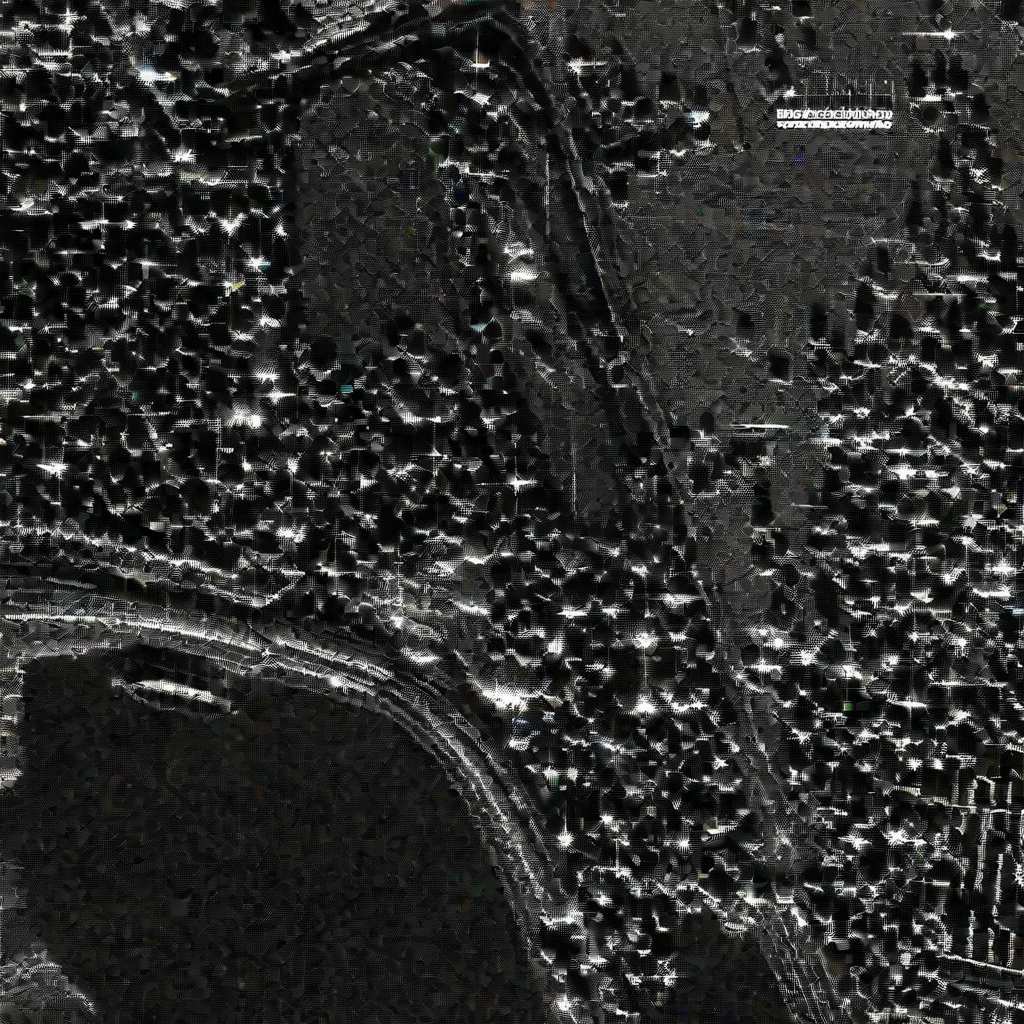} & \includegraphics[width=0.22\textwidth]{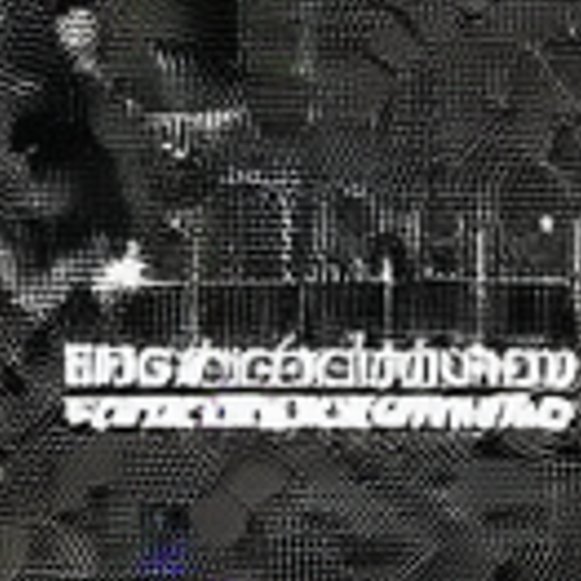} \\
        \textbf{smile-road-5} & \textbf{smile-road-5 close up view} \\

    \end{tabular}
    \end{adjustbox}
    \caption{Comparison of real and generated images (1024x1024px at 40cm) - at epoch 8 - with the same prompt: "A satellite view of a dynamic city with several buildings and a network of roads."}
    \label{fig:image_smile_road}
\end{figure}

As shown in Table~\ref{tab:training_lora_full}, full fine-tuning of the UNet consistently improves alignment and realism, as reflected in lower CLIP rank mean and variance. Although the model \textit{smile-road-5} achieves the best KL divergence score—indicating good statistical similarity to real SAR distributions—its high rank variance and weak alignment suggest instability. Indeed, visual inspection (Figure~\ref{fig:image_smile_road}) reveals that this model sometimes generates unrealistic features, such as handwritten-like artifacts, which do not match the prompt. This also impacts the general understanding of scene composition of the city compared to \textit{soleil-up-7}.

This instability may be attributed to the fact that Text Encoder 2 (TE2) was frozen during training. As mentioned in Table ~\ref{tab:sdxl_text_encoders}, TE2 provides a global caption embedding that complements the token-level embeddings from Text Encoder 1 (TE1). Since TE2 was not trained, the model may have lacked the full integration of the global context provided by the text embeddings from TE2, leading to the observed inconsistencies and unrealistic features in the generated images.

As shown in Appendix~\ref{appendix:training-images}, Figure~\ref{fig:subplot_models_full}, when we fully fine-tune all components of the model (\textit{mummy-pen-8}), we observe that the model does not converge, and color artifacts persist even after 8 epochs. In contrast, we see a large difference between all configurations when we fully train the UNet with a LoRA on both Text Encoders (\textit{soleil-up-7}), compared to the other configurations. This impacts both scene composition (with more details in the forest, realistic buildings in the city, and patterns of real mountains), resulting in more coherent and accurate scene generation.
Quantitative results in Table~\ref{tab:training_lora_full} show also a lower CLIP rank and good KL distance. 

To better understand which parts of the UNet contribute most to learning, we analyze the magnitude of parameter updates relative to the pretrained weights. As shown in Appendix~\ref{appendix:magnitude-ratio} (Figure~\ref{fig:mawc}), the largest changes are observed in the first ResNet blocks of the downsampling path and the final layers of the upsampling path. This suggests that early convolutional layers are critical for encoding SAR-specific noise and structure, further justifying the need for full fine-tuning of the UNet.

\subsection{Effect of LoRA Rank and Scaling}

We study how the LoRA rank \(r\) and scaling factor \(\alpha\) affect adaptation. With LoRA-based approach, the frozen base weight \(W\in\mathbb{R}^{m\times n}\) of the generative foundation model is updated additively:
\begin{equation}
W' \;=\; W + \Delta W, \qquad 
\Delta W \;=\; \frac{\alpha}{r}\, A B
\end{equation}

with trainable low-rank factors \(A\in\mathbb{R}^{m\times r}\) and \(B\in\mathbb{R}^{r\times n}\). By construction \(\mathrm{rank}(\Delta W)\le r\), so \(r\) controls the update capacity (degrees of freedom) and the adapter parameter count \(r(m+n)\), while the ratio \(\alpha/r\) sets the effective update strength. In practice, \(A\) is randomly initialized and \(B\) is often initialized to zero so that \(AB=0\) at the start of training.

We set \(\alpha = r/2\), keeping \(\alpha/r=0.5\) fixed. This allows us to vary \(r\) to study capacity without changing the overall update magnitude. Intuitively, small \(r\) may underfit (too few degrees of freedom), whereas very large \(r\) may overfit and erase pretrained priors.

\begin{figure}
    \centering

    \begin{minipage}[t]{0.48\textwidth}
        \scriptsize
        \renewcommand{\arraystretch}{1.2}
        \begin{adjustbox}{max width=\textwidth}
        \begin{tabular}{p{2cm} p{0.4cm}
                        >{\centering\arraybackslash}p{0.3cm} >{\centering\arraybackslash}p{0.3cm}
                        >{\centering\arraybackslash}p{0.3cm} >{\centering\arraybackslash}p{0.3cm}
                        >{\centering\arraybackslash}p{0.45cm} >{\centering\arraybackslash}p{0.45cm} >{\centering\arraybackslash}p{0.45cm}}
            \toprule
            \multirow{2}{*}{\textbf{Train ID}} 
            & \multirow{2}{*}{\textbf{UNet}} 
            & \multicolumn{2}{c}{\makecell{\textbf{TE1}\\\textbf{LoRA}}} 
            & \multicolumn{2}{c}{\makecell{\textbf{TE2}\\\textbf{LoRA}}} 
            & \multicolumn{2}{c}{\makecell{\textbf{CLIP}\\\textbf{Rank}}$\downarrow$} 
            & \multirow{2}{*}{\textbf{KL} $\downarrow$} \\
            \cmidrule(lr){3-4} \cmidrule(lr){5-6} \cmidrule(lr){7-8}
            & & \textbf{r} & \textbf{a} & \textbf{r} & \textbf{a} & $\nu$ & $\sigma$ & \\
            \midrule
            soleil-up-7       & \flame & 8     & 4     & 8     & 4   & \textbf{1.61} & \textbf{1.17} & 0.42 \\
            apple-color-6     & \flame & 64    & 32    & 64    & 32    & 1.74    & 1.77 & \textbf{0.37}   \\
            fiber-network-6   & \flame & 128   & 64    & 128   & 64     & 1.77    & 1.59  & 0.42  \\
            screen-light-4    & \flame & 256   & 128   & 256   & 128    & 1.70    & 1.65 & \textbf{0.37}   \\
            \bottomrule
        \end{tabular}
        \end{adjustbox}
        \captionof{table}{Comparison of LoRA configurations – at epoch 8 – for Text Encoders 1 and 2 (r: rank and a: alpha) and UNet full fine-tuning (\flame).}
        \label{tab:merged_lora_rank_alpha}
    \end{minipage}
    \hfill
    \begin{minipage}[t]{0.48\textwidth}
        \includegraphics[width=\textwidth]{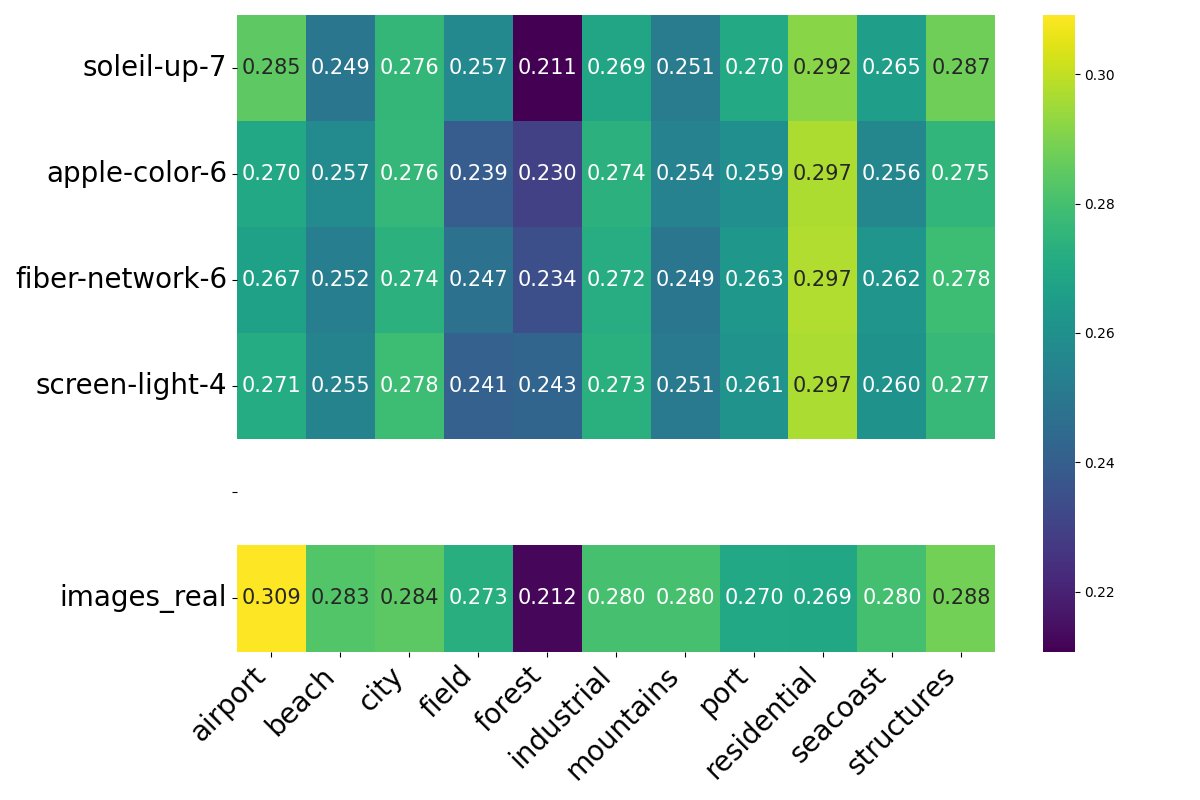}
        \captionof{figure}{Mean cosine distance between image-text pairs, compared to real ones using SAR-CLIP Model.}
        \label{fig:heatmap-_lora_rank}
    \end{minipage}

\end{figure}

In our experiments, we fully fine-tune the UNet (\flame) and apply LoRA to both text encoders (TE1/TE2) with different \((r,\alpha)\) pairs under this rule. Results are summarized in Table~\ref{tab:merged_lora_rank_alpha}.

Contrary to what has been observed in the literature, our results suggest that increasing the LoRA \textit{rank} beyond certain values does not enhance vision–language alignment in the SAR generation setting. As shown in Table~\ref{tab:merged_lora_rank_alpha}, the best (lowest) CLIP Rank score is achieved with a relatively low-rank setting (\(r=8\), \(\alpha=4\)) used in the \textit{soleil-up-7} model, whereas larger ranks (e.g., \textit{fiber-network-6}, \textit{screen-light-4}) do not produce higher scores.

From the equation below, and since we set $\alpha = r/2$, increasing $r$ increases the adapter parameter count $r(m+n)$ and raises the capacity of $\Delta W$ (because $\mathrm{rank}(\Delta W)\le r$), while the effective update scale $\alpha/r$ remains constant. The results indicate that, for text-encoder adaptation to SAR, adding degrees of freedom to \(\Delta W\) beyond a moderate level does not translate into better vision–language alignment on our data; the low-rank setting preserves alignment more effectively for the same update magnitude.

Figure~\ref{fig:heatmap-_lora_rank} is consistent with this interpretation: cosine similarities between image–prompt pairs for \textit{soleil-up-7} are closer to those for real SAR–caption pairs, indicating more faithful semantic alignment under low-rank updates.



\subsection{Study on Batch Size}

In our dataset, SAR images exhibit considerable variability in contrast due to differences in scene content and acquisition conditions. For instance, forested areas may appear brighter when wet, such as after rainfall or when near water bodies. Agricultural fields show distinct backscatter signatures depending on crop type, growth stage, or soil moisture. Urban regions also vary depending on building materials and orientation. This intrinsic variability presents challenges in modeling consistent textural and radiometric patterns.

\begin{figure}[htb]
\centering
\begin{minipage}[t]{0.48\textwidth}
    \centering
    \scriptsize
    \renewcommand{\arraystretch}{1.1}
    \setlength{\tabcolsep}{4pt}

    \begin{adjustbox}{max width=\textwidth}
    \begin{tabular}{p{2.1cm} p{0.4cm}
                    >{\centering\arraybackslash}p{0.3cm} >{\centering\arraybackslash}p{0.3cm}
                    >{\centering\arraybackslash}p{0.3cm} >{\centering\arraybackslash}p{0.3cm}
                    >{\centering\arraybackslash}p{0.45cm}
                    >{\centering\arraybackslash}p{0.45cm} >{\centering\arraybackslash}p{0.45cm}
                    >{\centering\arraybackslash}p{0.45cm}}
        \toprule
        \multirow{2}{*}{\textbf{Train ID}} 
        & \multirow{2}{*}{\textbf{UNet}} 
        & \multicolumn{2}{c}{\makecell{\textbf{TE1}\\\textbf{LoRA}}} 
        & \multicolumn{2}{c}{\makecell{\textbf{TE2}\\\textbf{LoRA}}} 
        & \multirow{2}{*}{\makecell{\textbf{Batch}\\\textbf{Size}}}
        & \multicolumn{2}{c}{\makecell{\textbf{CLIP}\\\textbf{Rank}} $\downarrow$} 
        & \multirow{2}{*}{\textbf{KL} $\downarrow$} \\
        \cmidrule(lr){3-4} \cmidrule(lr){5-6} \cmidrule(lr){8-9}
        & & \textbf{r} & \textbf{a} & \textbf{r} & \textbf{a} & & $\nu$ & $\sigma$ & \\
        \midrule
        soleil-up-7       & \flame & 8 & 4 & 8 & 4 & 16 & \textbf{1.61} & 1.17 & 0.43 \\
        king-elephant-9   & \flame & 8 & 4 & 8 & 4 & 32 & 1.63 & \textbf{1.05} & 0.42 \\
        whale-north-8     & \flame & 8 & 4 & 8 & 4 & 64 & 1.79 & 1.48 & \textbf{0.35} \\
        boad-see-9        & \flame & 8 & 4 & 8 & 4 & 128 & 1.79 & 1.45 & 0.42 \\
        \bottomrule
    \end{tabular}
    \end{adjustbox}
    \captionof{table}{Comparison of batch size values at epoch 8 with a fixed training configuration (UNet full fine-tuning \flame, LoRA for Text Encoders).}
    \label{tab:merged_batchsize_metrics}
\end{minipage}

\vspace{1em}

\begin{minipage}[t]{0.48\textwidth}
    \centering
    \begin{adjustbox}{max width=\textwidth}
    \begin{tabular}{cc}
        \includegraphics[width=0.45\textwidth]{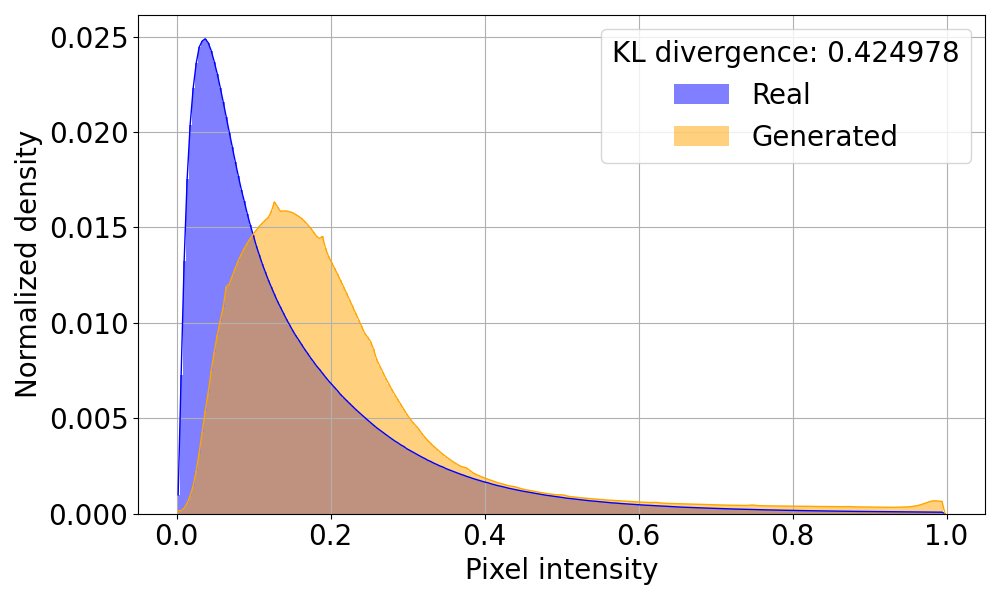} &
        \includegraphics[width=0.45\textwidth]{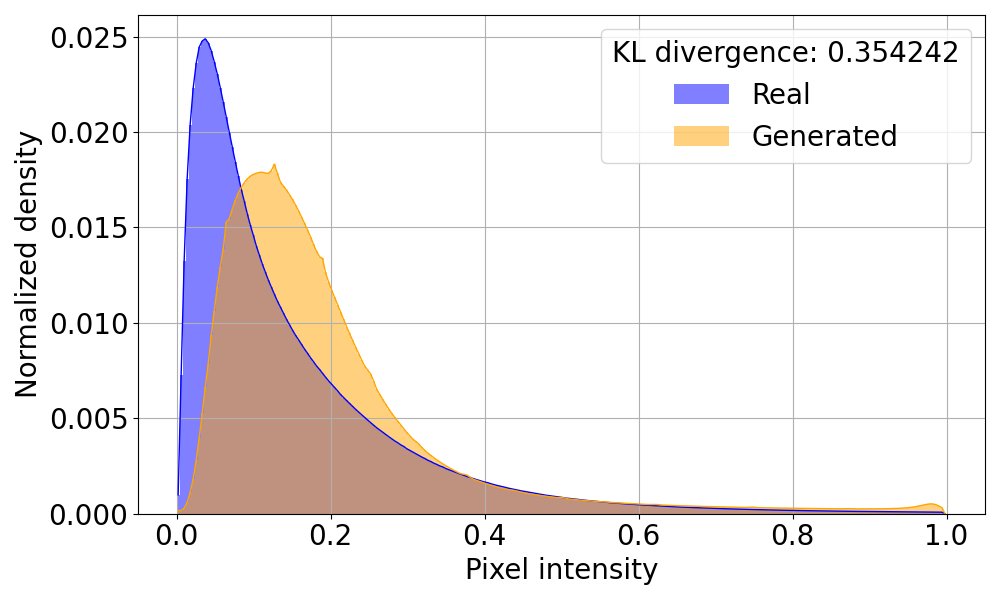} \\
        \textbf{soleil-up-7 (KL = 0.43)} & \textbf{whale-north-8 (KL = 0.35)}
    \end{tabular}
    \end{adjustbox}
    \captionof{figure}{Comparison of KL distance distributions between 330 real and generated flattened images.}
    \label{fig:histos_king-whale}
\end{minipage}
\end{figure}

To mitigate overfitting to the amplitude distribution of small batches and promote better generalization across diverse SAR characteristics, we investigate the impact of batch size on model performance. Larger batches are expected to provide more statistically representative samples within each optimization step. Due to hardware limitations, we simulate larger batch sizes using gradient accumulation. Results are presented in Table~\ref{tab:merged_batchsize_metrics}.

As shown in Table~\ref{tab:merged_batchsize_metrics} and Figure~\ref{fig:histos_king-whale}, increasing the batch size generally improves the similarity between generated and real SAR distributions, as measured by the KL divergence. Larger batches expose the model to a broader diversity of textures and amplitude statistics, helping it learn more robust and representative features. The best KL score is observed for the \textit{whale-north-8} model (batch size = 64), suggesting a potential trade-off: extremely large batches (e.g., 128) may lead to diminishing returns or underfitting, while mid-sized batches offer an optimal balance between generalization and convergence.

\subsection{Study on VAE Decoder}

While most studies using SDXL take the original Variational Autoencoder (VAE) without modification, we hypothesize that fine-tuning its encoder and/or decoder components may improve latent representation quality for SAR images. This is motivated by the fact that SAR data differ significantly from natural images in terms of statistical structure, noise characteristics (e.g., speckle), and semantic content.

Fine-tuning the pretrained VAE used in SDXL, however, is particularly challenging when transferring to a new domain such as SAR. In our experiments, fine-tuning the VAE encoder often led to latent space instability and mode collapse. In practice, the pretrained VAE already reconstructs SAR images with low error, and the decoded outputs follow a Rayleigh-like amplitude distribution that closely resembles real SAR data. Attempts to adjust the encoder disrupted this alignment and led to overfitting.

Moreover, training separately the VAE from the rest of the model (UNet and text encoders) proved to be ineffective, because coherence across components is essential in foundation model pipelines. To address this, we chose to fine-tune only the VAE decoder jointly with the UNet and both text encoders during a final refinement phase.

Despite the VAE’s strong reconstruction ability in pixel space, generating SAR images from pure Gaussian noise during inference remains challenging. While UNet fine-tuning helps, the decoder often fails to fully reproduce the textures and amplitude dynamics typical of SAR data.

To mitigate this, we perform a short refinement of the VAE decoder together with the UNet and the text encoders. As noted in Section~\ref{sec:SDXL}, the model is conditioned with timesteps $t\in[0,1000]$ sampled uniformly. To improve performance while preserving the pretrained semantic prior, we bias fine-tuning toward the final stages of the reverse diffusion (low-noise regime), which correspond to early timesteps of the forward process. Concretely, we run a single epoch and restrict training timesteps to the last 15\% of the reverse-diffusion schedule, which concentrates learning near reconstruction and reduces latent drift, improving SAR image fidelity.

In addition, we add a Kullback--Leibler (KL) divergence term to the loss to minimize the divergence between the amplitude distribution of the generated image $\hat{x}$ and that of the target image $x$:
{\small
\begin{equation}
\mathcal{L}_{\mathrm{KL}}
\;=\;
D_{\mathrm{KL}}\!\big(P_{\mathrm{real}}(x)\,\|\,P_{\mathrm{gen}}(\hat{x})\big).
\end{equation}
}
The total loss used during refinement is
{\small
\begin{equation}
\mathcal{L}_{\mathrm{refine}}
\;=\;
\mathcal{L}_{\mathrm{base}} \;+\; \lambda_{\mathrm{KL}}\,\mathcal{L}_{\mathrm{KL}},
\end{equation}
}
where $\mathcal{L}_{\mathrm{base}}$ denotes the standard diffusion (noise-prediction) loss used in our setup, and $\lambda_{\mathrm{KL}}$ controls the weight of the distribution-matching term.

\begin{figure}
    \centering
    \begin{minipage}[t]{0.48\textwidth}
        \small
        \renewcommand{\arraystretch}{1.1}
        \setlength{\tabcolsep}{4pt}
    
        \begin{adjustbox}{max width=\textwidth}
        \begin{tabular}{p{3.5cm} p{0.4cm}
                        >{\centering\arraybackslash}p{0.3cm} >{\centering\arraybackslash}p{0.3cm}
                        >{\centering\arraybackslash}p{0.3cm} >{\centering\arraybackslash}p{0.3cm}
                        >{\centering\arraybackslash}p{0.3cm} >{\centering\arraybackslash}p{0.3cm}
                        >{\centering\arraybackslash}p{0.45cm} >{\centering\arraybackslash}p{0.45cm}
                        >{\centering\arraybackslash}p{0.45cm}}
            \toprule
            \multirow{2}{*}{\textbf{Train ID}} 
            & \multirow{2}{*}{\textbf{UNet}} 
            & \multicolumn{2}{c}{\textbf{VAE}} 
            & \multicolumn{2}{c}{\makecell{\textbf{TE1}\\\textbf{LoRA}}} 
            & \multicolumn{2}{c}{\makecell{\textbf{TE2}\\\textbf{LoRA}}} 
            & \multicolumn{2}{c}{\makecell{\textbf{CLIP}\\\textbf{Rank}} $\downarrow$}
            & \multirow{2}{*}{\textbf{KL} $\downarrow$} \\
            \cmidrule(lr){3-4} \cmidrule(lr){5-6} \cmidrule(lr){7-8} \cmidrule(lr){9-10}
            & & \textbf{E} & \textbf{D} & \textbf{r} & \textbf{a} & \textbf{r} & \textbf{a} & $\nu$ & $\sigma$ & \\
            \midrule
            whale-north-8 & \flame & \xmark & \xmark & 8 & 4 & 8 & 4 & 1.79 & 1.48 & 0.35 \\
            whale-north-8-refined & \snowflake & \xmark & \cmark & 8 & 4 & 8 & 4 & 1.79 & 1.82 & 0.33 \\
            
            \bottomrule
        \end{tabular}
        \end{adjustbox}
    
        \captionof{table}{Study on VAE Decoder fine-tuning — with UNet full fine-tuning (\flame) and fixed LoRA for both Text Encoders.}
        \label{tab:vae_training_configs}
    \end{minipage}
\end{figure}

    \begin{figure}

        \begin{minipage}[t]{0.48\textwidth}
            \centering
            \begin{adjustbox}{max width=\textwidth}
            \begin{tabular}{cc}
                \includegraphics[width=0.45\textwidth]{histo_kl/whale-north-8_histo_global.jpg} &
                \includegraphics[width=0.45\textwidth]{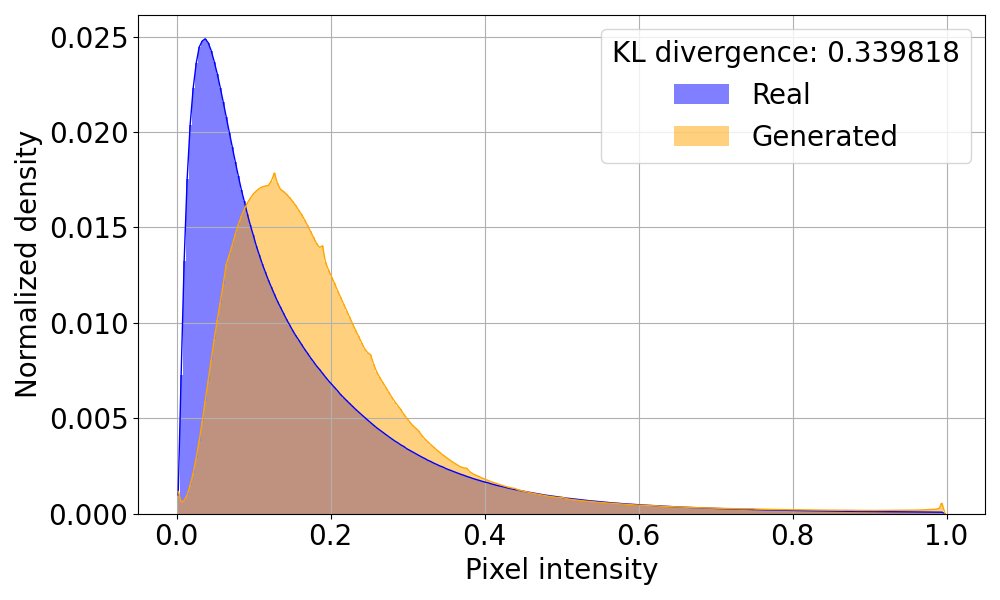} \\
                \textbf{whale-north-8 (KL = 0.35)} & \textbf{whale-north-8-refined (KL = 0.33)}
            \end{tabular}
            \end{adjustbox}
            \captionof{figure}{Comparison of KL distance distributions between 330 real and generated flattened images - with a refining training strategy on the last 15 \% of the denoising process.}
            \label{fig:histos_king-whale2}
        \end{minipage}
\end{figure}

\begin{figure}
        \hfill
        \begin{minipage}[t]{0.5\textwidth}
            \scriptsize
            \centering
            \begin{adjustbox}{max width=\textwidth}
            \begin{tabular}{cc}
                \includegraphics[width=0.38\textwidth]{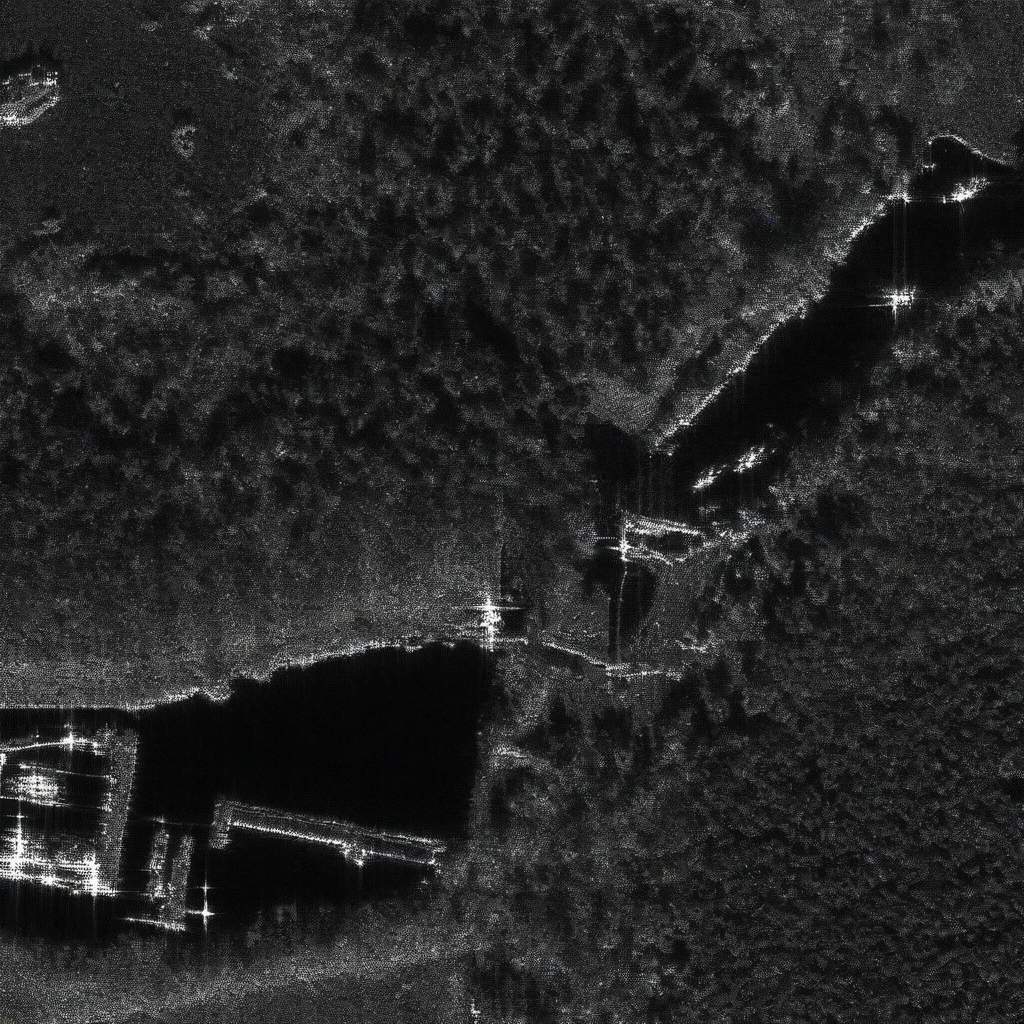} &
                \includegraphics[width=0.38\textwidth]{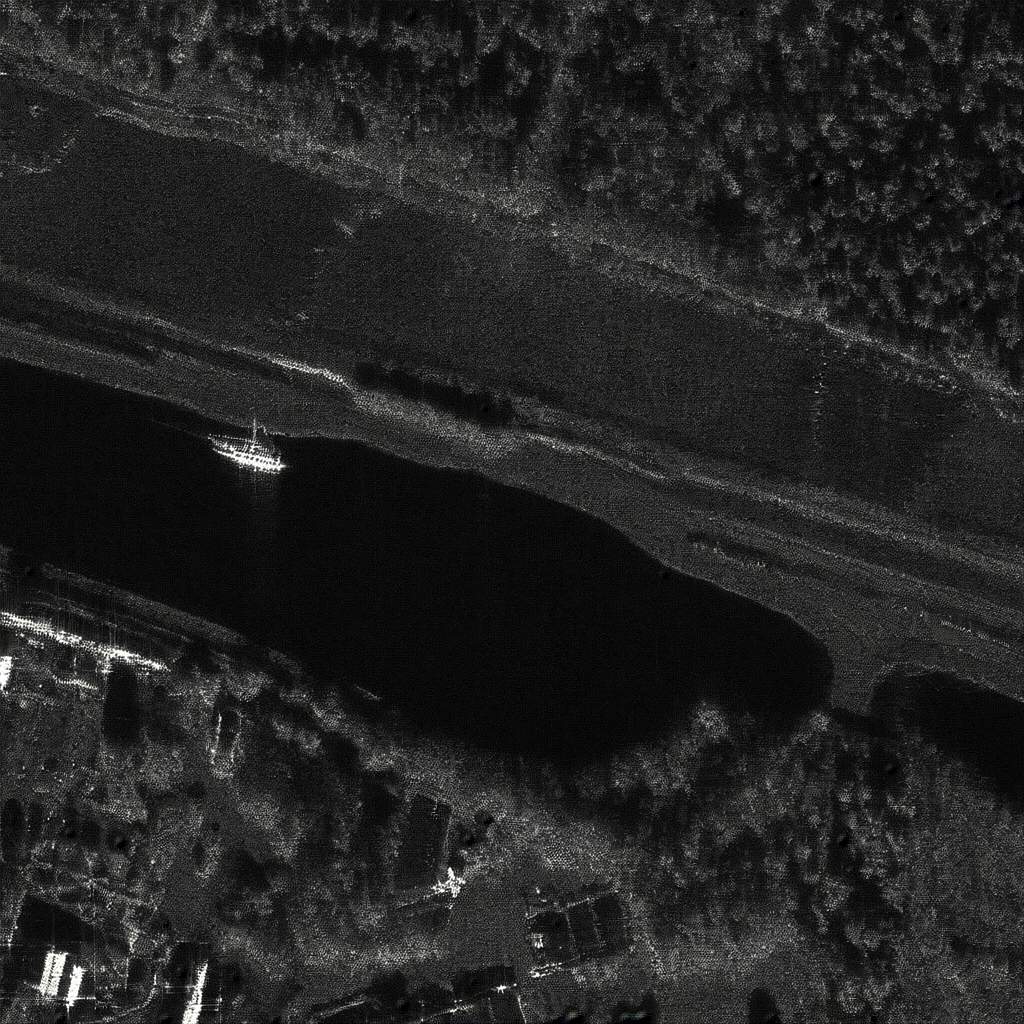} \\
                
                \includegraphics[width=0.38\textwidth]{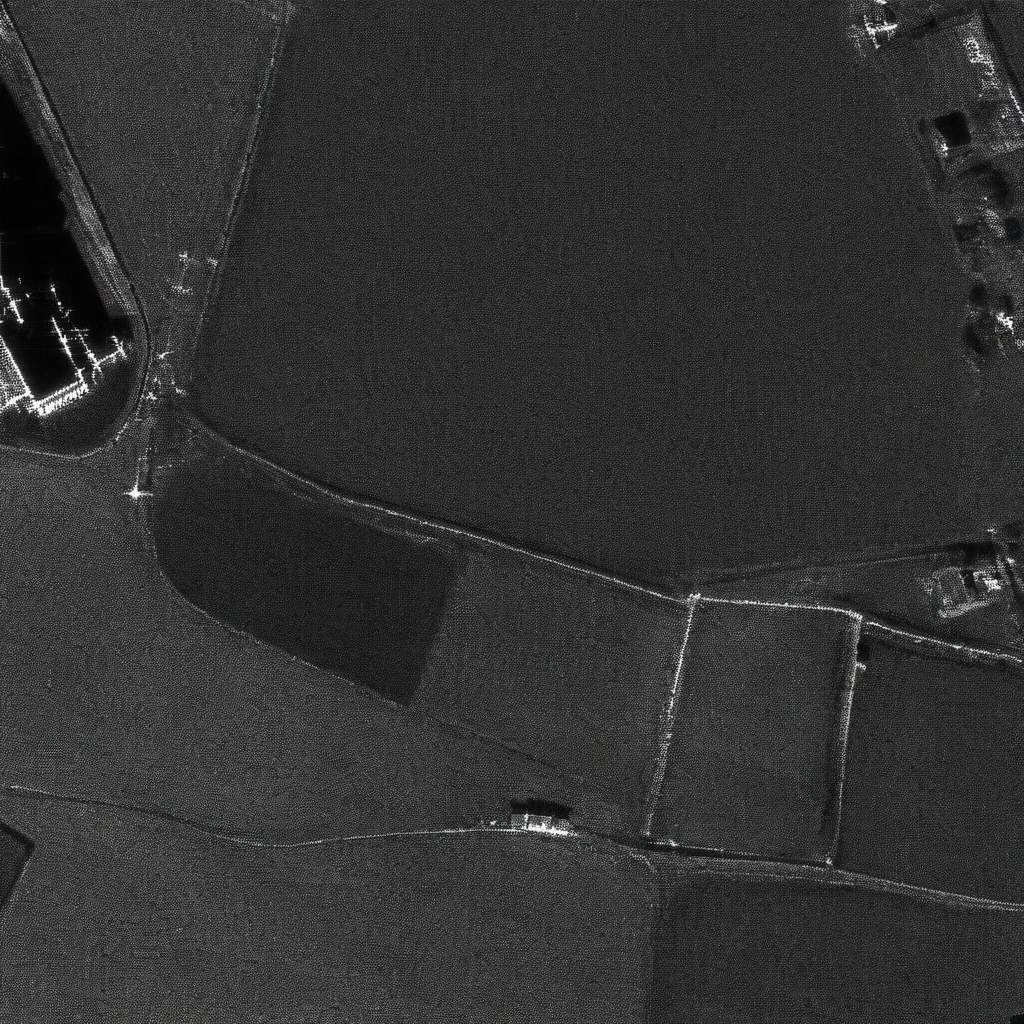} &
                \includegraphics[width=0.38\textwidth]{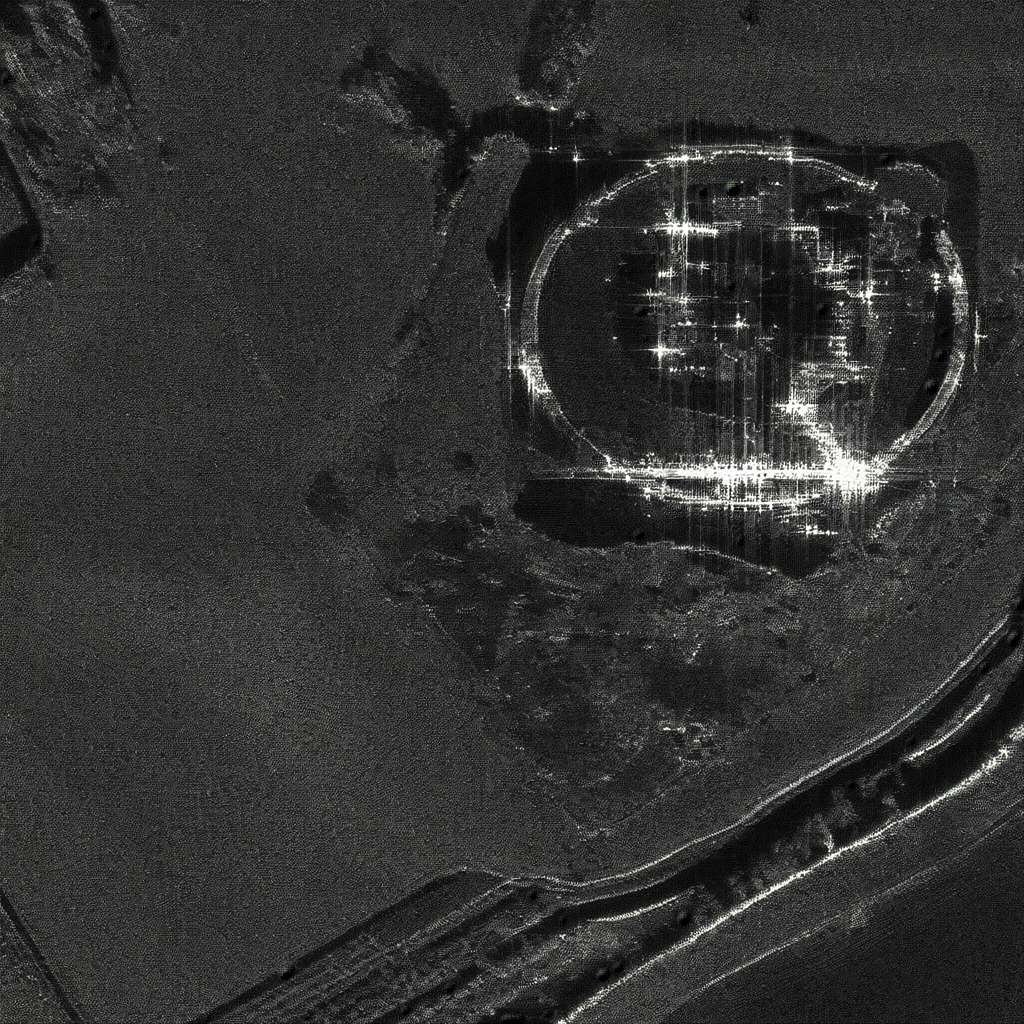} \\
    
                \includegraphics[width=0.38\textwidth]{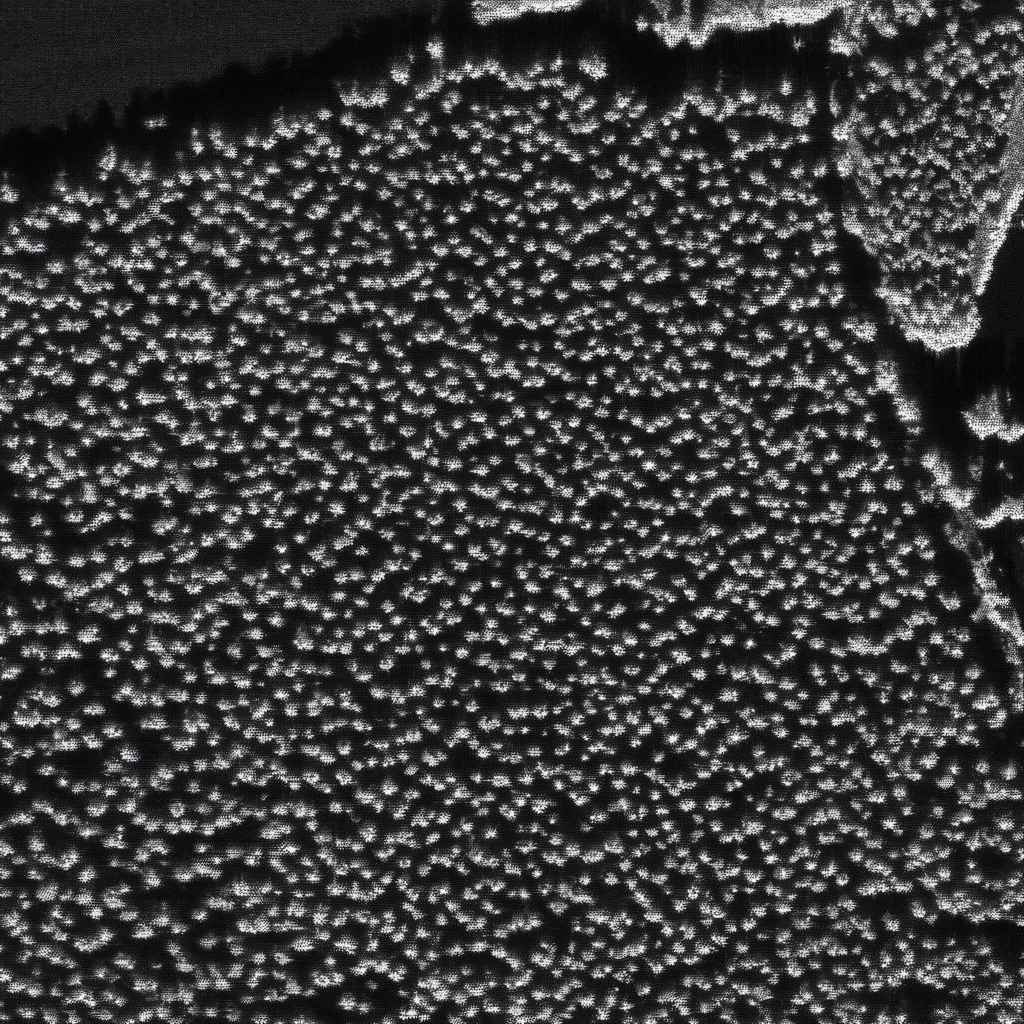} &
                \includegraphics[width=0.38\textwidth]{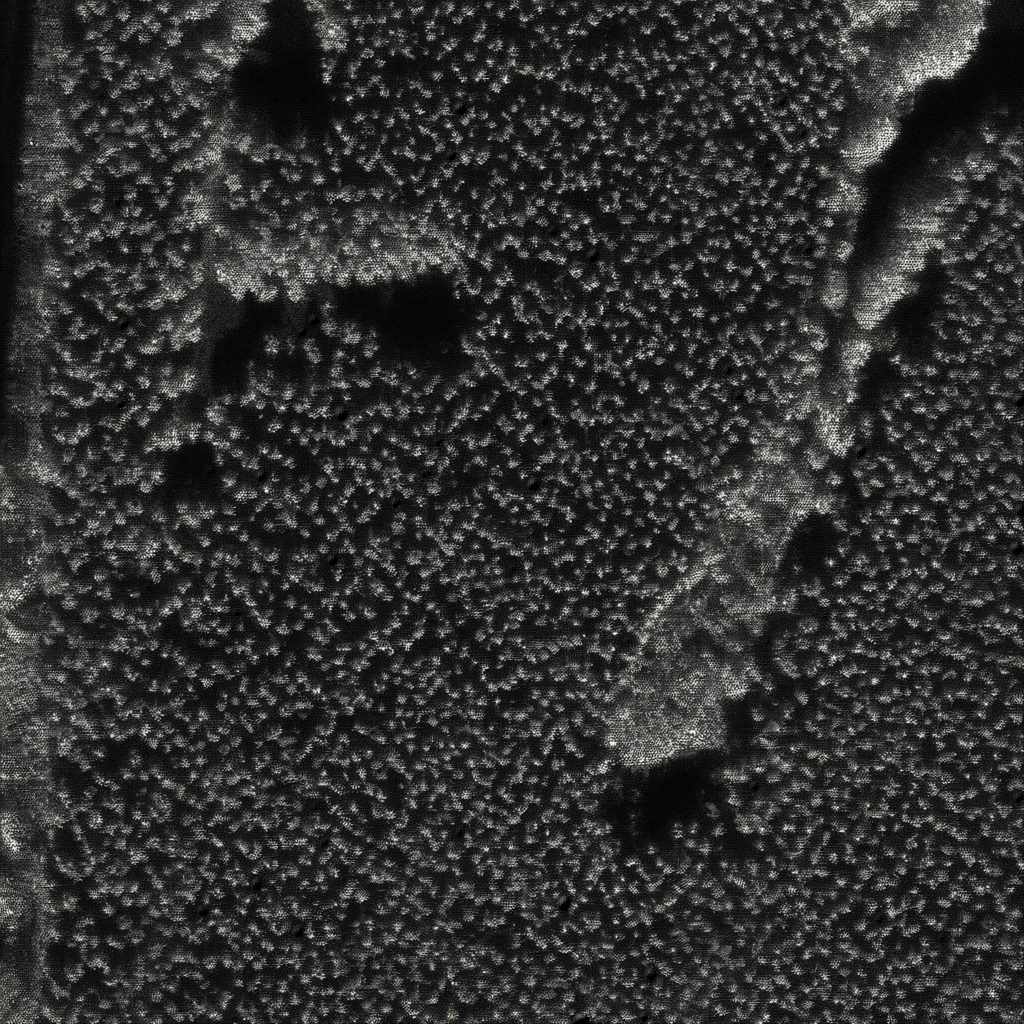} \\
                
                \textbf{whale-north-8} & \textbf{whale-north-8-refined} \\

            \end{tabular}
            \end{adjustbox}

            \caption{Comparison of generated images (1024x1024px at 40cm) — at epoch 8 — with the same prompts: (1) "A satellite view of a port with a boat in the water and a forest nearby." (2) "A satellite view of a vast expanse of land with a circular structure and a few isolated buildings." (3) "A satellite view of a dense forest with a river in the center of the forest."}
            \label{fig:image-boad-whale-tour}
        \end{minipage}

\end{figure}

Overall, we observe that the model \textit{whale-north-8-refined} has both SAR-specific fidelity and prompt-conditioned semantic competence. Beyond producing SAR  imagery, it uses semantic knowledge of the pretrained base model to assemble out-of-distribution scene configurations. As illustrated in ~\ref{fig:image-boad-whale-tour}, it successfully adds a boat within a harbor, synthesizes a circular installation in open terrain, and delineates a river through a forest—each consistent with the textual prompt in a manner that is physically consistent with SAR imaging.

\subsection{Study on <SAR> Token Embedding Learning}

To improve transfer to the SAR domain, we adopt a token-learning strategy inspired by textual inversion. We extend the tokenizer vocabulary with a new token \texttt{<SAR>} and assign it a learnable embedding optimized jointly with the model parameters.

During training, we modify the prompts by replacing general descriptions like “A satellite view of...” with SAR-specific expressions such as “A \texttt{<SAR>} image of...”. This encourages the model to associate the \texttt{<SAR>} token with the statistical and structural patterns specific to radar imagery. Gradients from the diffusion objective (and our distributional terms) update the token embedding together with the UNet and the text encoders, aligning the textual representation with SAR-consistent latent visual features.

\begin{figure}[htb]
    \centering
    \begin{minipage}[t]{0.48\textwidth}
        \small
        \renewcommand{\arraystretch}{1.1}
        \setlength{\tabcolsep}{4pt}
    
        \begin{adjustbox}{max width=\textwidth}
        \begin{tabular}{p{3.5cm} p{0.4cm}
                        >{\centering\arraybackslash}p{0.3cm} >{\centering\arraybackslash}p{0.3cm}
                        >{\centering\arraybackslash}p{0.3cm} >{\centering\arraybackslash}p{0.3cm}
                        >{\centering\arraybackslash}p{0.3cm} >{\centering\arraybackslash}p{0.3cm}
                        >{\centering\arraybackslash}p{0.45cm} >{\centering\arraybackslash}p{0.45cm}
                        >{\centering\arraybackslash}p{0.45cm}}
            \toprule
            \multirow{2}{*}{\textbf{Train ID}} 
            & \multirow{2}{*}{\textbf{UNet}} 
            & \multicolumn{2}{c}{\textbf{VAE}} 
            & \multicolumn{2}{c}{\makecell{\textbf{TE1}\\\textbf{LoRA}}} 
            & \multicolumn{2}{c}{\makecell{\textbf{TE2}\\\textbf{LoRA}}} 
            & \multicolumn{2}{c}{\makecell{\textbf{CLIP}\\\textbf{Rank}} $\downarrow$}
            & \multirow{2}{*}{\textbf{KL} $\downarrow$} \\
            \cmidrule(lr){3-4} \cmidrule(lr){5-6} \cmidrule(lr){7-8} \cmidrule(lr){9-10}
            & & \textbf{E} & \textbf{D} & \textbf{r} & \textbf{a} & \textbf{r} & \textbf{a} & $\nu$ & $\sigma$ & \\
            \midrule
            whale-north-8 & \flame & \xmark & \xmark & 8 & 4 & 8 & 4 & 1.79 & 1.48 & 0.35 \\
            tour-reine-2 & \flame & \xmark & \cmark & 8 & 4 & 8 & 4 & 1.74 & 1.35 & 0.34 \\
            heart-rose-2 & \flame & \xmark & \cmark & 8 & 4 & 8 & 4 & \textbf{1.68} & \textbf{1.31} & \textbf{0.23} \\
            
            \bottomrule
        \end{tabular}
        \end{adjustbox}
    
        \captionof{table}{Study on <SAR> Token Embedding Learning — with UNet full fine-tuning (\flame) and fixed LoRA for both Text Encoders.}
        \label{tab:emb_training_configs}
    \end{minipage}
    \end{figure}

\begin{figure}
    \centering
    
    \begin{minipage}[t]{0.48\textwidth}
        \centering
        \small
        \begin{adjustbox}{max width=\textwidth}
        \begin{tabular}{cc}
            \includegraphics[width=0.45\textwidth]{histo_kl/soleil-up-7-cosine0_histo_global.jpg} &
            \includegraphics[width=0.45\textwidth]{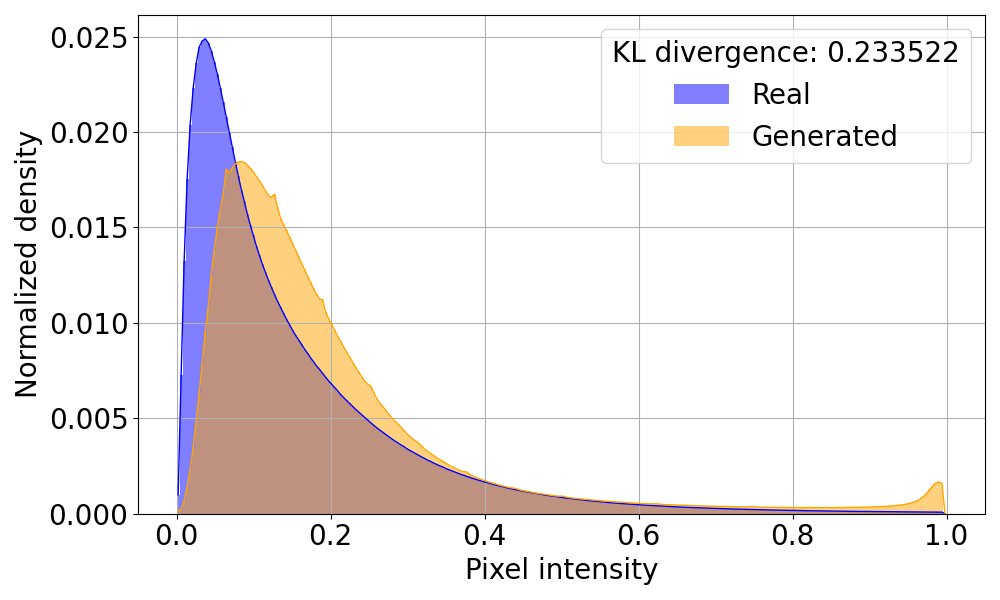} \\
            \textbf{whale-north-8 (KL = 0.35)} & \textbf{heart-rose-2 (KL = 0.23)}
        \end{tabular}
        \end{adjustbox}
        \captionof{figure}{Comparison of KL distance distributions between 330 real and generated flattened images.}
        \label{fig:histos_whale-heart}
    \end{minipage}
\end{figure}

\begin{figure}[htb]
    \centering
    \small
    \begin{minipage}[t]{0.48\textwidth}
        \scriptsize
        \centering
        \begin{adjustbox}{max width=\textwidth}
        \begin{tabular}{cc}
            \includegraphics[width=0.45\textwidth]{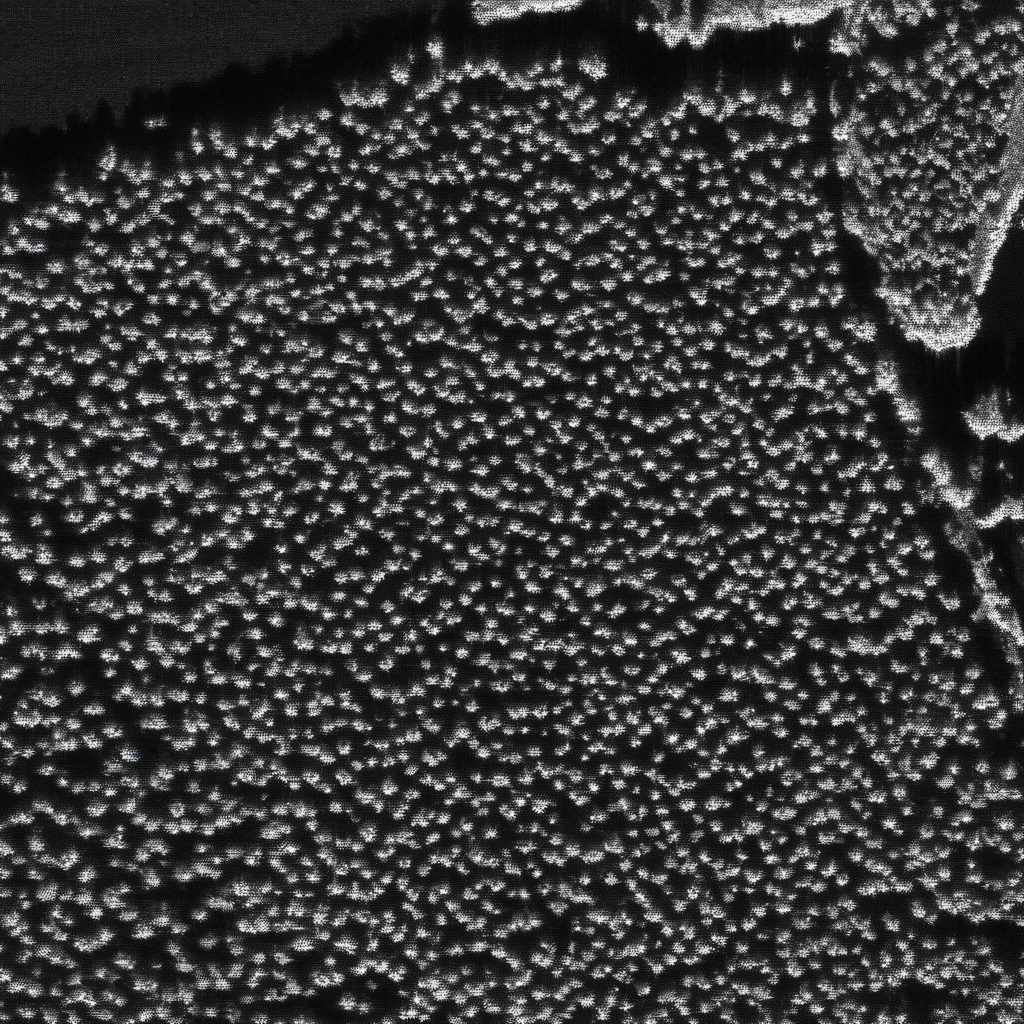} &
            \includegraphics[width=0.45\textwidth]{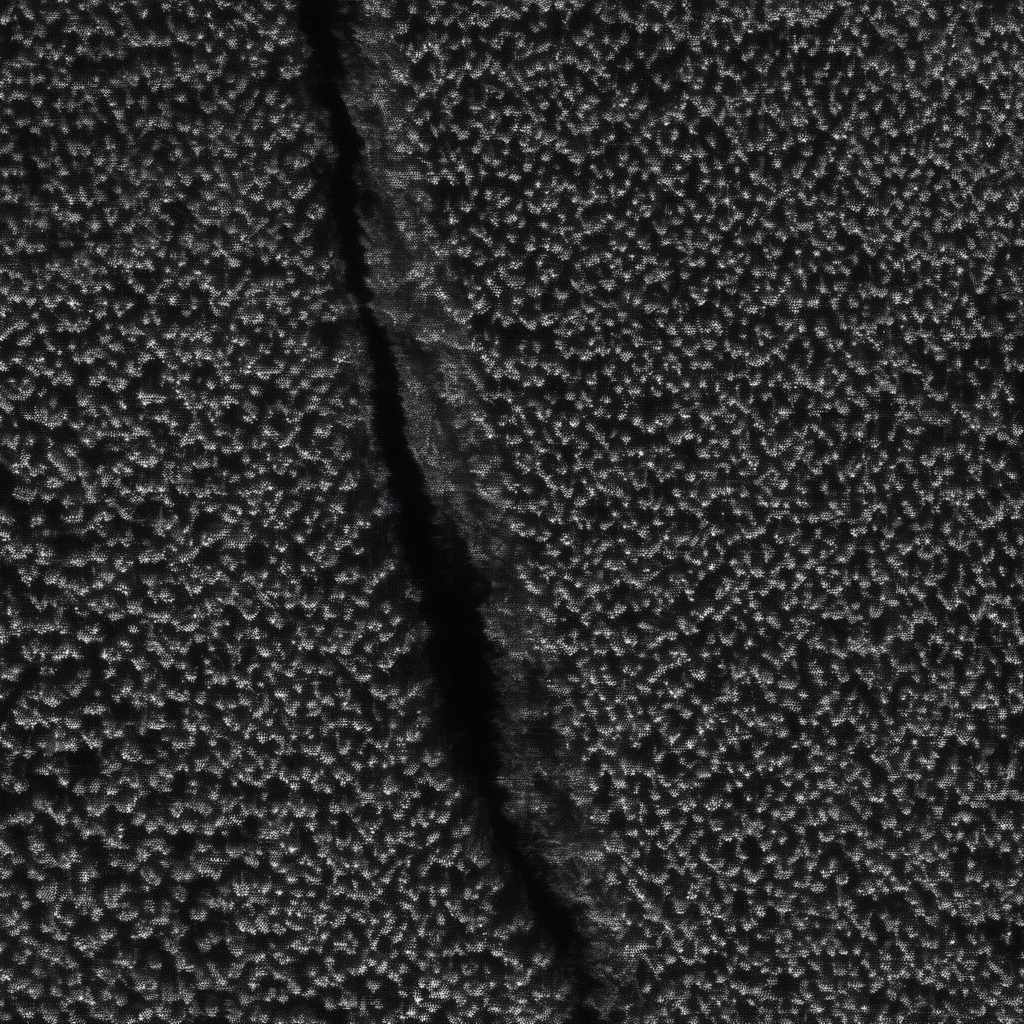} \\
            \includegraphics[width=0.45\textwidth]{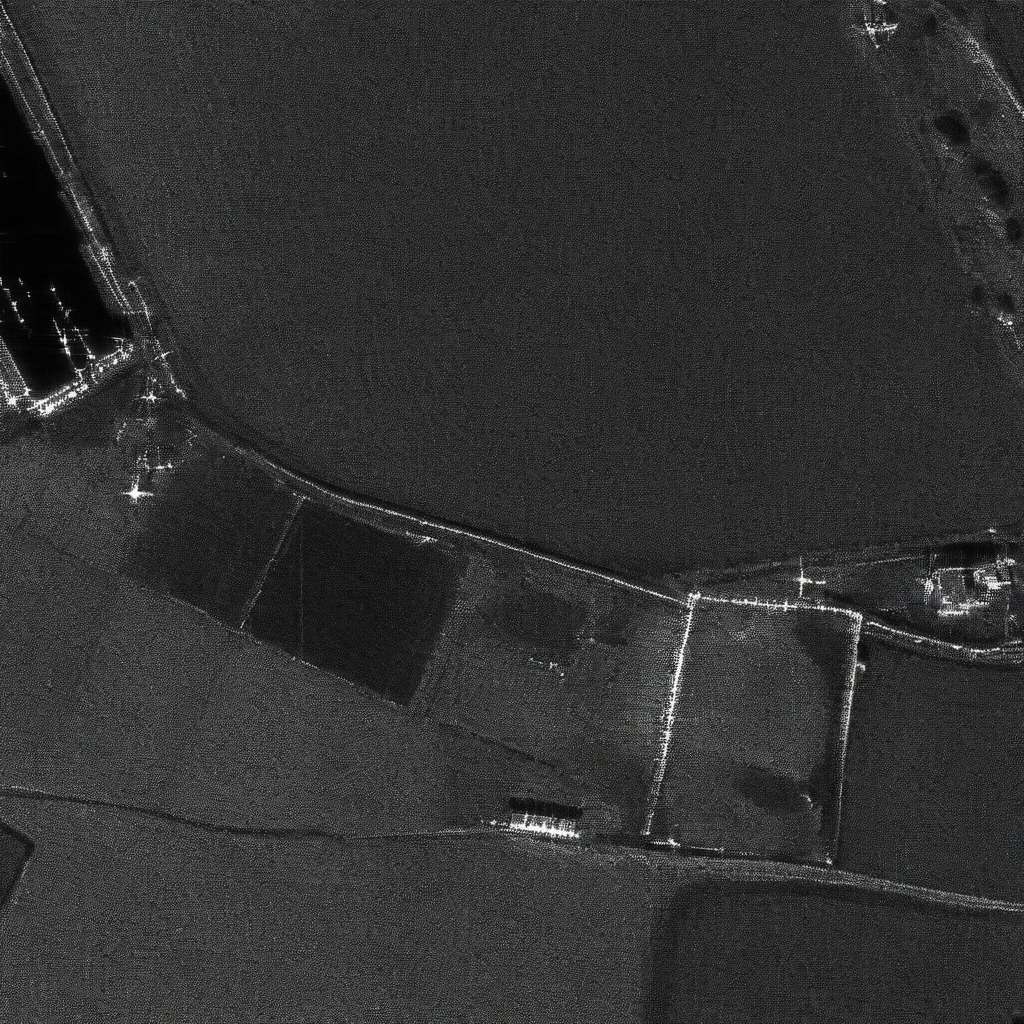} &
            \includegraphics[width=0.45\textwidth]{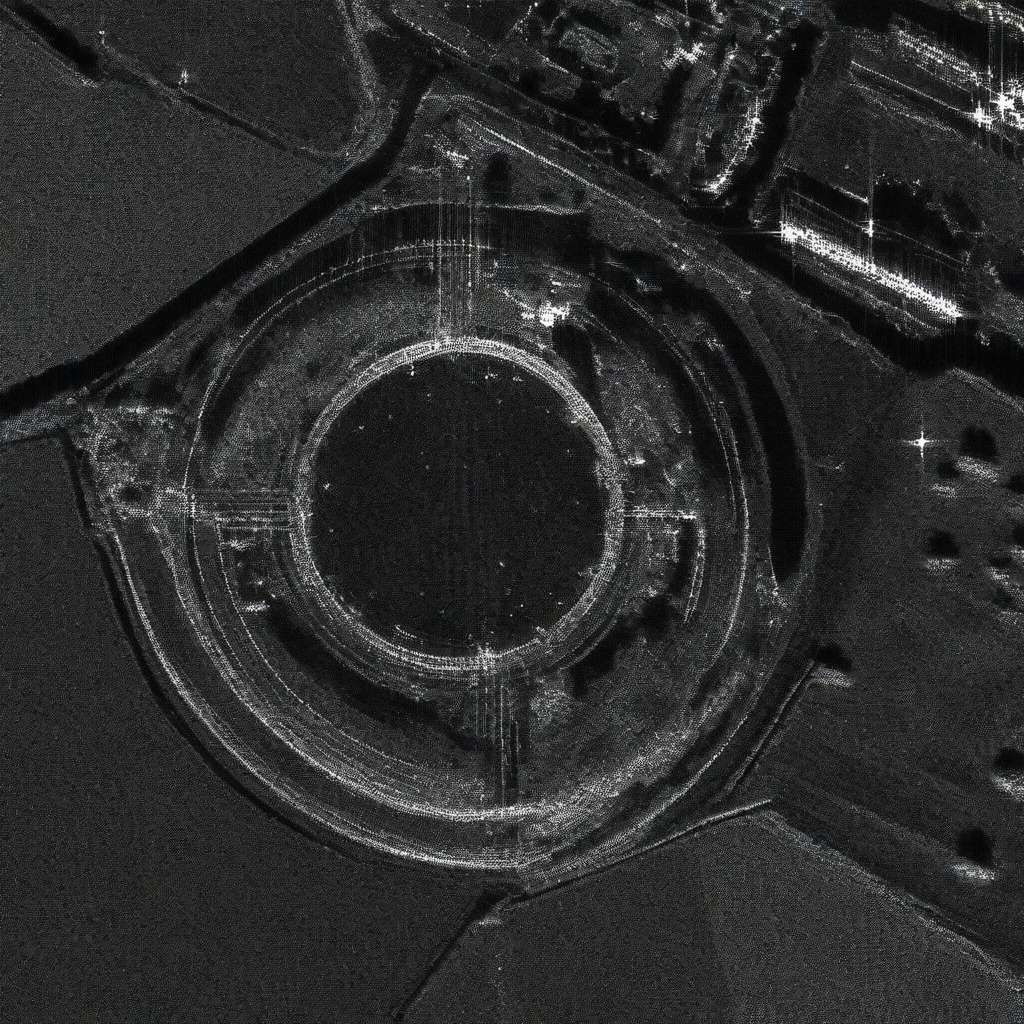} \\
            \textbf{whale-north-8} & \textbf{heart-rose-2} \\
        \end{tabular}
        \end{adjustbox}
        \captionof{figure}{Comparison of generated images (1024x1024px at 40cm) — at epoch 8 — with the same prompts: (1) "A SAR image of a dense forest with a river in the center of the forest." (2) "A SAR image of a vast expanse of land with a circular structure and a few isolated buildings."}
        \label{fig:image-boad-whale-heart}
    \end{minipage}
\end{figure}

The model \textit{heart-rose-2} achieves the lowest KL divergence (0.23) and optimal CLIP Rank scores (1.68 and 1.31 for $\nu$ and $\sigma$, respectively), indicating superior performance in generating realistic SAR images (Table \ref{tab:emb_training_configs}). Histogram comparison (Figure \ref{fig:histos_whale-heart}) reveals that the probability density distribution of \textit{heart-rose-2} more closely captures the inherent dynamics of SAR data, indicating that this model learns better representational dynamics of Synthetic Aperture Radar imagery.
And, it is capable of composing scenes that were never explicitly observed during training like “a circular structure in a city center” or “a river in the middle of a dense forest”  (Figure~\ref{fig:image-boad-whale-heart}). In general, \textit{heart-rose-2} generates high-quality SAR images with realistic spatial structures and texture patterns, as illustrated in Figure~\ref{fig:heart_rose_grid}. 

Given the class imbalance in our dataset (e.g., Forest, Airport, City, etc.), combined with the fact that SAR images have distinct distribution dynamics depending on the observed scene, we further evaluated model behavior per semantic category:

\begin{table}[htp]
    \centering
    \scriptsize
    \renewcommand{\arraystretch}{1.2}
    \setlength{\tabcolsep}{4pt}
    
    \begin{adjustbox}{max width=\columnwidth} 
    \begin{tabular}{l
                    >{\centering\arraybackslash}p{0.6cm}
                    >{\centering\arraybackslash}p{0.6cm}
                    >{\centering\arraybackslash}p{0.6cm}
                    >{\centering\arraybackslash}p{0.6cm}
                    >{\centering\arraybackslash}p{0.6cm}
                    >{\centering\arraybackslash}p{0.67cm}}
        \toprule
        \multirow{3}{*}{\textbf{Category}} 
        & \multicolumn{3}{c}{\textbf{whale-north-8}} 
        & \multicolumn{3}{c}{\textbf{heart-rose-2}} \\
        \cmidrule(lr){2-4} \cmidrule(lr){5-7}
        & \multicolumn{2}{c}{\makecell{\textbf{CLIP}\\\textbf{Rank}} $\downarrow$} 
        & \multirow{2}{*}{\textbf{KL} $\downarrow$} 
        & \multicolumn{2}{c}{\makecell{\textbf{CLIP}\\\textbf{Rank}} $\downarrow$} 
        & \multirow{2}{*}{\textbf{KL} $\downarrow$} \\
        & $\nu$ & $\sigma$ & & $\nu$ & $\sigma$ & \\
        \midrule
        Forest       & 2.27 & 1.80 & 1.30 & \textbf{1.77} & \textbf{0.65} & \textbf{1.23} \\
        City         & 1.73 & 1.60 & 0.35 & \textbf{1.37} & \textbf{0.37} & \textbf{0.23} \\
        Field        & 2.37 & 2.50 & 0.17 & \textbf{2.07} & \textbf{1.60} & \textbf{0.16} \\
        Port         & \textbf{2.20} & \textbf{1.89} & 1.02 & 2.60 & 4.97 & \textbf{0.42} \\
        Airport      & 2.37 & 3.03 & 0.18 & \textbf{2.27} & \textbf{2.80} & \textbf{0.16} \\
        Mountains    & 2.47 & 3.32 & 0.73 & \textbf{1.73} & \textbf{1.53} & \textbf{0.28} \\
        Structures   & \textbf{2.50} & \textbf{3.05} & 0.30 & 2.60 & 3.57 & \textbf{0.24} \\
        Seacoast     & 2.57 & 3.65 & 0.36 & \textbf{1.57} & \textbf{0.65} & \textbf{0.31} \\
        Beach        & \textbf{1.97} & \textbf{2.43} & 0.31 & \textbf{1.97} & 2.70 & \textbf{0.24} \\
        Industrial   & \textbf{1.93} & \textbf{1.60} & 0.23 & 2.00 & 2.00 & \textbf{0.22} \\
        Residential  & 1.43 & \textbf{0.51} & 0.14 & \textbf{1.40} & 0.84 & \textbf{0.08} \\
        \bottomrule
    \end{tabular}
    \end{adjustbox}
    \caption{Per-category CLIP Rank scores and KL divergence for two models \textit{whale-north-8} and \textit{heart-rose-2}.}
    \label{tab:model_summary_emb}
\end{table}

As shown in Table~\ref{tab:model_summary_emb}, the \textit{heart-rose-2} model outperforms \textit{whale-north-8} in nearly all categories, particularly in forested, mountainous, and seacoast scenes, which typically exhibit more complex scattering patterns. Specifically, the \textit{heart-rose-2} model achieves the lowest KL distance across all categories and the best CLIP Rank scores in 8 out of 11 categories.

\begin{figure}[H]
    \centering
    \scriptsize
    \renewcommand{\arraystretch}{1.2}
    \begin{tabular}{cc}
        \textbf{Real SAR images} & \textbf{Generated SAR images} \\
        \includegraphics[width=0.45\linewidth]{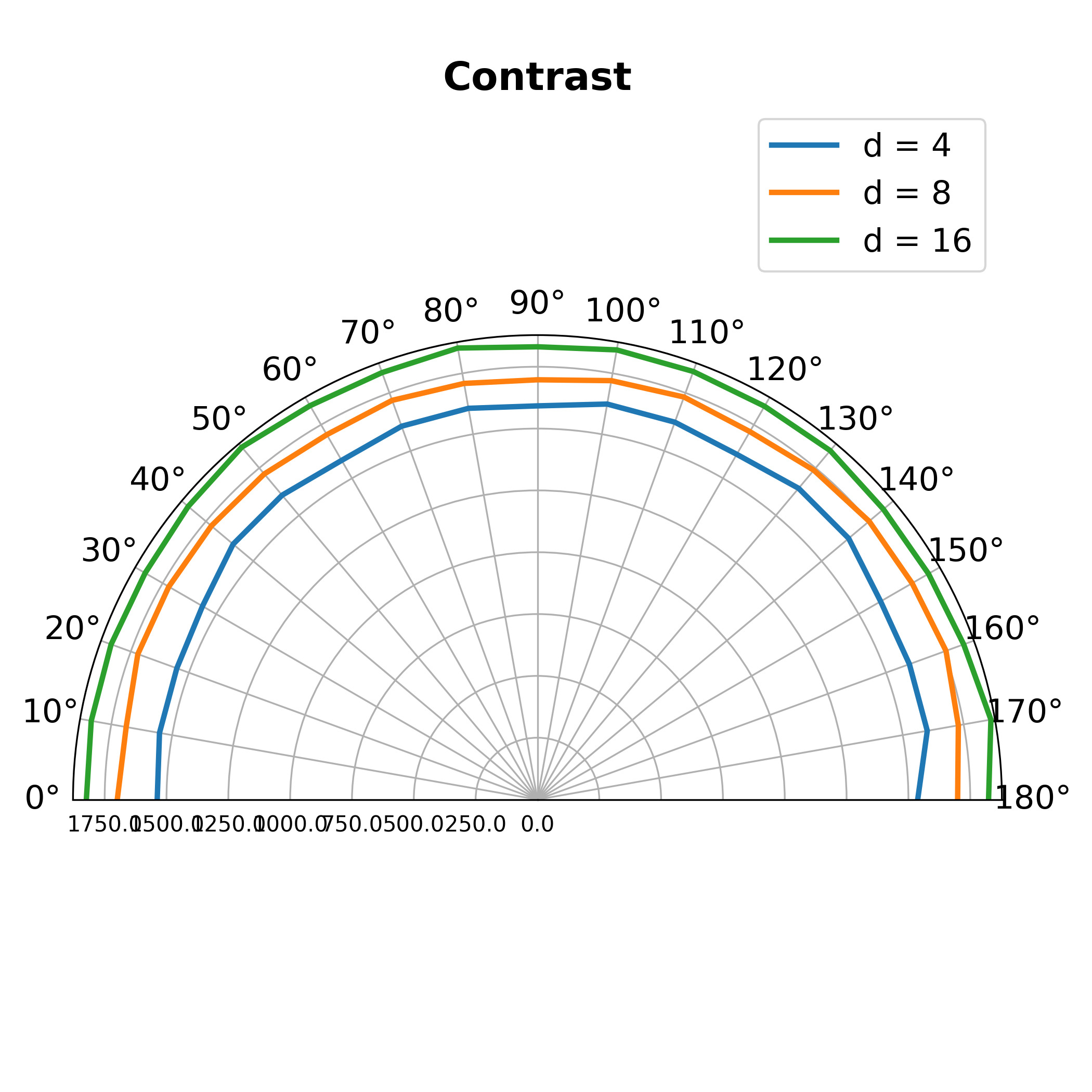} &
        \includegraphics[width=0.45\linewidth]{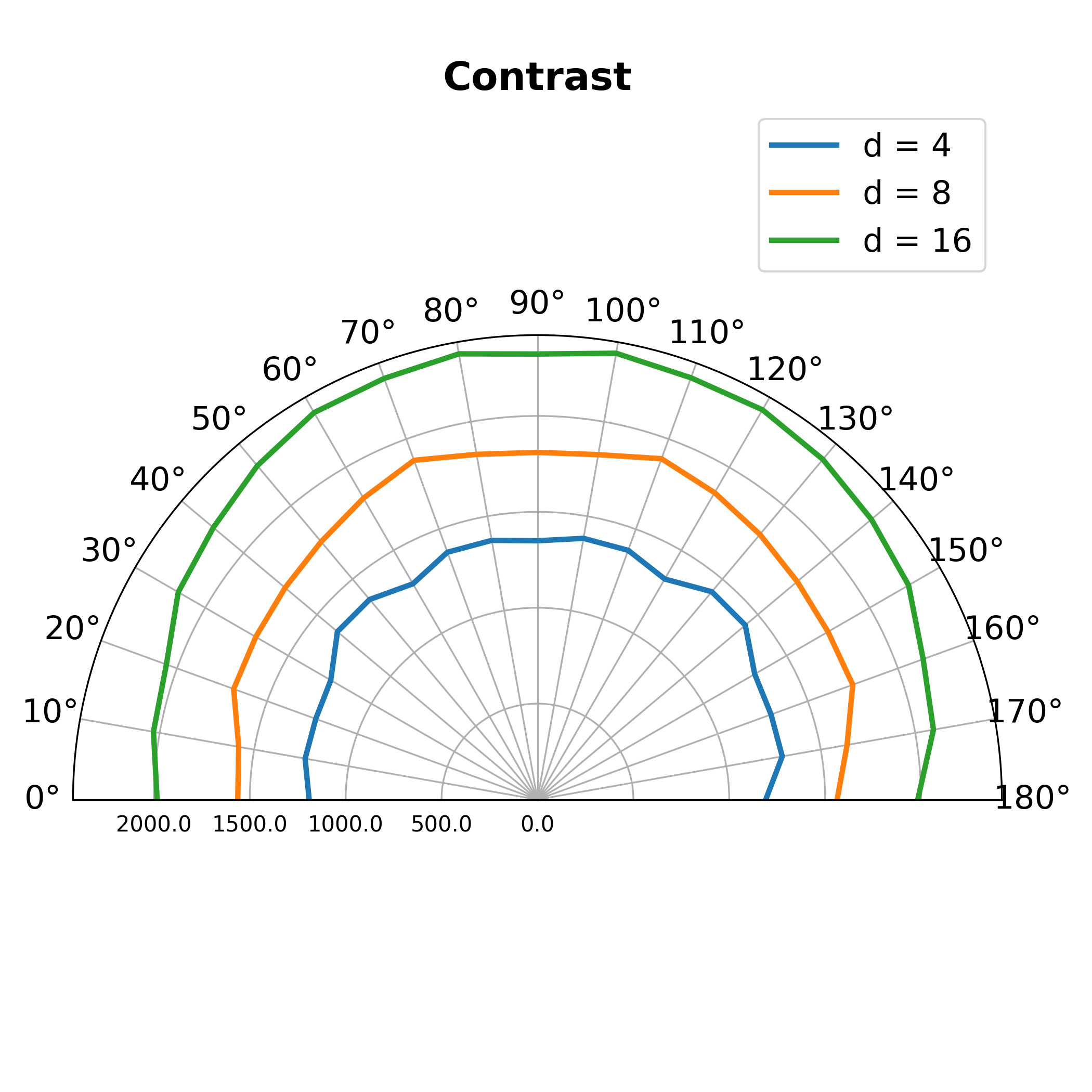} \\
        \includegraphics[width=0.45\linewidth]{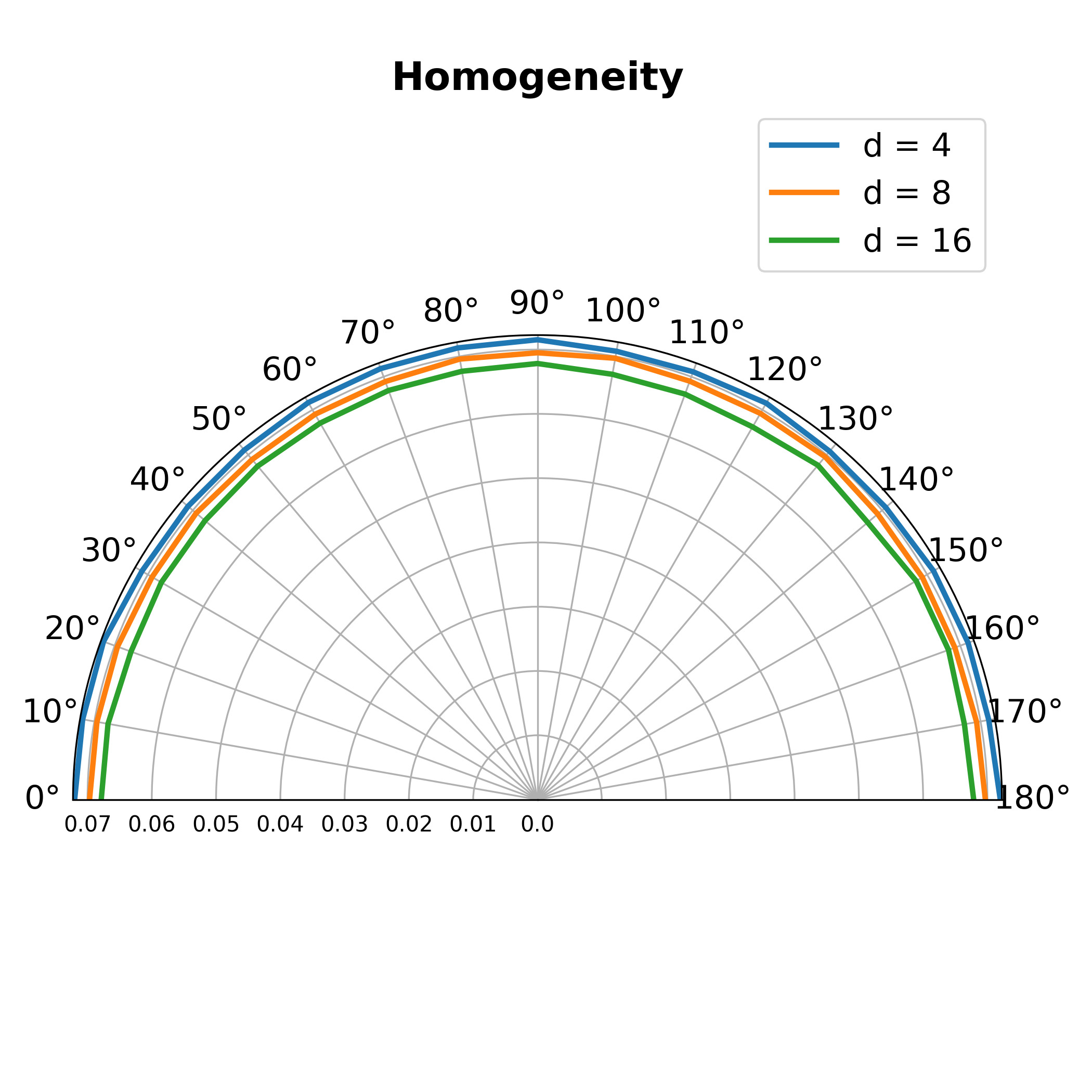} &
        \includegraphics[width=0.45\linewidth]{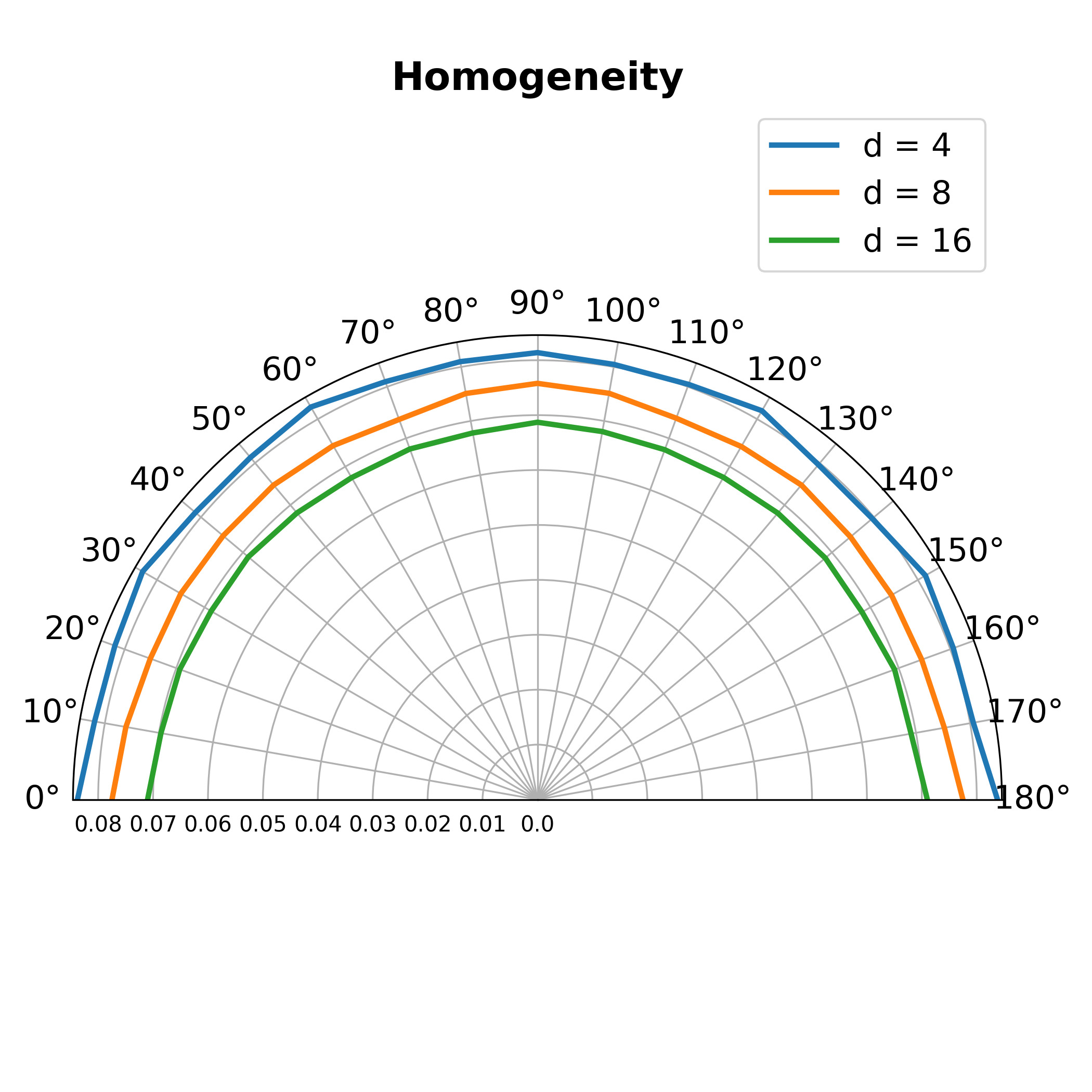} \\
        \includegraphics[width=0.45\linewidth]{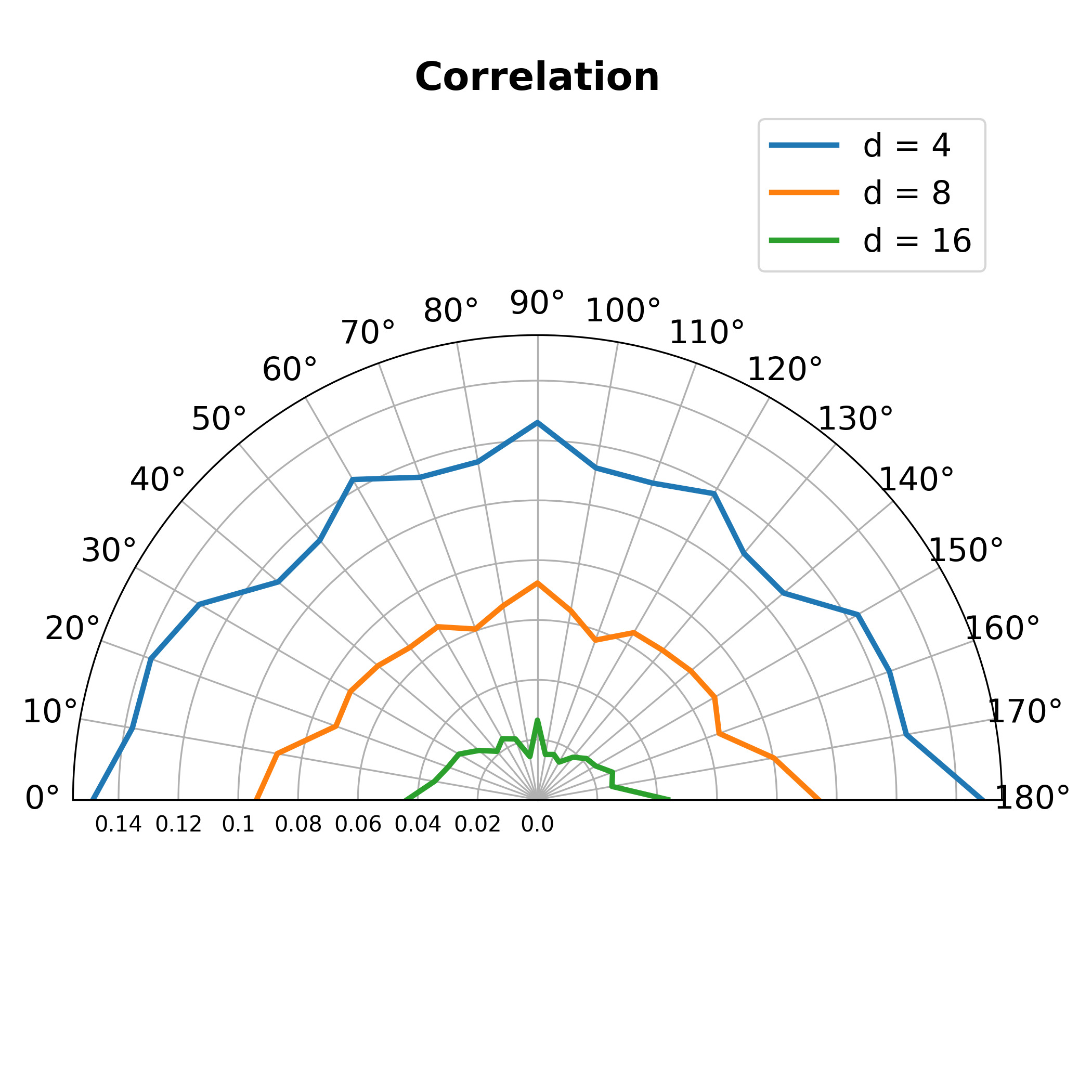} &
        \includegraphics[width=0.45\linewidth]{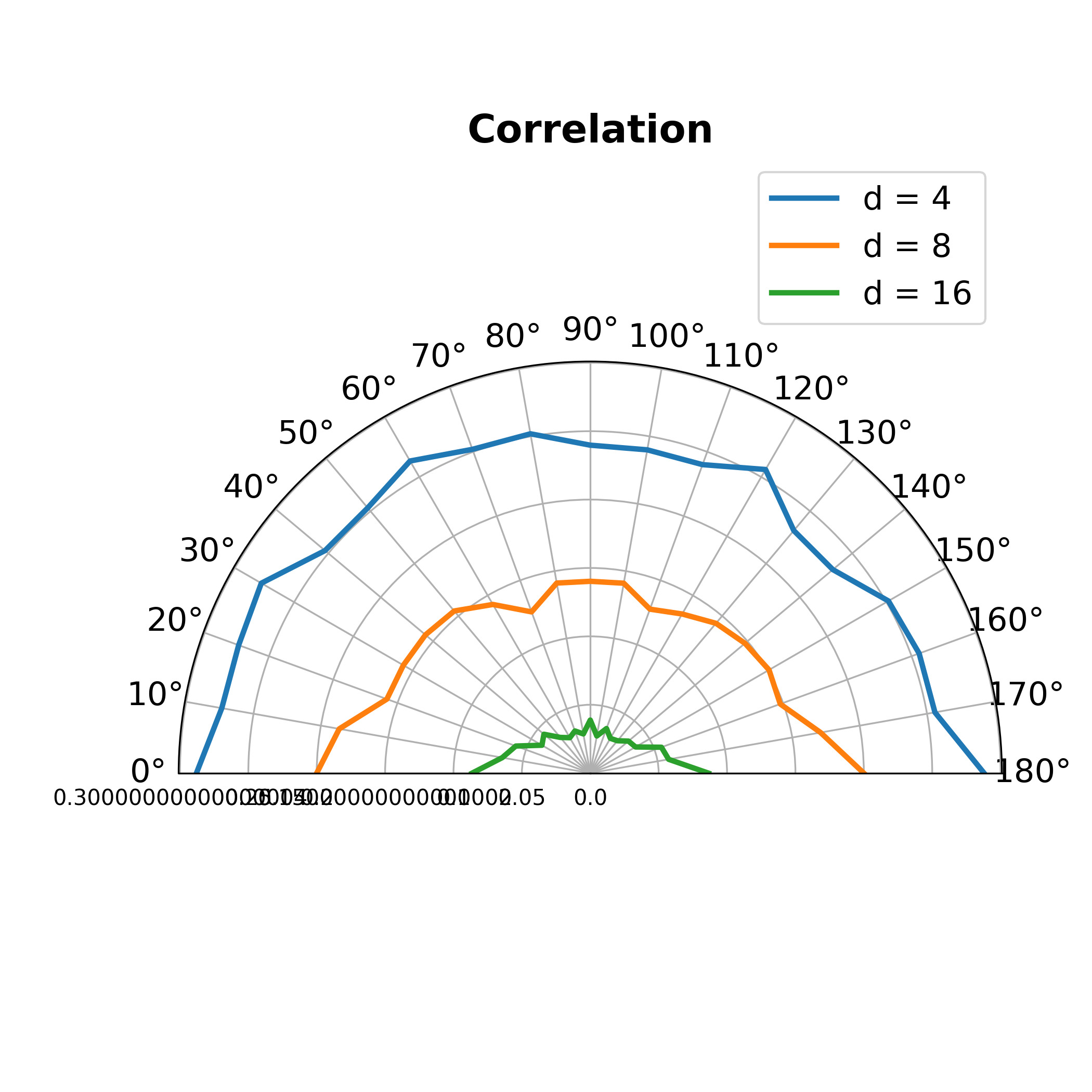} \\
        \includegraphics[width=0.45\linewidth]{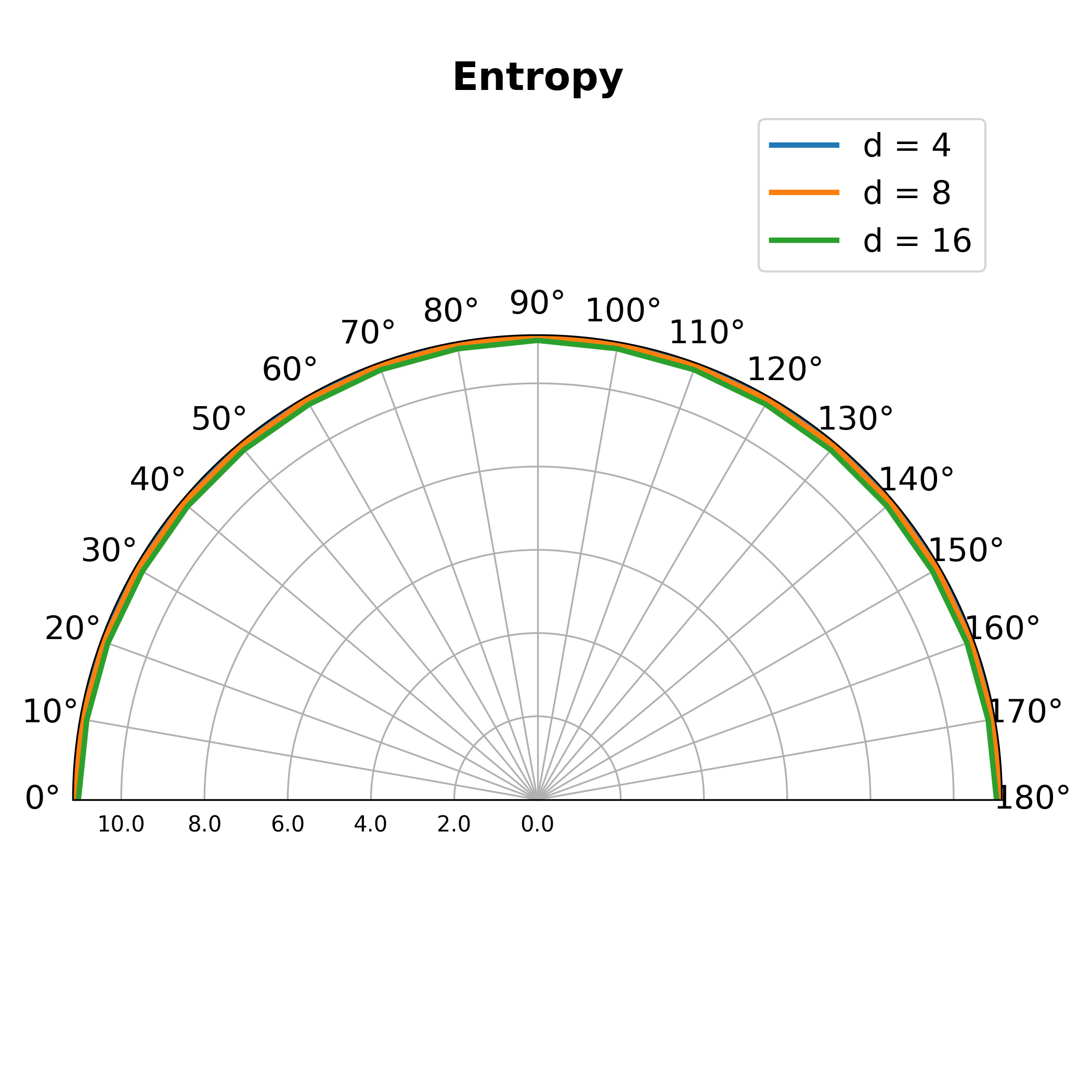} &
        \includegraphics[width=0.45\linewidth]{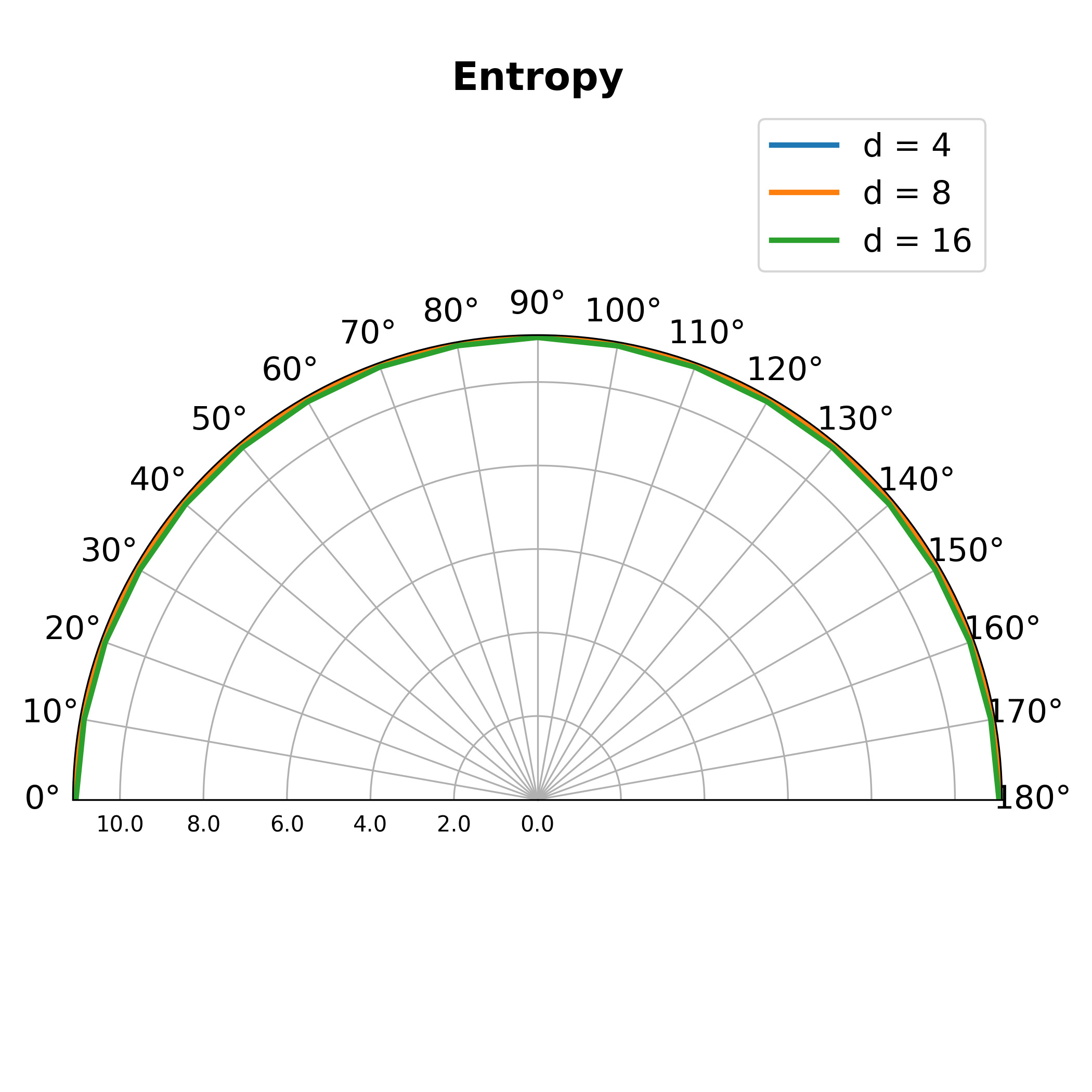} \\
    \end{tabular}
    \caption{GLCM texture metrics (epoch 10) for real vs generated SAR images at different distances and rotation angles. Each row shows one metric: contrast, homogeneity, correlation, and entropy.}
    \label{fig:glcm_epoch10_comparison}
\end{figure}

As detailed in Section~\ref{sec:evaluation-models-images}, GLCM-based texture metrics provide further insight into the model's generative realism. Figure~\ref{fig:glcm_epoch10_comparison} shows that directional patterns, particularly for metrics like entropy, are well reproduced. However, contrast is less accurate across all angles. Indeed, in Figure \ref{fig:histos_whale-heart}, the KL divergence of the \textit{heart-rose-2} model shows that the generated pixel distribution is smoother compared to the real Rayleigh distribution, with fewer black pixels (low-intensity) and more white pixels (high-intensity) in the generated images. In terms of spatial correlation of texture, which corresponds to large-scale spatial dependencies related to SAR geometry, the model \textit{heart-rose-2} performs well, although some variations are observed across different angles.

\section{Applications}

Our fine-tuned Stable Diffusion XL model for SAR imagery offers several practical applications. A significant advantage of this model is its ability to generate novel data beyond the training domain, creating new and unique content not present in the original dataset. 
\begin{figure*}
    \centering
    \scriptsize
    \setlength{\tabcolsep}{1pt}
    \renewcommand{\arraystretch}{1.1}
    \begin{adjustbox}{max width=0.95\textwidth}

    \begin{tabular}{*{5}{>{\centering\arraybackslash}m{0.18\textwidth}}}
        \includegraphics[width=\linewidth]{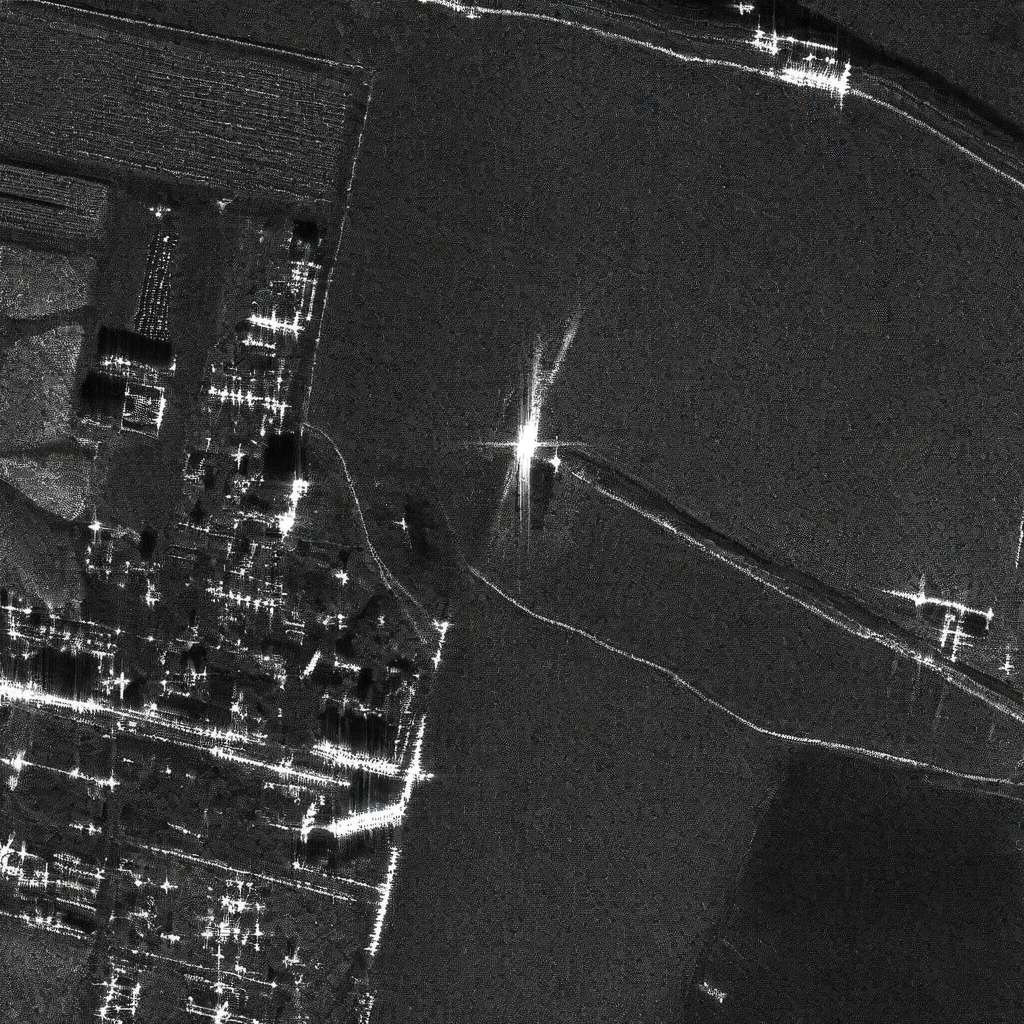}\par(a) &
        \includegraphics[width=\linewidth]{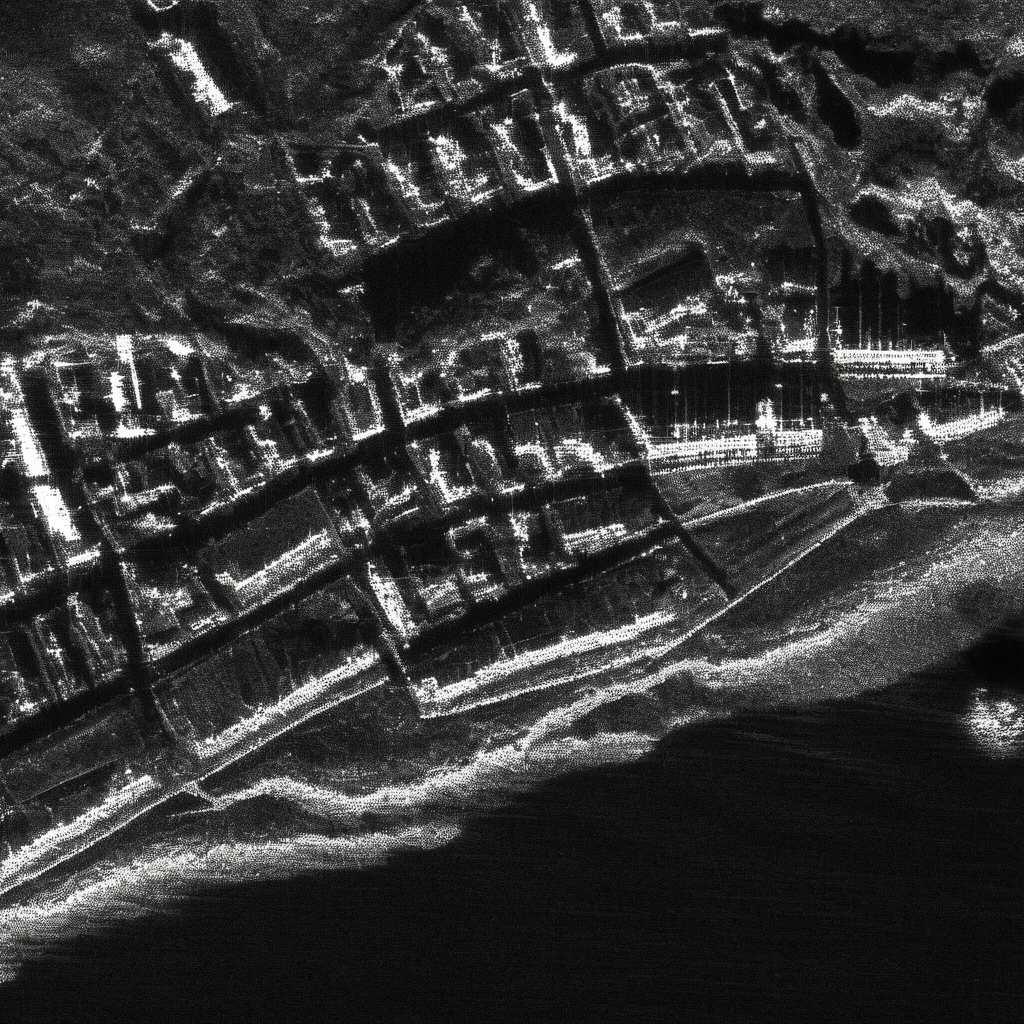}\par(b) &
        \includegraphics[width=\linewidth]{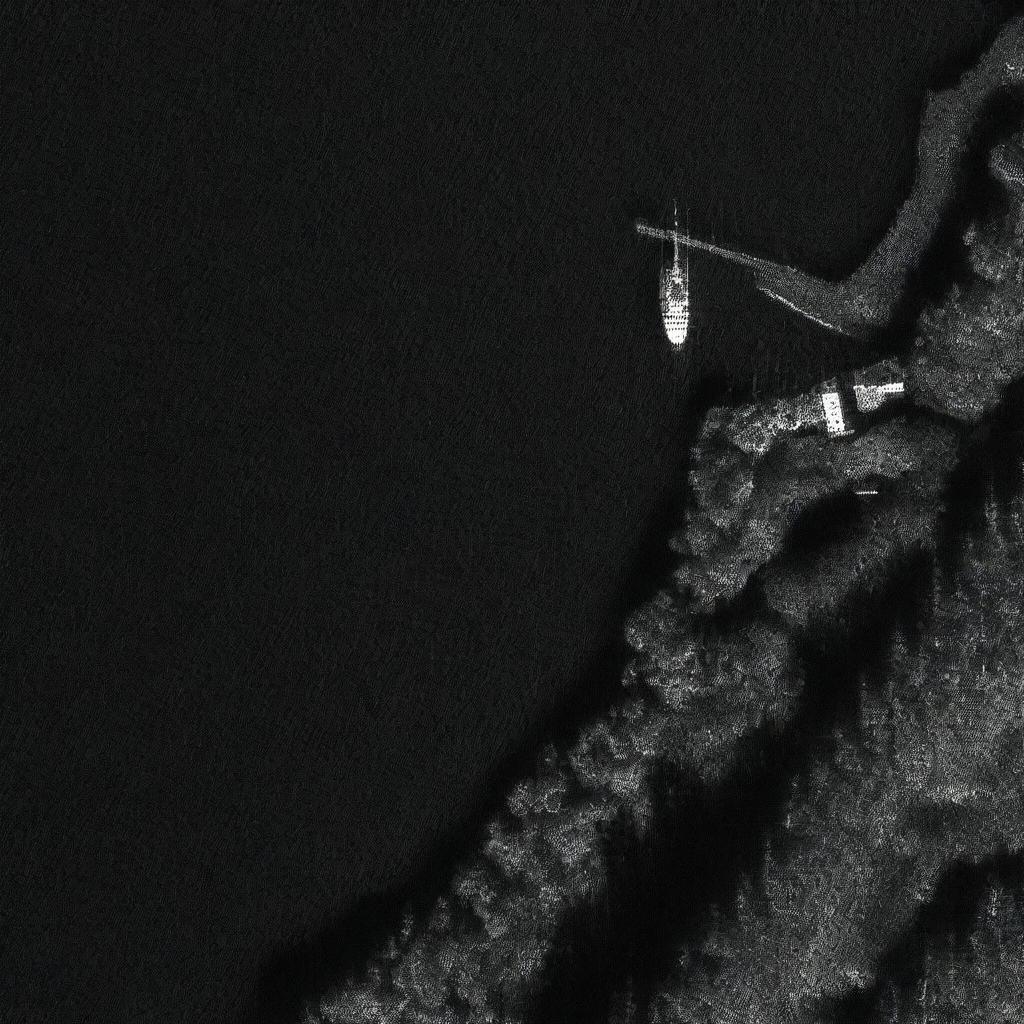}\par(c) &
        \includegraphics[width=\linewidth]{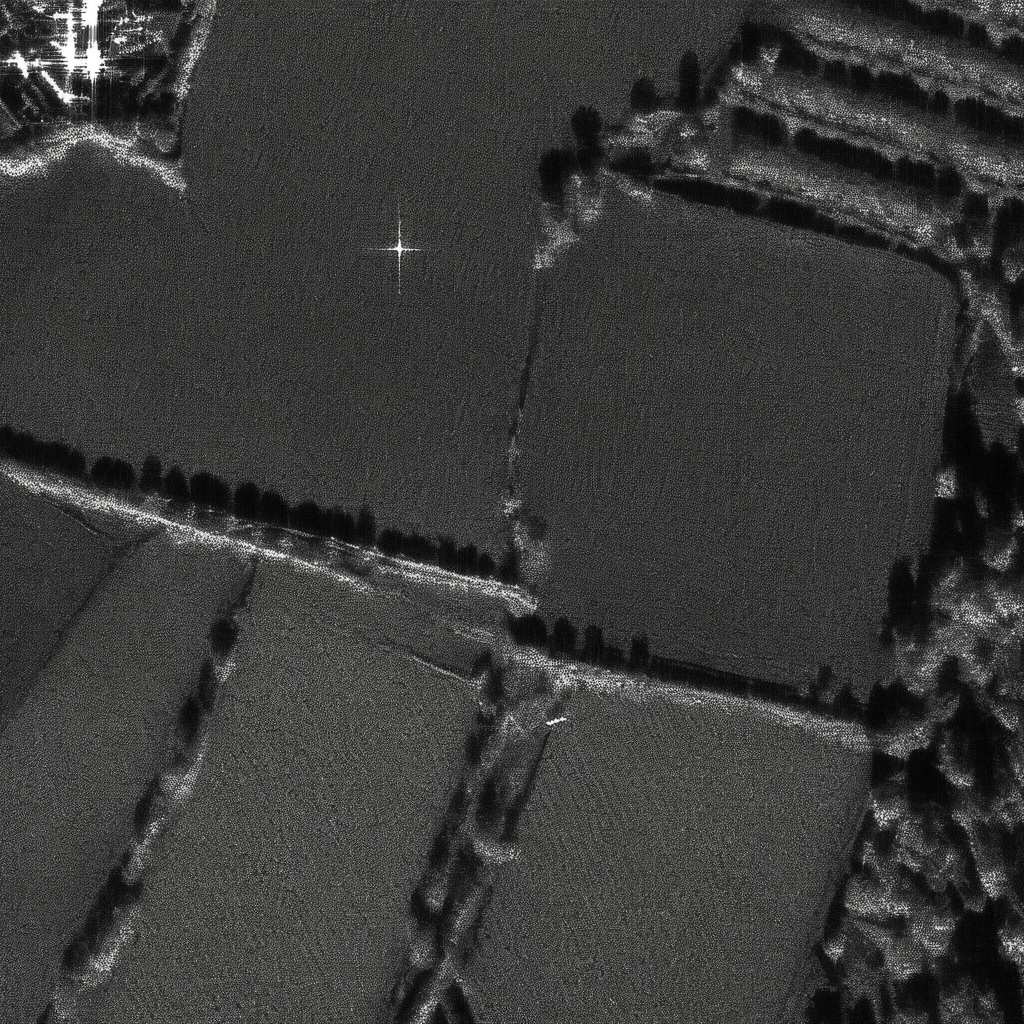}\par(d) &
        \includegraphics[width=\linewidth]{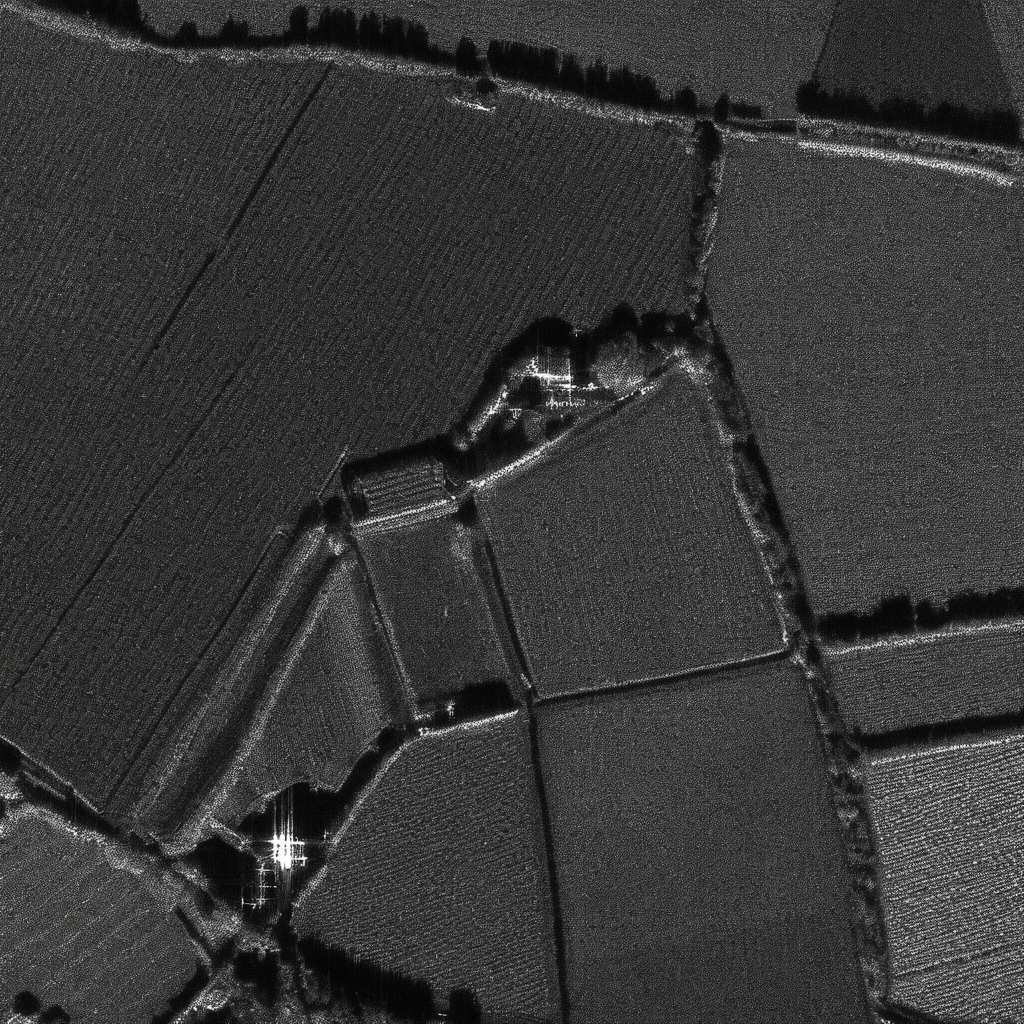}\par(e) \\

        \includegraphics[width=\linewidth]{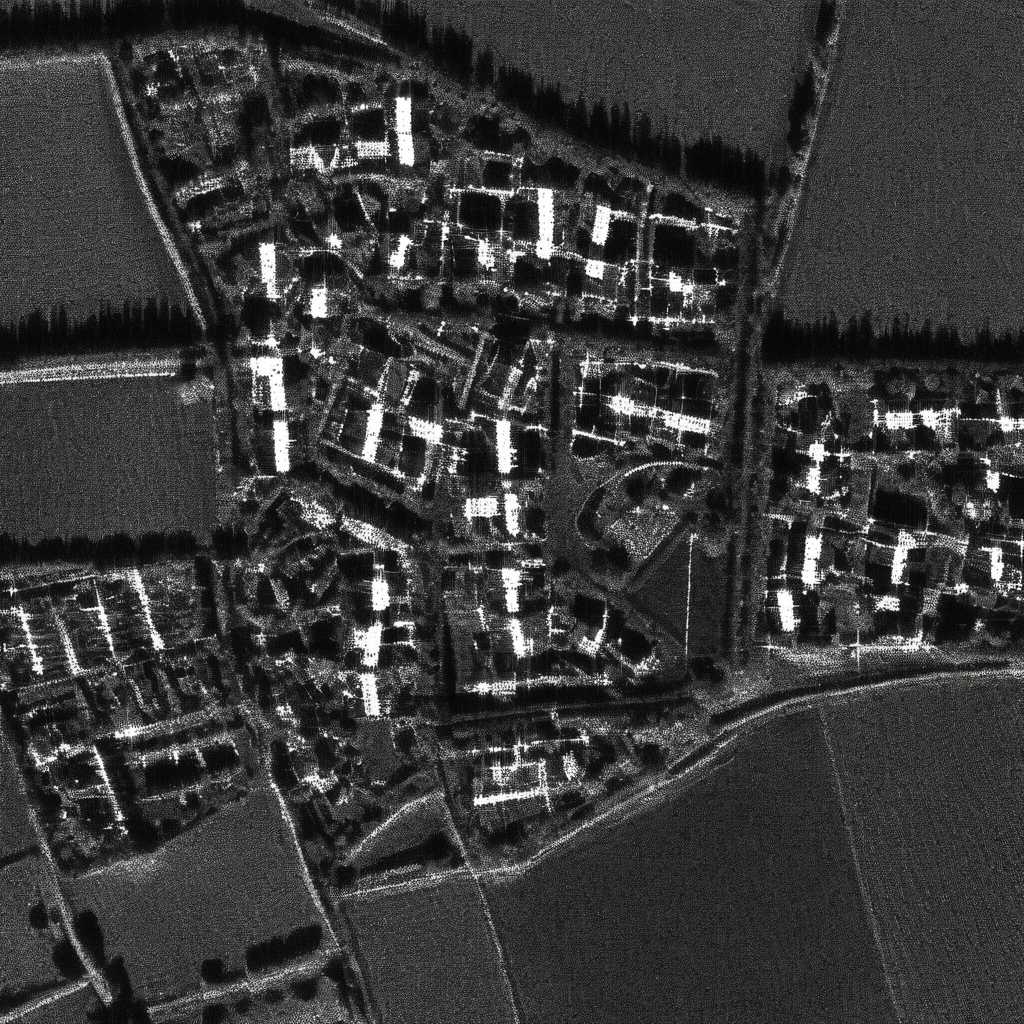}\par(f) &
        \includegraphics[width=\linewidth]{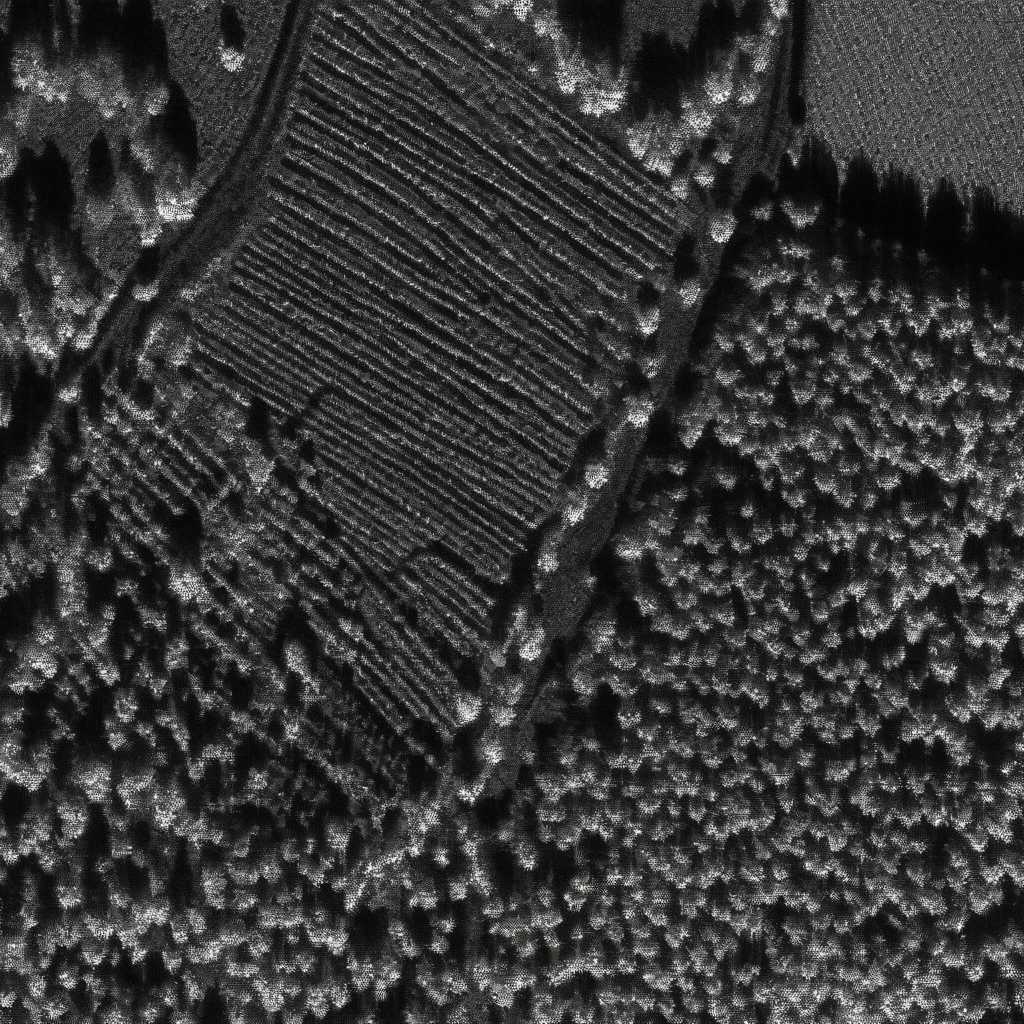}\par(g) &
        \includegraphics[width=\linewidth]{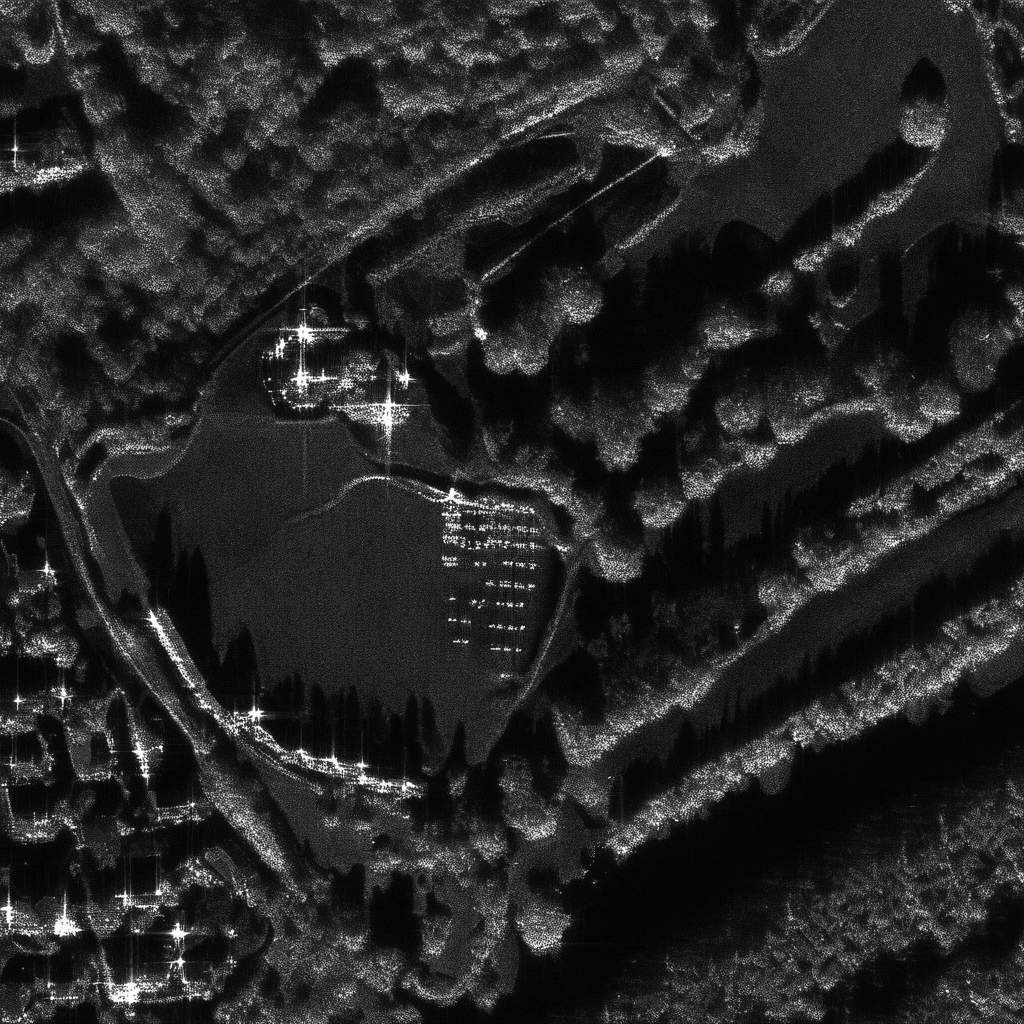}\par(h) &
        \includegraphics[width=\linewidth]{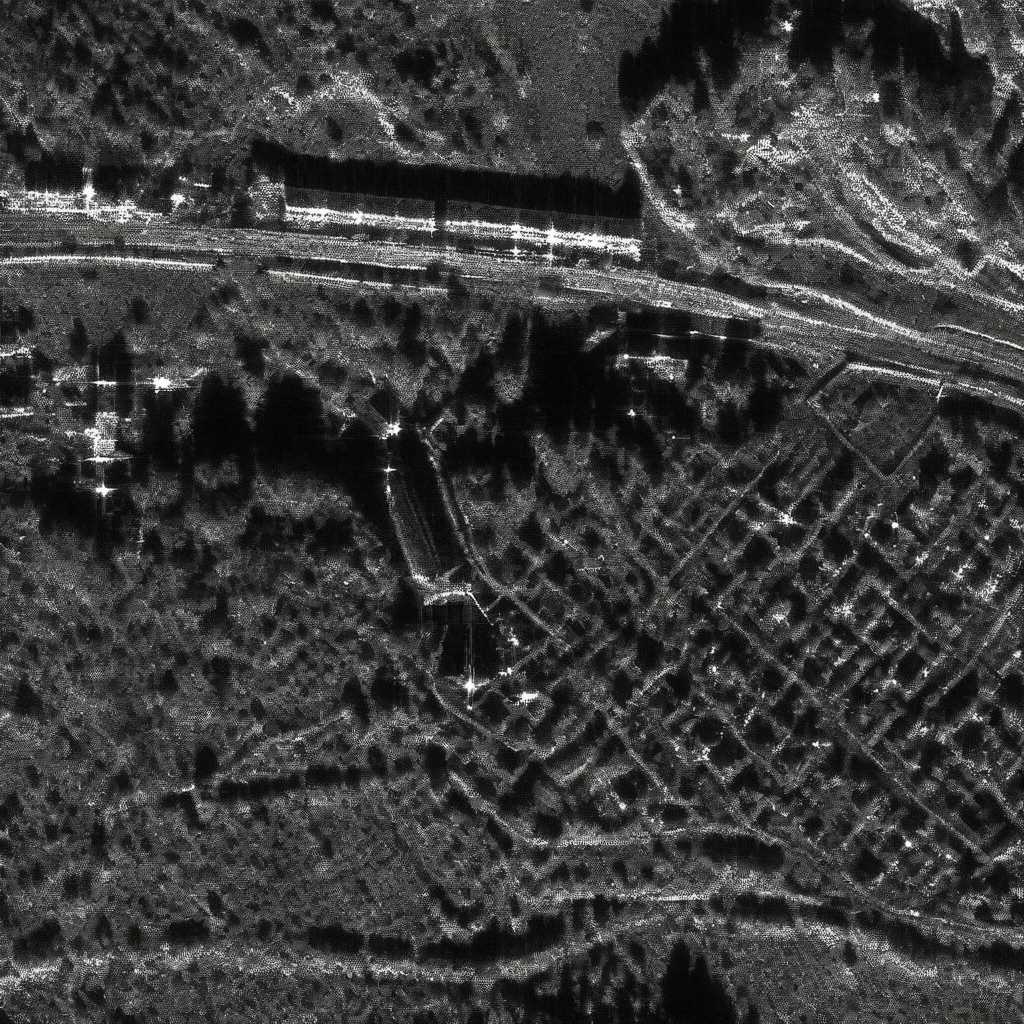}\par(i) &
        \includegraphics[width=\linewidth]{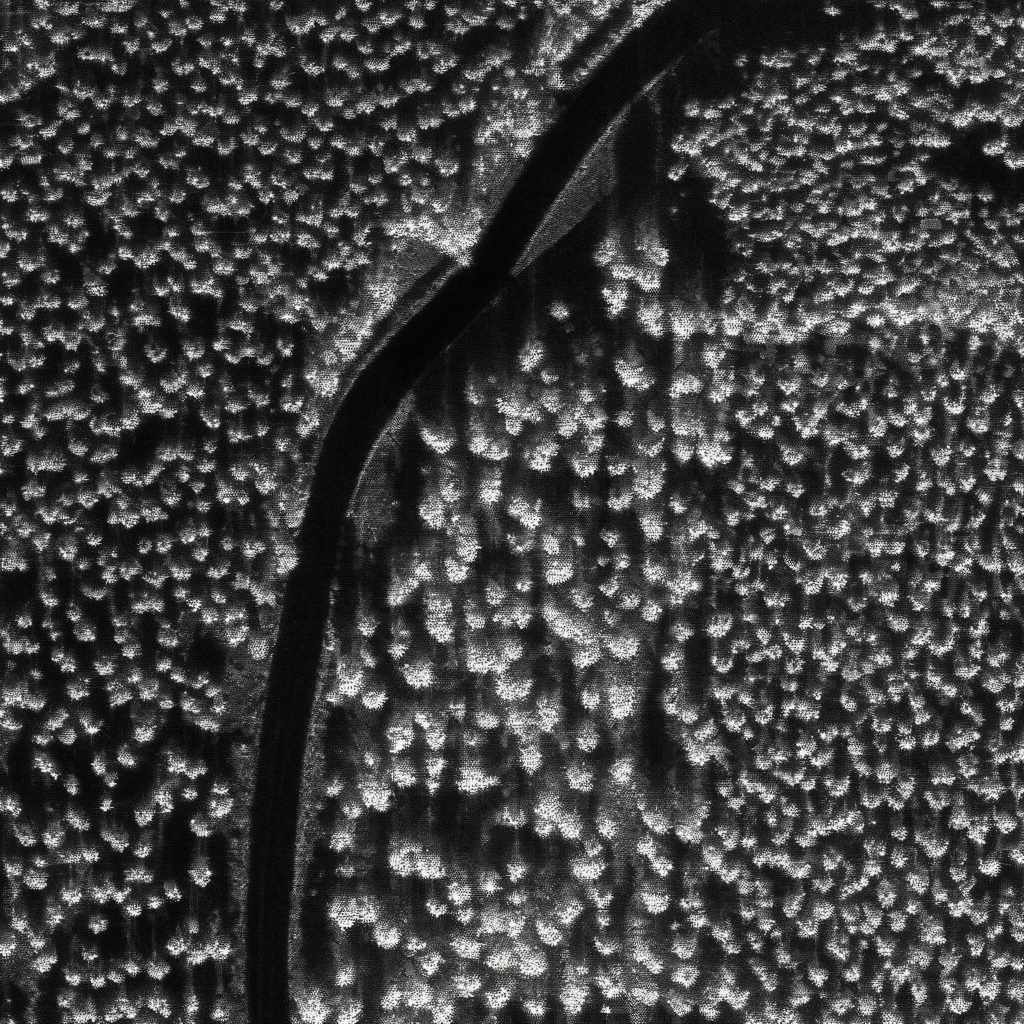}\par(j) \\

        \includegraphics[width=\linewidth]{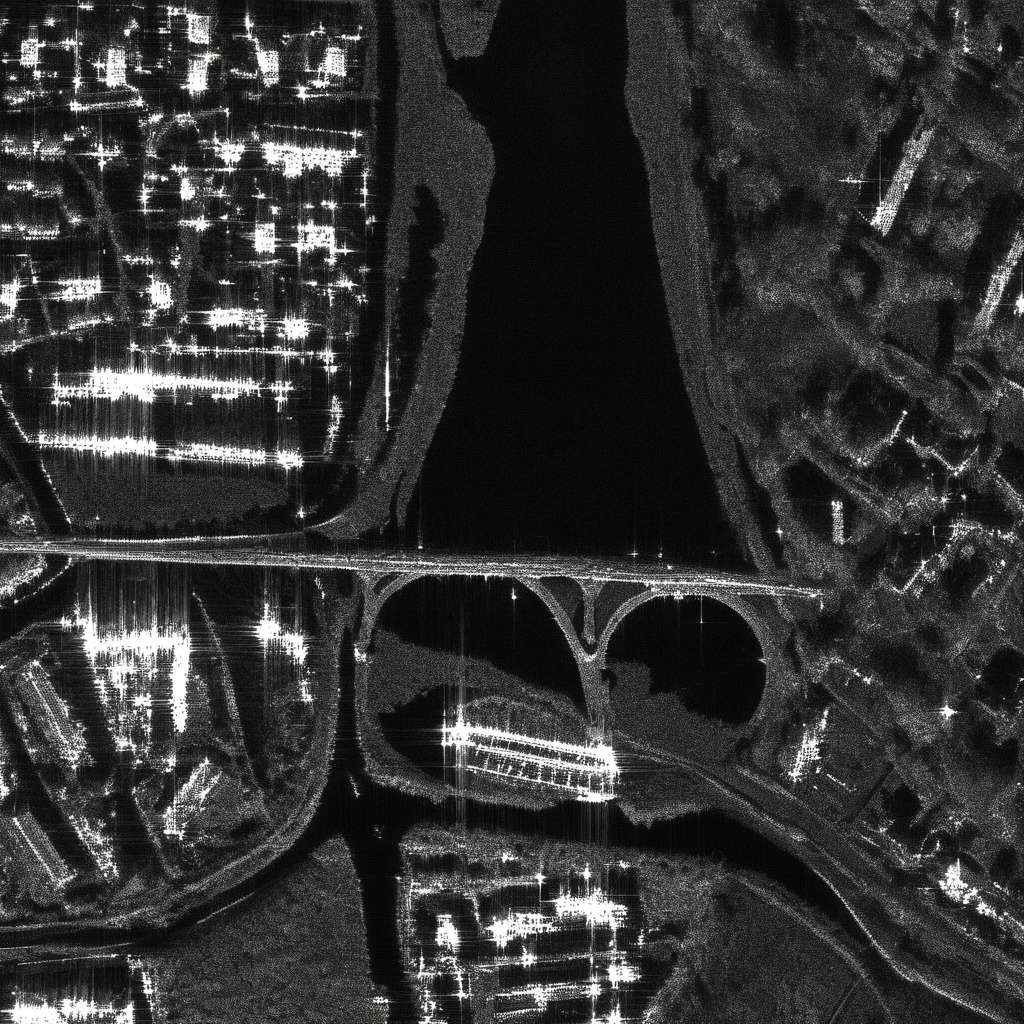}\par(k) &
        \includegraphics[width=\linewidth]{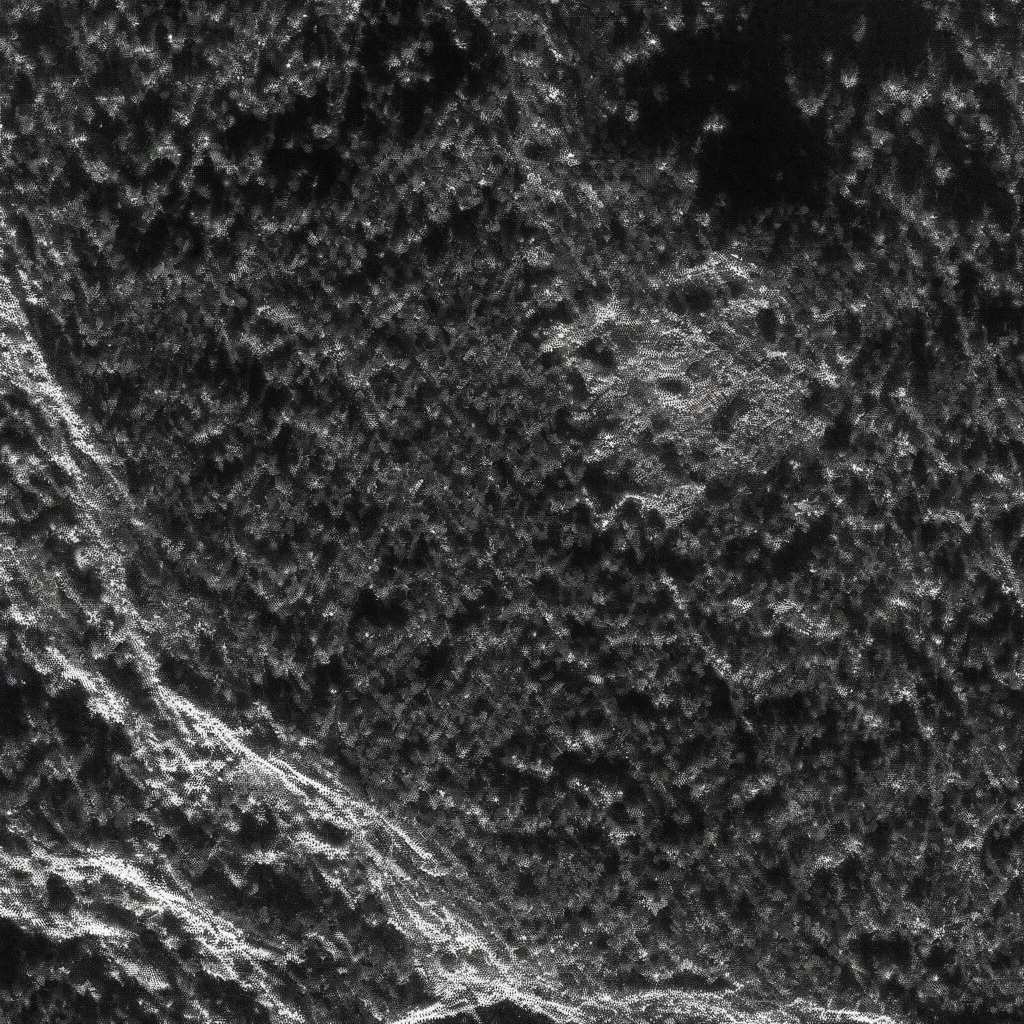}\par(l) &
        \includegraphics[width=\linewidth]{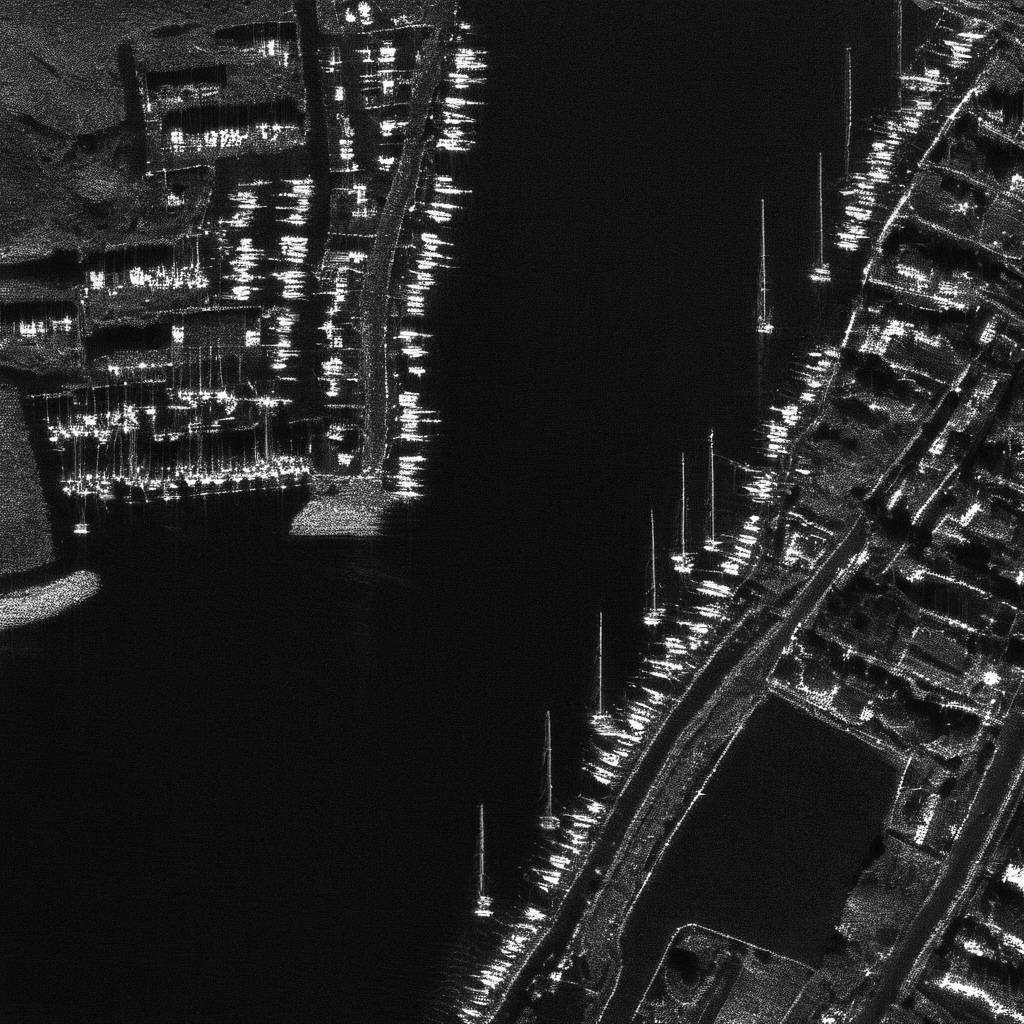}\par(m) &
        \includegraphics[width=\linewidth]{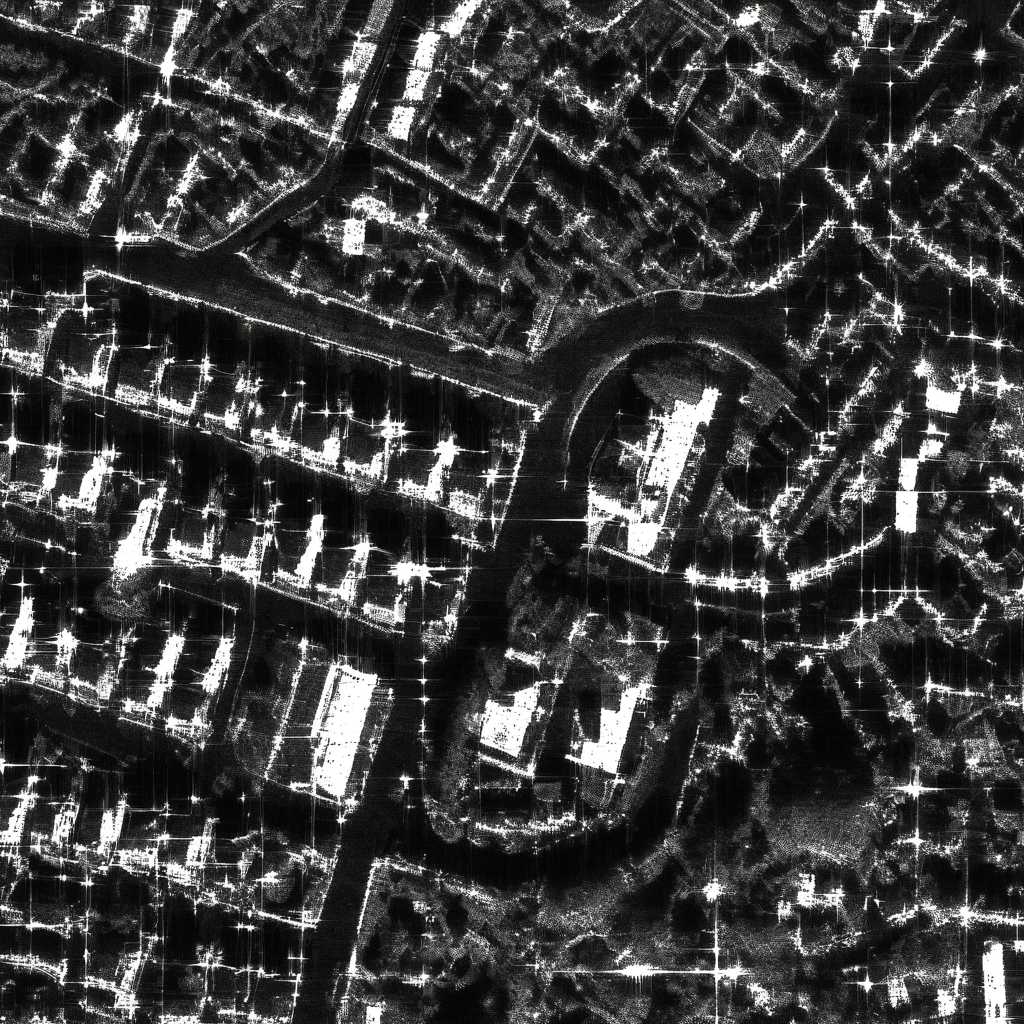}\par(n) &
        \includegraphics[width=\linewidth]{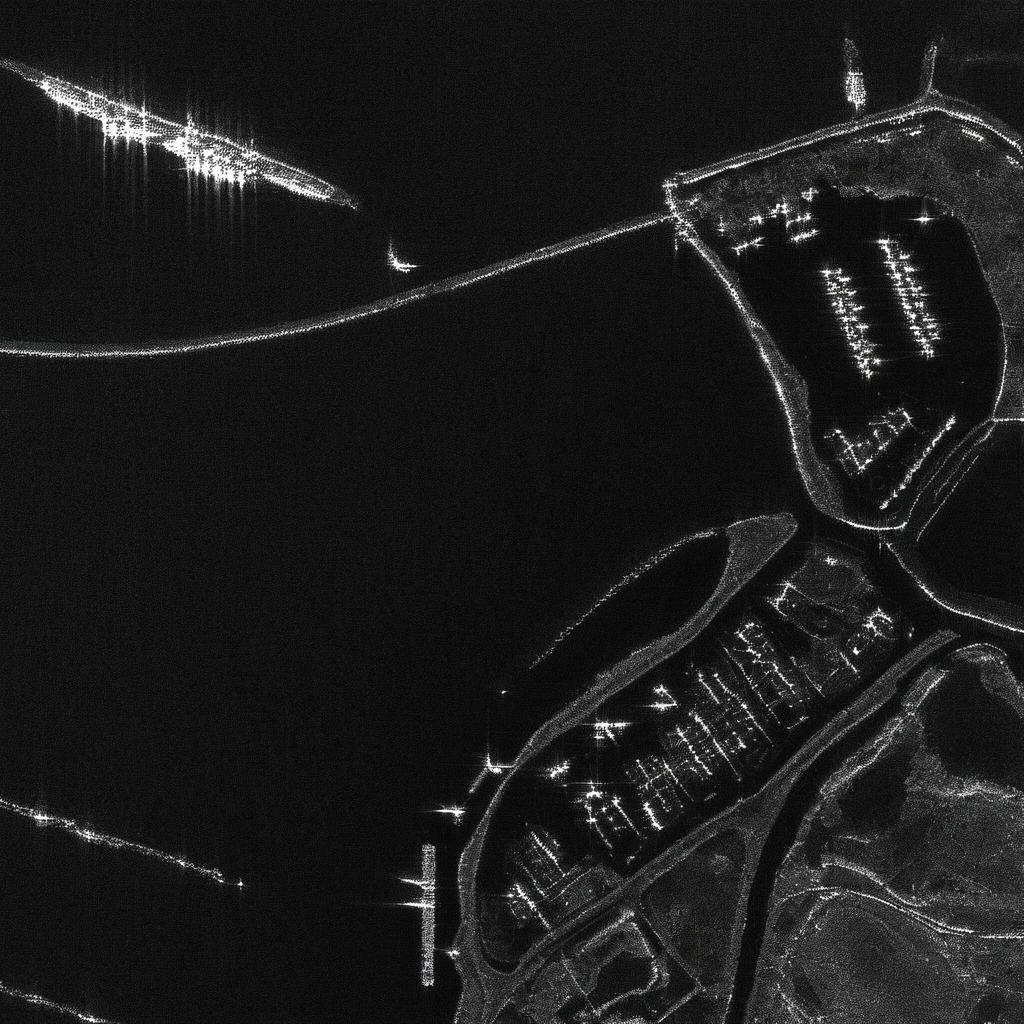}\par(o) \\
    \end{tabular}

    \end{adjustbox}
    
    \caption{
    Generated images (1024x1024px at 40cm) from the \textbf{heart-rose-2} model. Each image corresponds to a distinct textual prompt (see Appendix~\ref{appendix:heart-rose-2-images}).
    }
    \label{fig:heart_rose_grid}
\end{figure*}

As illustrated in Figure~\ref{fig:heart_rose_grid}, we observe that we are able to generate various types of landscapes, such as fields, forests, seacoasts, or mountains, that the model has never seen during its training. This includes scenarios like a boat near a forested coastline (c) or a bridge with a particular design in the middle of a landscape (k). These results demonstrate the model's capacity to produce creative and realistic scenarios from its language understanding.

This is possible because the model, based on a pretrained architecture, is capable of producing variations of objects or rare situations that are not overfitted to the limited examples seen during training. Consequently, the model can generate realistic and representative variants of these uncommon cases or objects, making it especially effective for tasks requiring creative generation of new SAR imagery.

For example, it can be used to perform image-to-image generation while preserving essential statistical properties of SAR data, such as speckle texture and reflectivity distributions. This makes the model particularly valuable for applications like adding spatial detail, or refining outputs from ONERA’s physics-based simulators.


More details can be found in our previous work (\cite{debuysere2024synthesizing}, \cite{debuysère2025spacebornairbornsarimage} and \cite{trouve2024synthesis}) where we demonstrated the effectiveness of our fine-tuned model in a conditional, multi-resolution ControlNet pipeline for large-scale scene generation. We also showed its ability to transform satellite TerraSAR-X data into high-resolution, airborne-like imagery with reduced sensor noise. These applications rely on ControlNet to guide generation using structural priors.

In this section, we illustrate two use cases: TerraSAR-X conditioned 40 cm synthesis and simulator-conditioned 80 cm synthesis.

\subsection{TerraSAR-X–Conditioned Synthesis at 40 cm}

We use our model to enhance TerraSAR-X satellite acquisitions. The original images (acquired at ~1.35~m resolution) are up-sampled to 40~cm in the frequency domain and passed through our generation pipeline.

\begin{figure}[htb]
    \centering

    \begin{minipage}[t]{0.48\textwidth}
        
        \centering
        \begin{adjustbox}{max width=\textwidth}
        \begin{tabular}{cc}
            \includegraphics[width=0.5\textwidth]{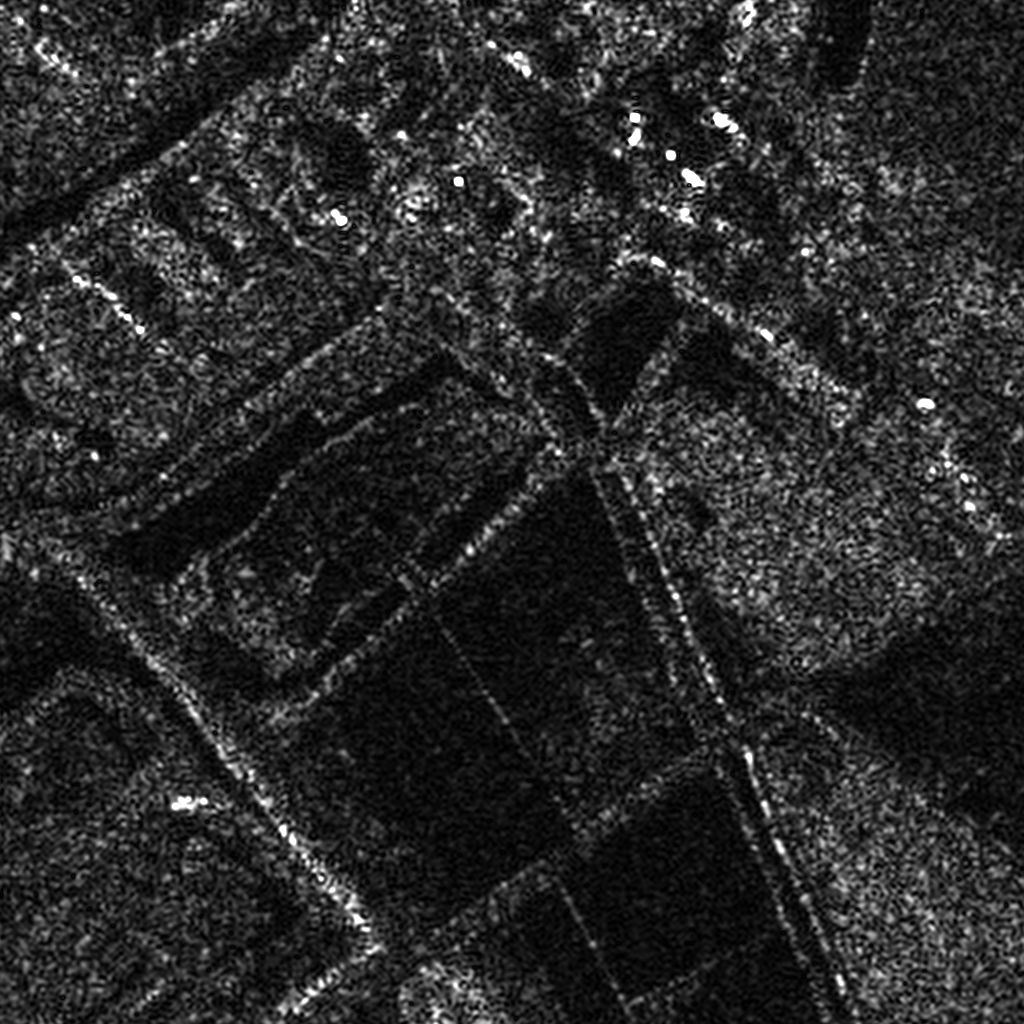} &
            \includegraphics[width=0.5\textwidth]{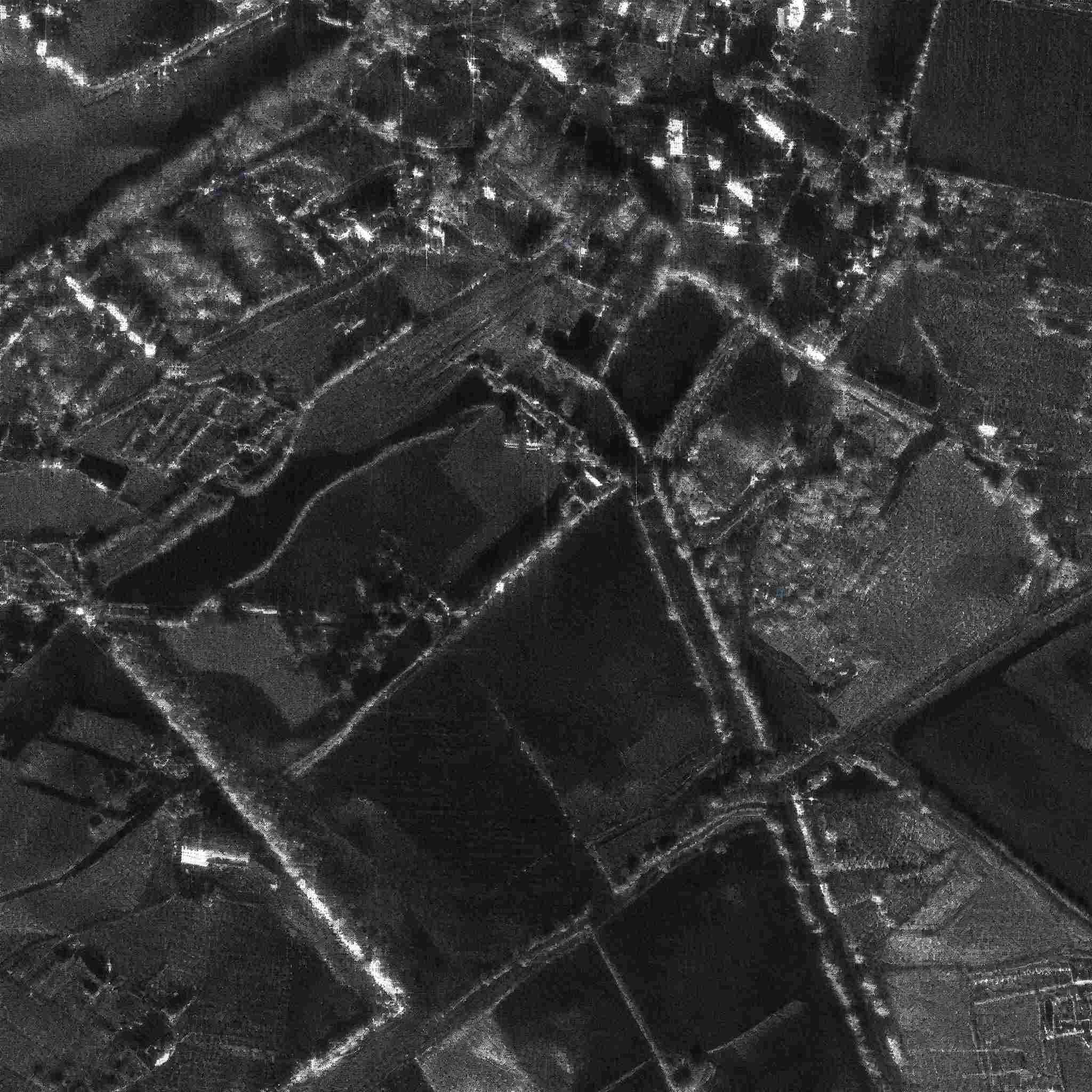} 
            \\
            \includegraphics[width=0.5\textwidth]{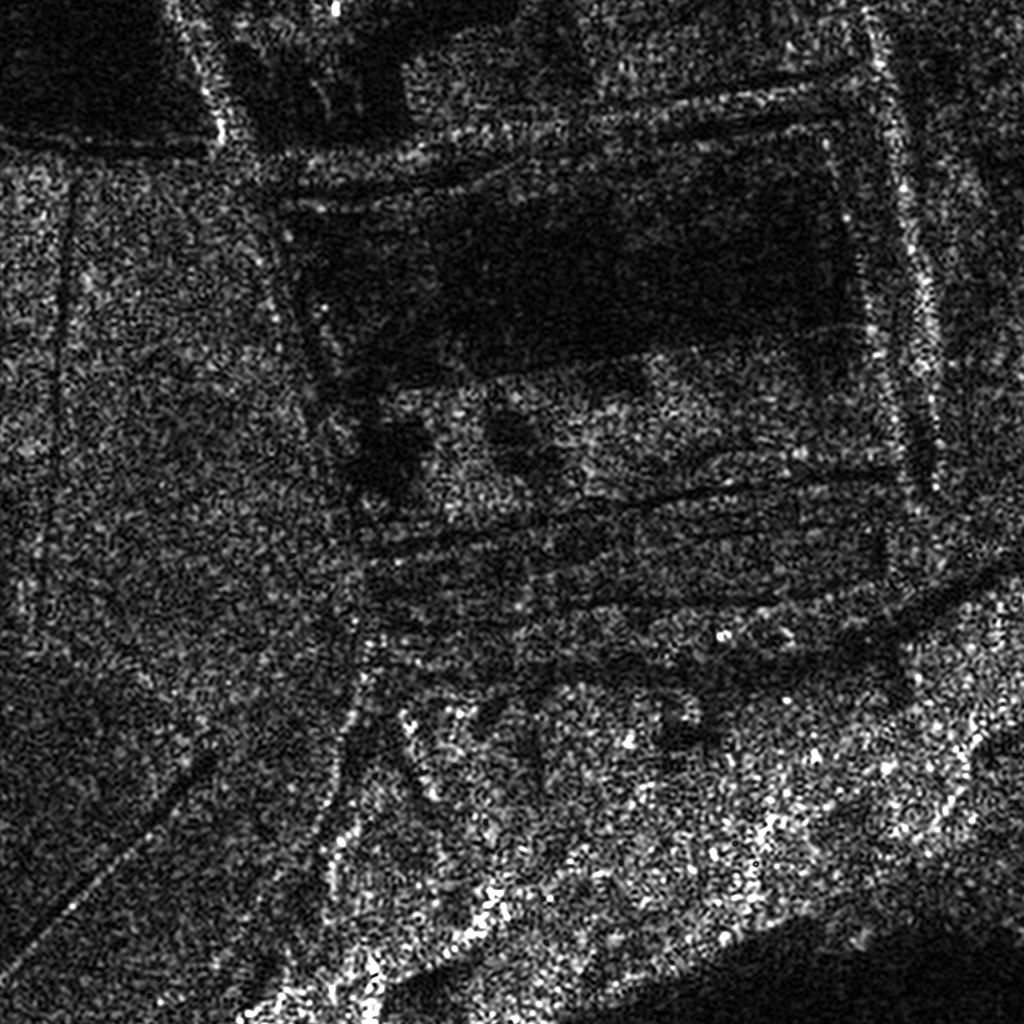} &
            \includegraphics[width=0.5\textwidth]{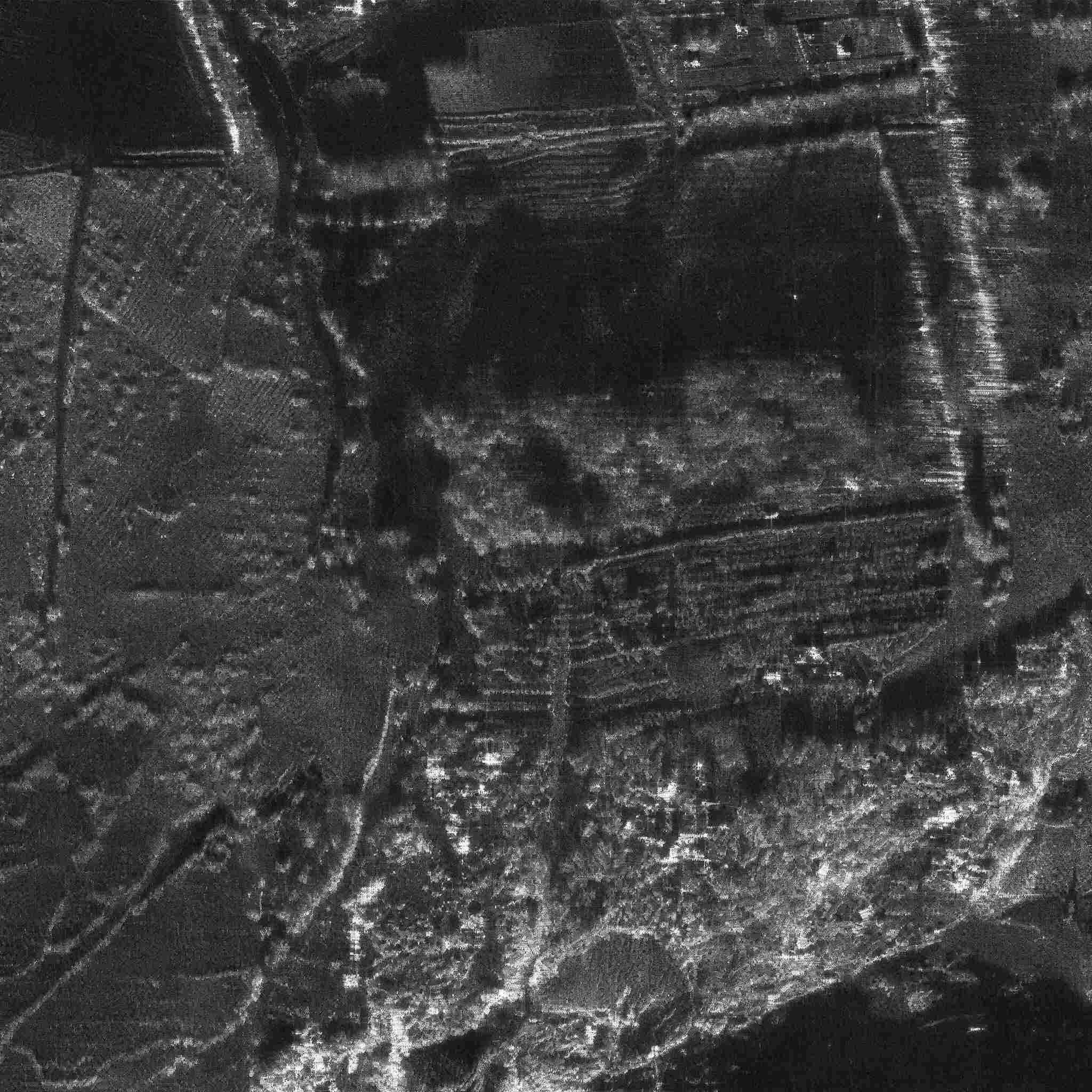} 
            \\
            \includegraphics[width=0.5\textwidth]{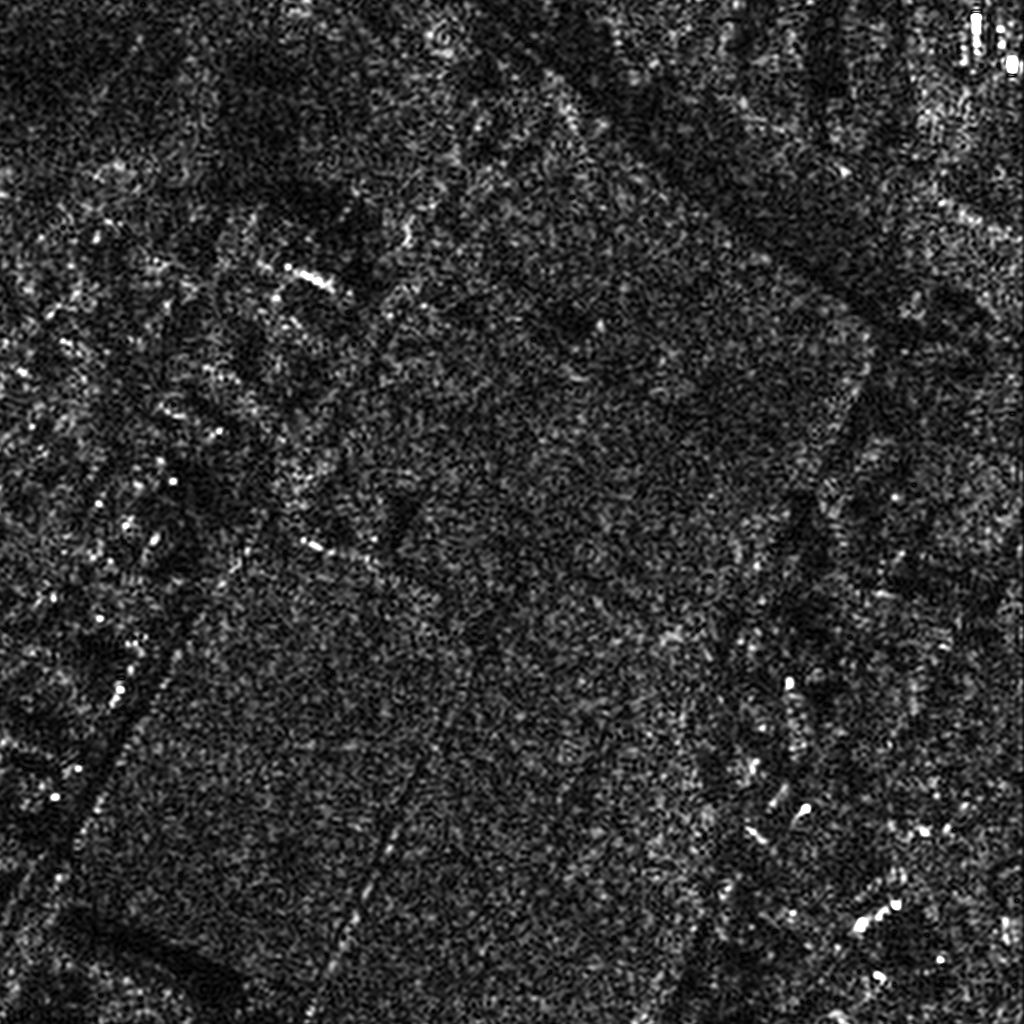} &
            \includegraphics[width=0.5\textwidth]{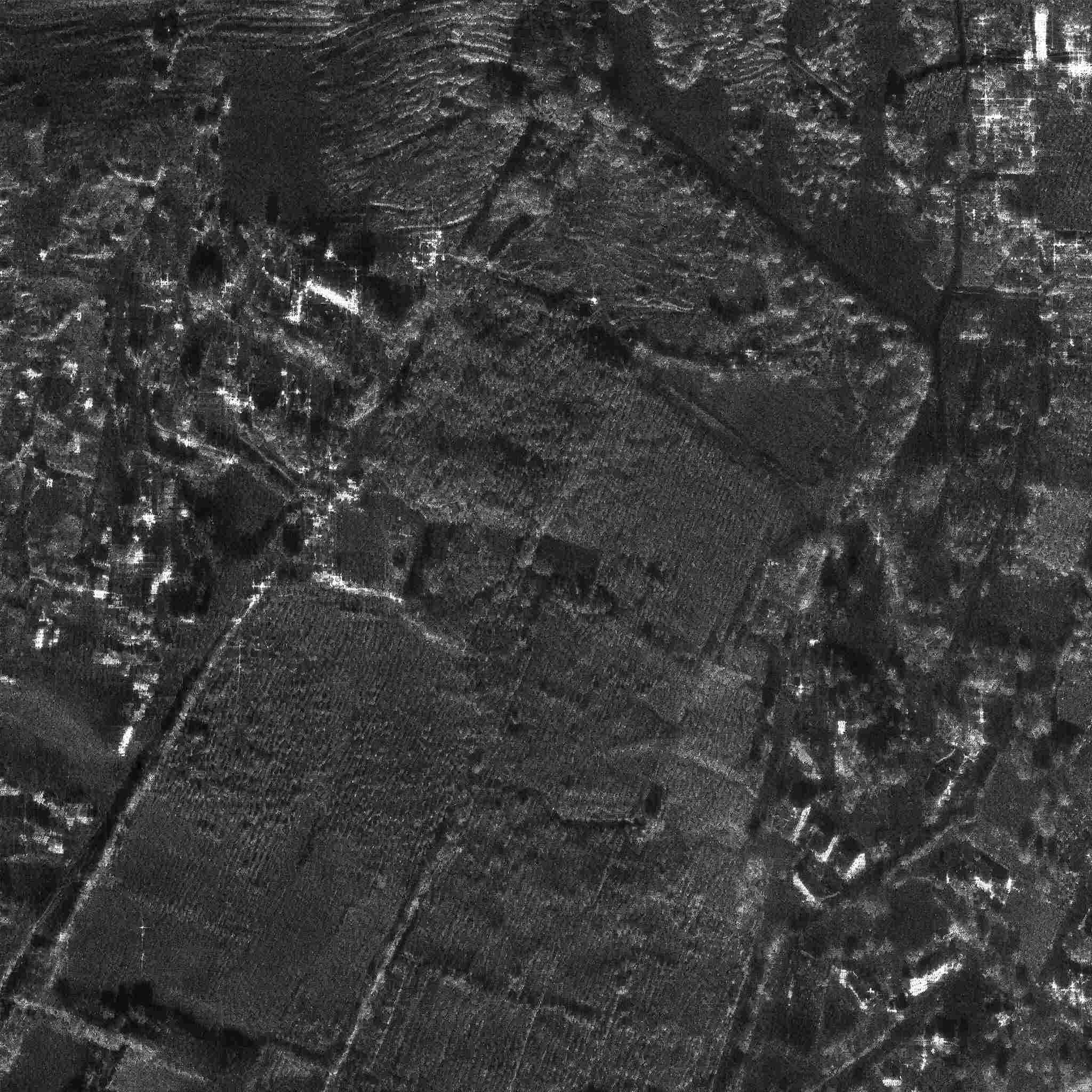} \\
            \textbf{TerraSAR-X image (1024x1024)} & \textbf{Generated image (2048x2048)} \\
        \end{tabular}
        \end{adjustbox}
        \captionof{figure}{Enhancing TerraSAR-X images with prompts (1): "A SAR image of various patches of crop fields with roads and a cluster of houses.", (2): "A SAR image a large rectangular field near a beach." and (3): "A SAR various patches of fields with vegetation and a few houses."}
        \label{fig:image-terrasarx}
    \end{minipage}
    \hfill
\end{figure}

As shown in Figure~\ref{fig:image-terrasarx}, the enhanced images exhibit sharper textures, improved contrast, and more homogeneous noise characteristics compared to the original TerraSAR-X inputs, while preserving realistic backscattering structures. In particular, the TerraSAR-X reference image at a resolution of around 1 m contains characteristic speckle patterns on which the model can diffuse to add realistic high-frequency detail.

\subsection{Simulated SAR Image-Conditioned Synthesis}

We also apply our model to ONERA’s EMPRISE outputs (80 cm). The pipeline produces 40 cm imagery with finer textures and more realistic spatial structure.

\begin{figure}[htb]
    \centering

    \begin{minipage}[t]{0.47\textwidth}
        
        \centering
        \begin{adjustbox}{max width=\textwidth}
        \begin{tabular}{cc}
            \includegraphics[width=0.5\textwidth]{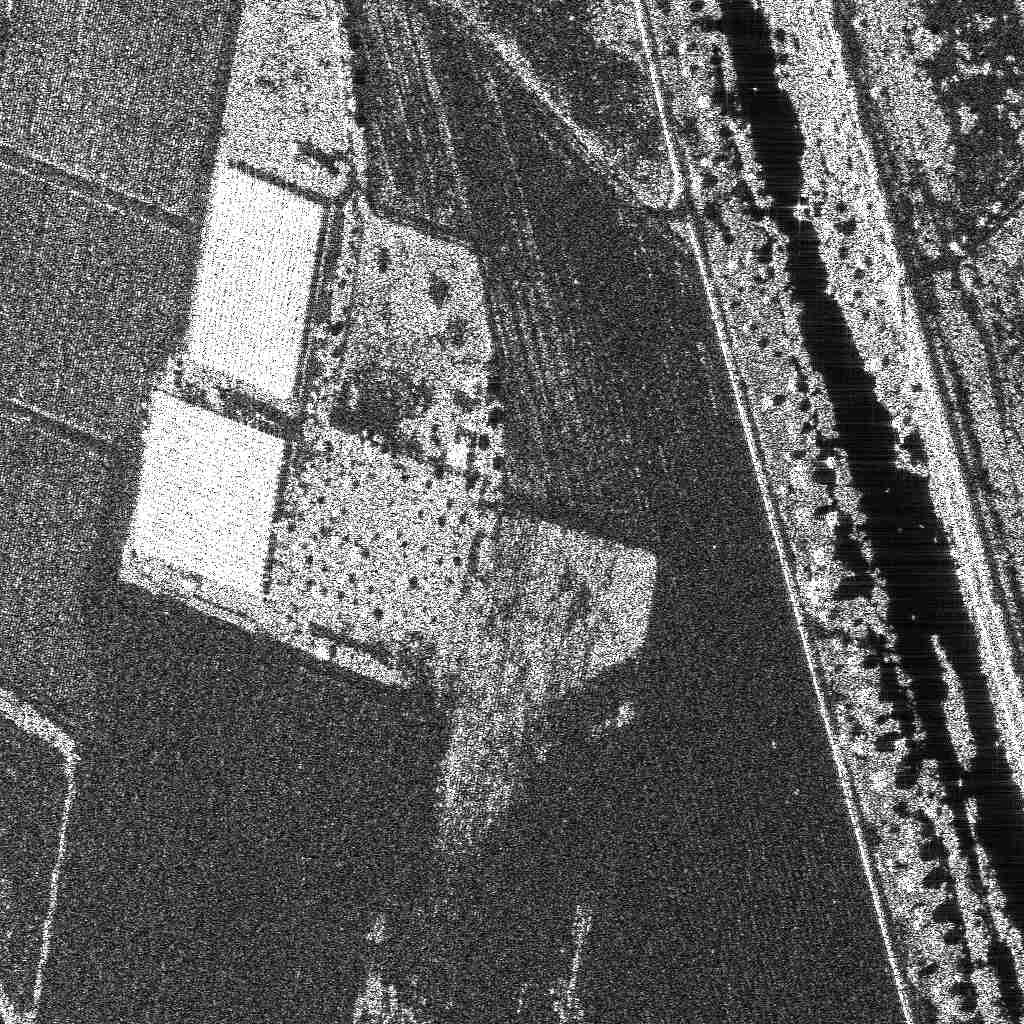} &
            \includegraphics[width=0.5\textwidth]{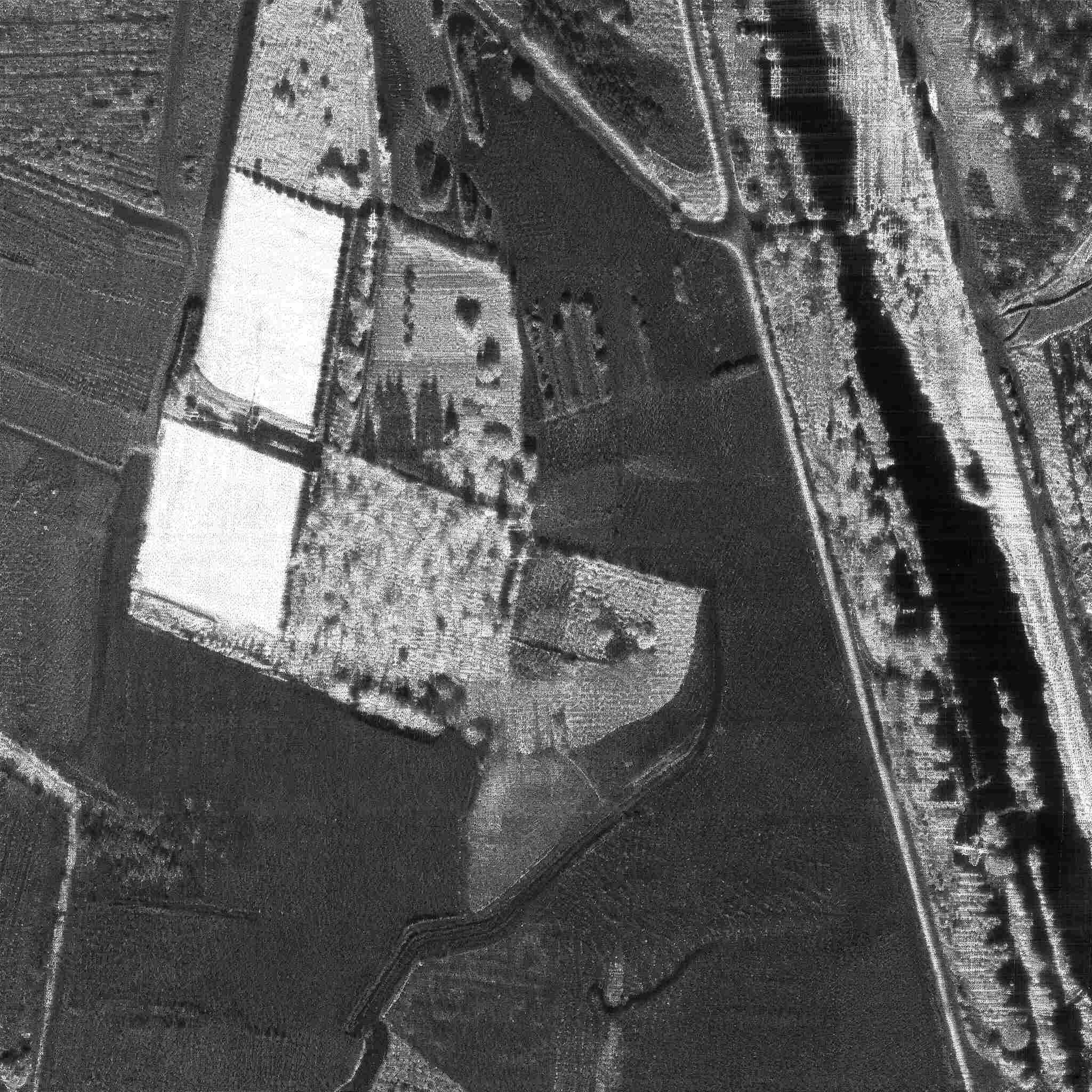} 
            \\
            \includegraphics[width=0.5\textwidth]{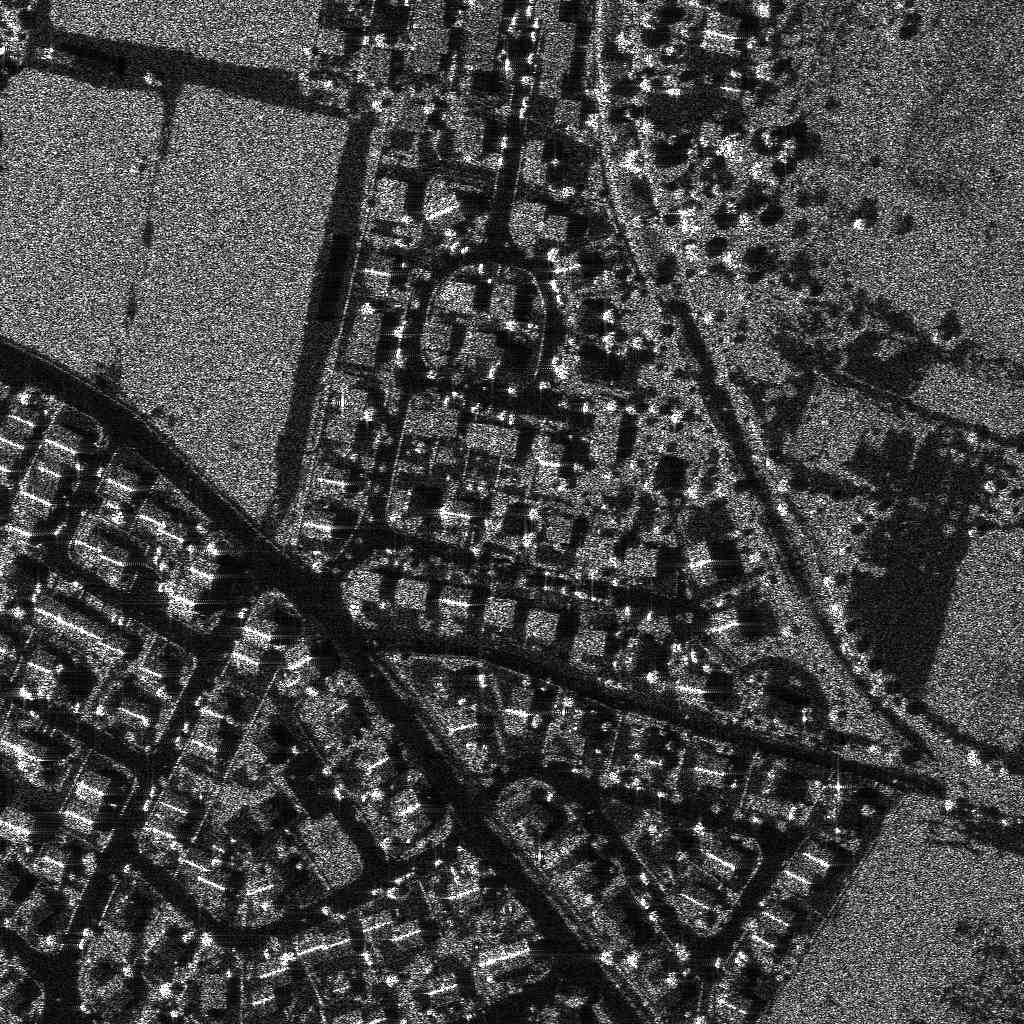} &
            \includegraphics[width=0.5\textwidth]{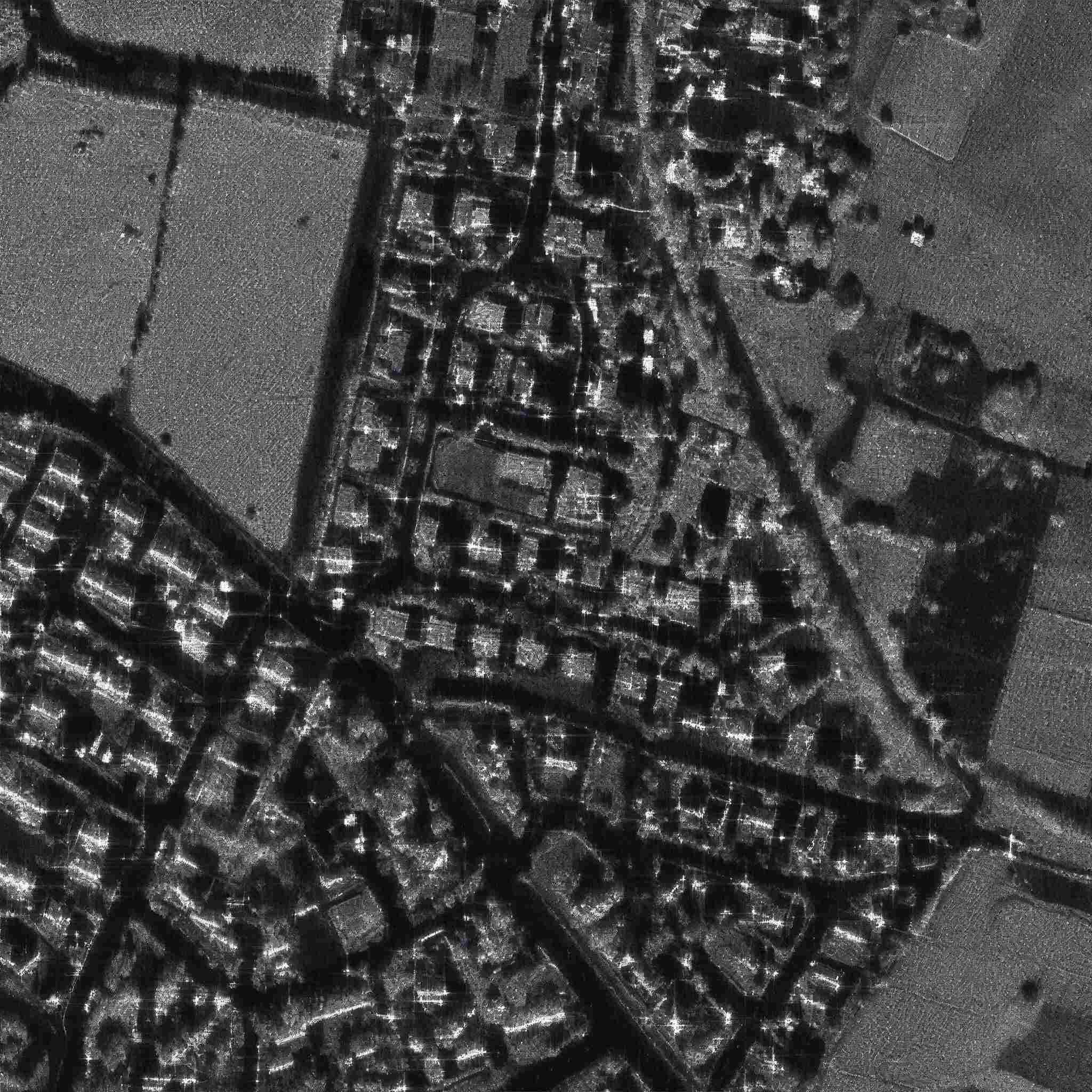} 
            \\
            \includegraphics[width=0.5\textwidth]{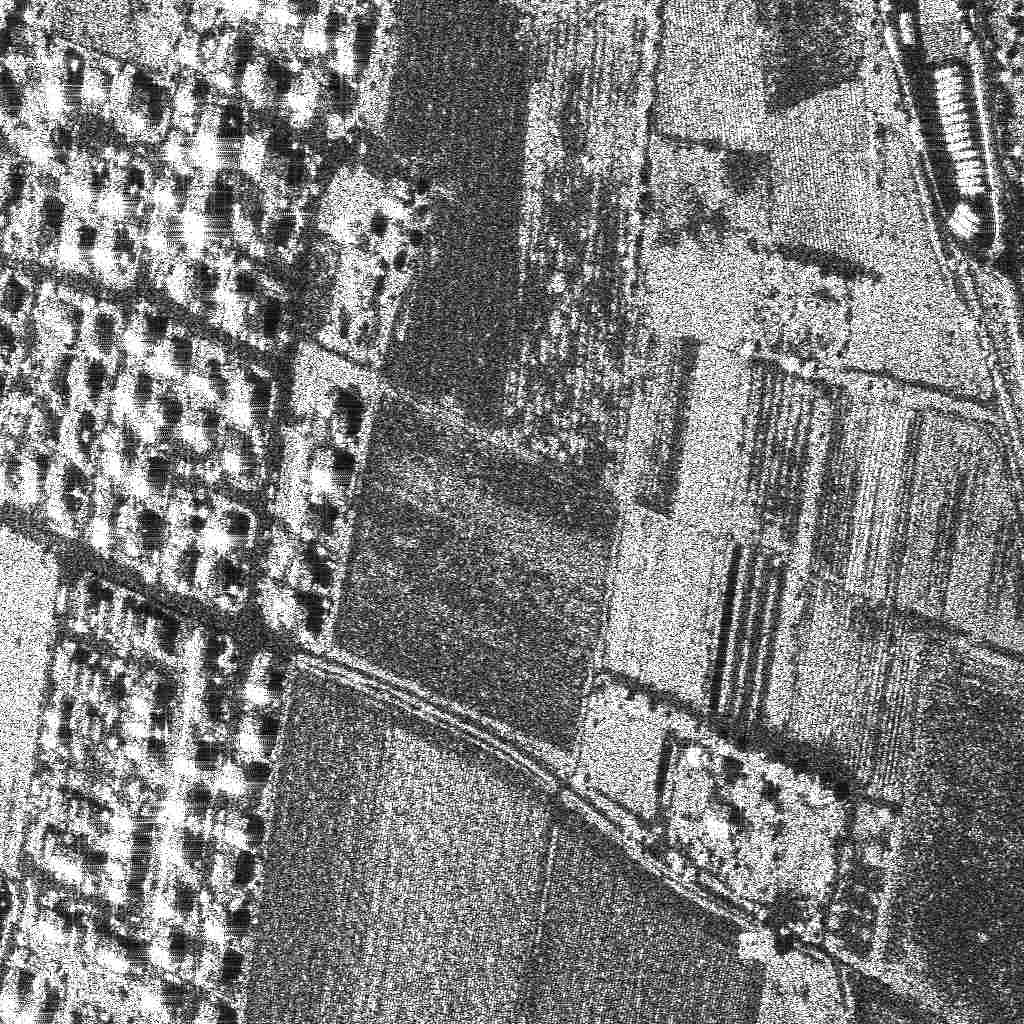} &
            \includegraphics[width=0.5\textwidth]{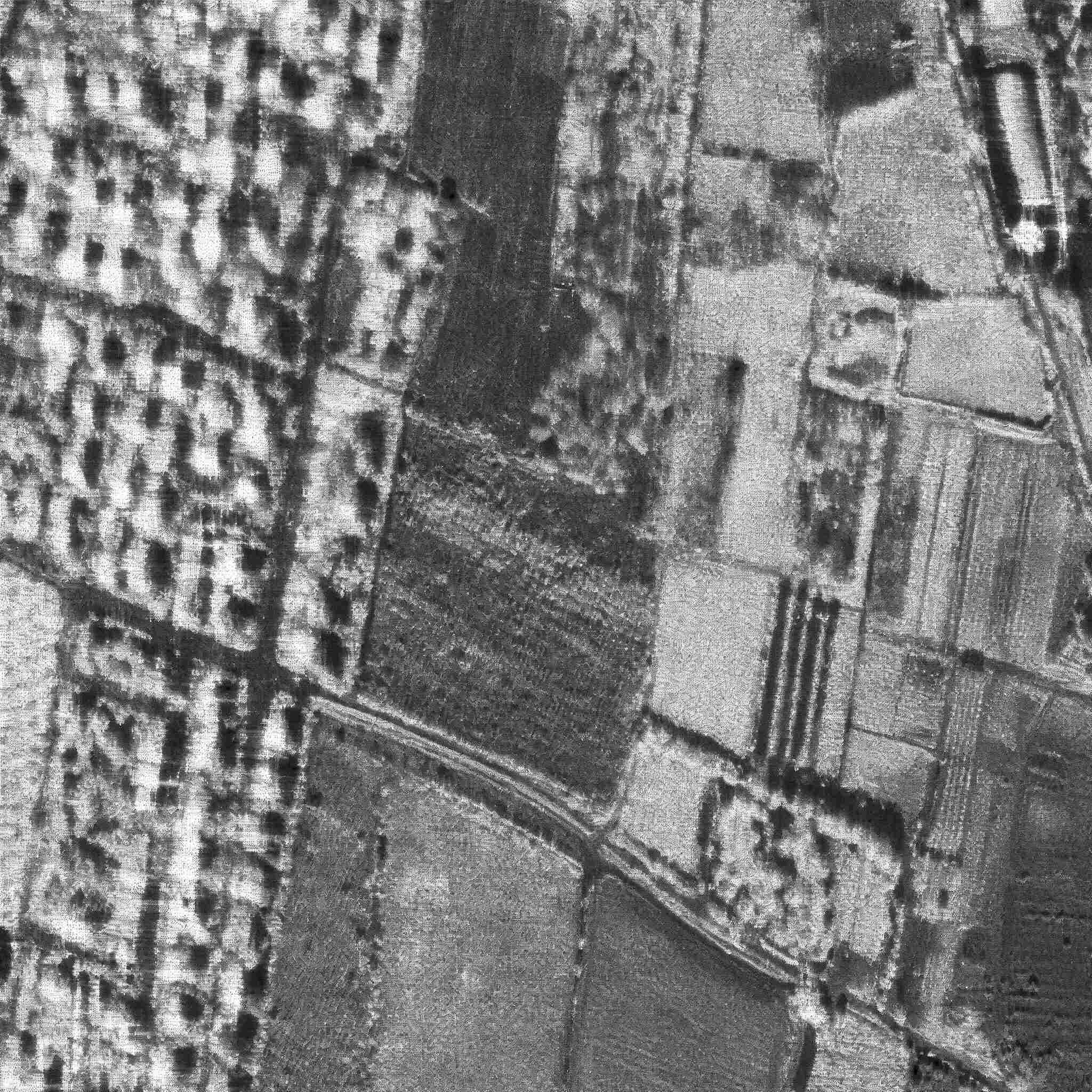} 
            \\
            \textbf{Simulated image (1024x1024)} & \textbf{Generated image (2048x2048)} \\
        \end{tabular}
        \end{adjustbox}
        \captionof{figure}{Enhancing ONERA's physics-based simulator EMPRISE Images with prompts (1): "A SAR image of  distinct patches of crop fields near a long river.", (2): "A SAR image of a dynamic city with buildings near highways and a few isolated fields." and (3): "A SAR image with crop fields near a city."}
        \label{fig:emprise}
    \end{minipage}
    
\end{figure}

Figure~\ref{fig:emprise} highlights the model’s ability to enhance simulated SAR images by adding spatially consistent fine textures, making them visually closer to real data and more suitable for human interpretation. In particular, it enriches vegetated areas with realistic structural details and interprets ambiguous shadow regions by adding plausible vegetation patterns.

\section{Discussion and Conclusion}

We presented, to our knowledge, the first comparative study of fine-tuning strategies for a large latent diffusion model—Stable Diffusion XL (SDXL)—on Synthetic Aperture Radar (SAR) imagery. While most prior work addresses optical domains, our results show that a vision–language foundation model can be adapted to generate physically grounded SAR scenes.

Experiments indicate that full UNet fine-tuning is most effective for learning SAR-specific structure, whereas text-encoder adapters benefit from low-rank updates and a learned \texttt{<SAR>} token to preserve prompt fidelity. A brief low-noise refinement of the VAE decoder further improves textural realism without destabilizing the latent space. Beyond unconditional synthesis, the model supports conditioned generation for practical use cases, including TerraSAR-X–guided 40\,cm synthesis and refinement of physics-based simulator outputs.

This capability enables scalable data augmentation and composition of rare or operationally relevant scenarios. Leveraging open-source foundation models also facilitates multimodal conditioning (e.g., segmentation, depth) and ControlNet-based guidance. 

Limitations include evaluation restricted to X-band at 40cm slant-range sampling distance and specific processing settings. Nonetheless, although our experiments target 40cm, we detail a parameter-efficient procedure to adapt the model to other resolutions using low computational resources (one GPU H100). Future directions include multimodal conditioning with elevation/optical inputs and extending learning to the complex domain (amplitude and phase).

In summary, this work highlights the potential of using generative AI for large-scale SAR scene generation, a domain where physics-based simulators are still underdeveloped. It sets a precedent for adapting pretrained generative models to unconventional data domains. It delivers insights for both the SAR and broader AI communities on how to transfer powerful foundation models to non-optical, physics-driven settings. Future directions include: multimodal conditioning (e.g., combining text with elevation or optical inputs), and learning in the complex SAR domain (amplitude and phase).

\section{CRediT authorship contribution statement}
\paragraph{Declaration of generative AI and AI-assisted technologies in the writing process}

During the preparation of this work the authors used ChatGPT to assist with the translation from French to English. After using this tool, the authors reviewed and edited the content as needed and take full responsibility for the content of the publication.

\paragraph{Declaration of competing interest}
The authors declare that they have no known competing financial interests or personal relationships that could have appeared to influence the work reported in this paper.

\section{Acknowledgments}
This work was supported by ONERA - The French Aerospace Lab, with financial support from the French Ministry of Defence (MoD). This research was conducted as part of Solène Debuysère's PhD thesis. The authors wish to express their sincere gratitude to all those who contributed to the success of this work.

\appendix

\section{Mean Absolute Weight Change (MAWC)}
\label{appendix:magnitude-ratio}
\setcounter{equation}{0}  
\renewcommand{\theequation}{A.\arabic{equation}}  
\setcounter{figure}{0}  
\renewcommand{\thefigure}{A.\arabic{figure}}  

We defined the Mean Absolute Weight Change (MAWC) as a metric that measures how much the weights of a model have changed between two checkpoints. Unlike metrics that simply count modified weights, MAWC also accounts for the magnitude of change.

\paragraph{Definition.}  
Let $w_i^{(0)}$ and $w_i^{(1)}$ denote the value of the $i$-th weight respectively, before and after fine-tuning. Let $W$ be the total number of weights in a given layer.

We define the absolute change in the $i$-th weight as:
\begin{equation}
\Delta w_i = \left| w_i^{(1)} - w_i^{(0)} \right|
\end{equation}

Then, the \textbf{Mean Absolute Weight Change (MAWC)} is the average absolute change across all weights:
\begin{equation}
\mathrm{MAWC} = \frac{1}{W} \sum_{i=1}^{W} \Delta w_i
\end{equation}

To analyze architectural patterns, we compute the MAWC for each layer individually and then report the mean across all layers belonging to the same sub-block (\textit{e.g.}, \texttt{ResNet 1}, \texttt{Attention 0}).

\begin{figure}
    \centering
    \includegraphics[width=0.99\linewidth]{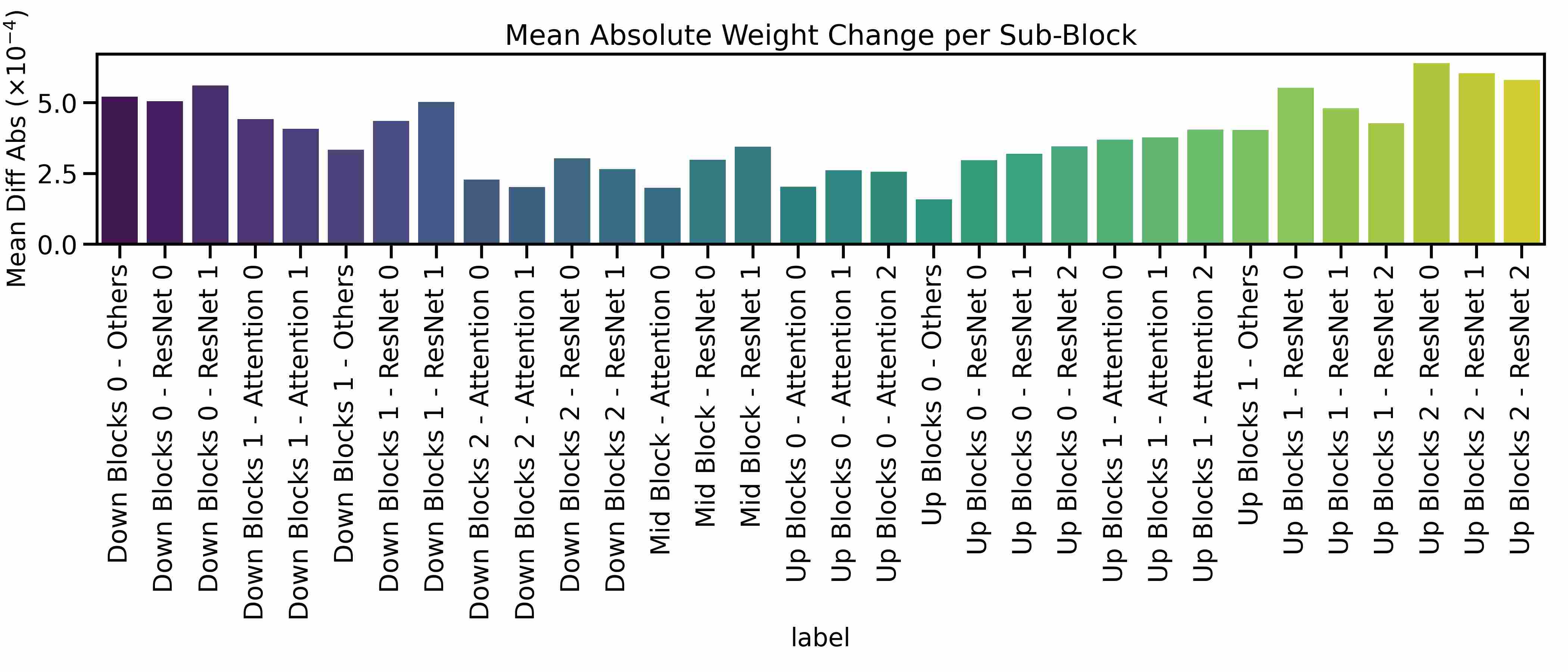}
    \caption{Mean Absolute Weight Difference per block and sub-block of the UNet during the training configuration \textit{soleil-up-7} with a resolution threshold of $5 \times 10^{-4}$}
    \label{fig:mawc}
\end{figure}

\section{Training images generation}
\label{appendix:training-images}
\setcounter{equation}{0}  
\renewcommand{\theequation}{B.\arabic{equation}}  
\setcounter{figure}{0}  
\renewcommand{\thefigure}{B.\arabic{figure}}  

To visually assess the quality and diversity of generations, we display side-by-side image samples produced by each model across six representative scene categories at epoch 8 using the same seed.

\begin{figure*}
    \centering
    \scriptsize
    \setlength{\tabcolsep}{1pt}
    \renewcommand{\arraystretch}{1.1}
    \begin{adjustbox}{max width=0.80\textwidth}
    
    \begin{tabular}{c@{\hskip 4pt}*{6}{>{\centering\arraybackslash}m{0.14\textwidth}}}
        & \textbf{Field} & \textbf{City} & \textbf{Forest} & \textbf{Seacoast} & \textbf{Airport} & \textbf{Mountains} \\

        \rotatebox[origin=c]{90}{\textbf{soleil-up-7}} &
        \includegraphics[width=\linewidth]{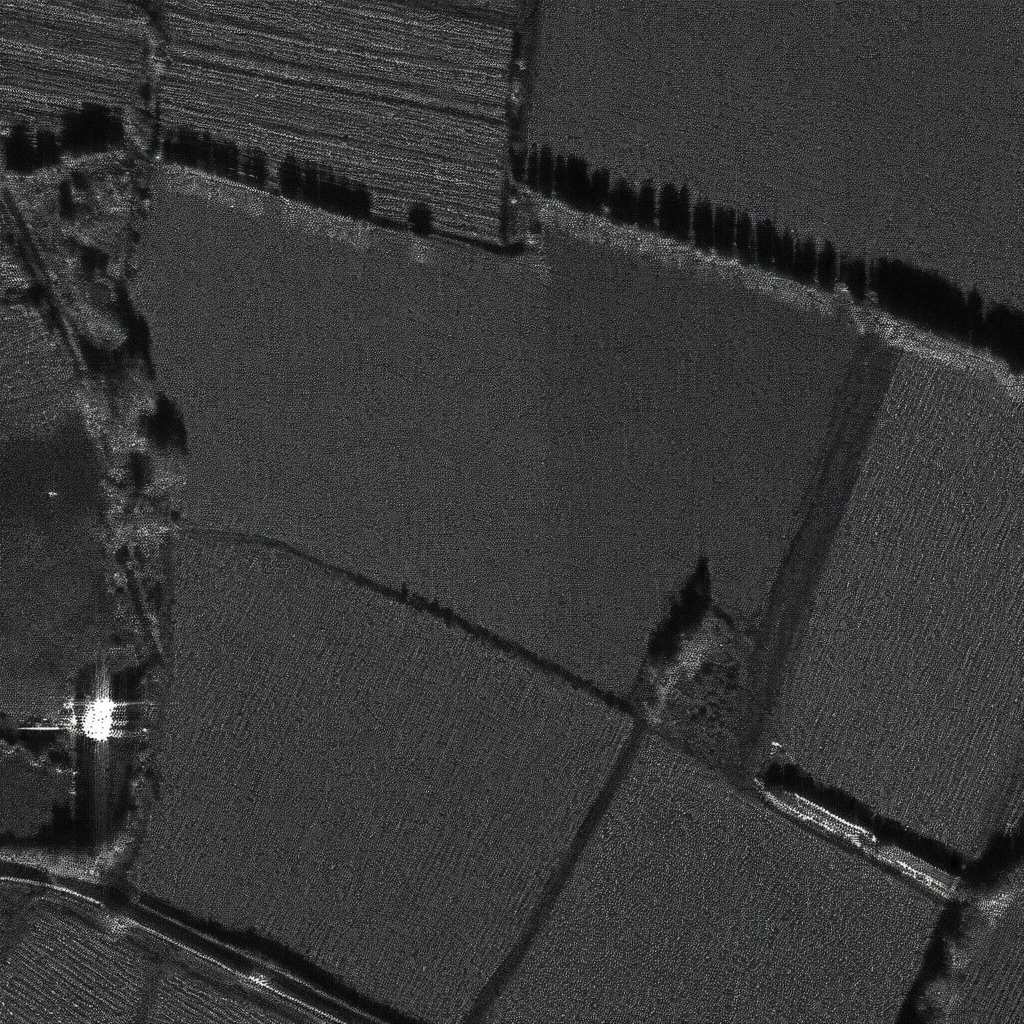} &
        \includegraphics[width=\linewidth]{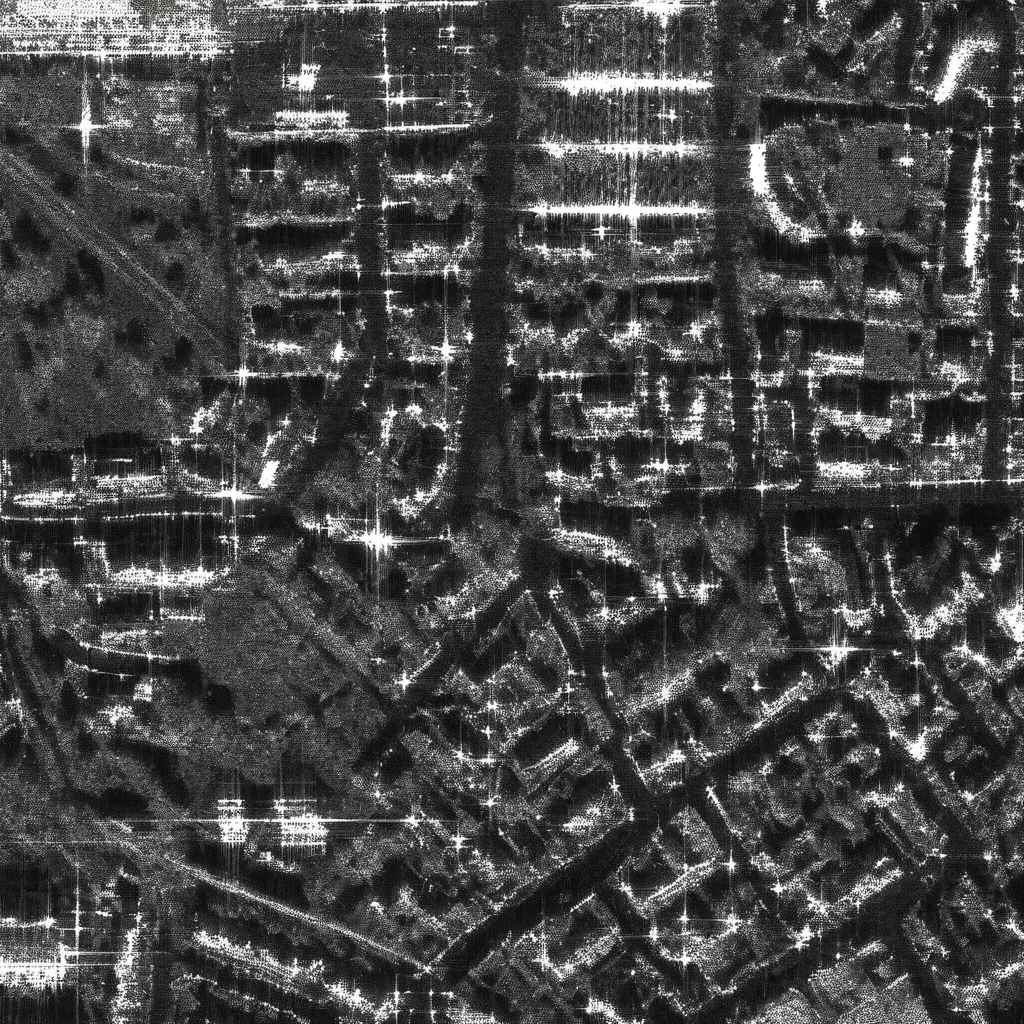} &
        \includegraphics[width=\linewidth]{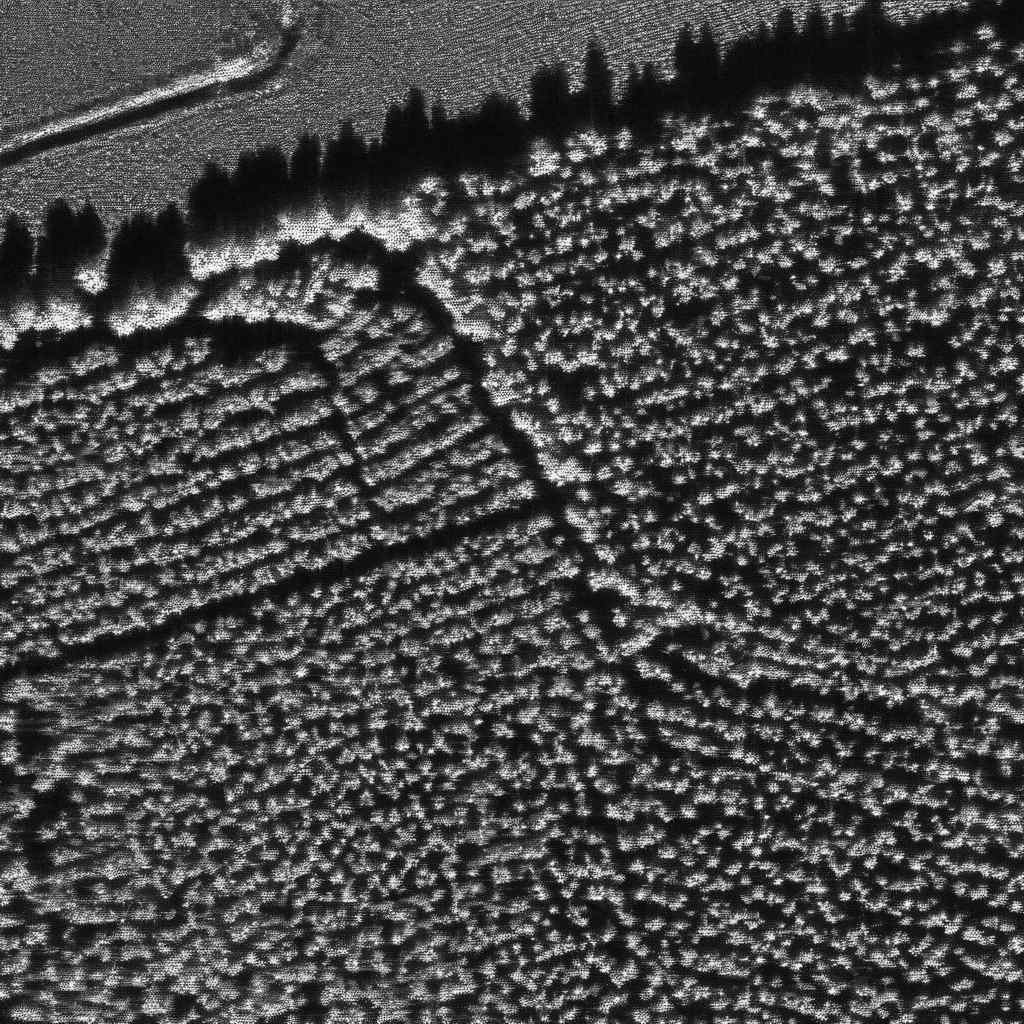} &
        \includegraphics[width=\linewidth]{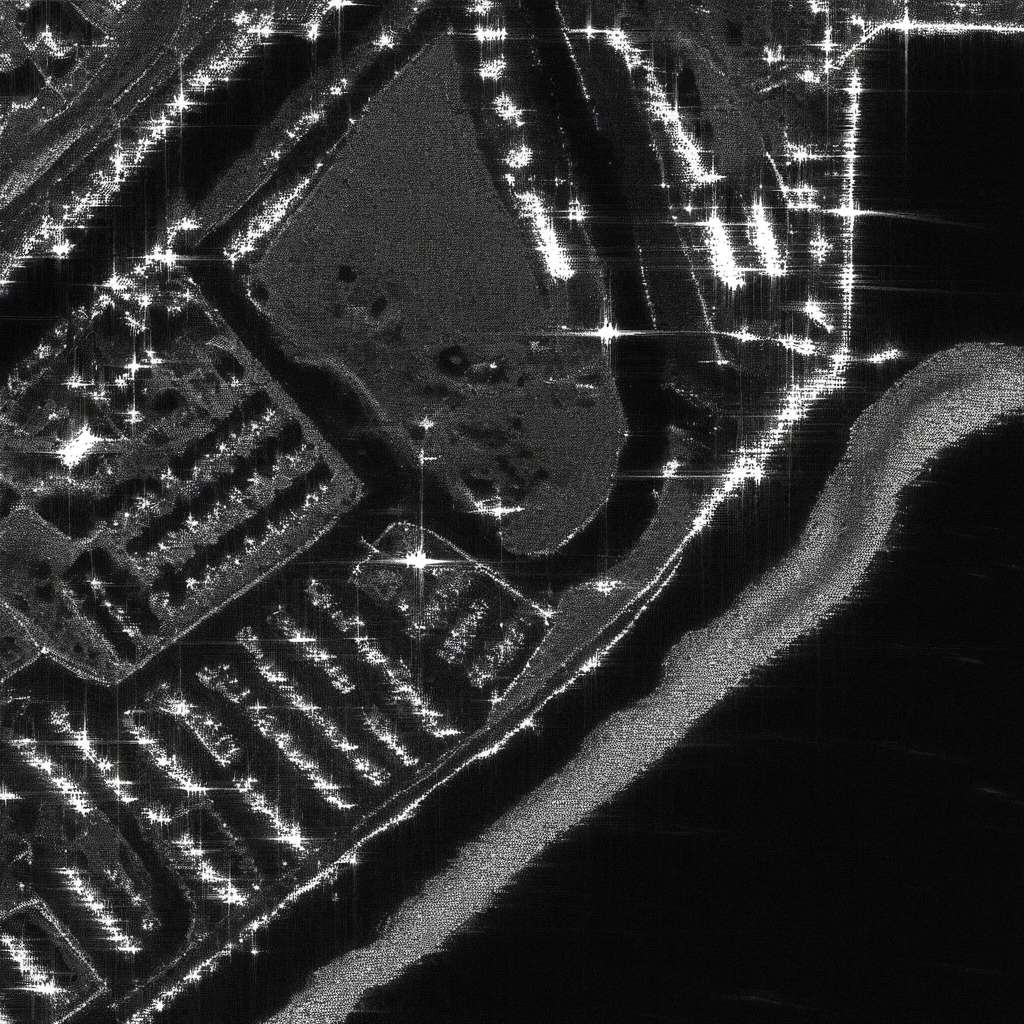} &
        \includegraphics[width=\linewidth]{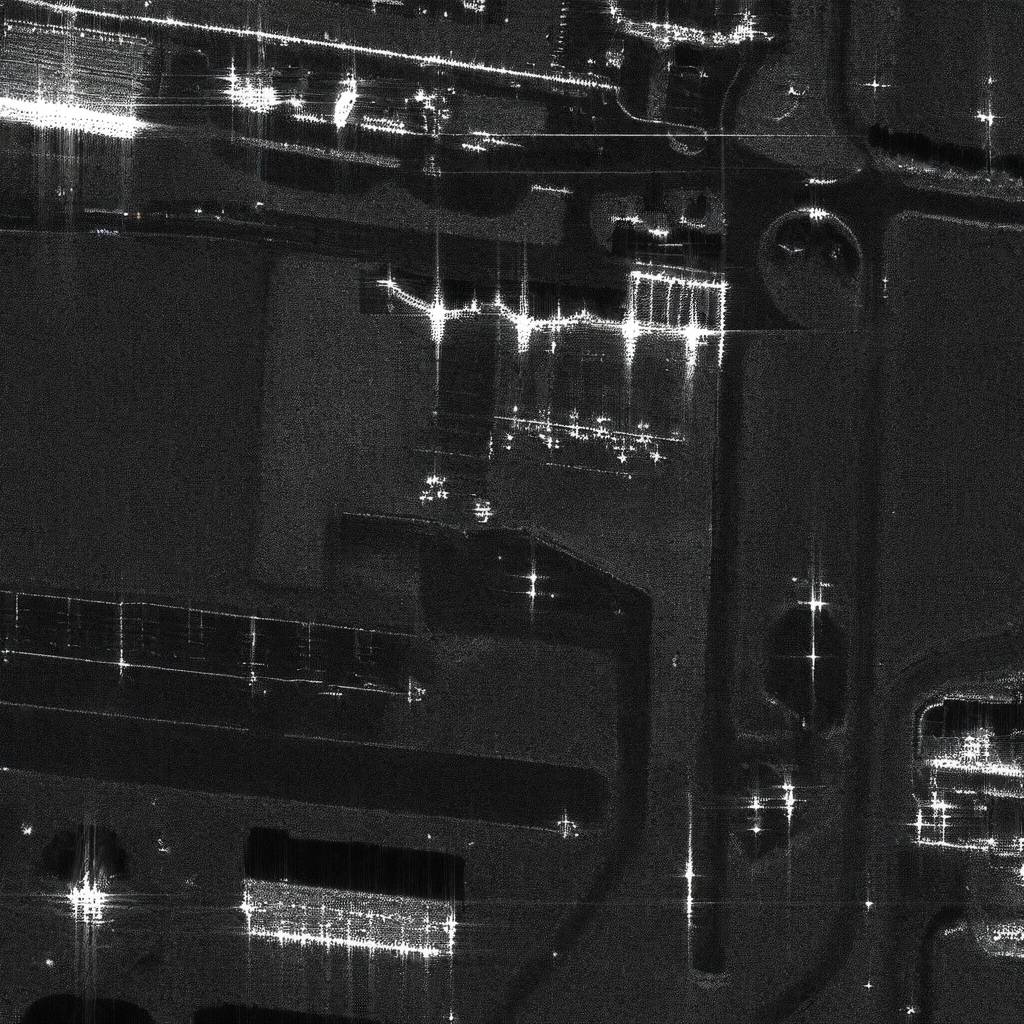} &
        \includegraphics[width=\linewidth]{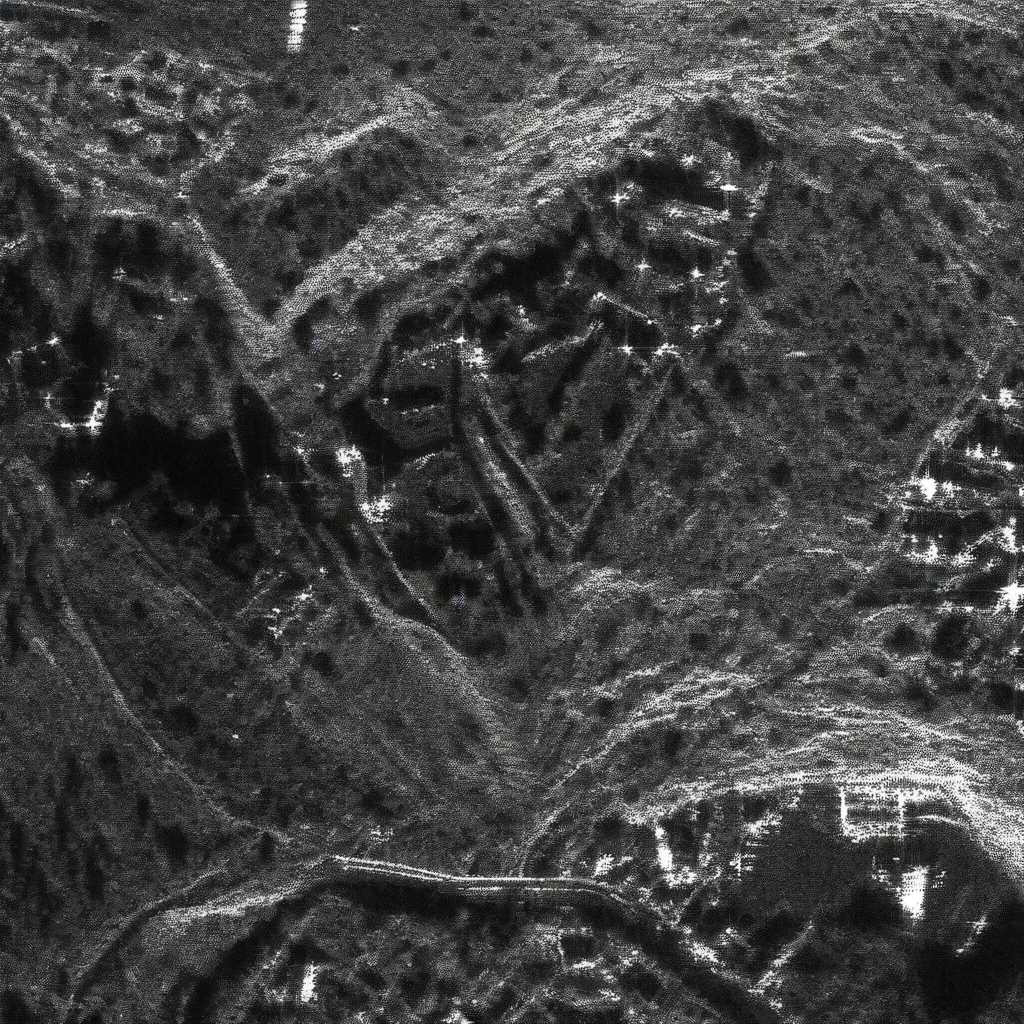} \\

        \rotatebox[origin=c]{90}{\textbf{umbrella-sand-8}} &
        \includegraphics[width=\linewidth]{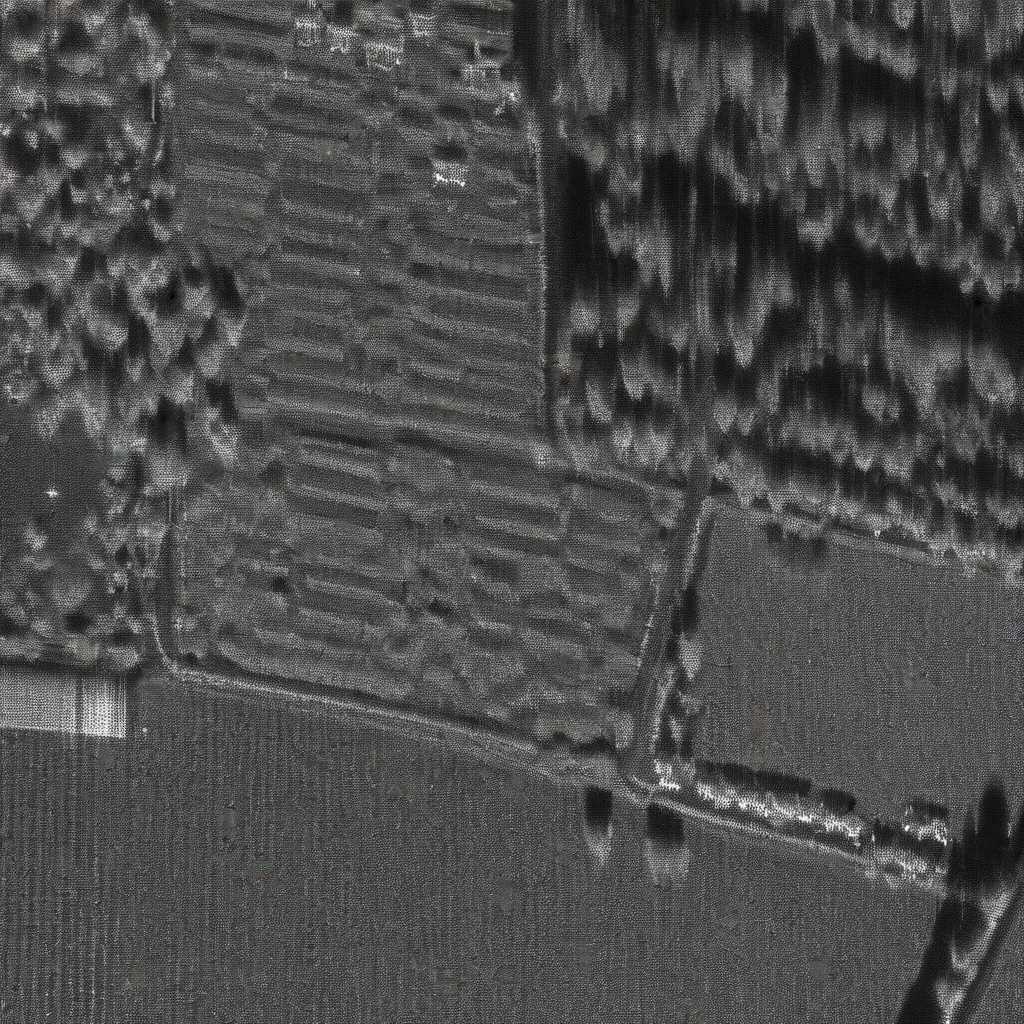} &
        \includegraphics[width=\linewidth]{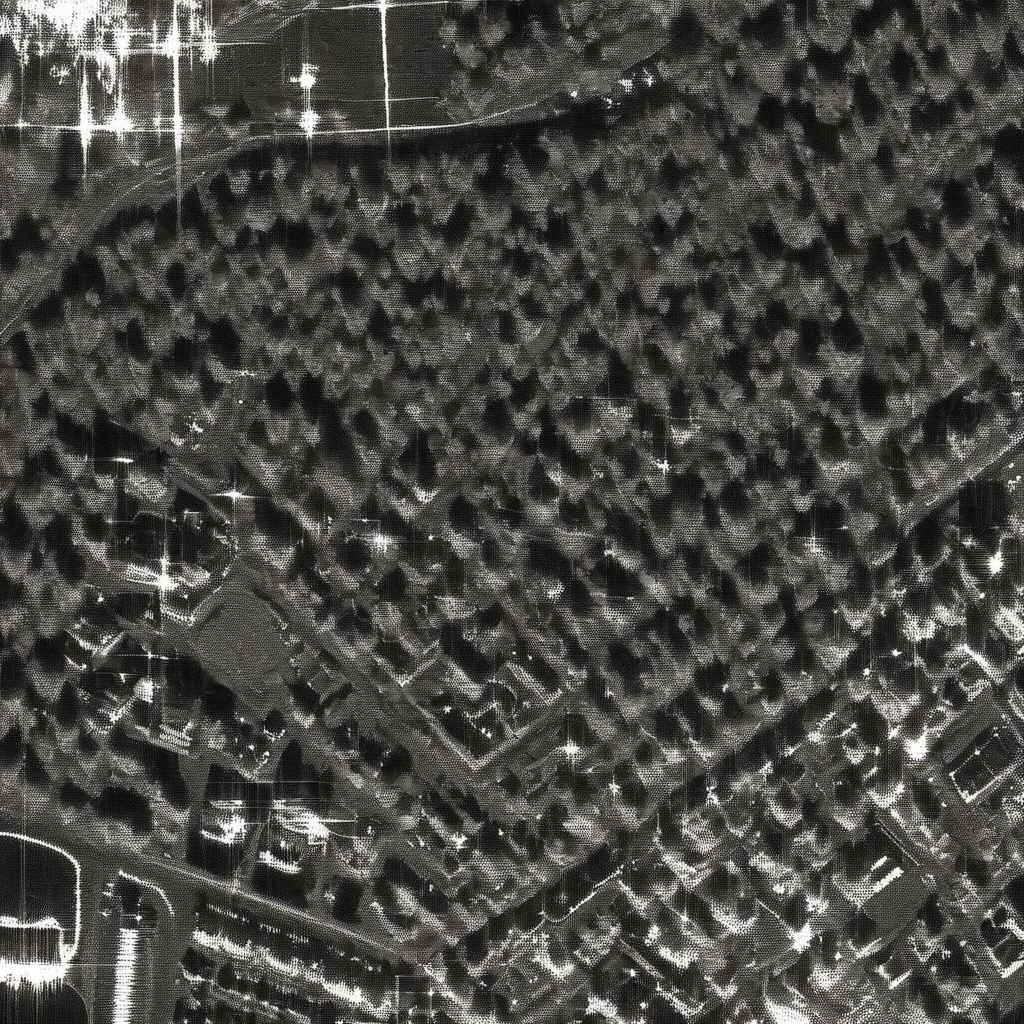} &
        \includegraphics[width=\linewidth]{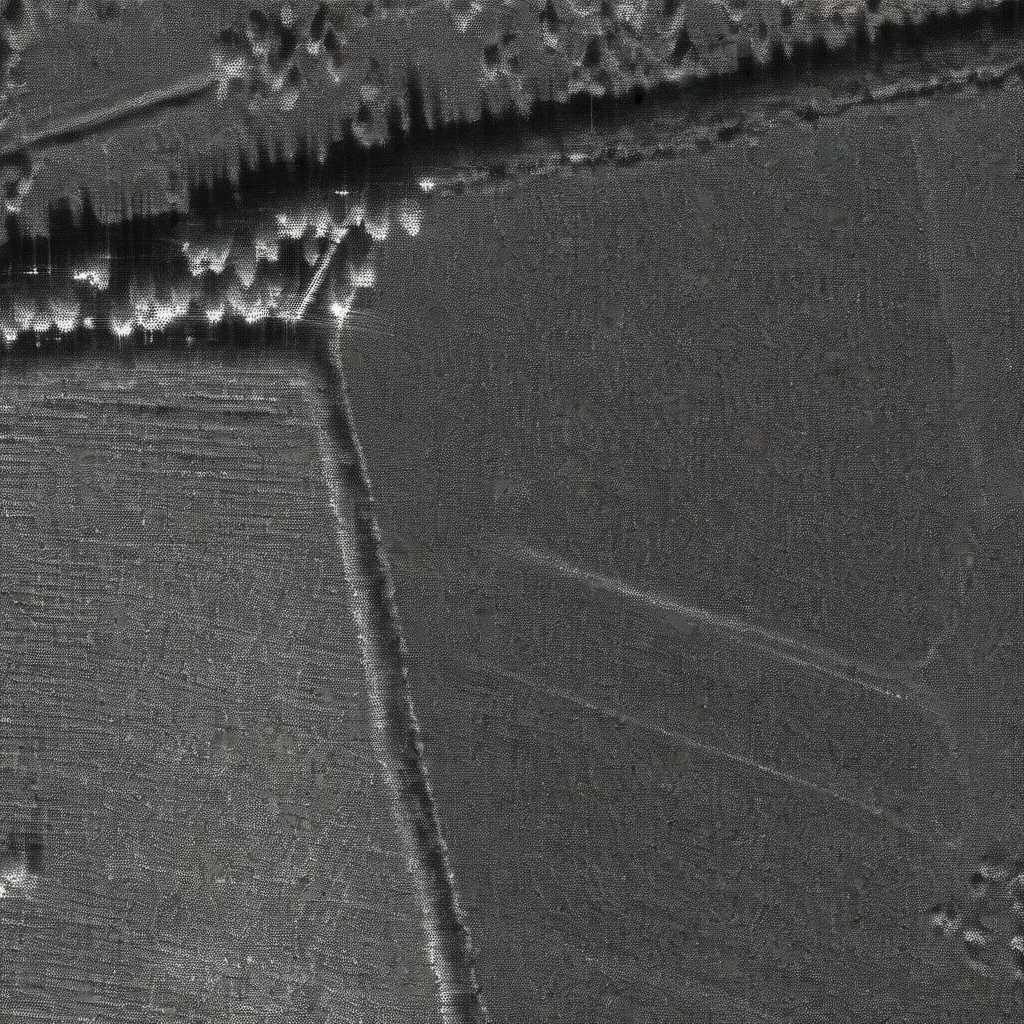} &
        \includegraphics[width=\linewidth]{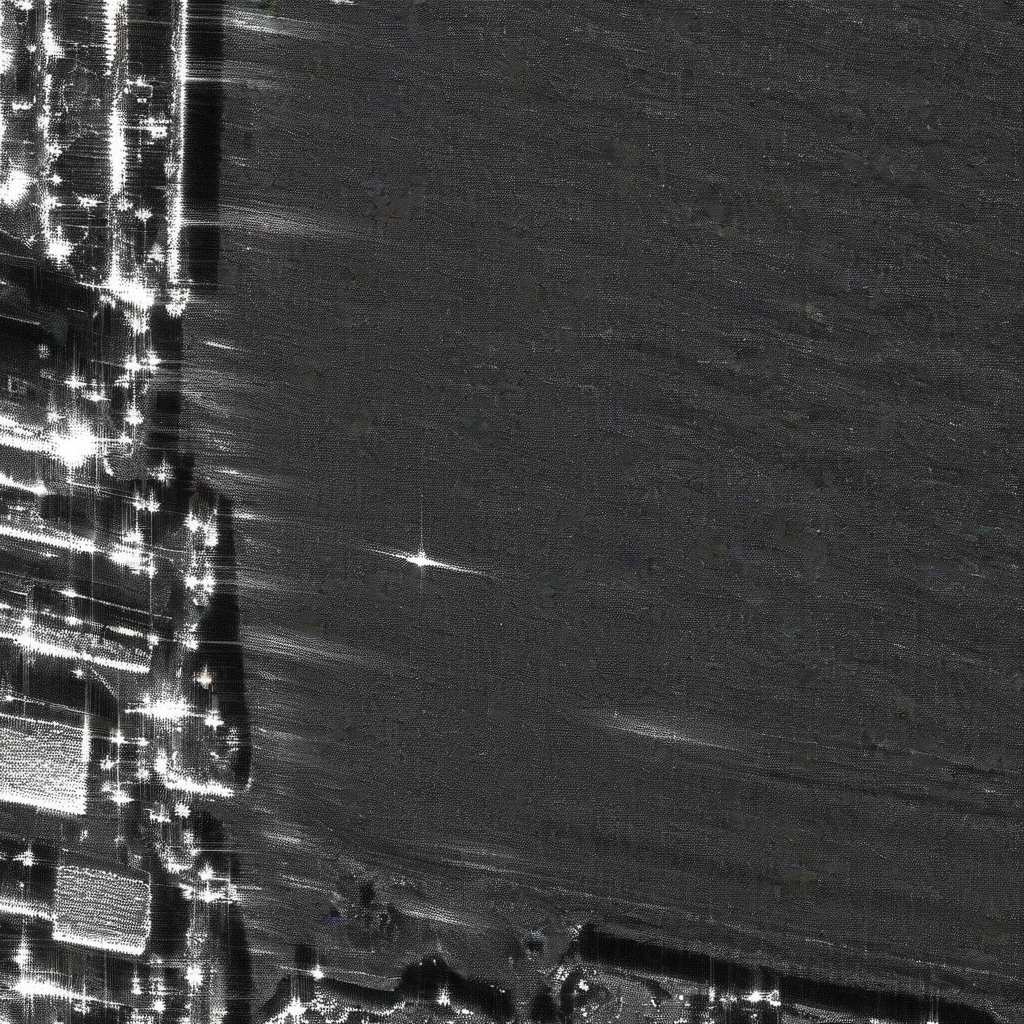} &
        \includegraphics[width=\linewidth]{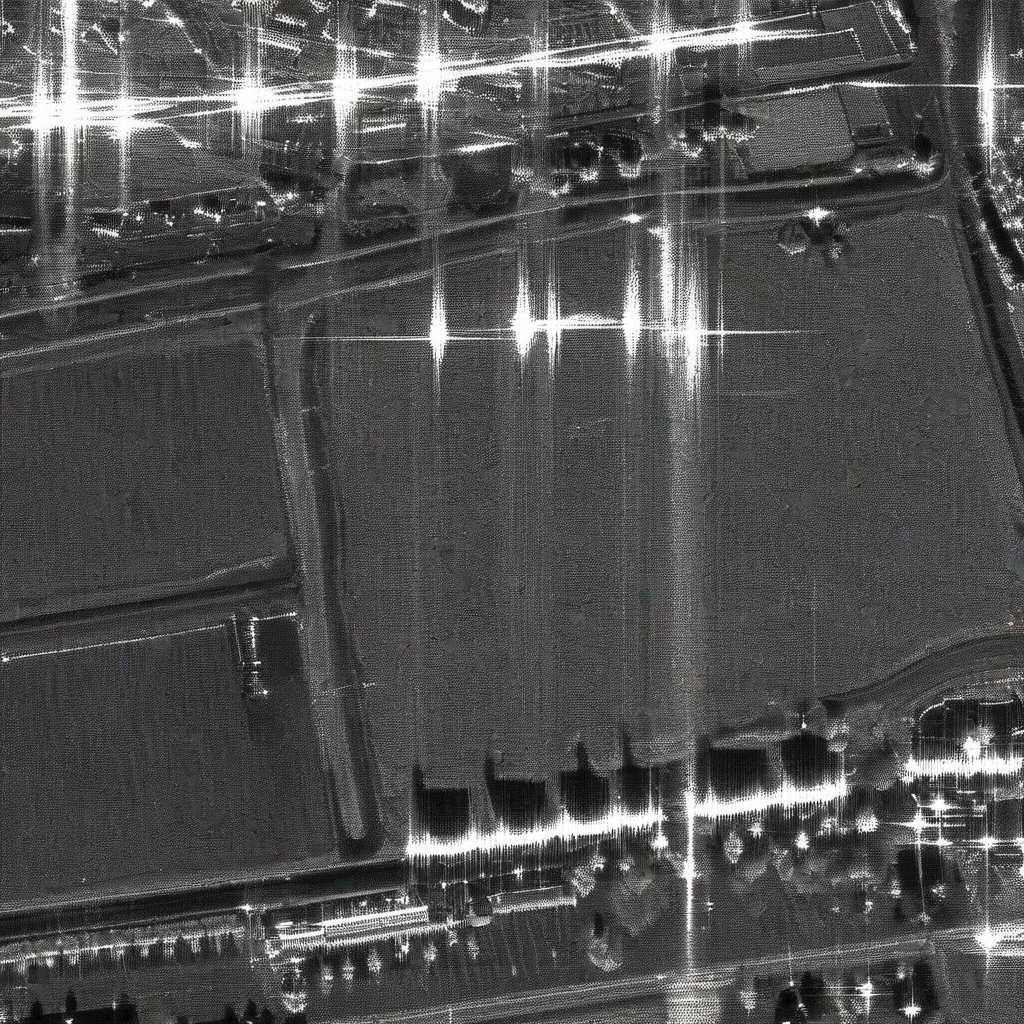} &
        \includegraphics[width=\linewidth]{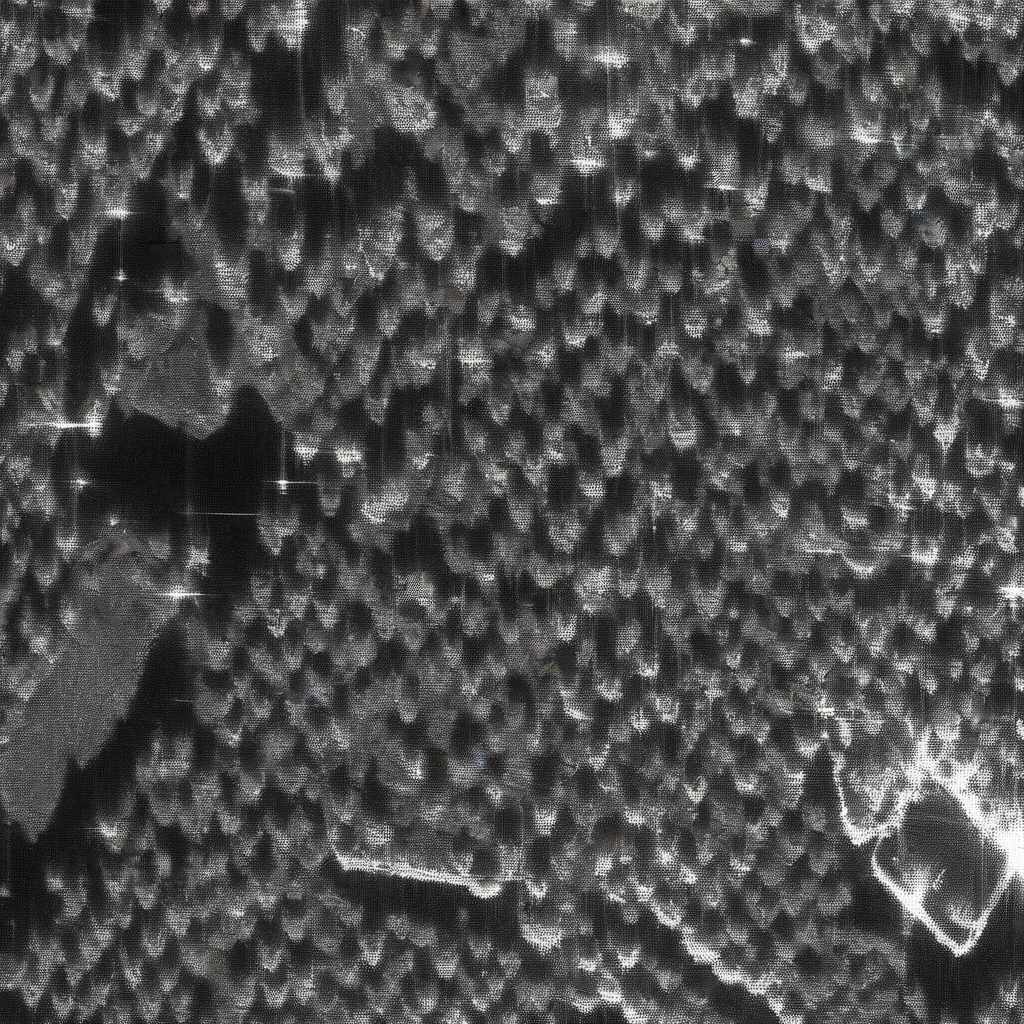} \\

        \rotatebox[origin=c]{90}{\textbf{rain-beach-6}} &
        \includegraphics[width=\linewidth]{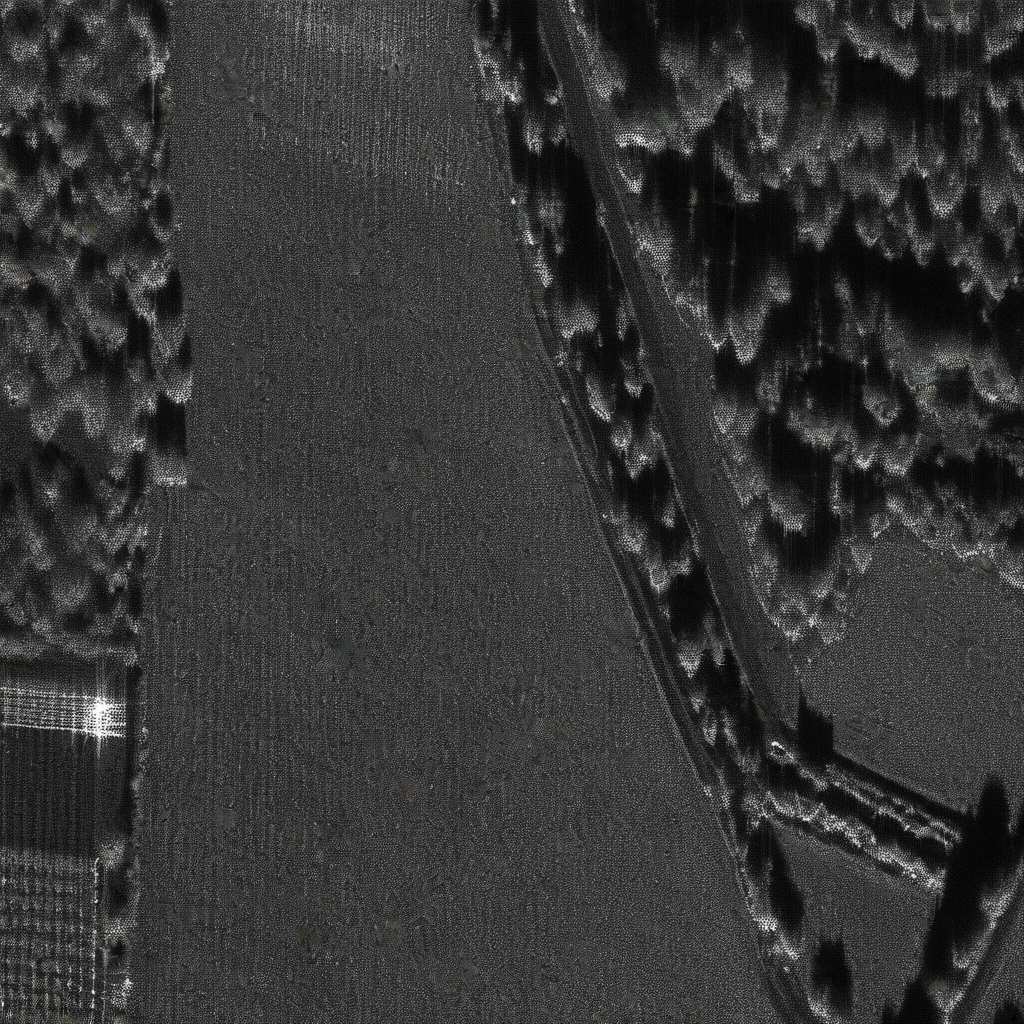} &
        \includegraphics[width=\linewidth]{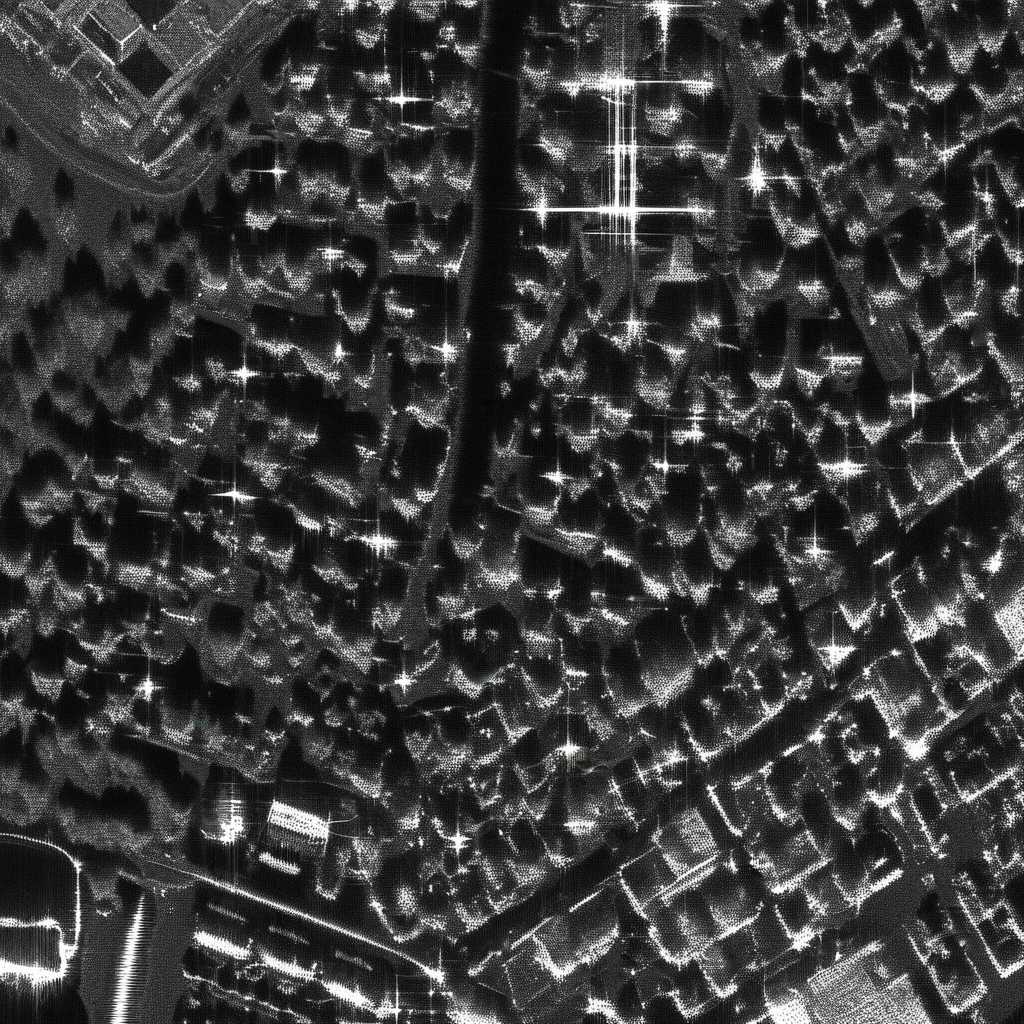} &
        \includegraphics[width=\linewidth]{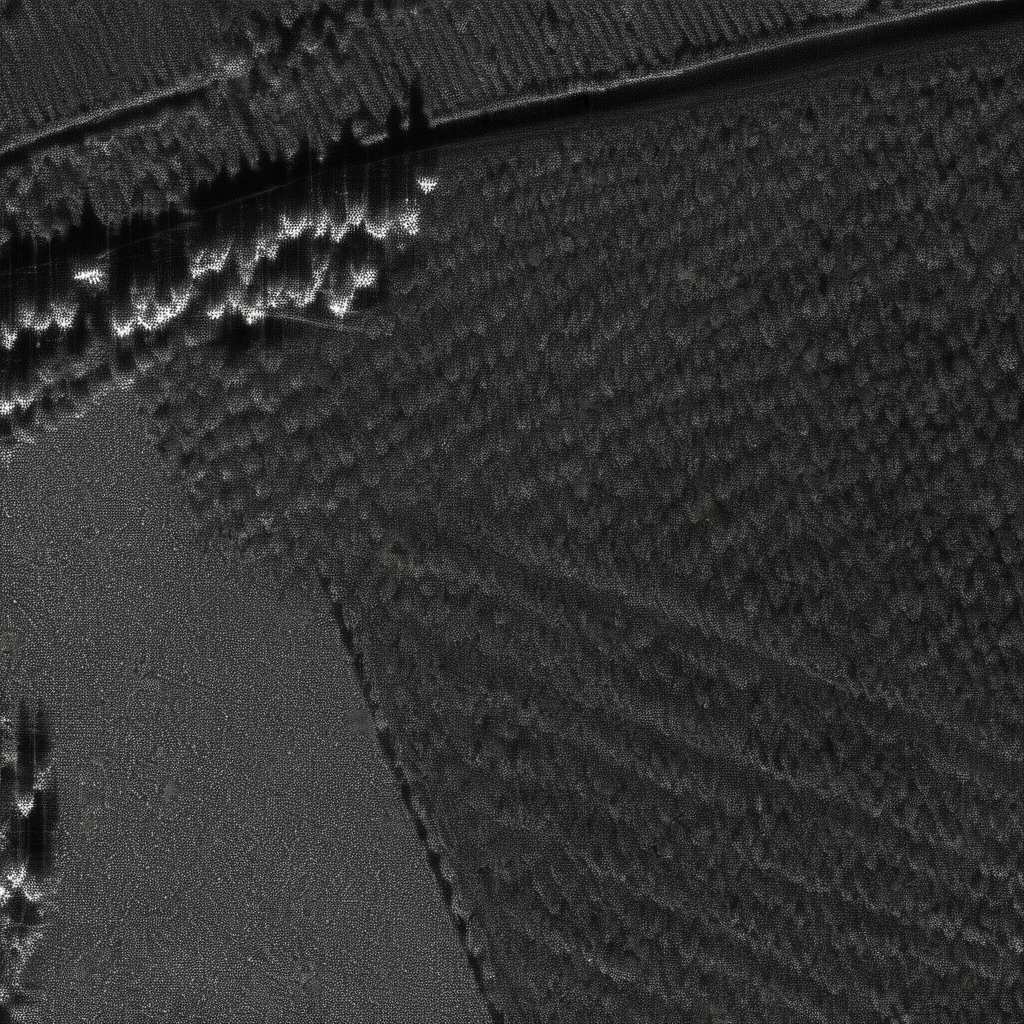} &
        \includegraphics[width=\linewidth]{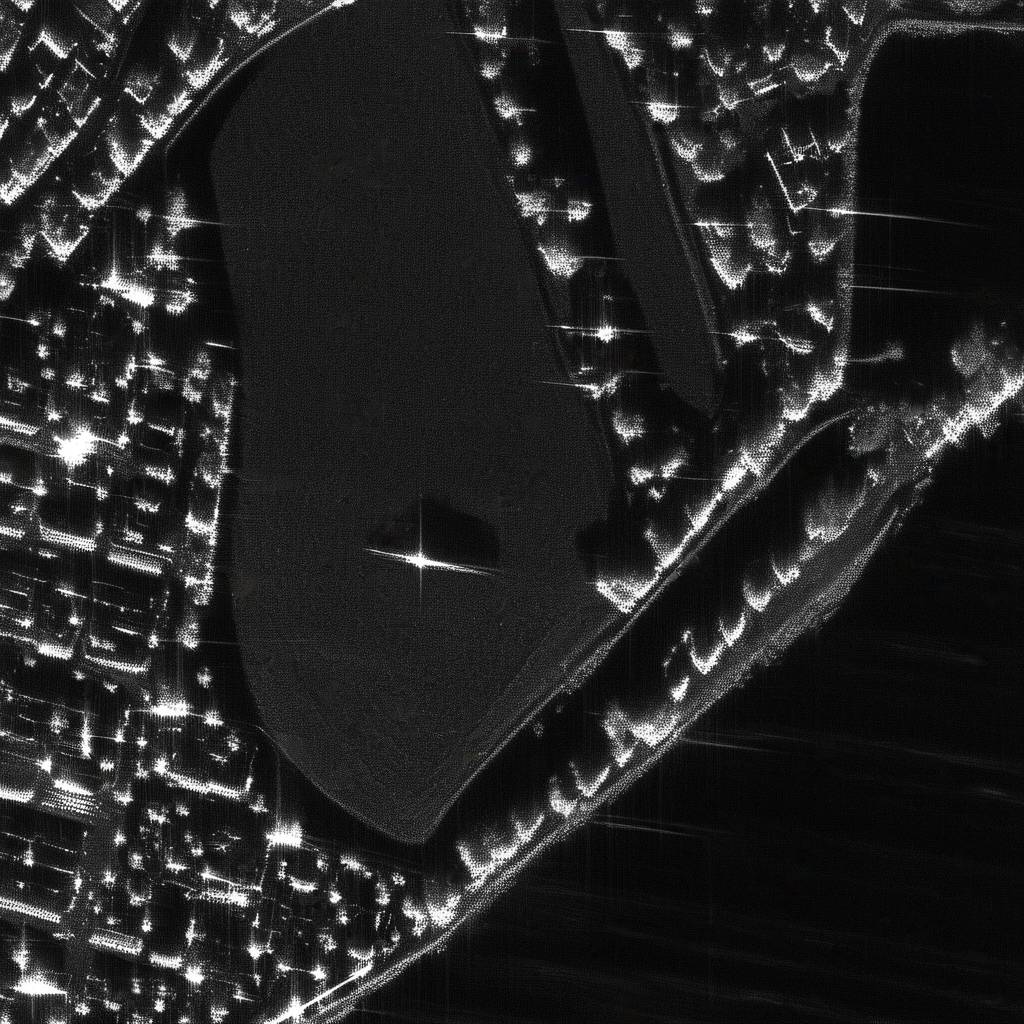} &
        \includegraphics[width=\linewidth]{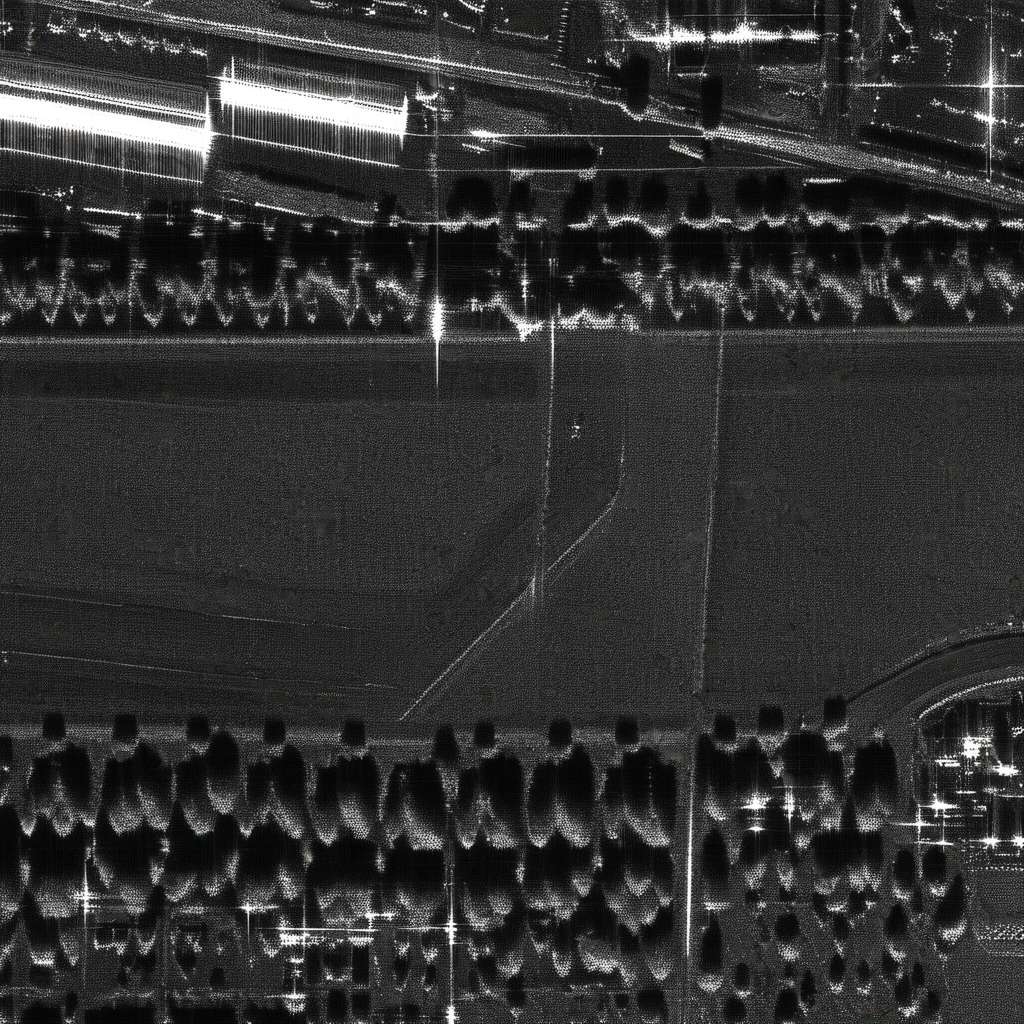} &
        \includegraphics[width=\linewidth]{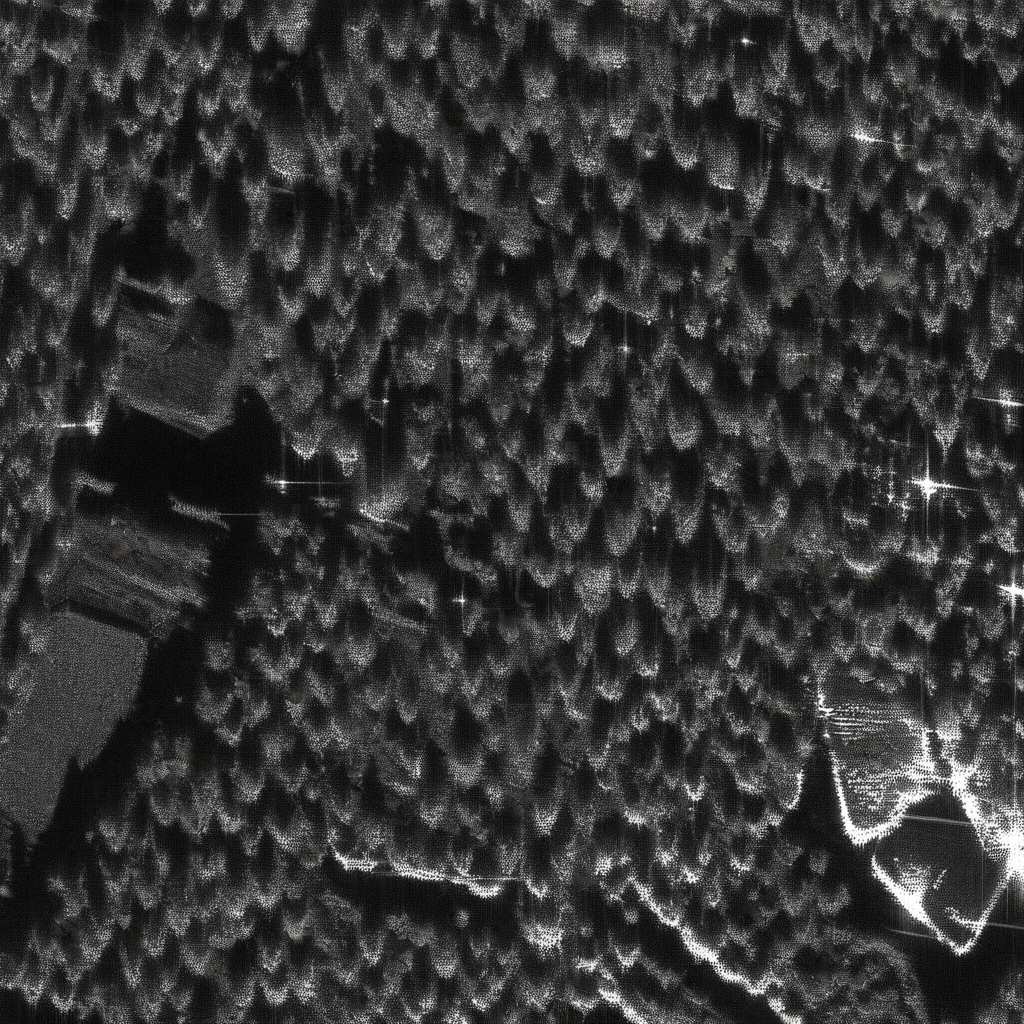} \\

        \rotatebox[origin=c]{90}{\textbf{lake-mont-9}} &
        \includegraphics[width=\linewidth]{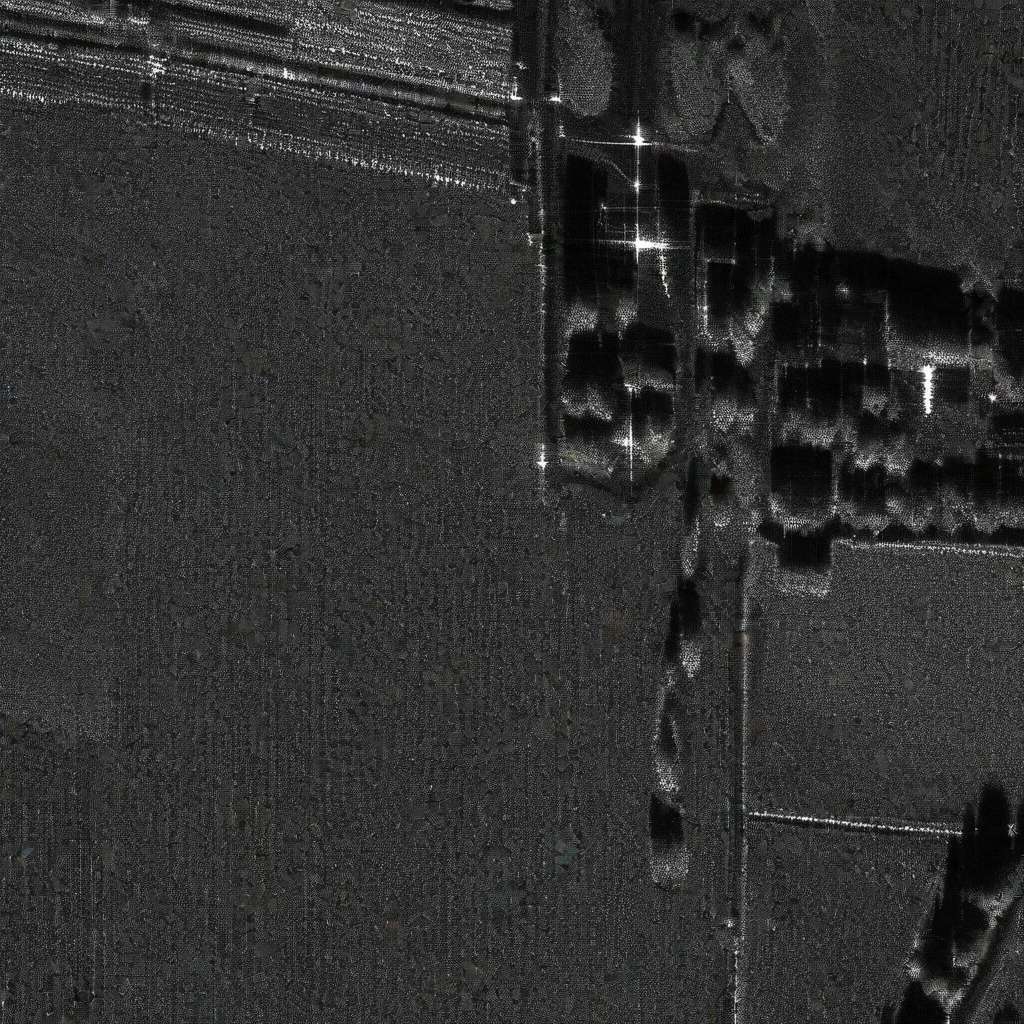} &
        \includegraphics[width=\linewidth]{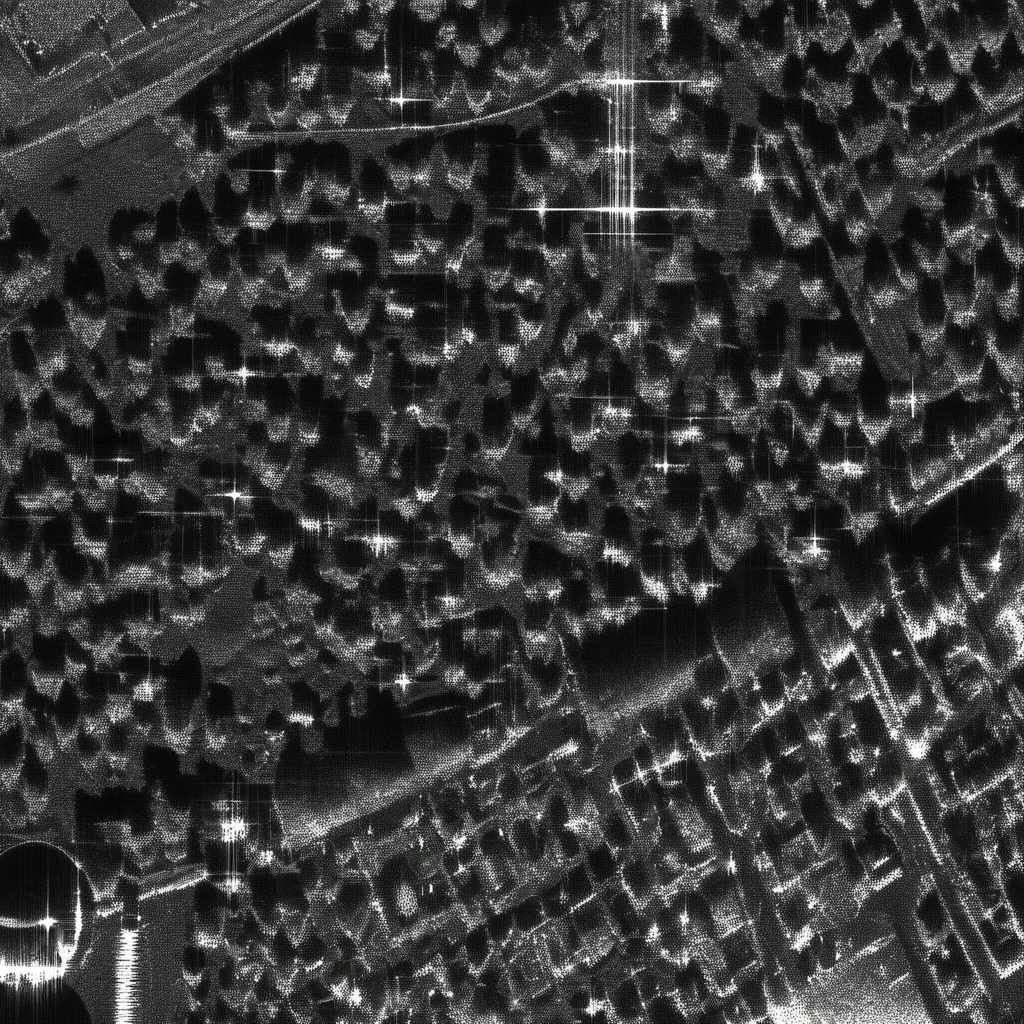} &
        \includegraphics[width=\linewidth]{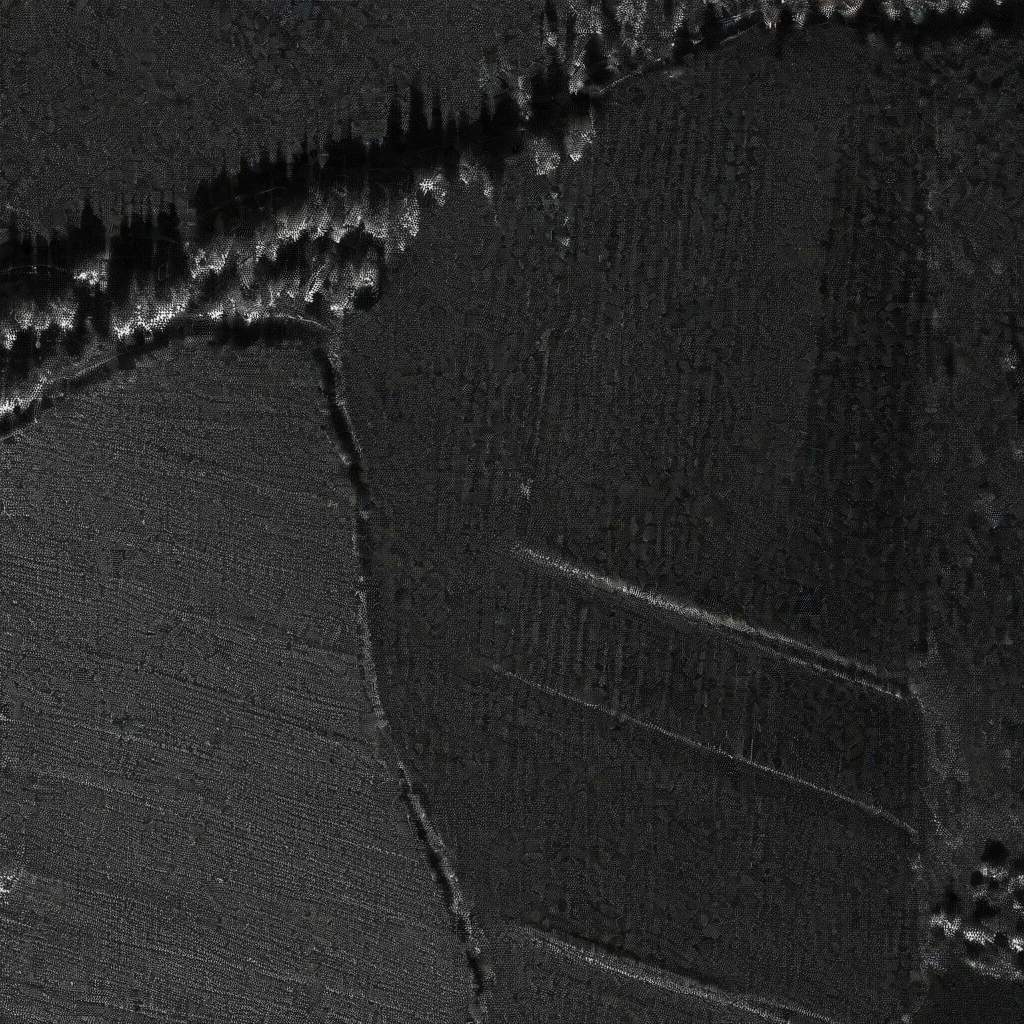} &
        \includegraphics[width=\linewidth]{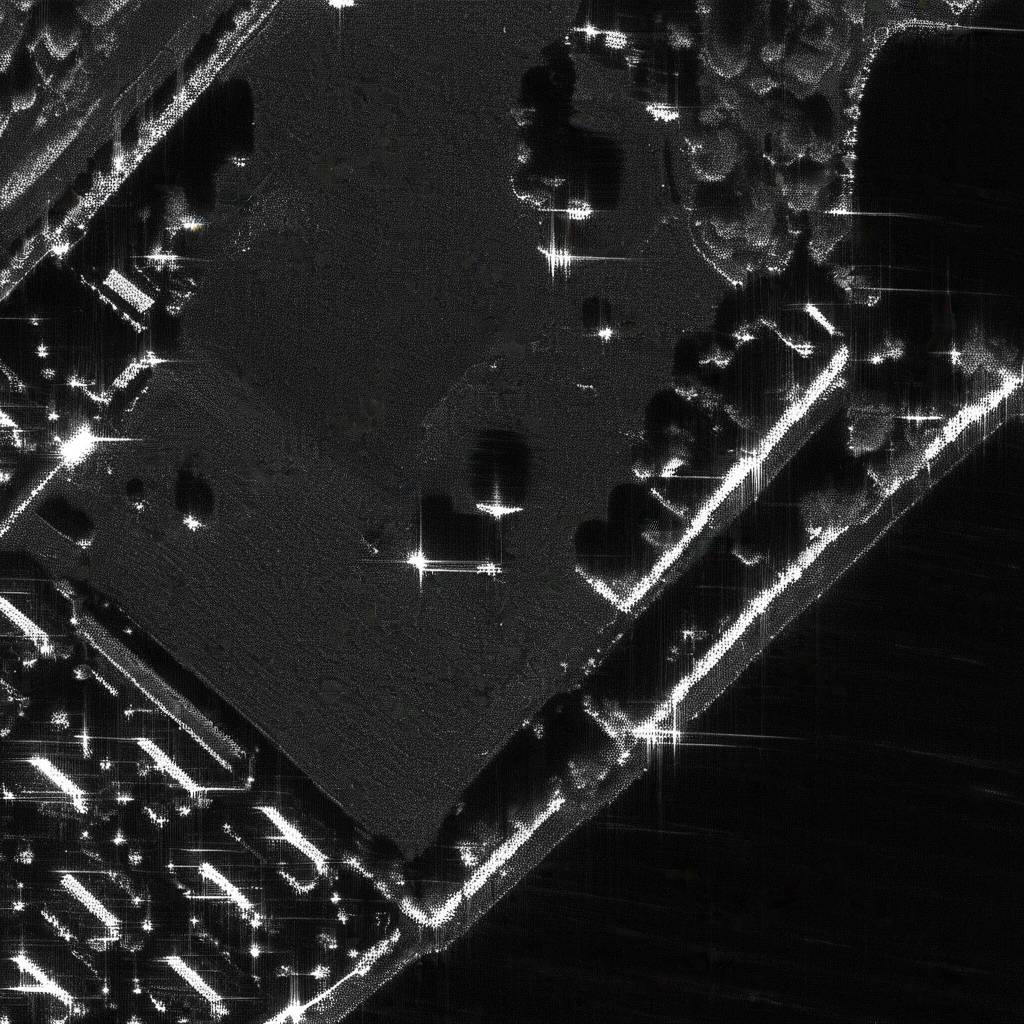} &
        \includegraphics[width=\linewidth]{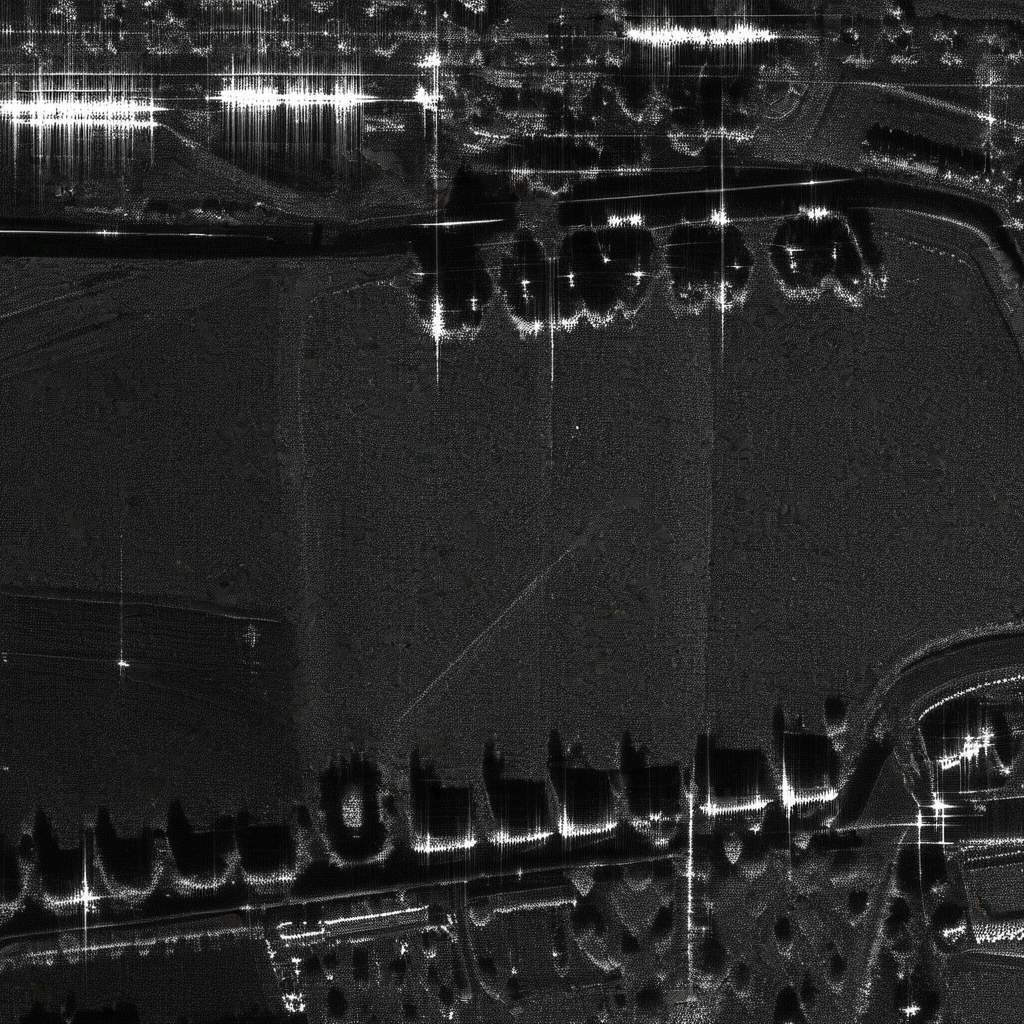} &
        \includegraphics[width=\linewidth]{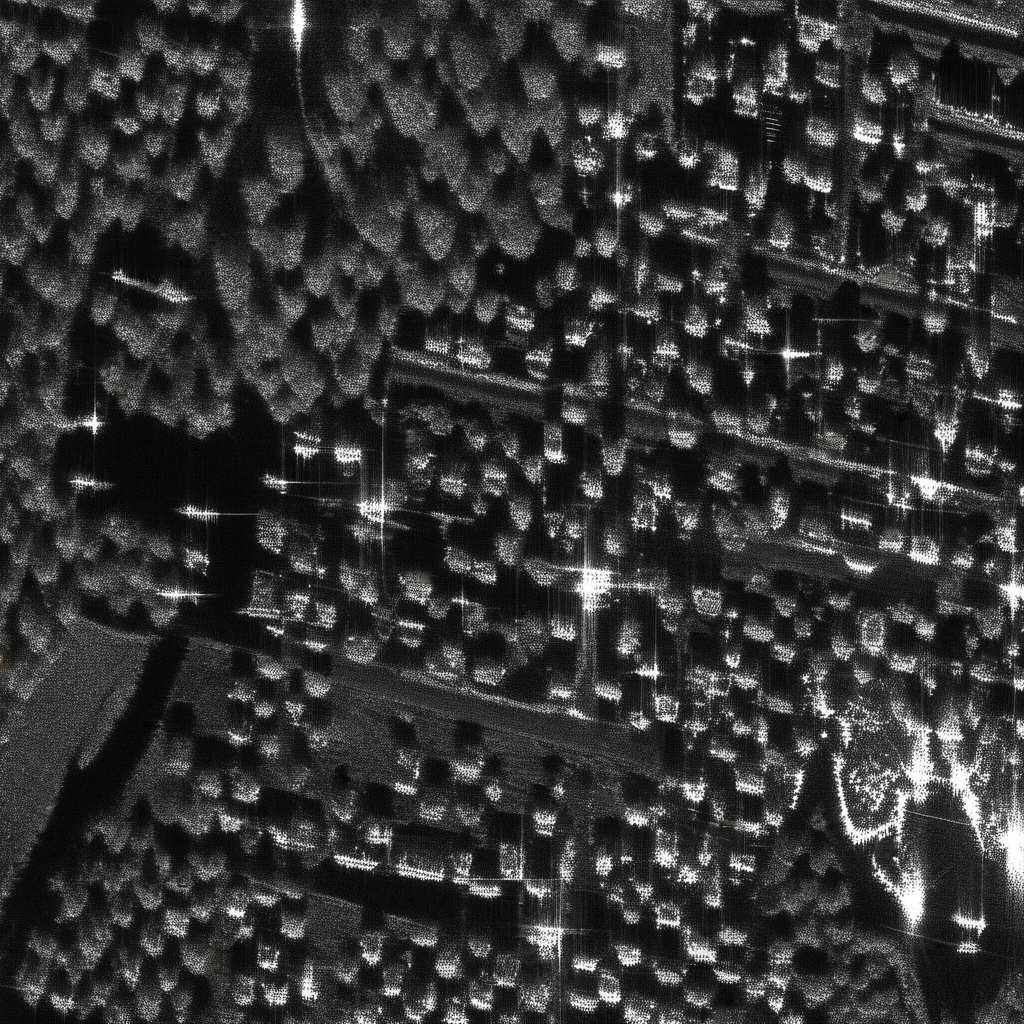} \\

        \rotatebox[origin=c]{90}{\textbf{eau-vie-4}} &
        \includegraphics[width=\linewidth]{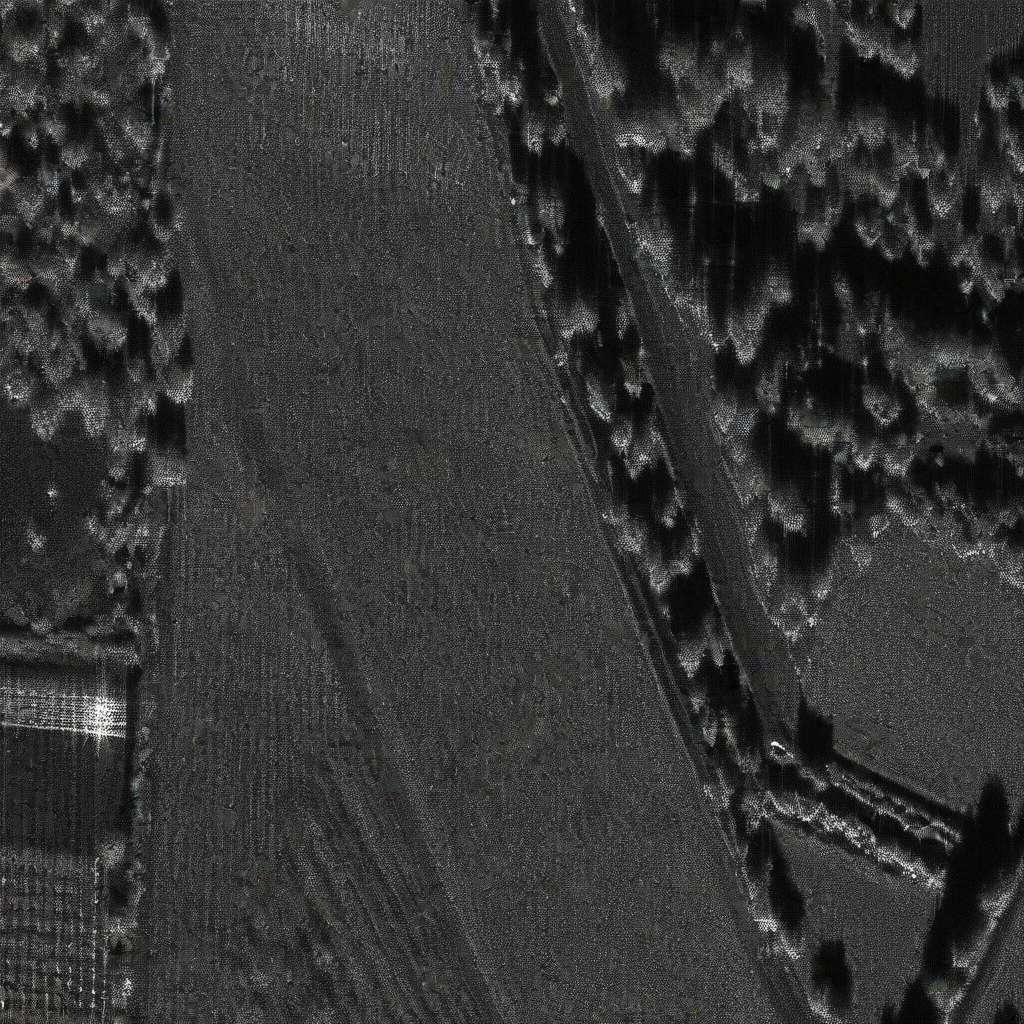} &
        \includegraphics[width=\linewidth]{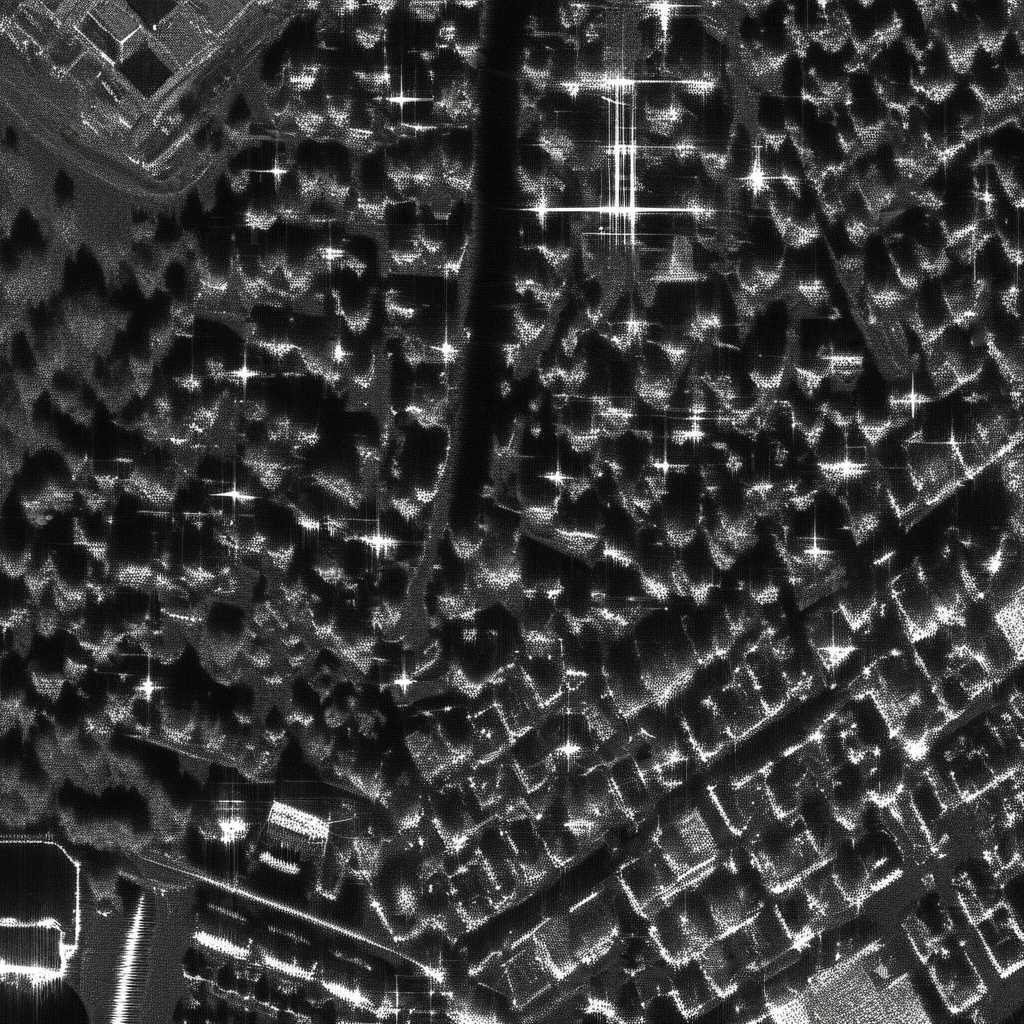} &
        \includegraphics[width=\linewidth]{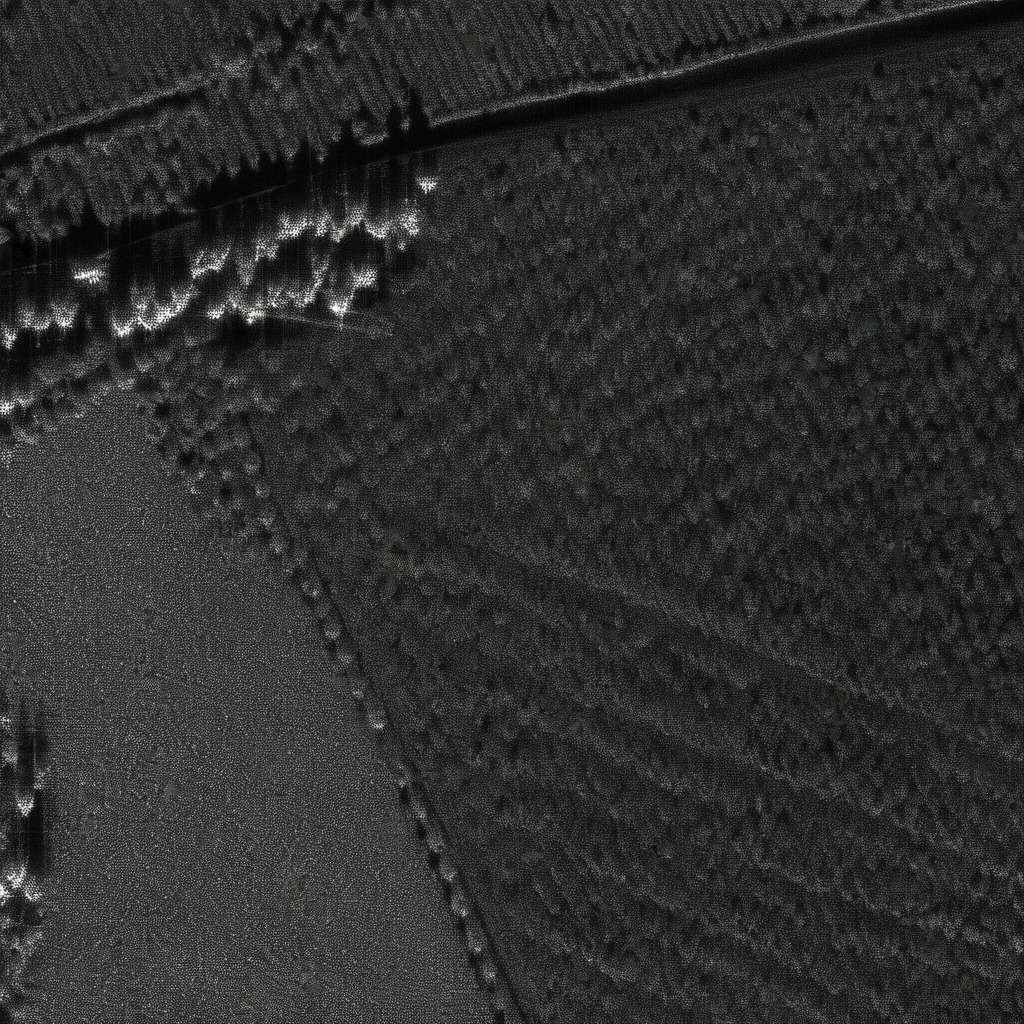} &
        \includegraphics[width=\linewidth]{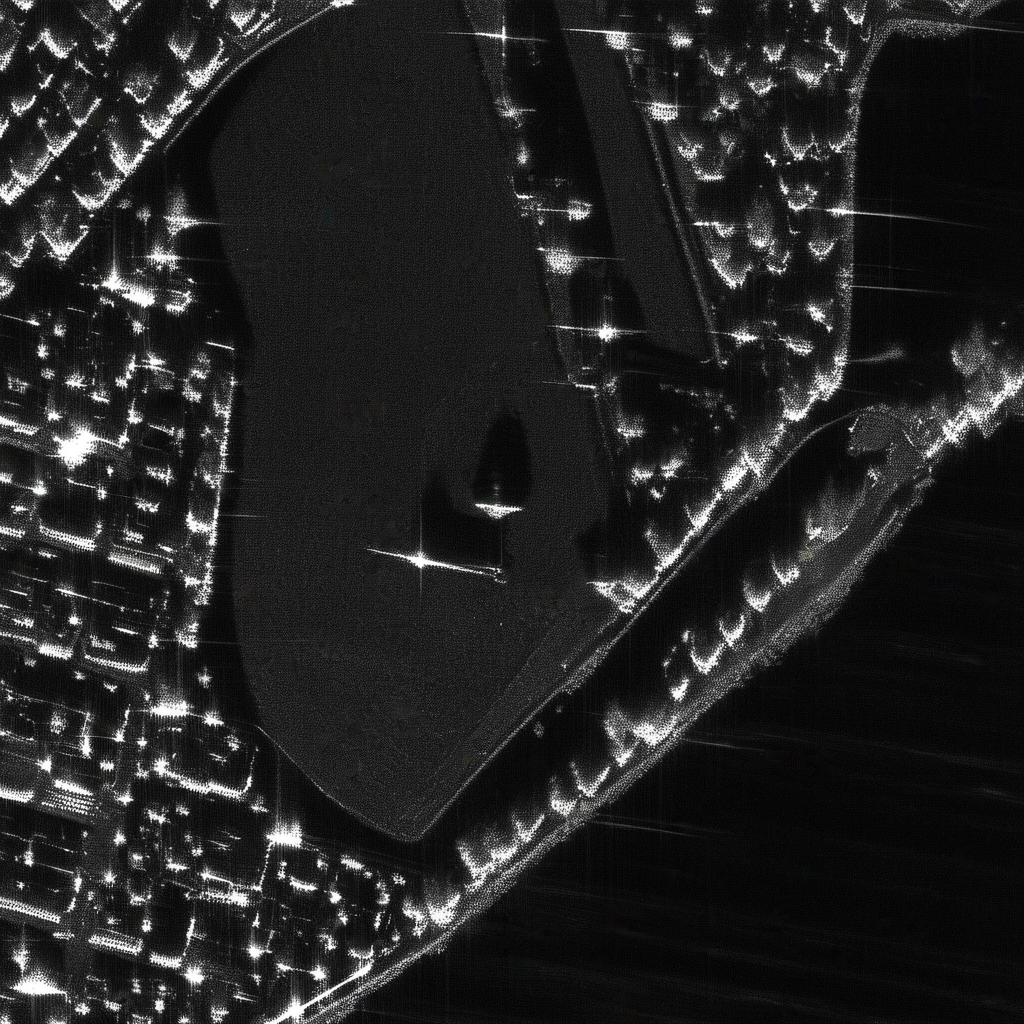} &
        \includegraphics[width=\linewidth]{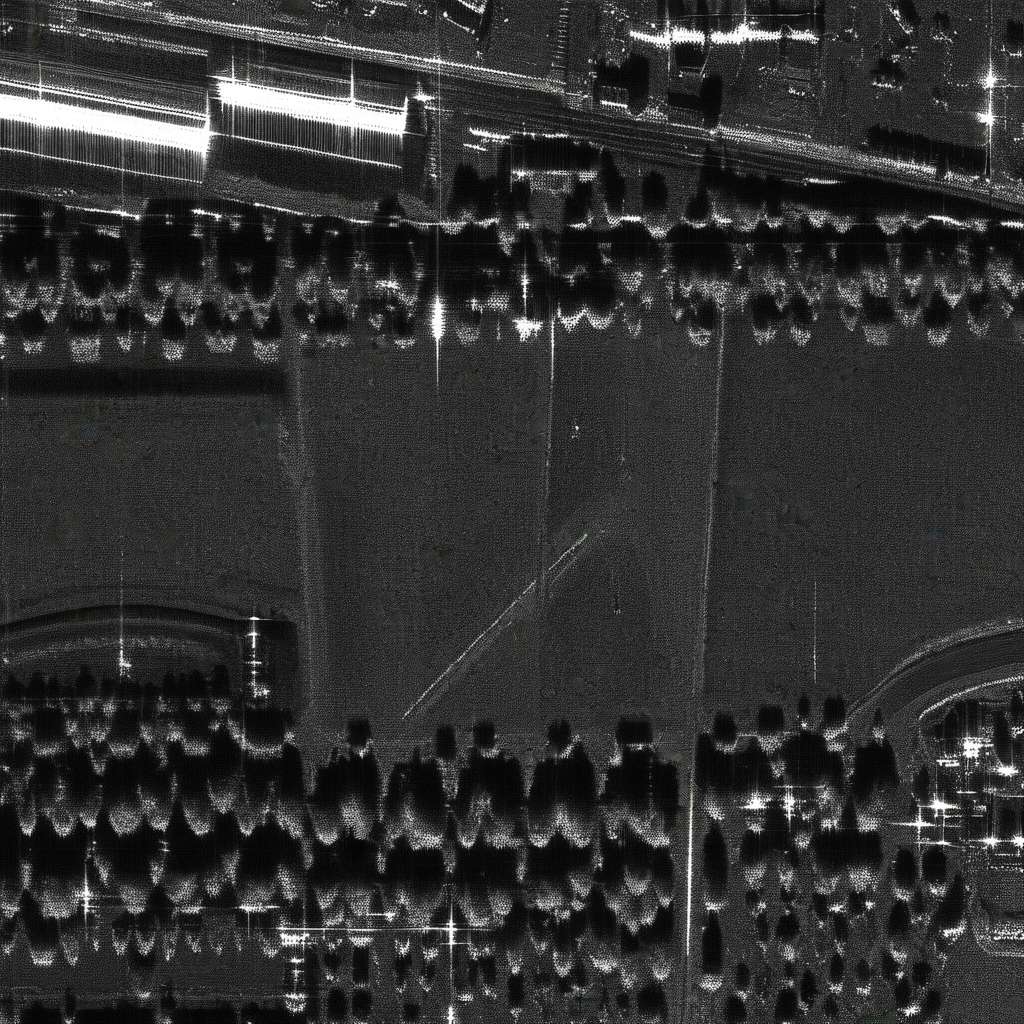} &
        \includegraphics[width=\linewidth]{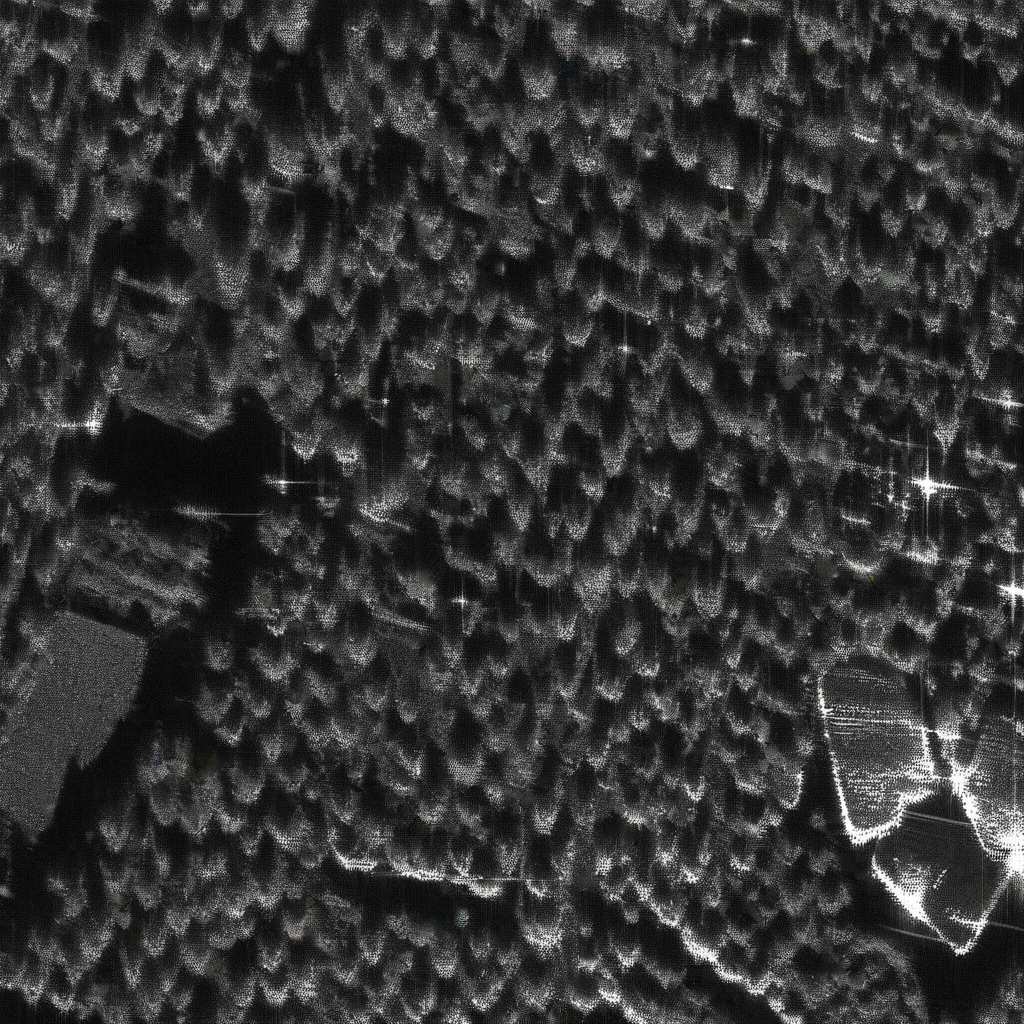} \\

        \rotatebox[origin=c]{90}{\textbf{king-kong-9}} &
        \includegraphics[width=\linewidth]{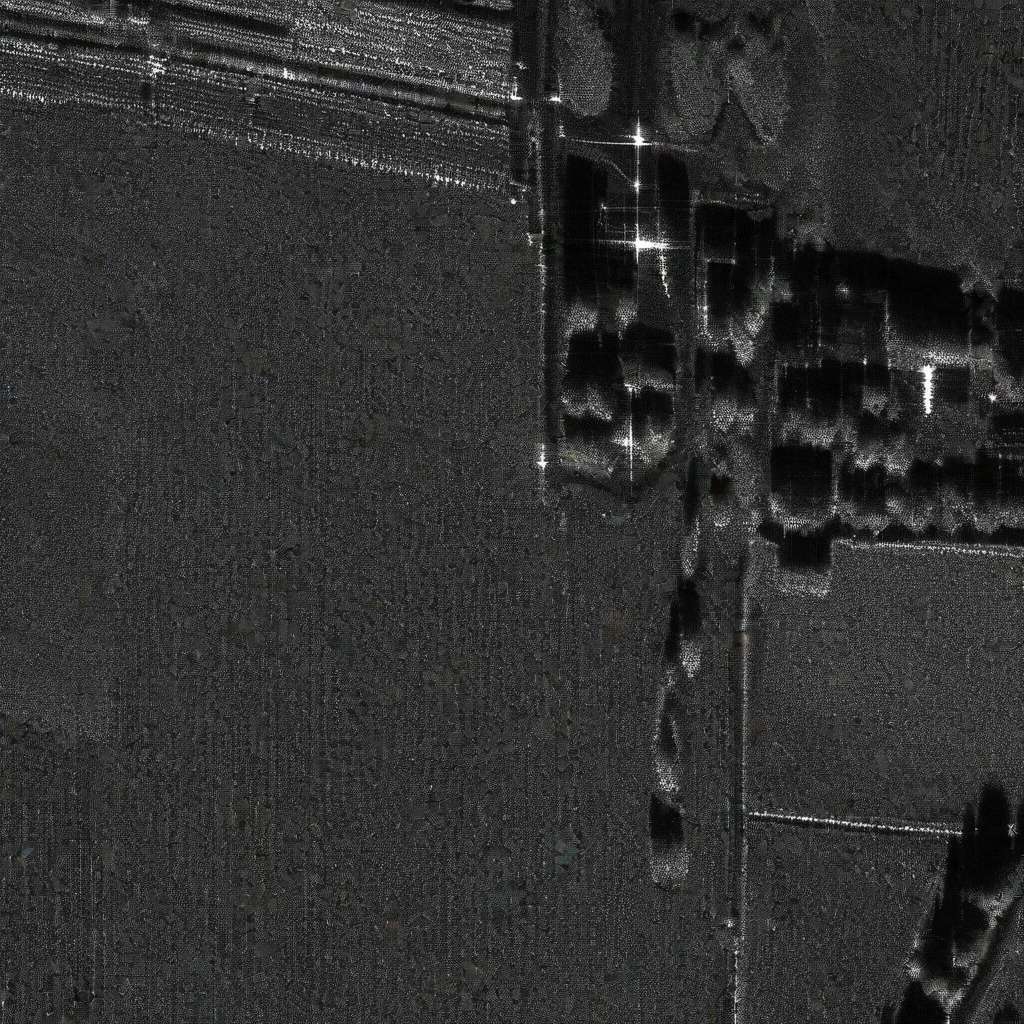} &
        \includegraphics[width=\linewidth]{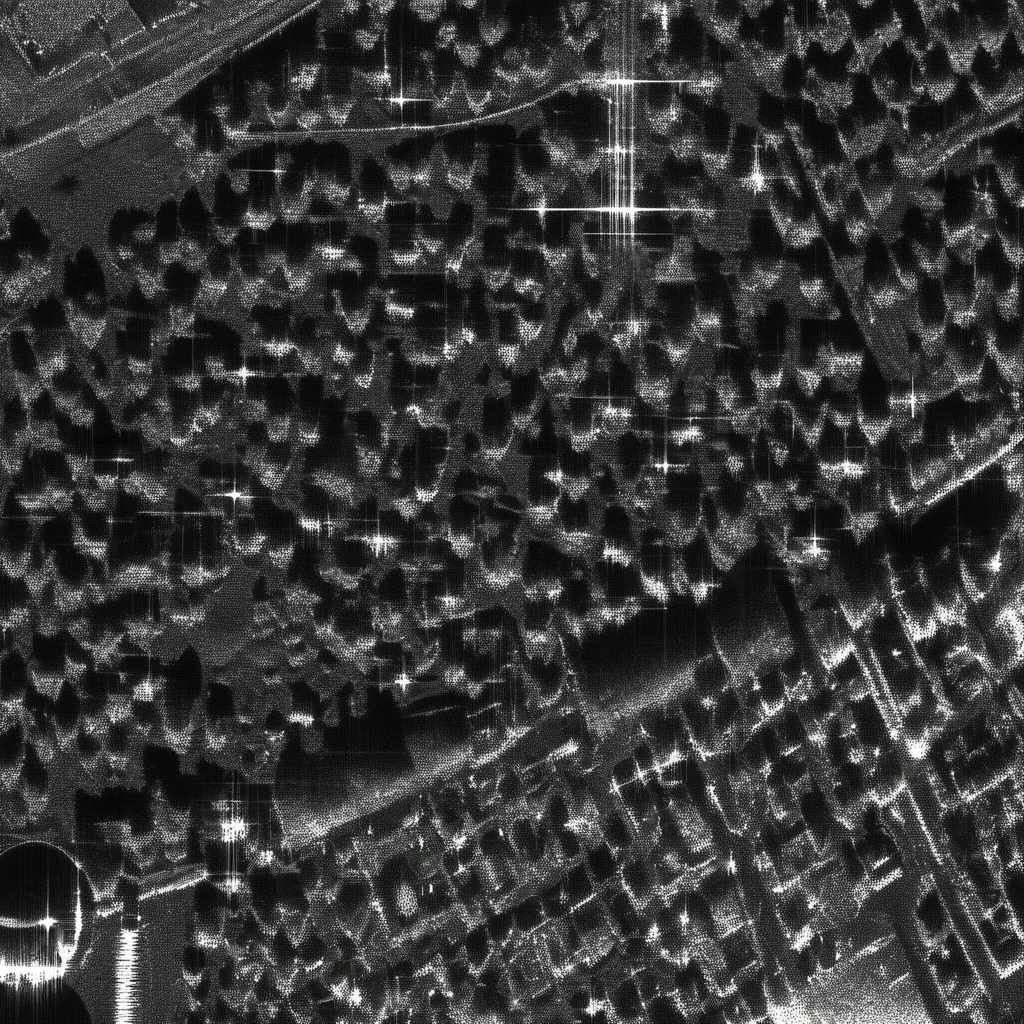} &
        \includegraphics[width=\linewidth]{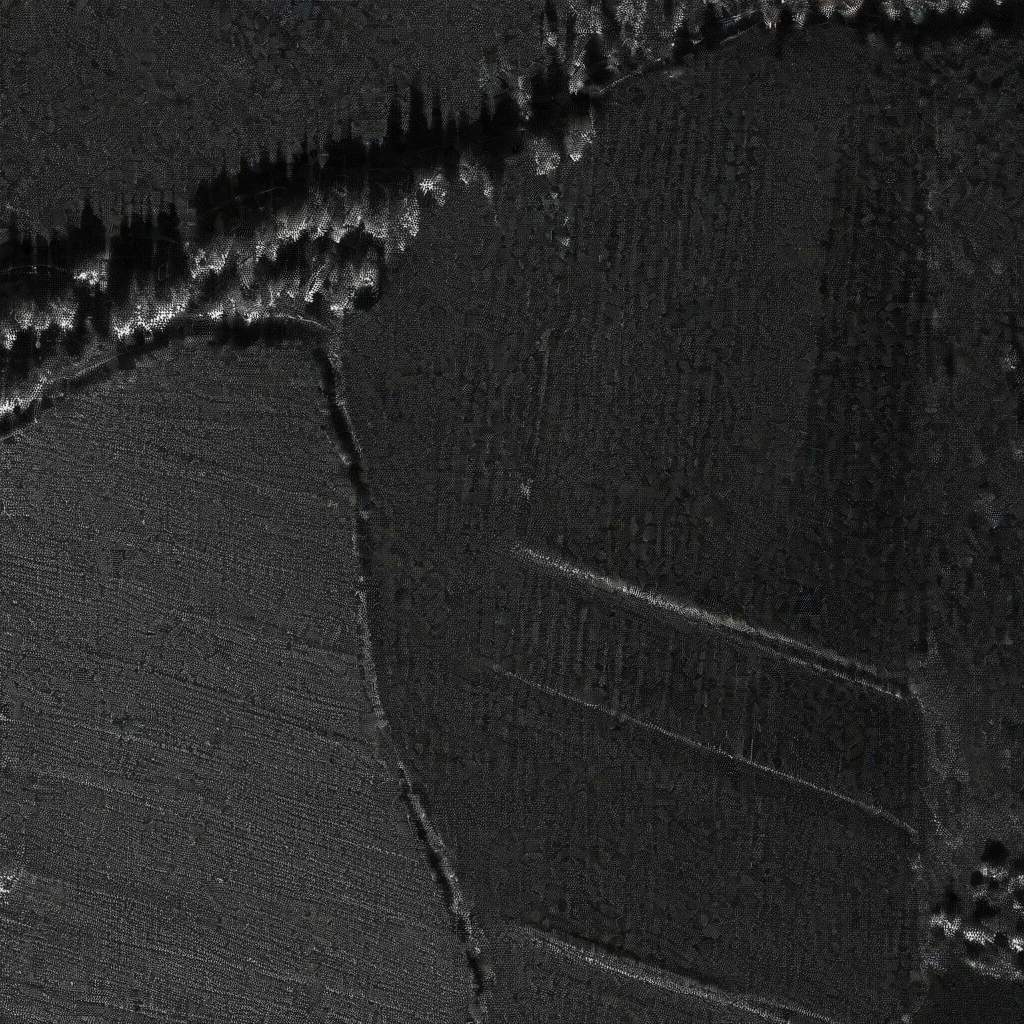} &
        \includegraphics[width=\linewidth]{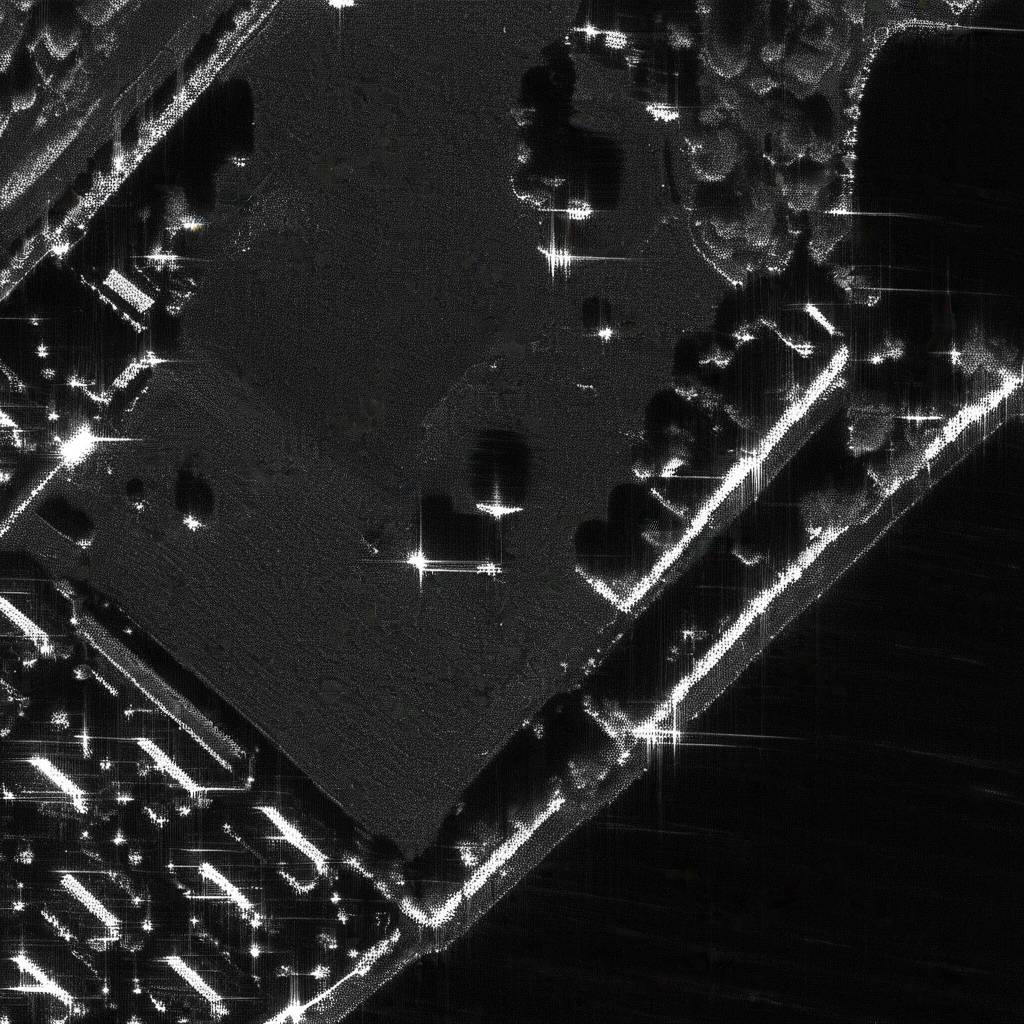} &
        \includegraphics[width=\linewidth]{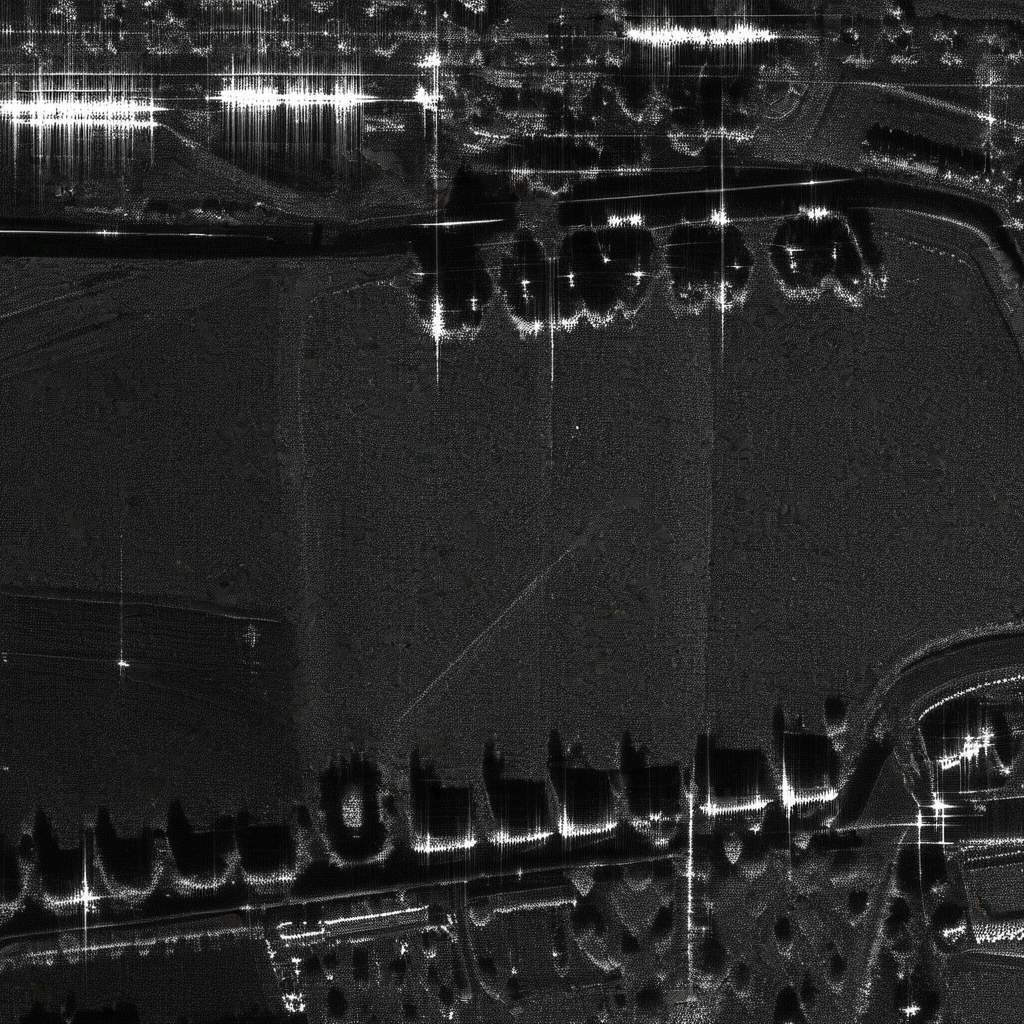} &
        \includegraphics[width=\linewidth]{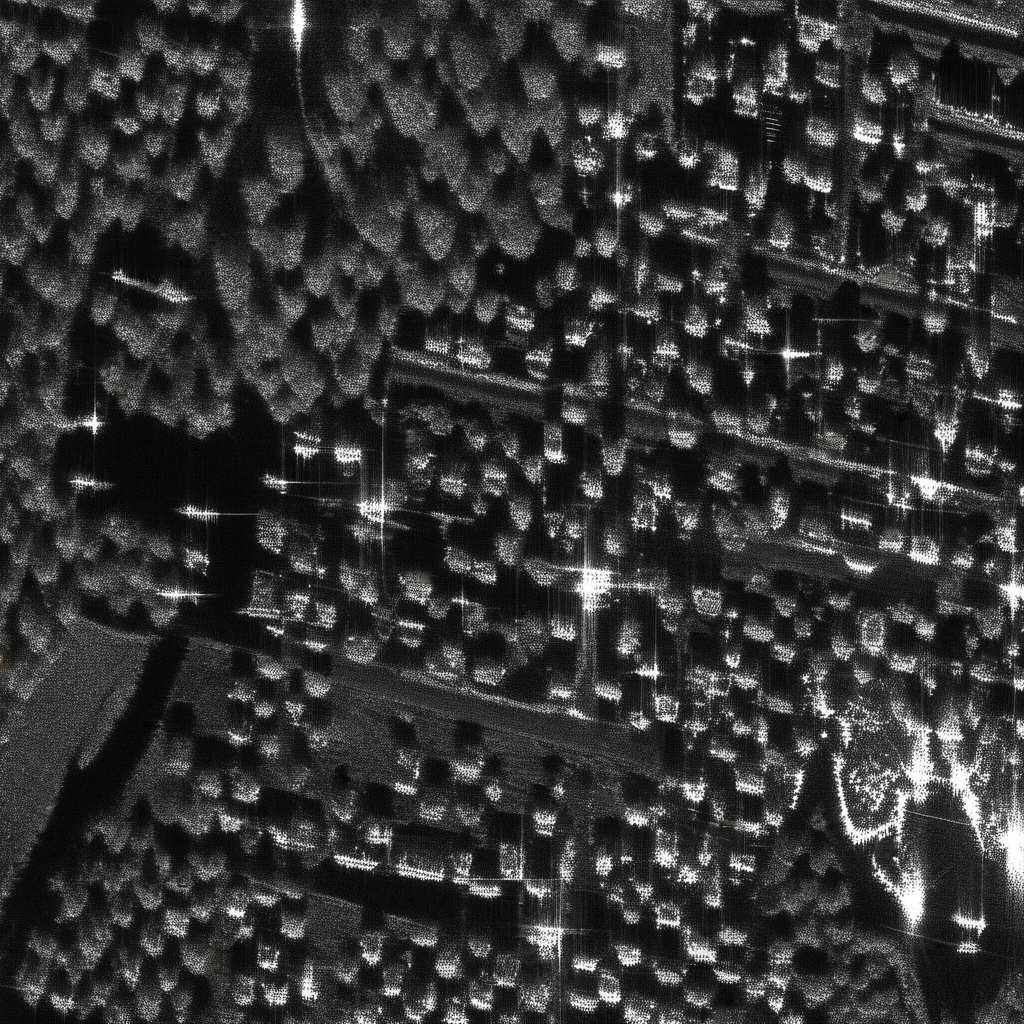} \\

        \rotatebox[origin=c]{90}{\textbf{mummy-pen-8}} &
        \includegraphics[width=\linewidth]{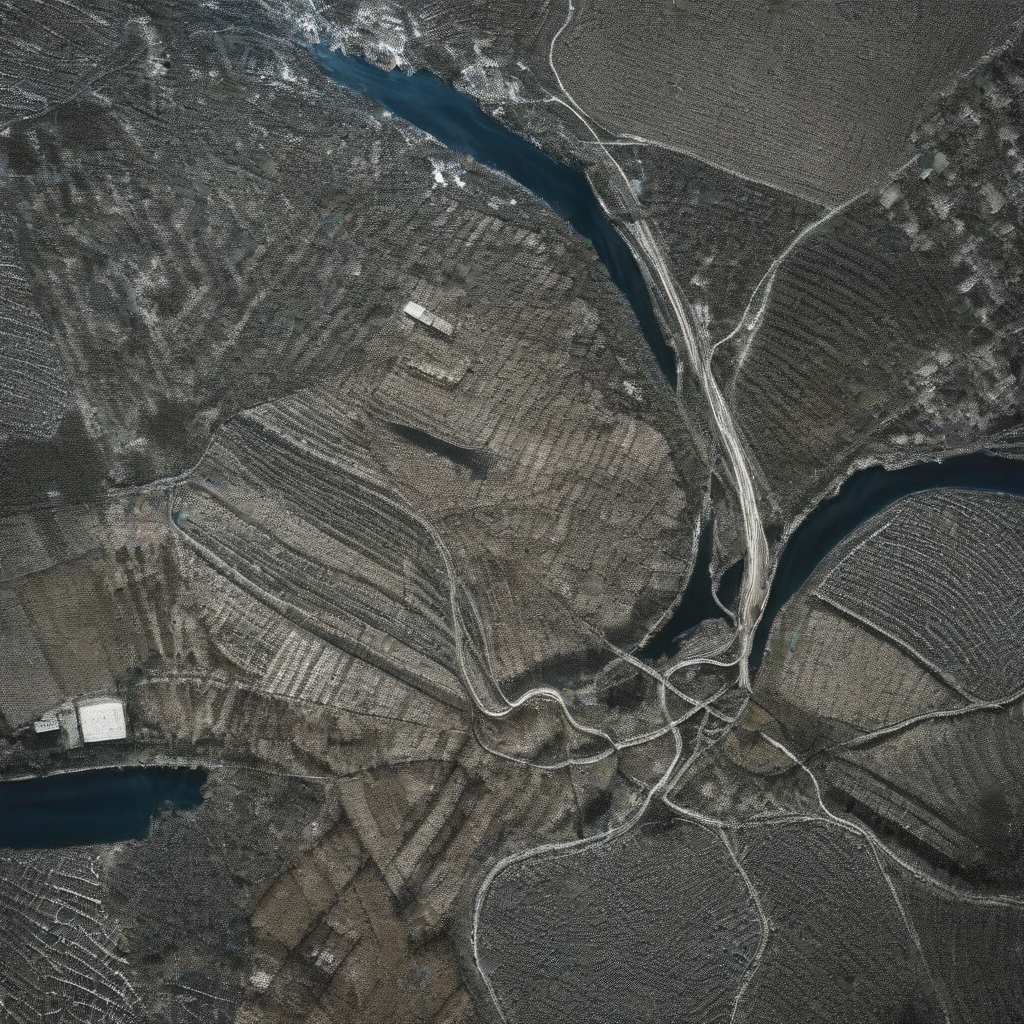} &
        \includegraphics[width=\linewidth]{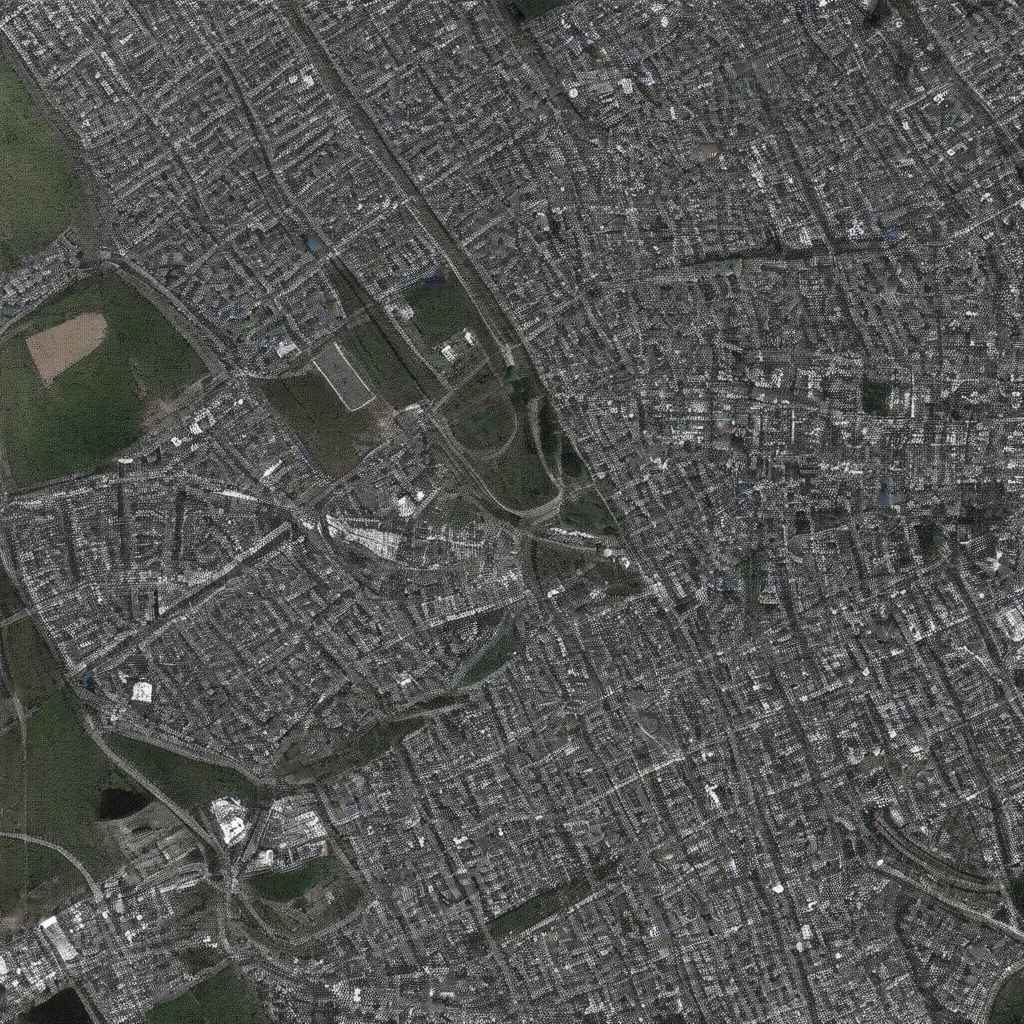} &
        \includegraphics[width=\linewidth]{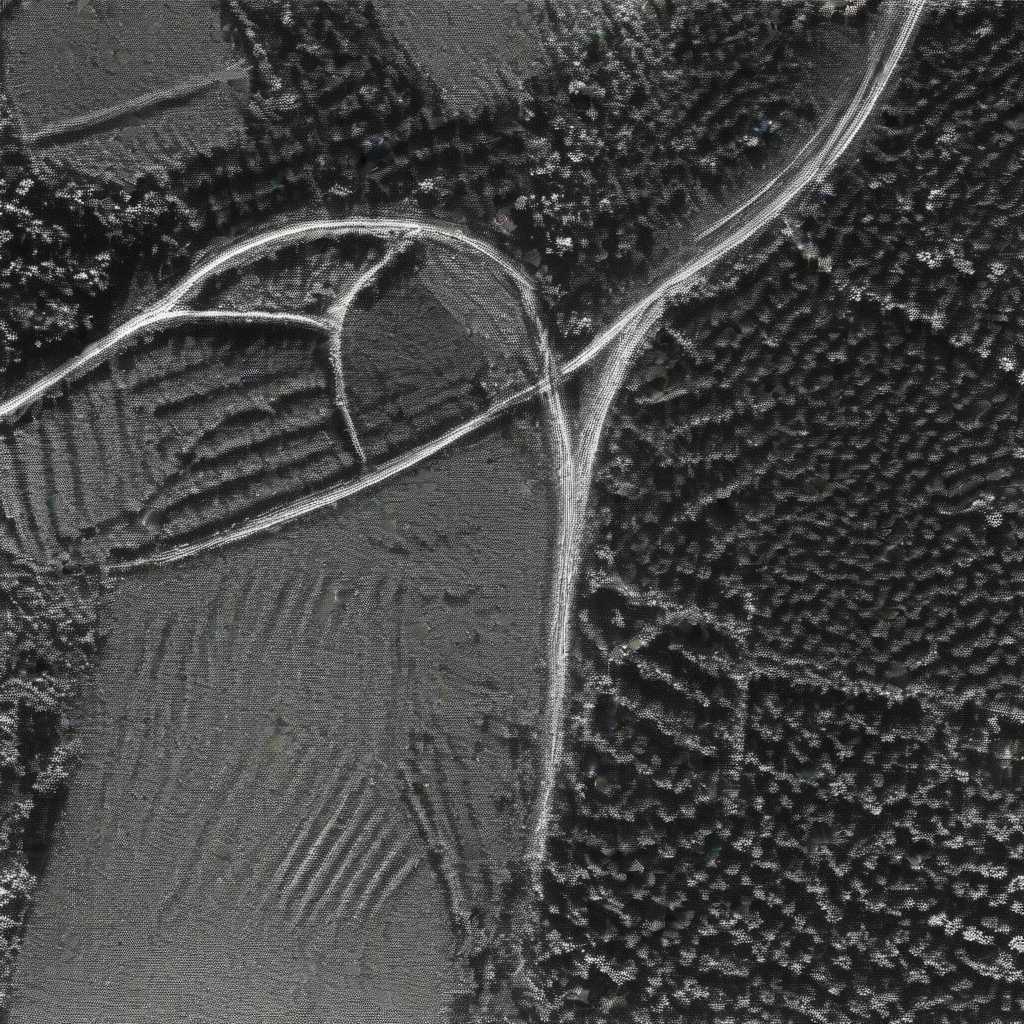} &
        \includegraphics[width=\linewidth]{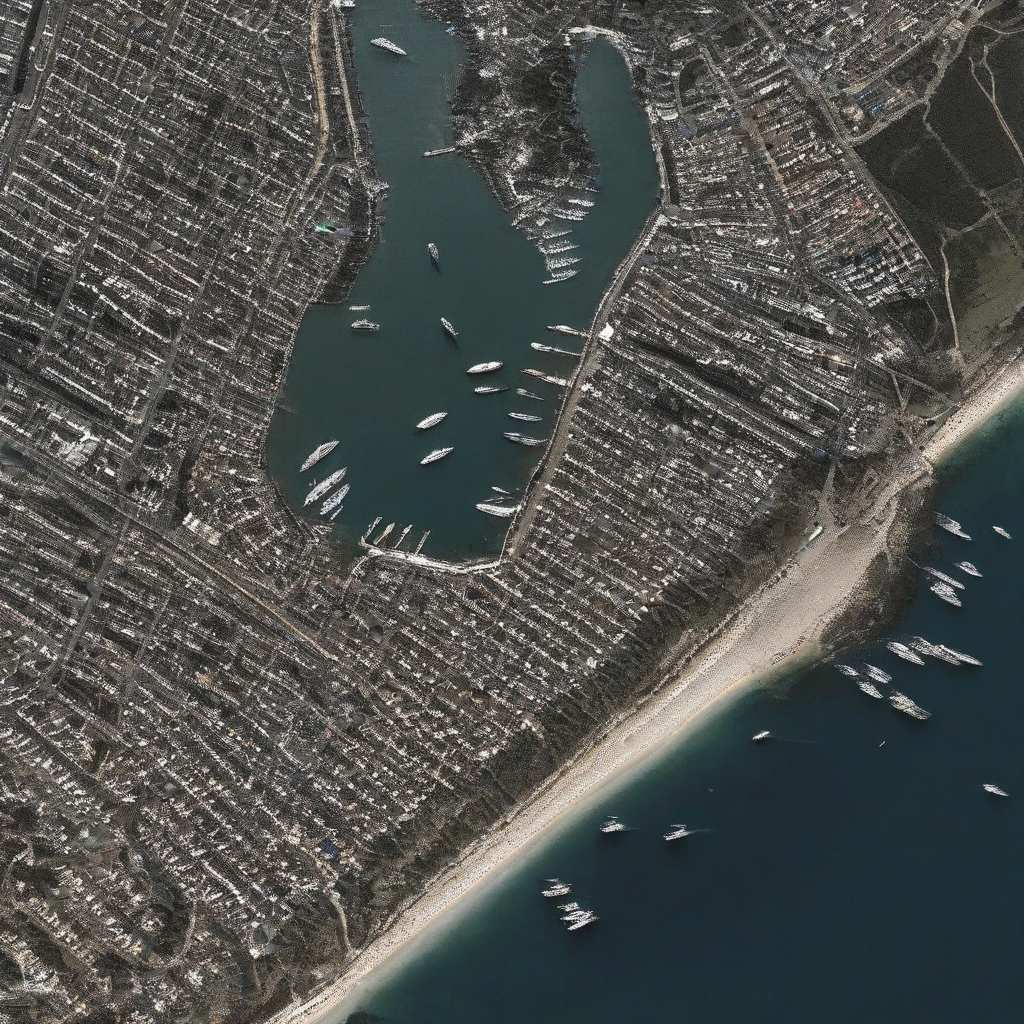} &
        \includegraphics[width=\linewidth]{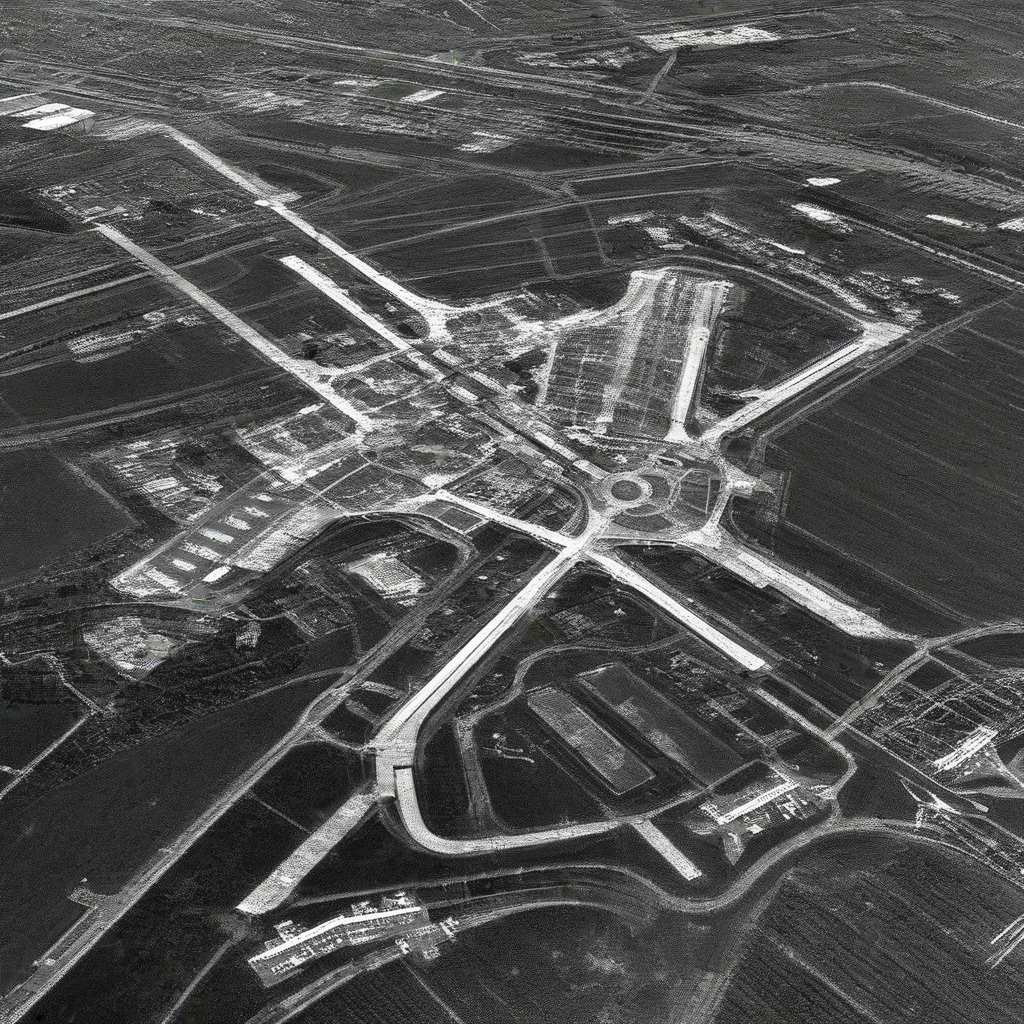} &
        \includegraphics[width=\linewidth]{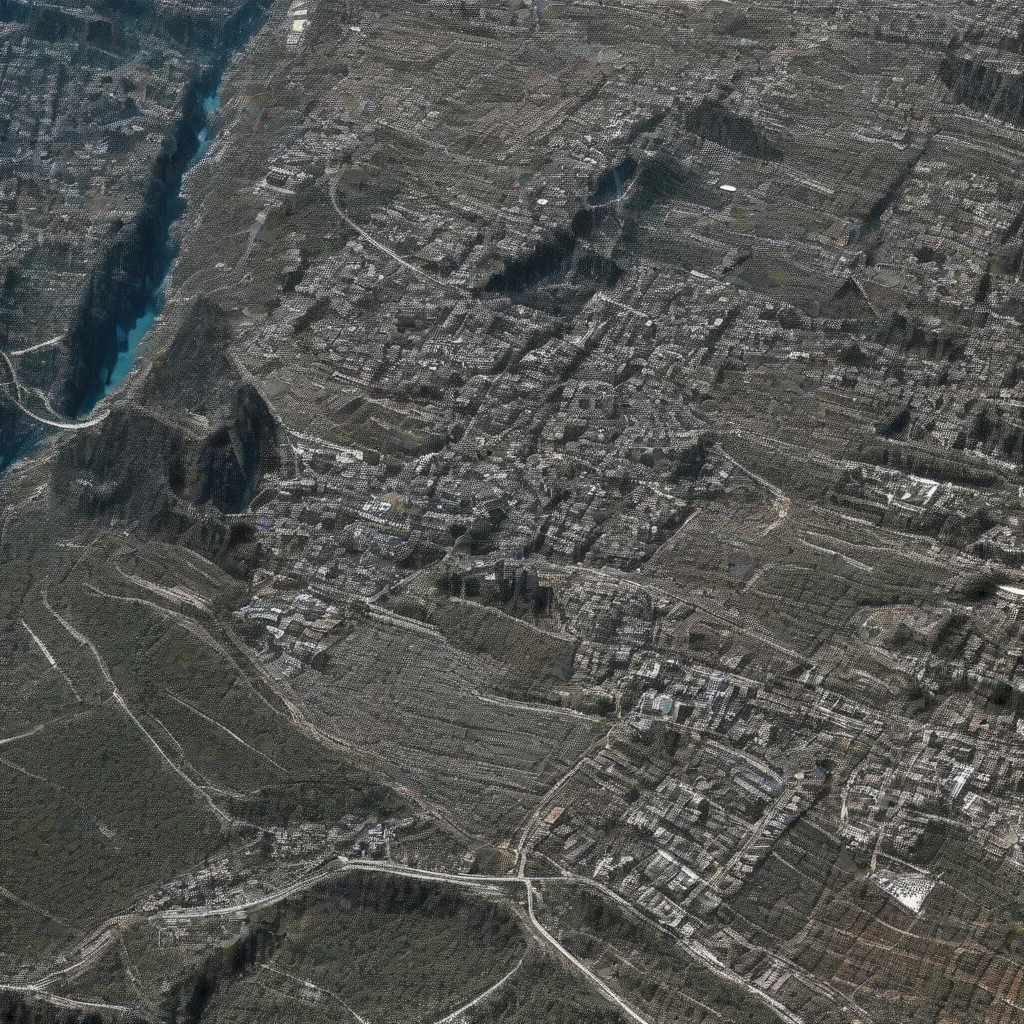} \\

        \rotatebox[origin=c]{90}{\textbf{smile-road-5}} &
        \includegraphics[width=\linewidth]{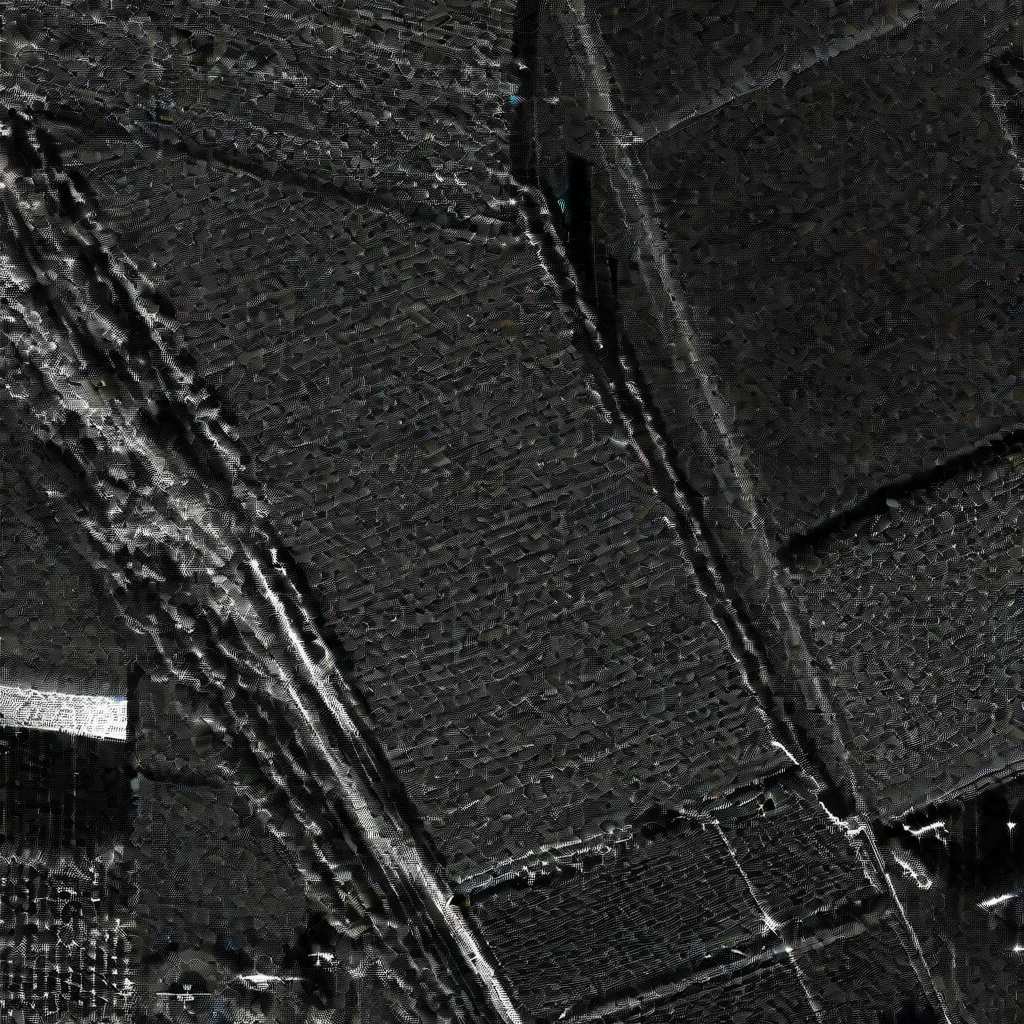} &
        \includegraphics[width=\linewidth]{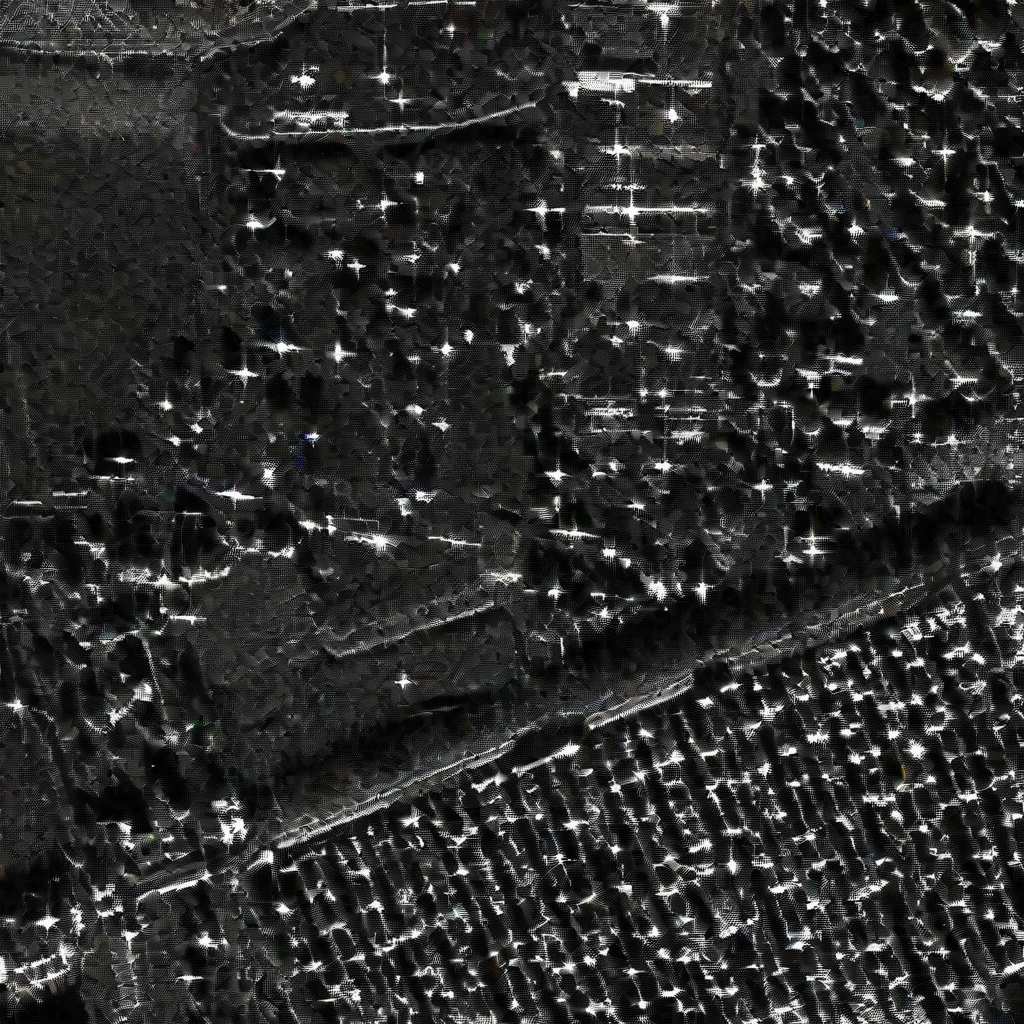} &
        \includegraphics[width=\linewidth]{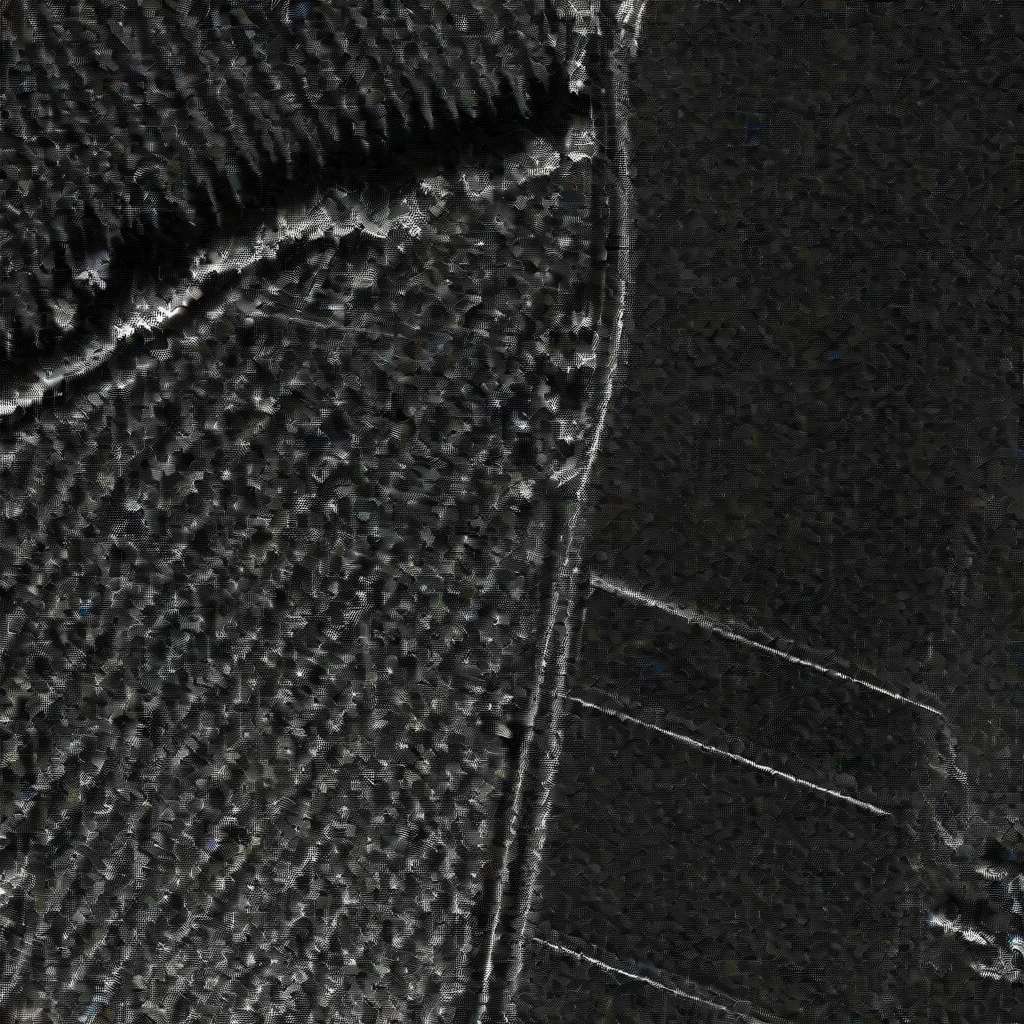} &
        \includegraphics[width=\linewidth]{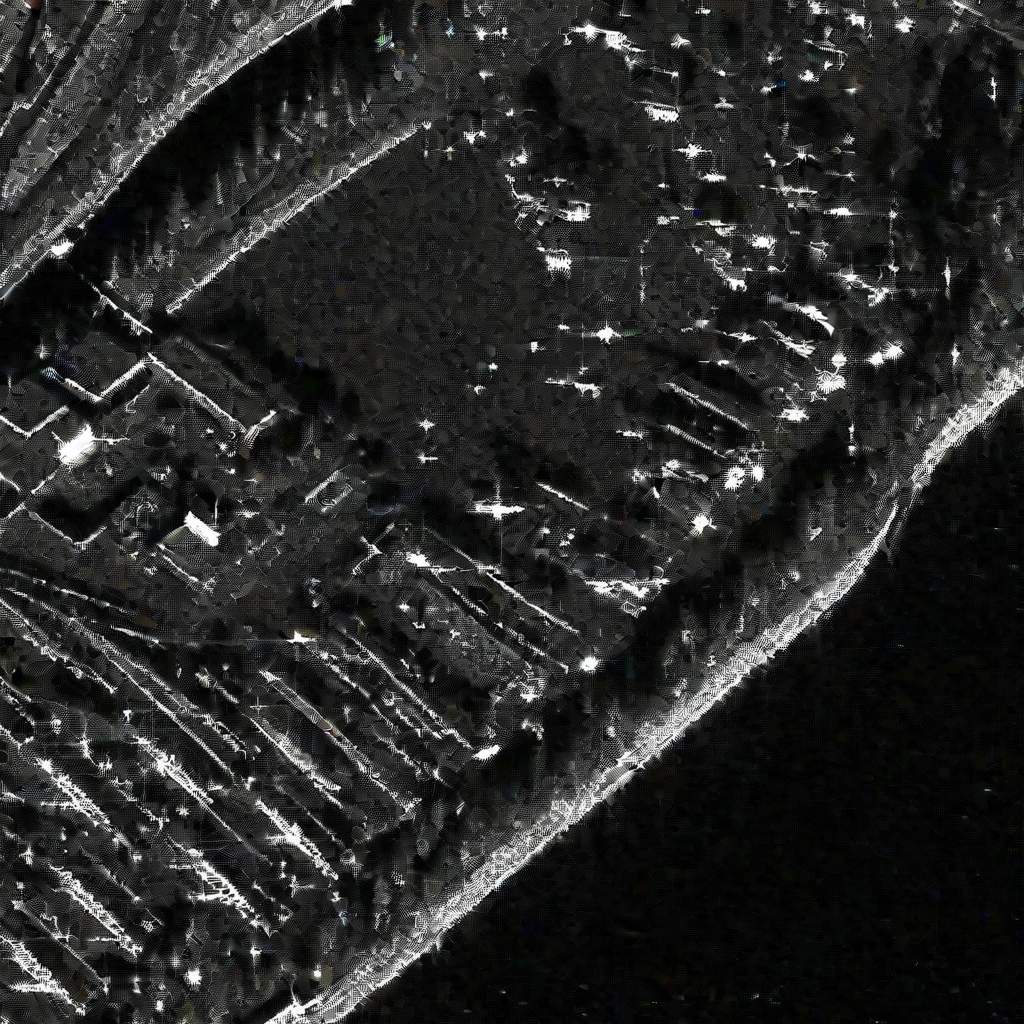} &
        \includegraphics[width=\linewidth]{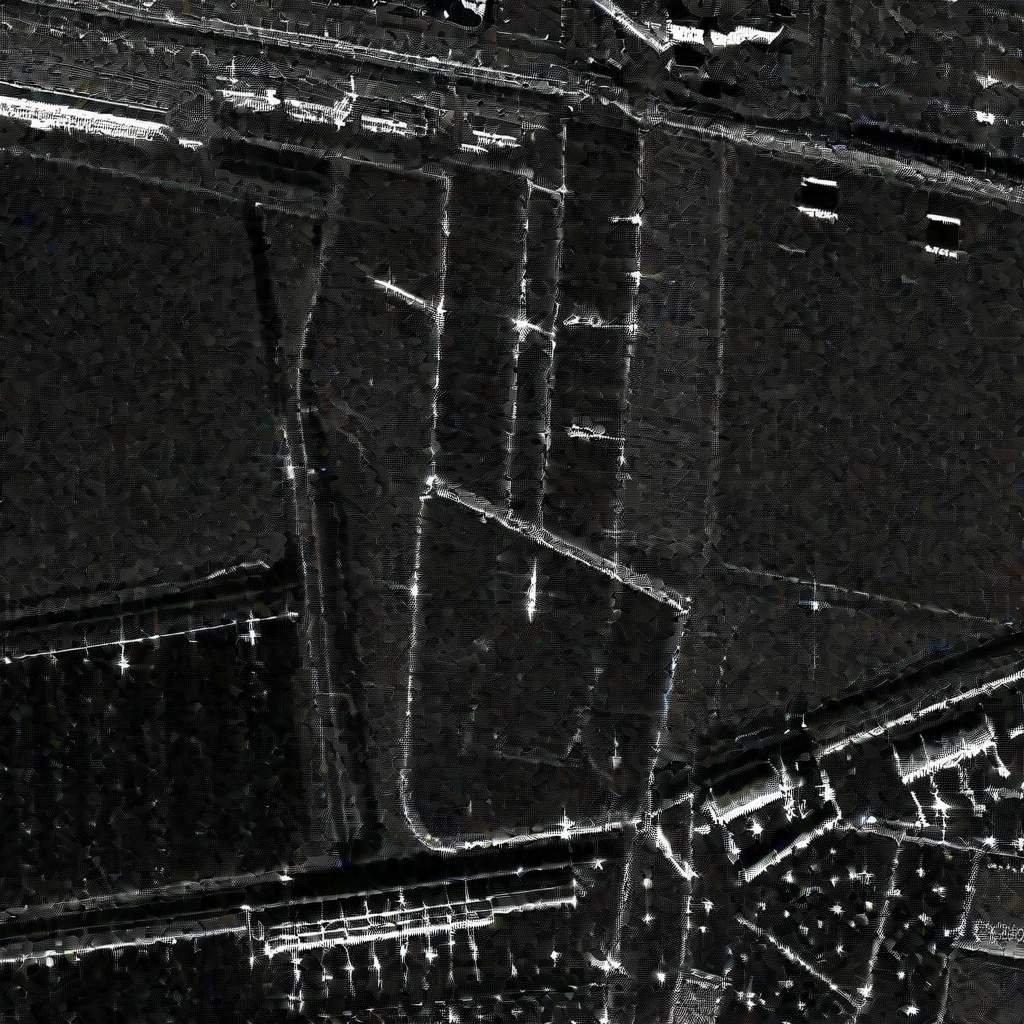} &
        \includegraphics[width=\linewidth]{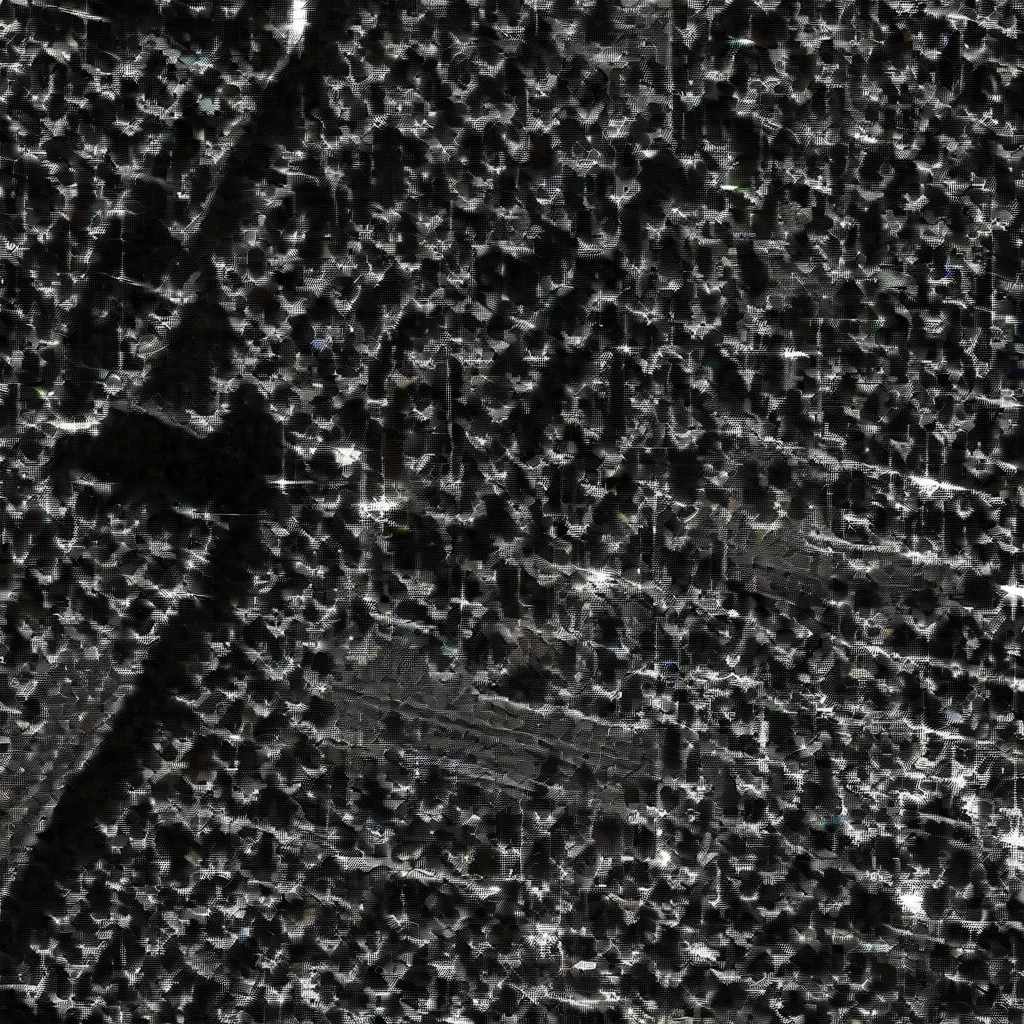} \\

        \rotatebox[origin=c]{90}{\textbf{super-bowl-2}} &
        \includegraphics[width=\linewidth]{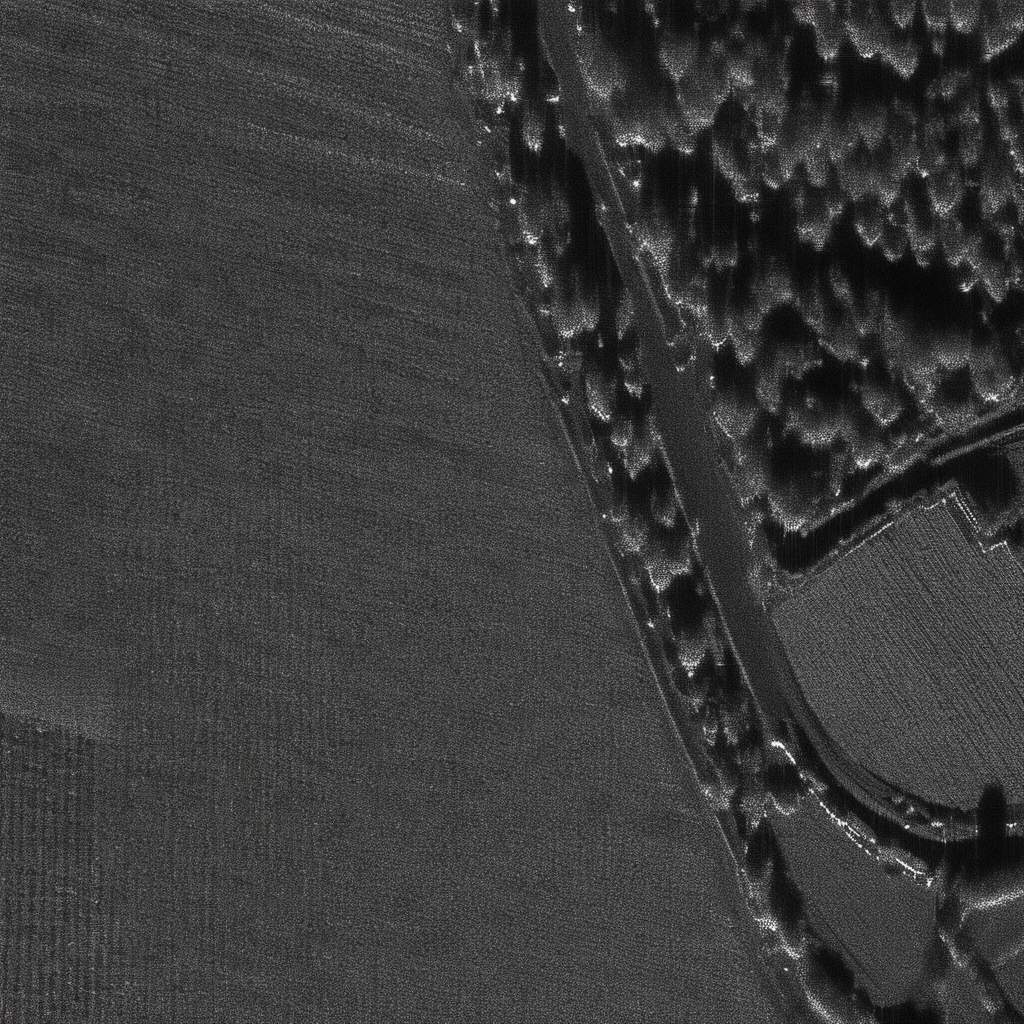} &
        \includegraphics[width=\linewidth]{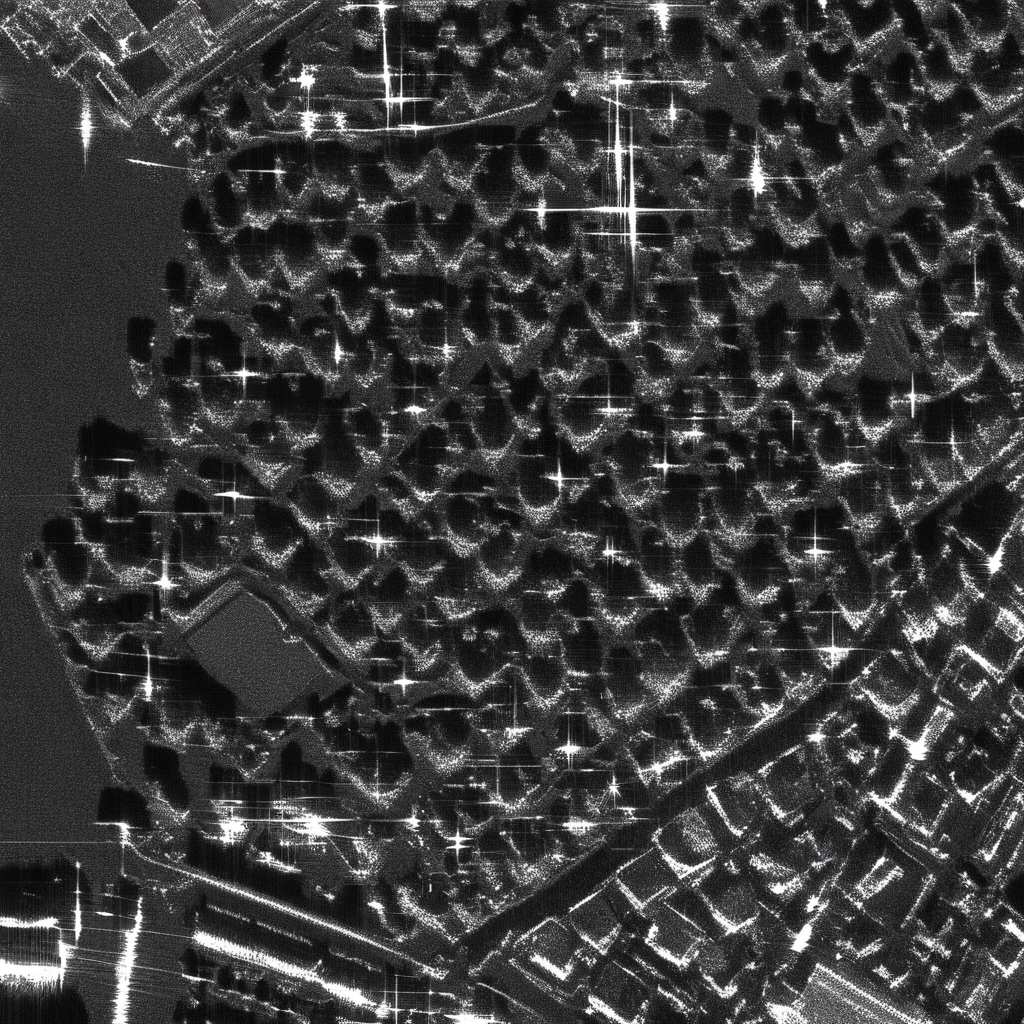} &
        \includegraphics[width=\linewidth]{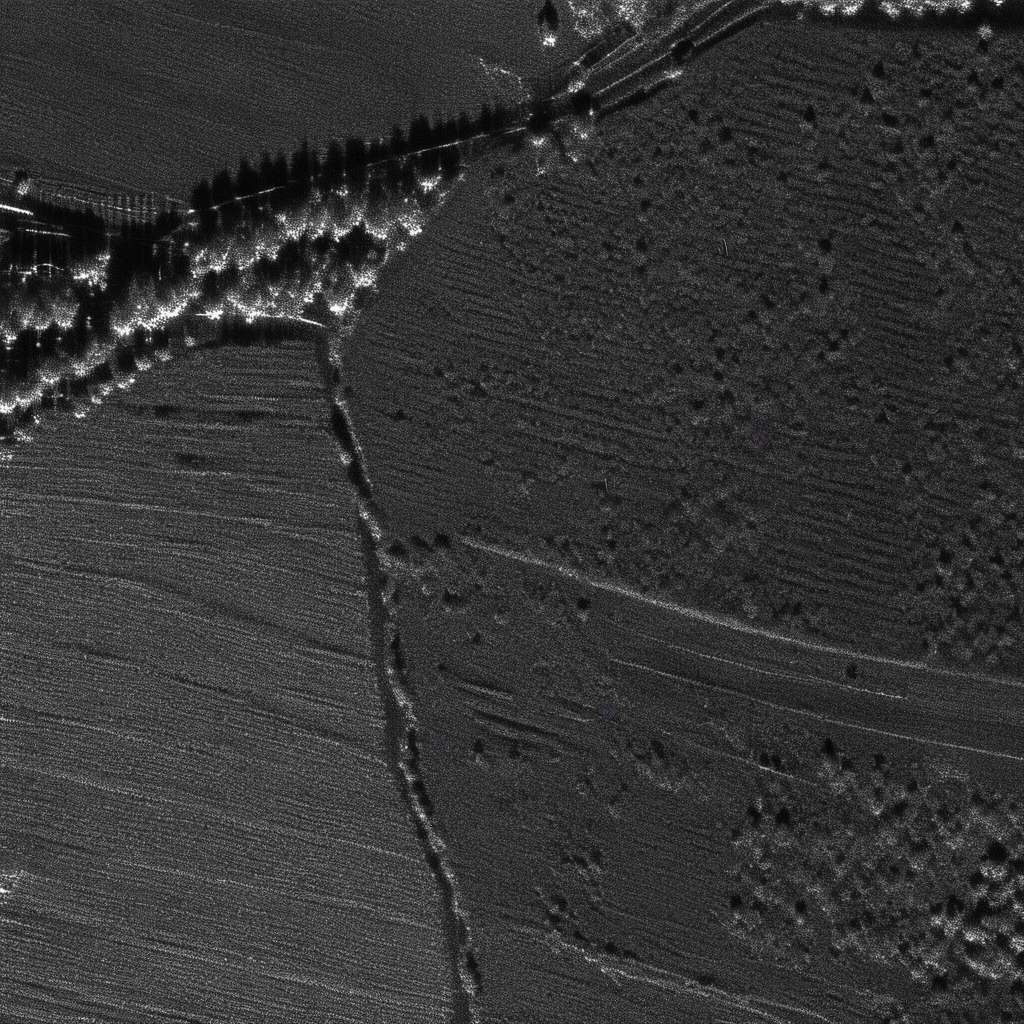} &
        \includegraphics[width=\linewidth]{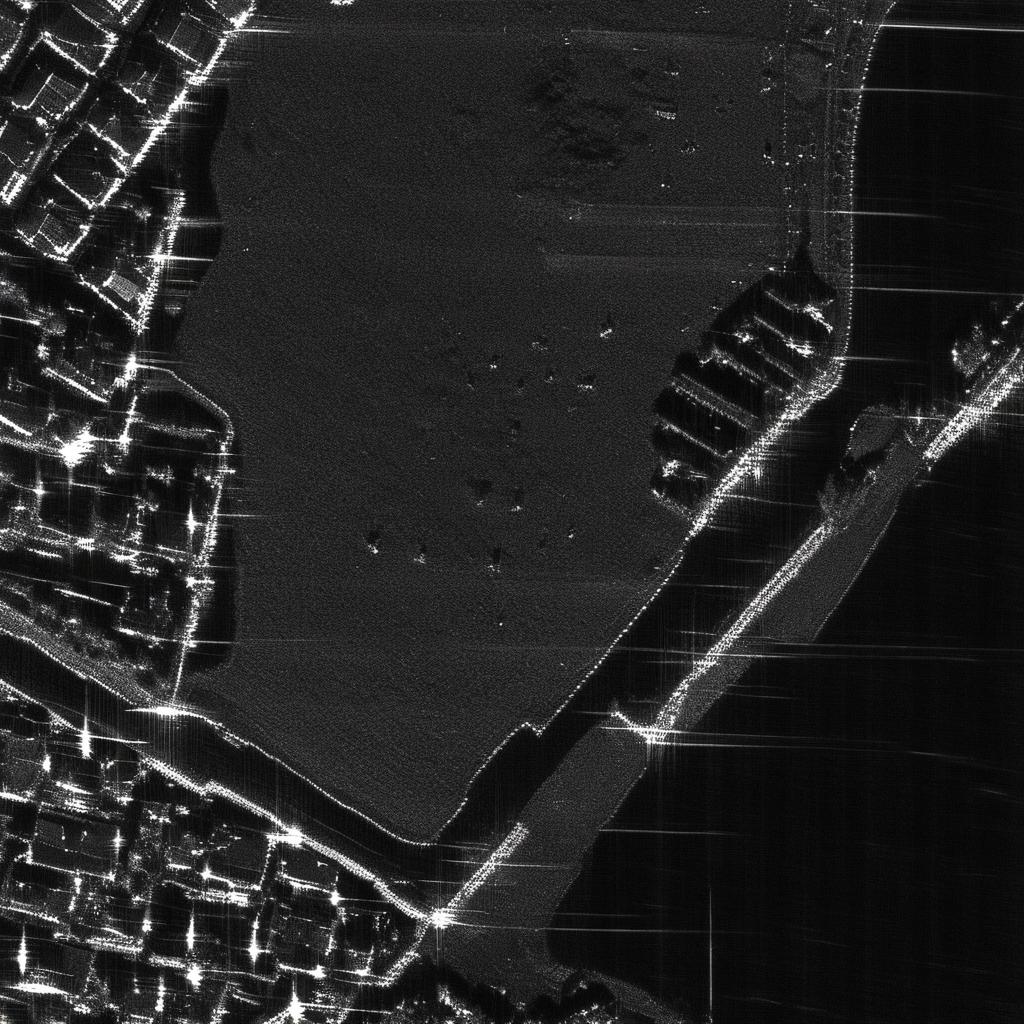} &
        \includegraphics[width=\linewidth]{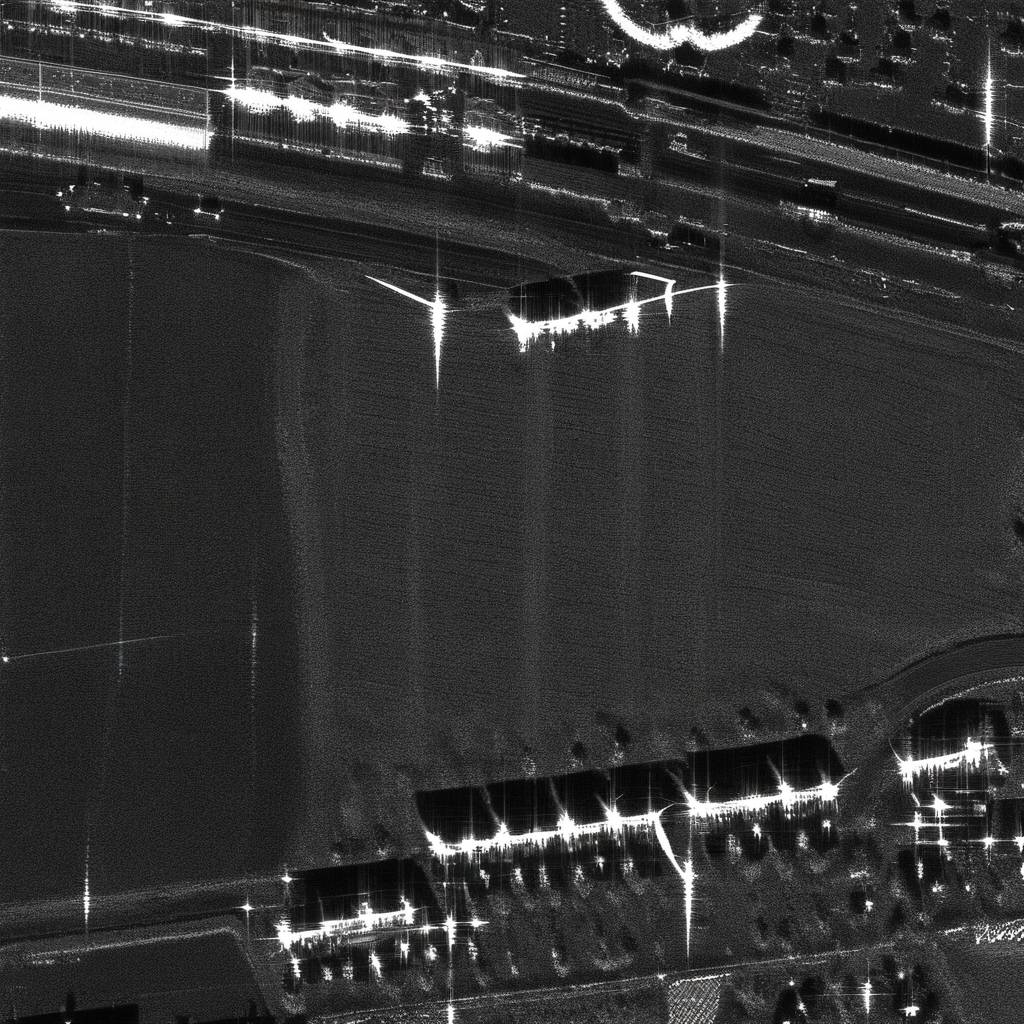} &
        \includegraphics[width=\linewidth]{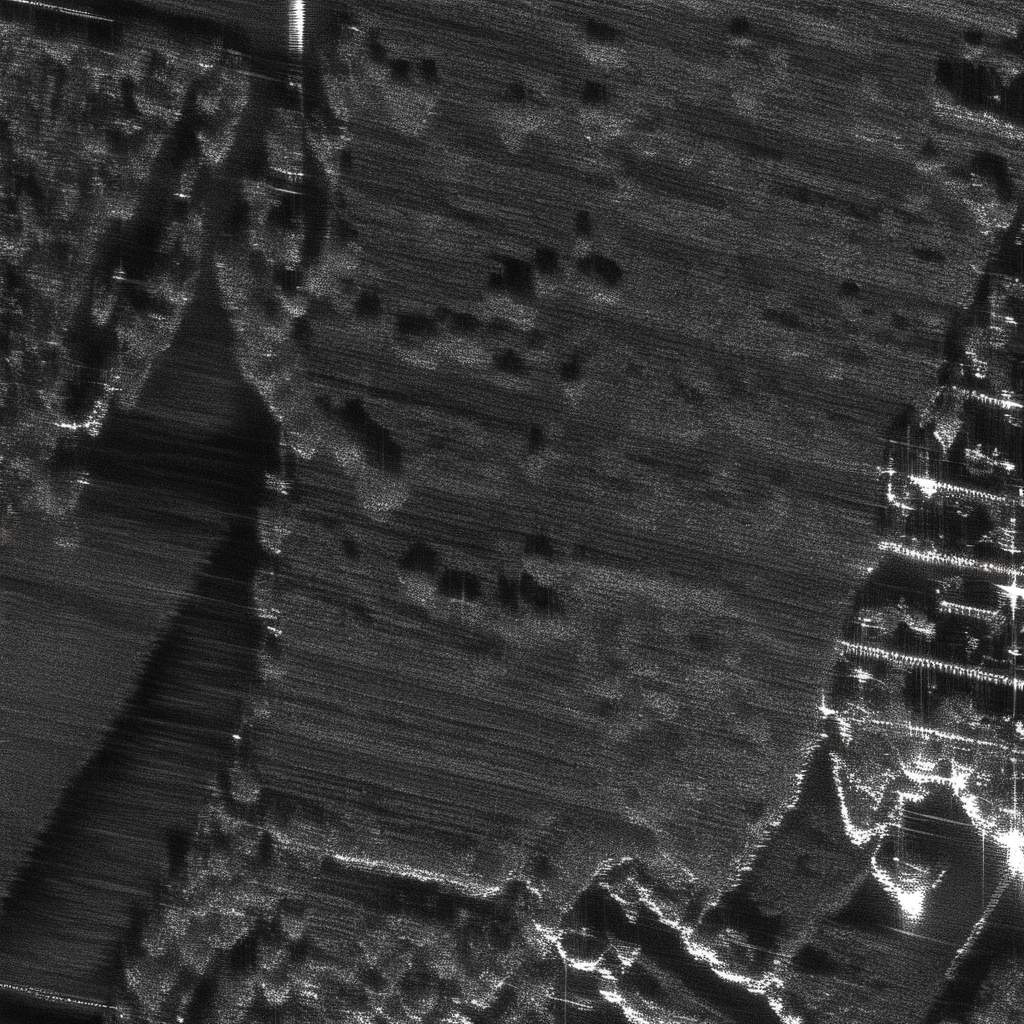} \\
    \end{tabular}
    \end{adjustbox}
    \caption{Study on the UNet, TE1 and TE2: Generated images (1024x1024px at 40cm) per category for 9 different models at epoch 8 (same seed training and evaluation)}
    \label{fig:subplot_models_full}
\end{figure*}

\begin{figure*}
    \centering
    \scriptsize
    \setlength{\tabcolsep}{1.3pt}
    \renewcommand{\arraystretch}{1.3}
    \begin{adjustbox}{max width=0.97\textwidth}
    
    \begin{tabular}{c@{\hskip 4pt}*{6}{>{\centering\arraybackslash}m{0.14\textwidth}}}
        & \textbf{Field} & \textbf{City} & \textbf{Forest} & \textbf{Seacoast} & \textbf{Airport} & \textbf{Mountains} \\

        \rotatebox[origin=c]{90}{\textbf{whale-north-8}} &
        \includegraphics[width=\linewidth]{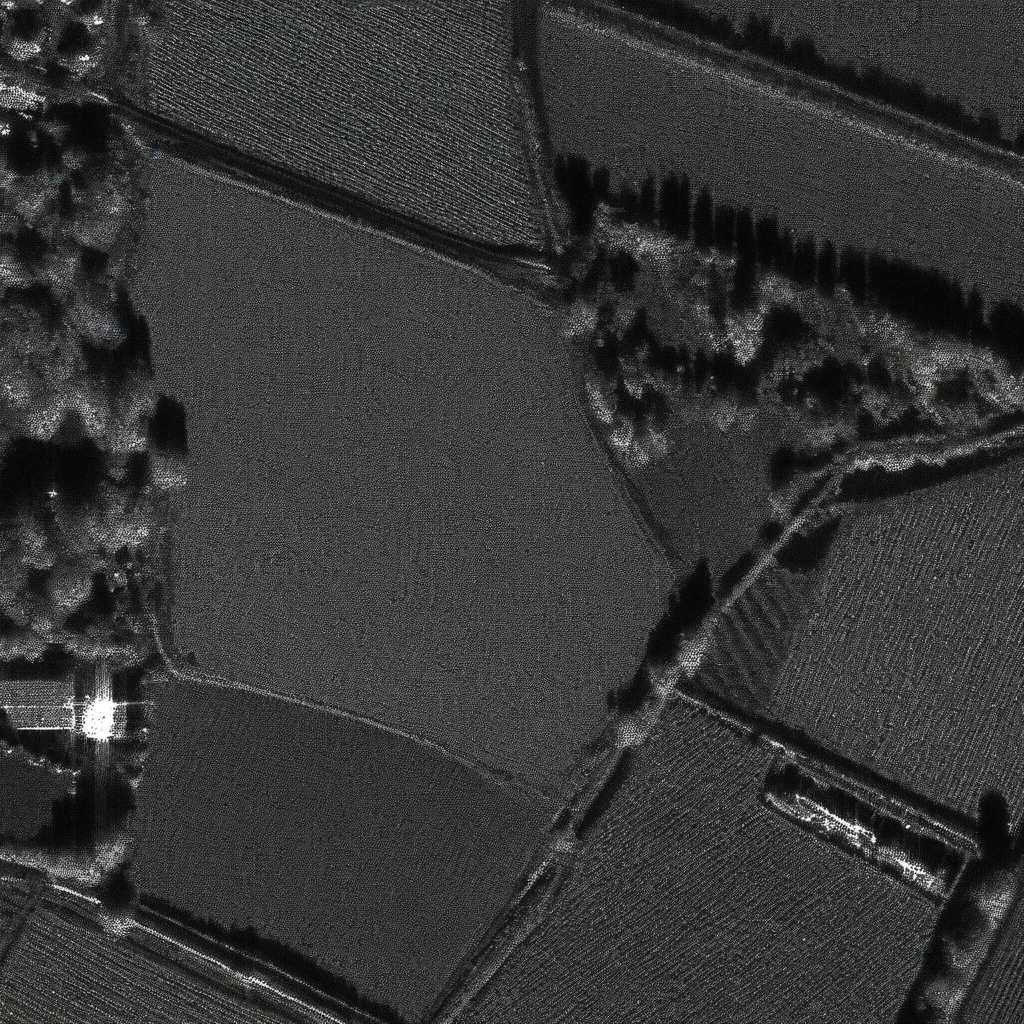} &
        \includegraphics[width=\linewidth]{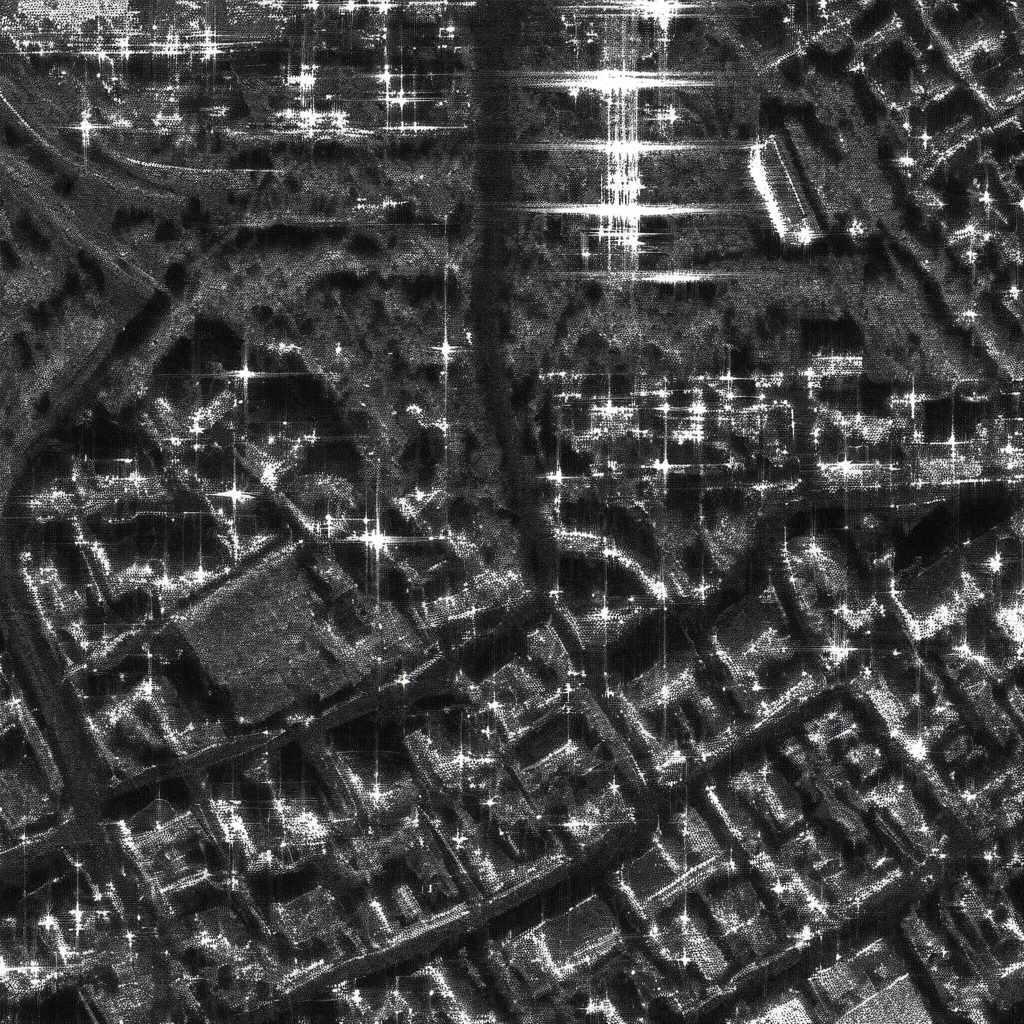} &
        \includegraphics[width=\linewidth]{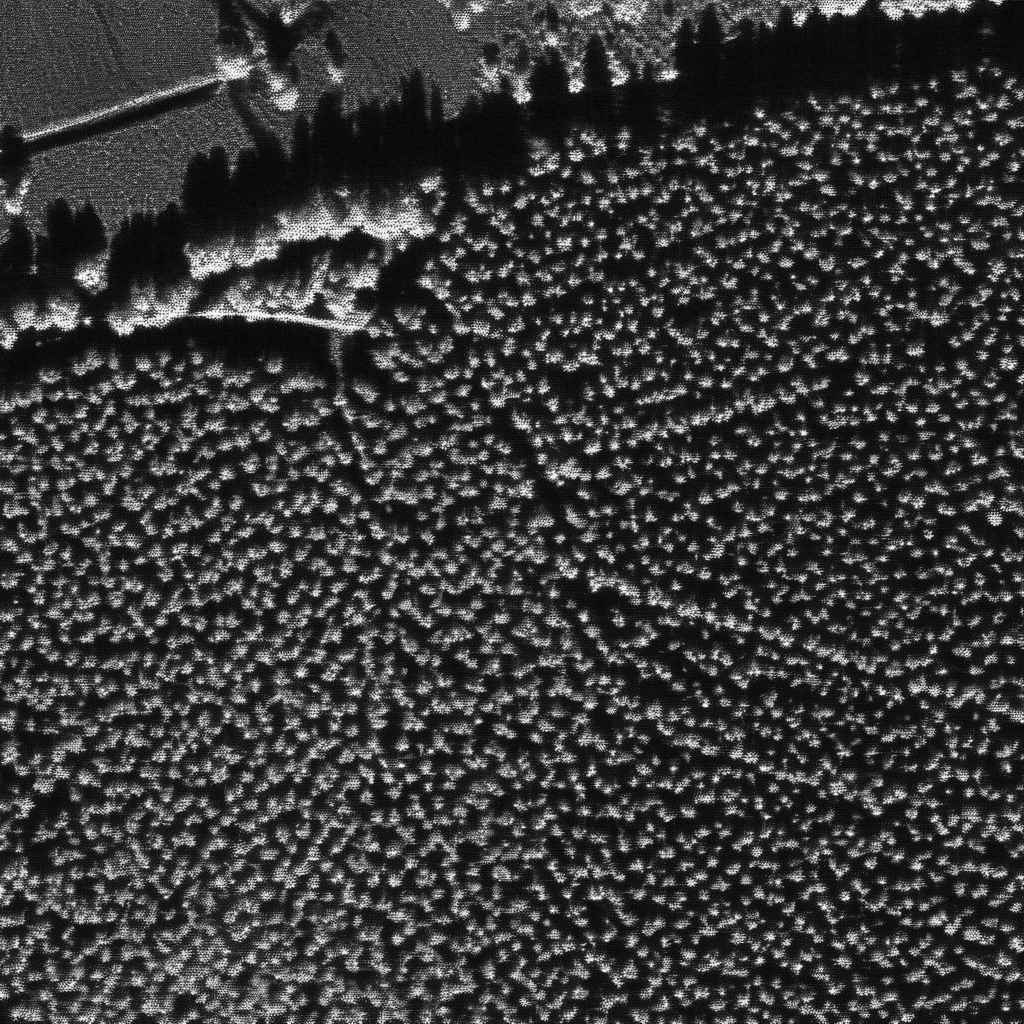} &
        \includegraphics[width=\linewidth]{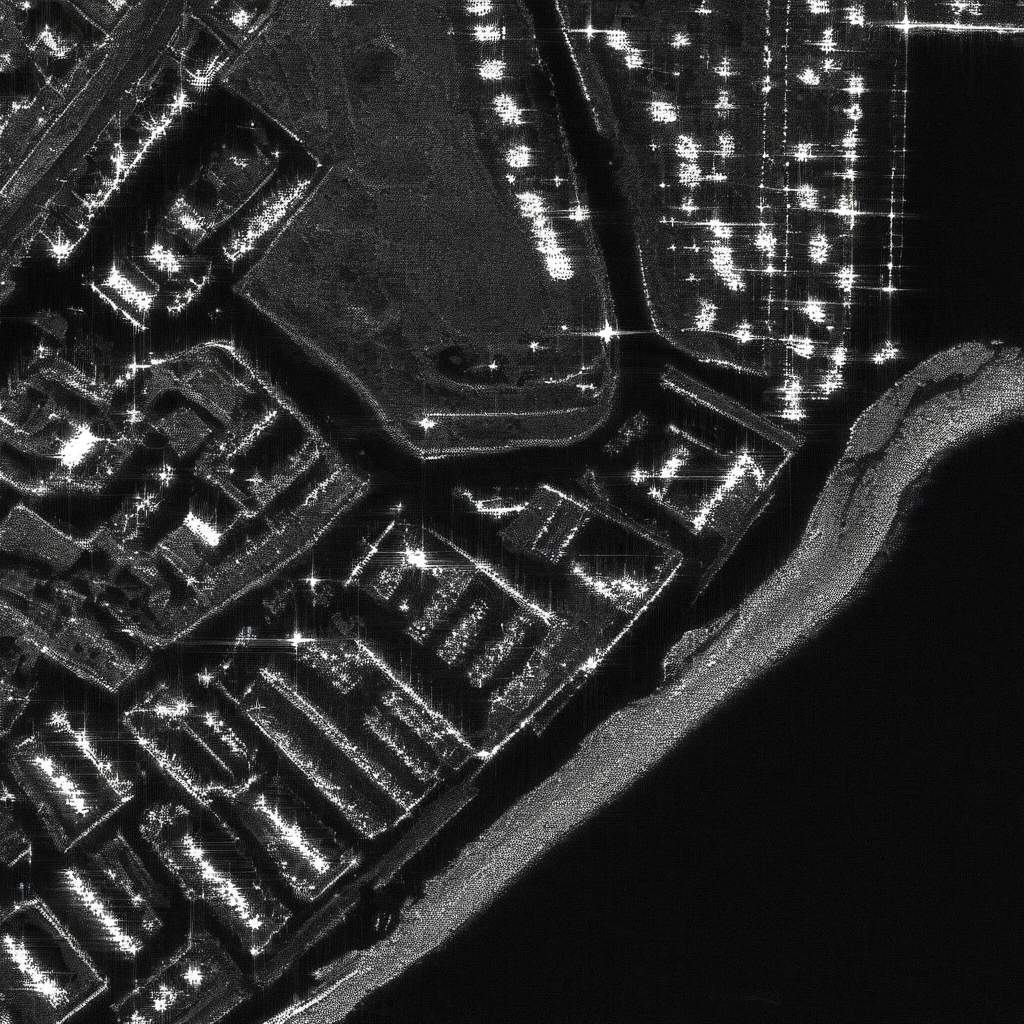} &
        \includegraphics[width=\linewidth]{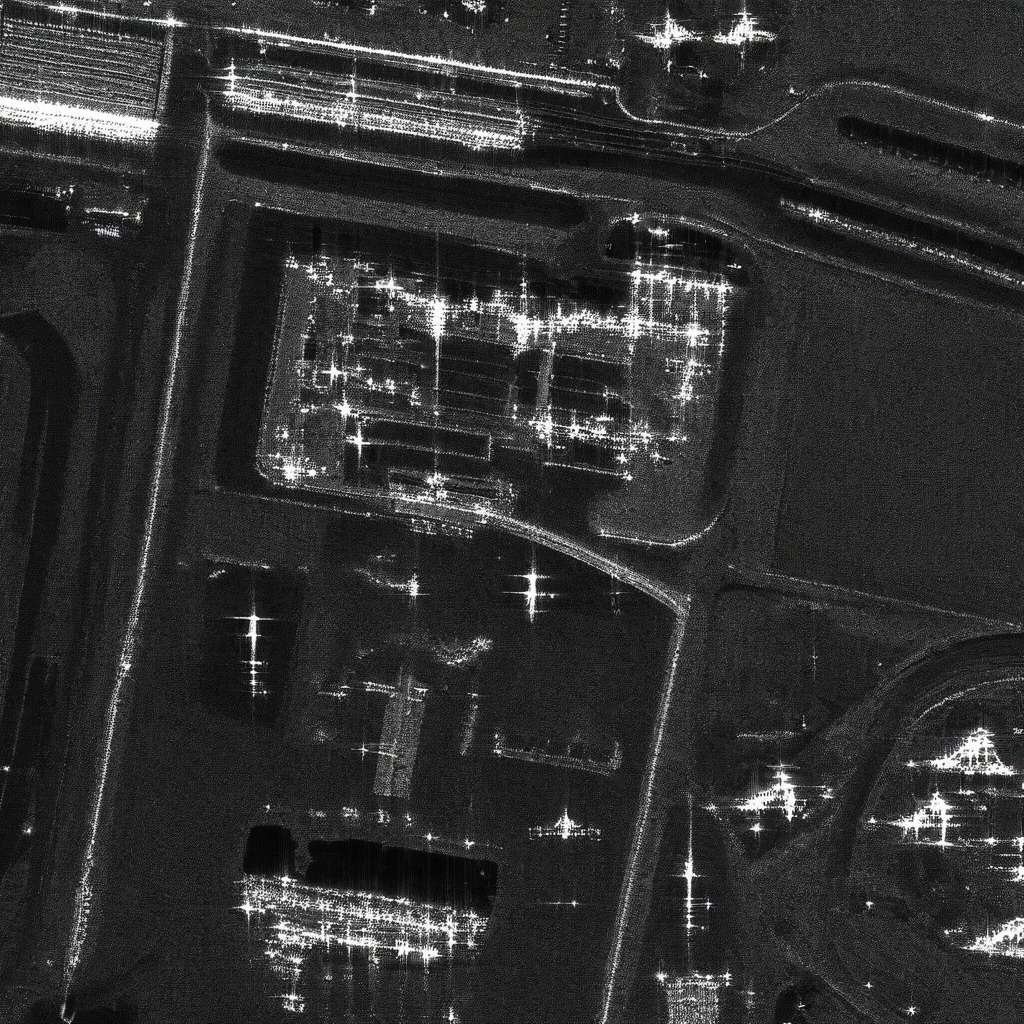} &
        \includegraphics[width=\linewidth]{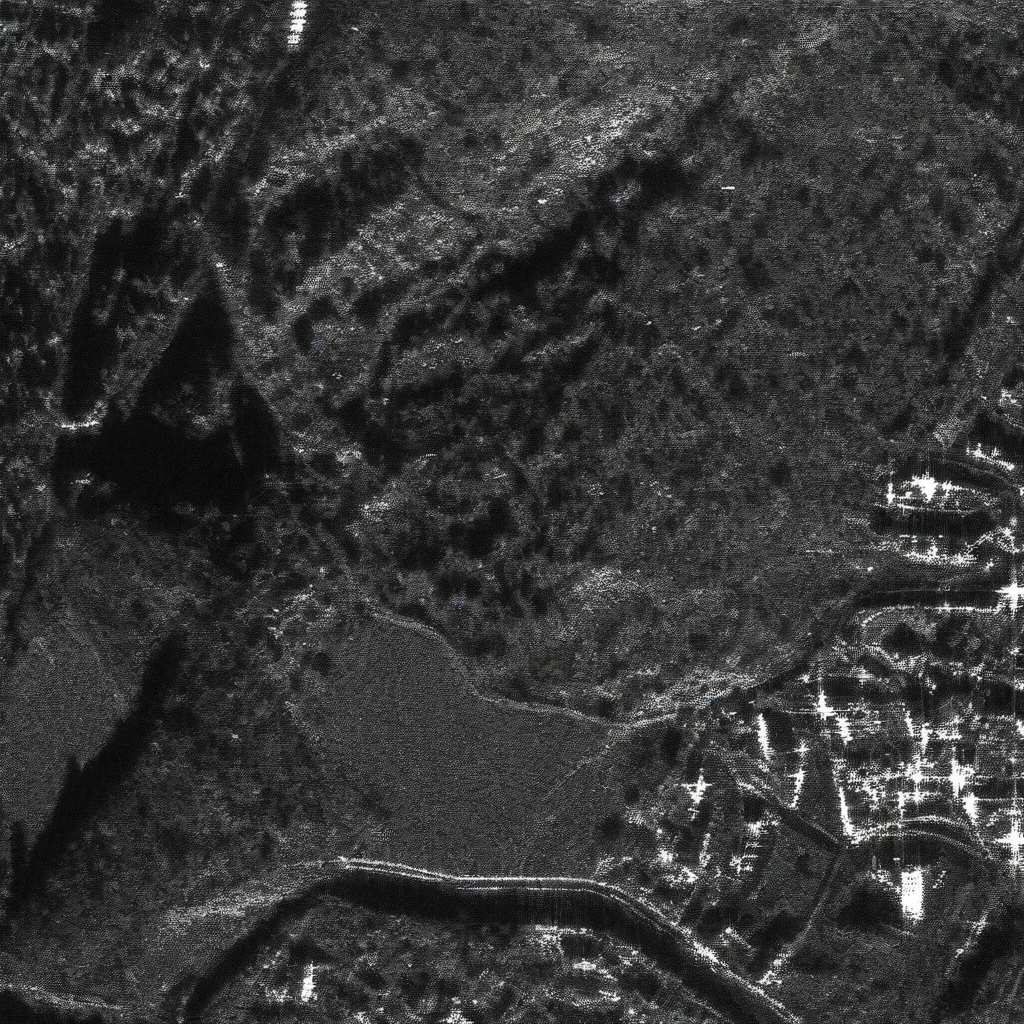} \\

        \rotatebox[origin=c]{90}{\textbf{whale-north-8-refined}} &
        \includegraphics[width=\linewidth]{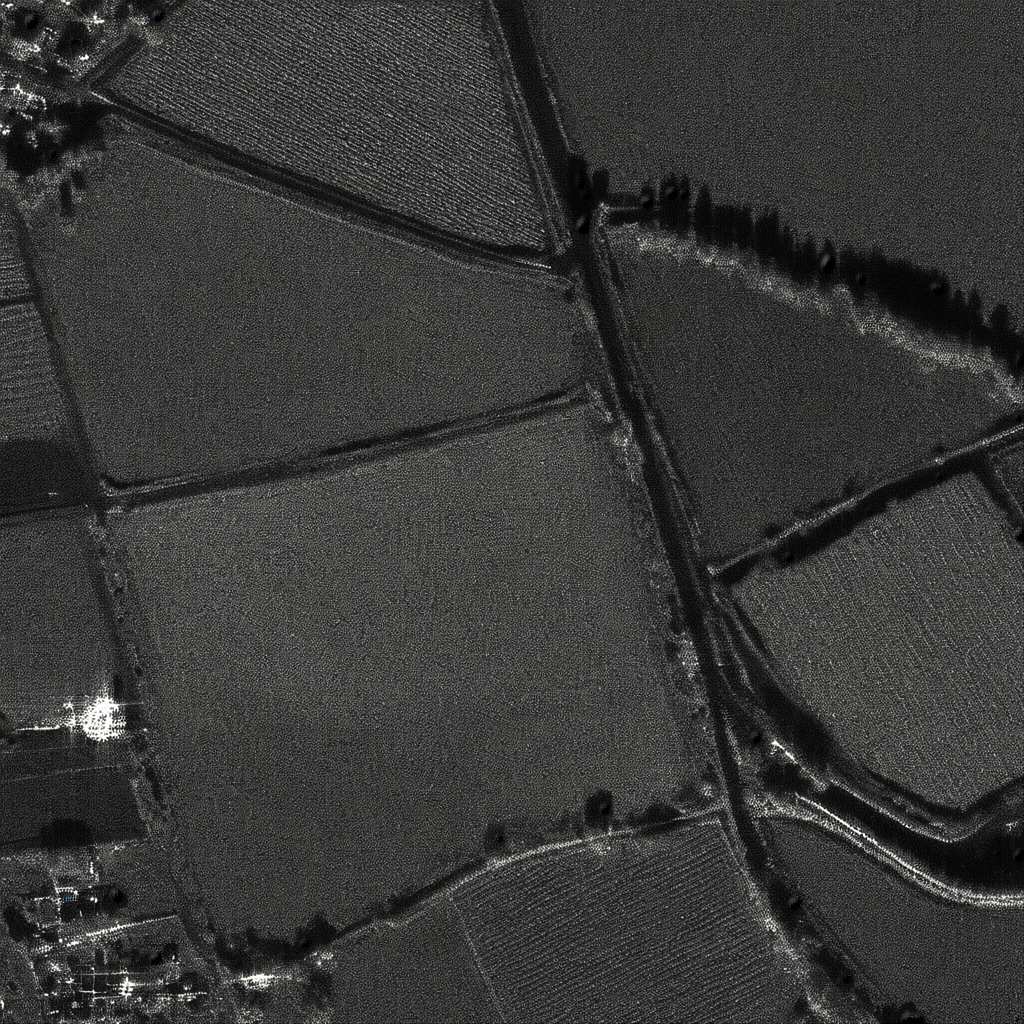} &
        \includegraphics[width=\linewidth]{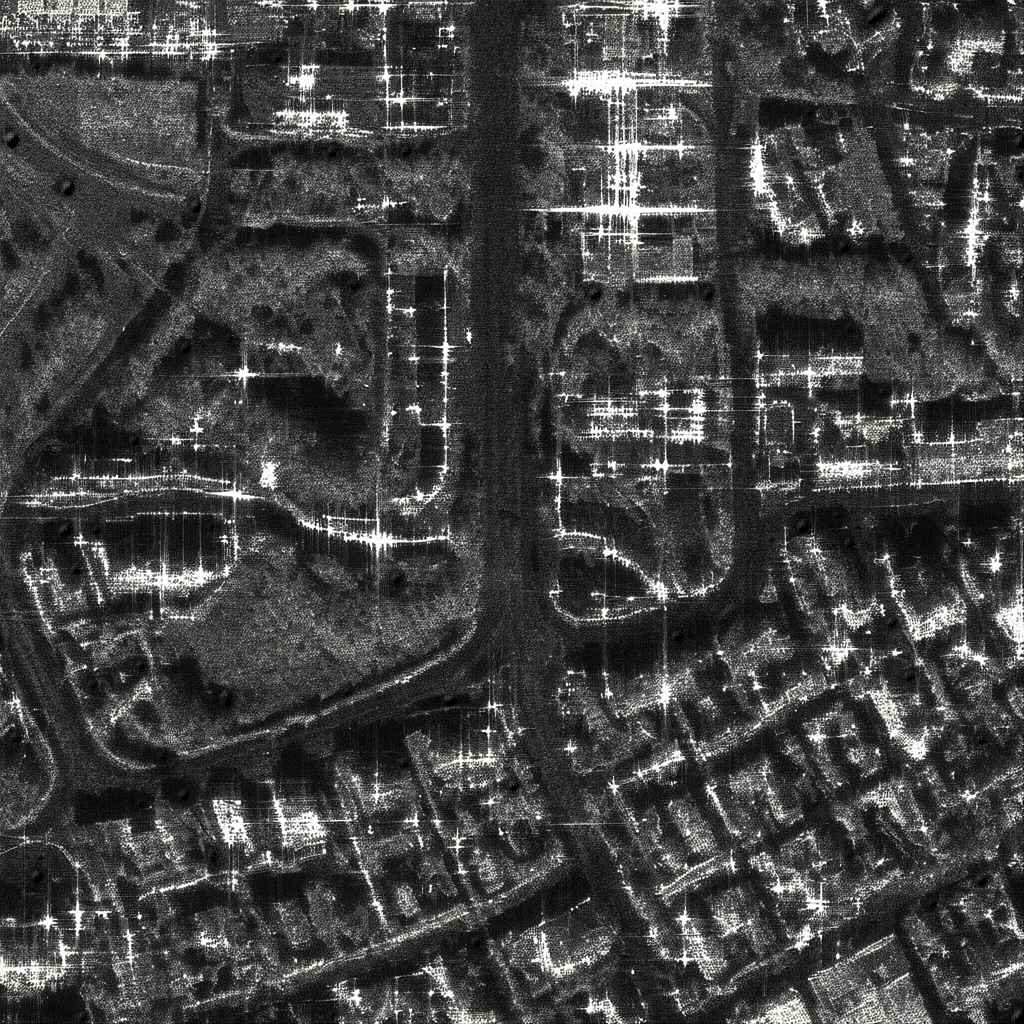} &
        \includegraphics[width=\linewidth]{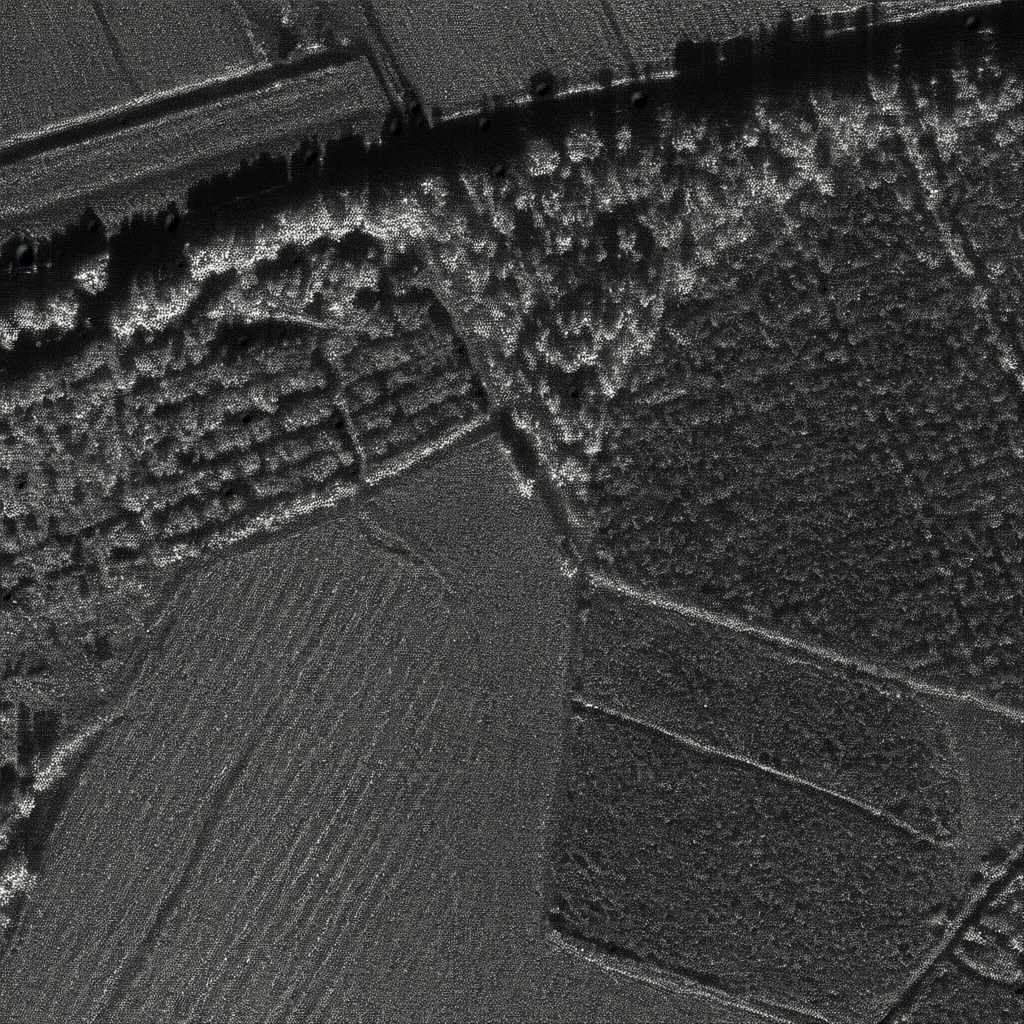} &
        \includegraphics[width=\linewidth]{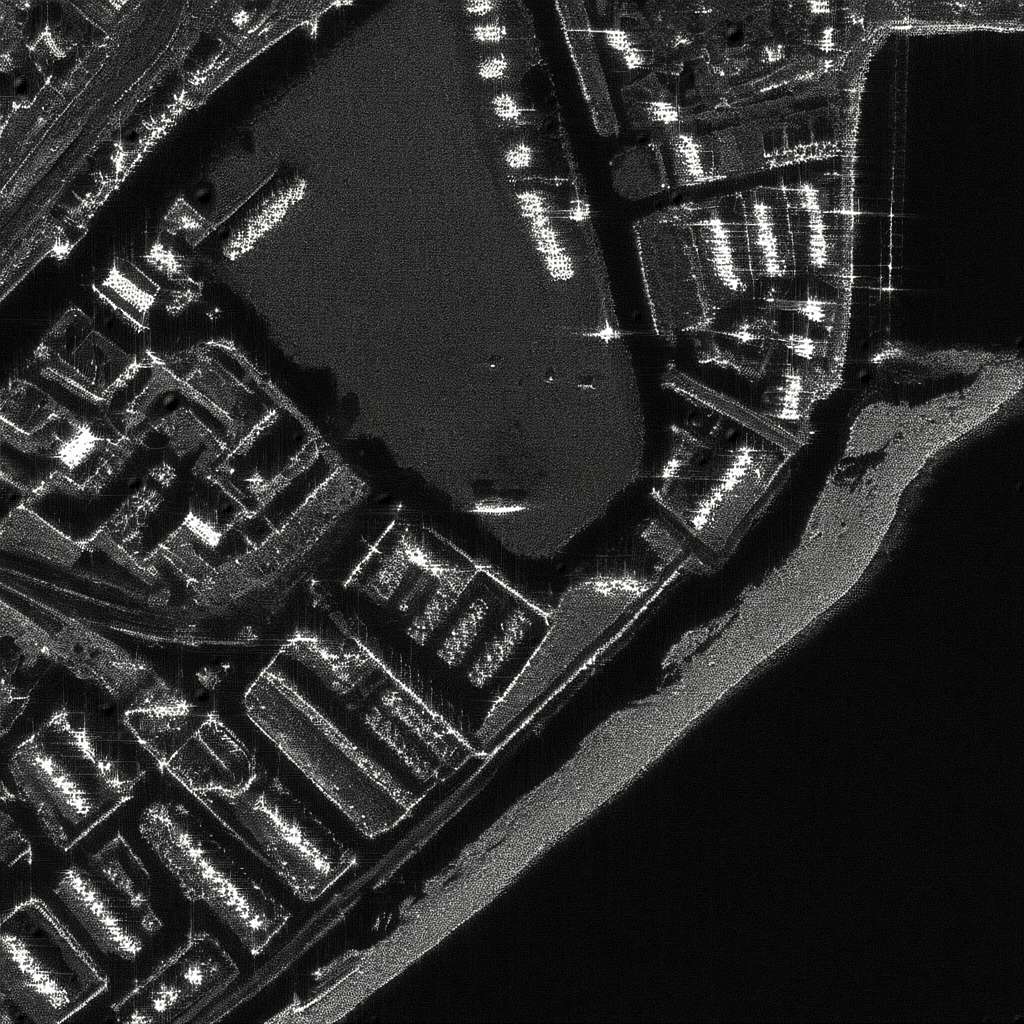} &
        \includegraphics[width=\linewidth]{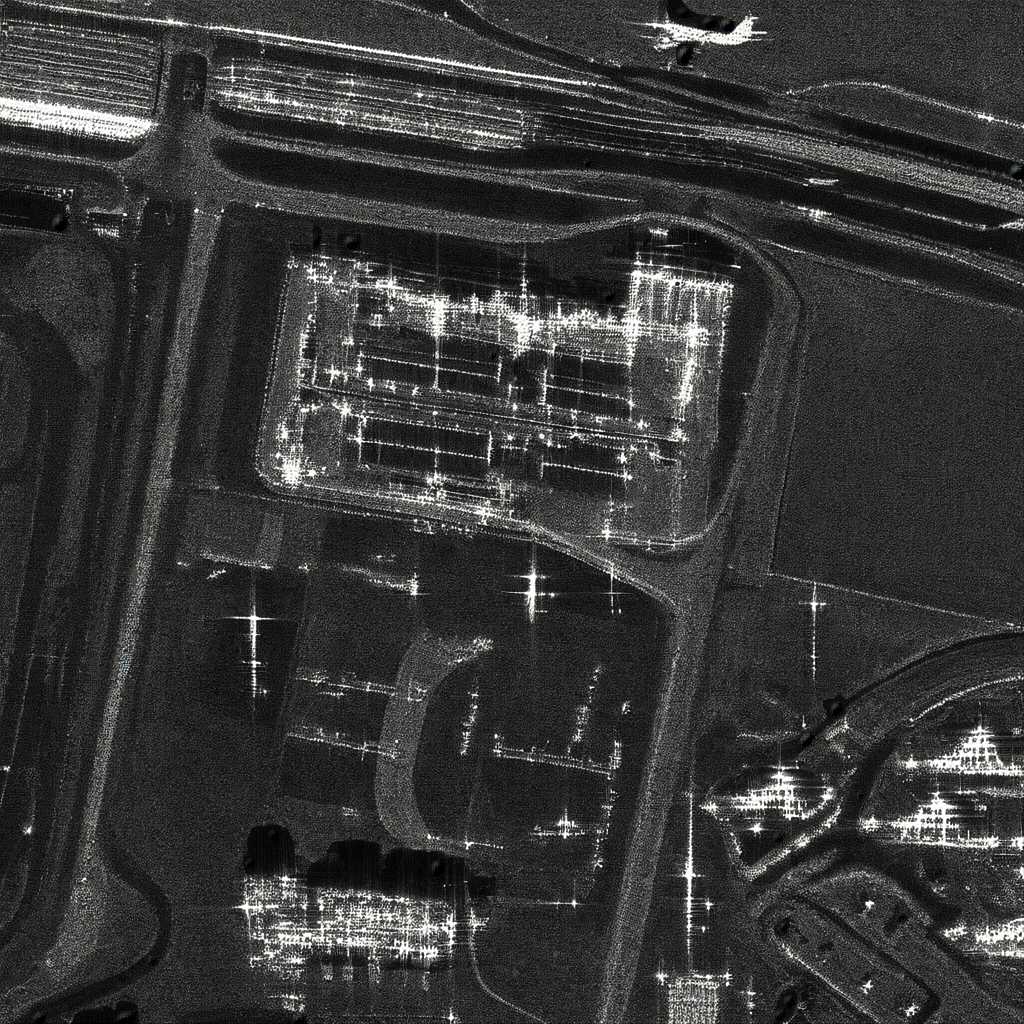} &
        \includegraphics[width=\linewidth]{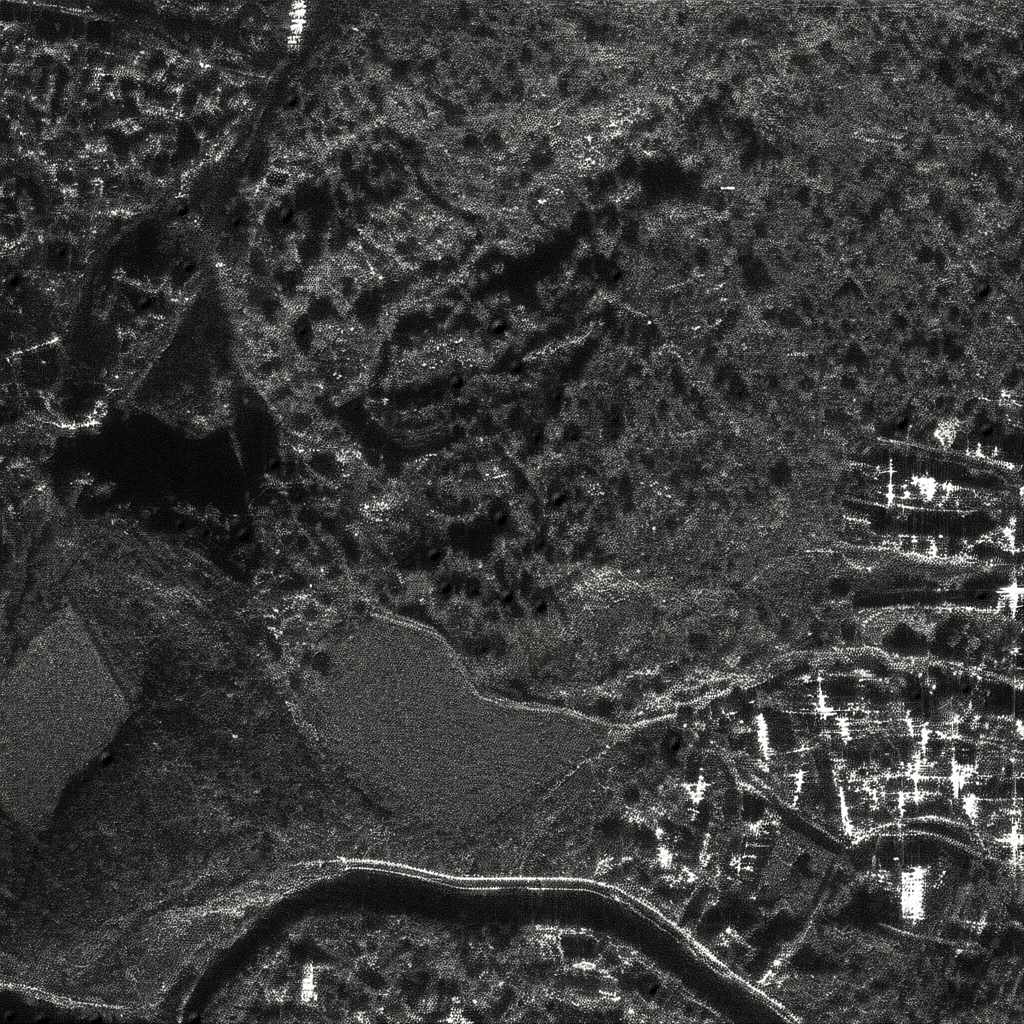} \\

        \rotatebox[origin=c]{90}{\textbf{heart-rose-2}} &
        \includegraphics[width=\linewidth]{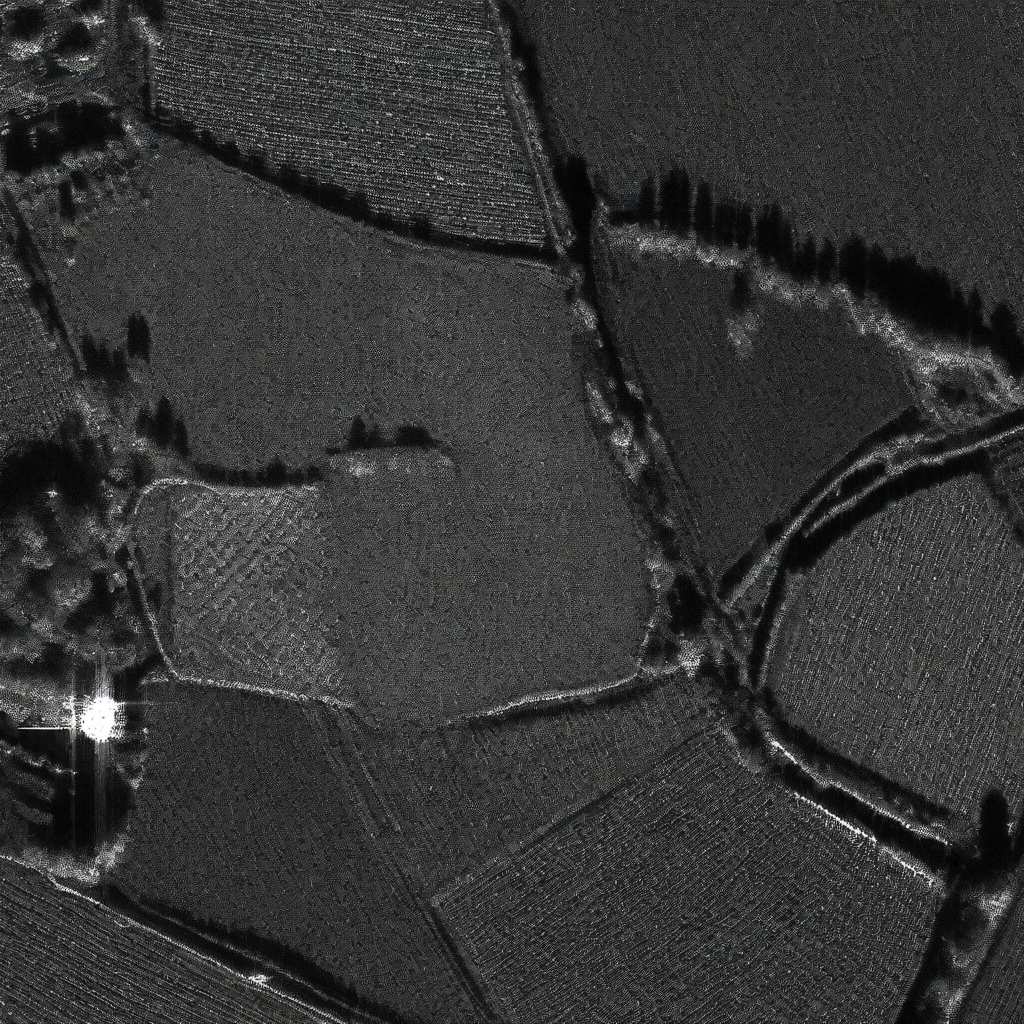} &
        \includegraphics[width=\linewidth]{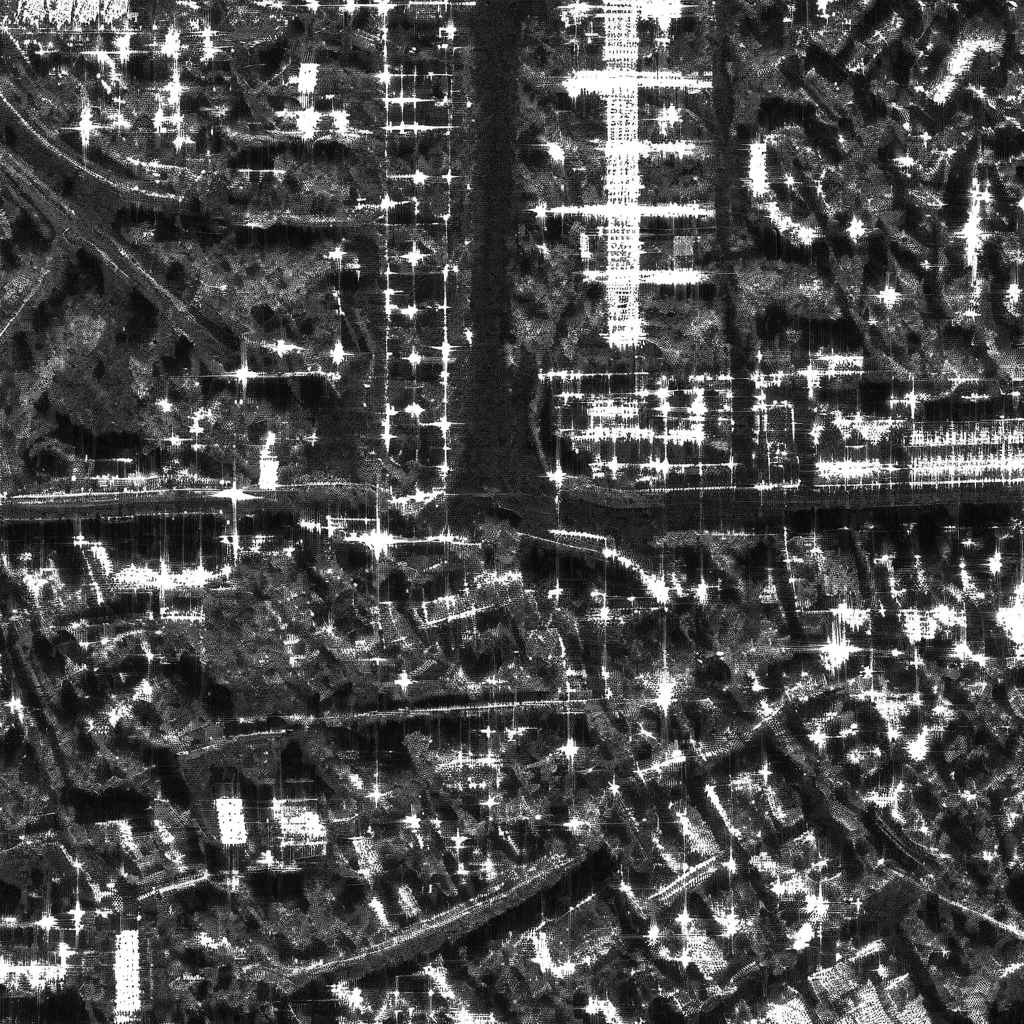} &
        \includegraphics[width=\linewidth]{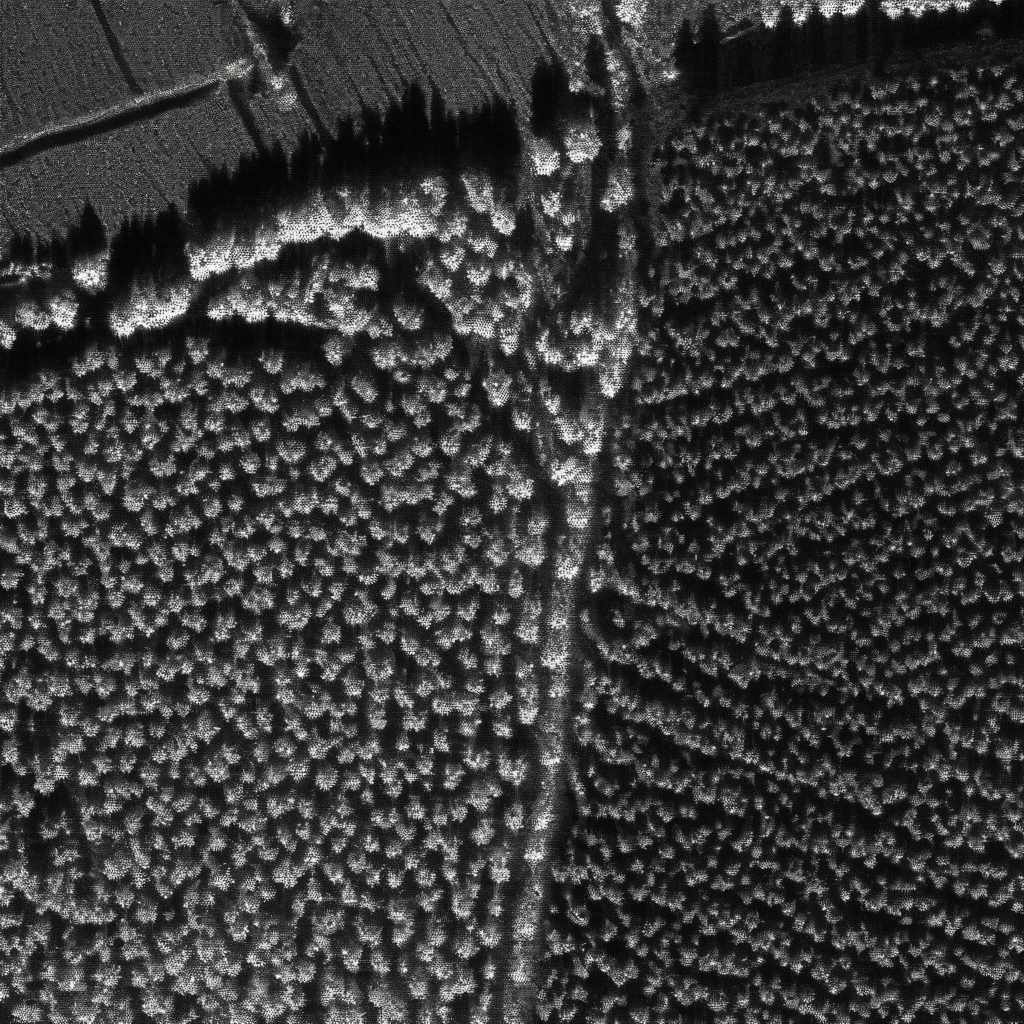} &
        \includegraphics[width=\linewidth]{images/whale-north-8-refined/seacoast_generated_image.jpg} &
        \includegraphics[width=\linewidth]{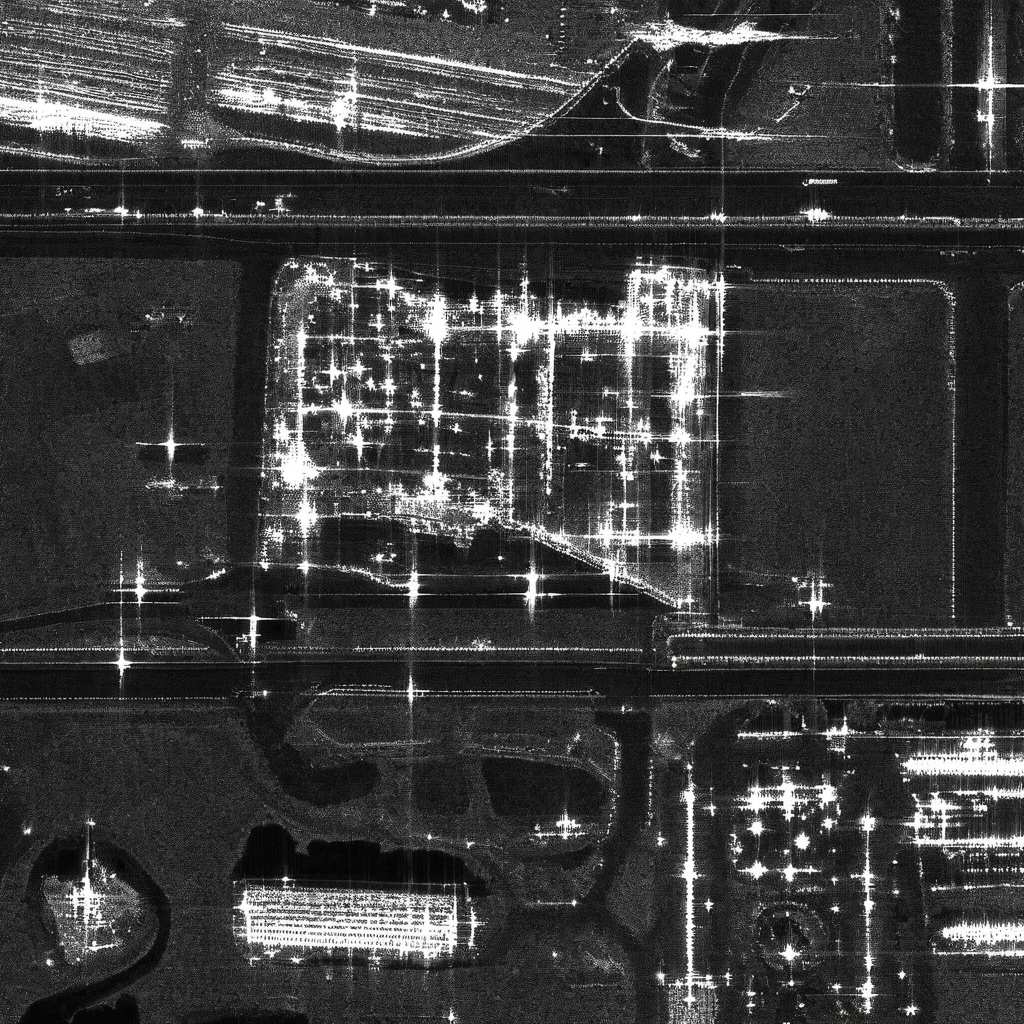} &
        \includegraphics[width=\linewidth]{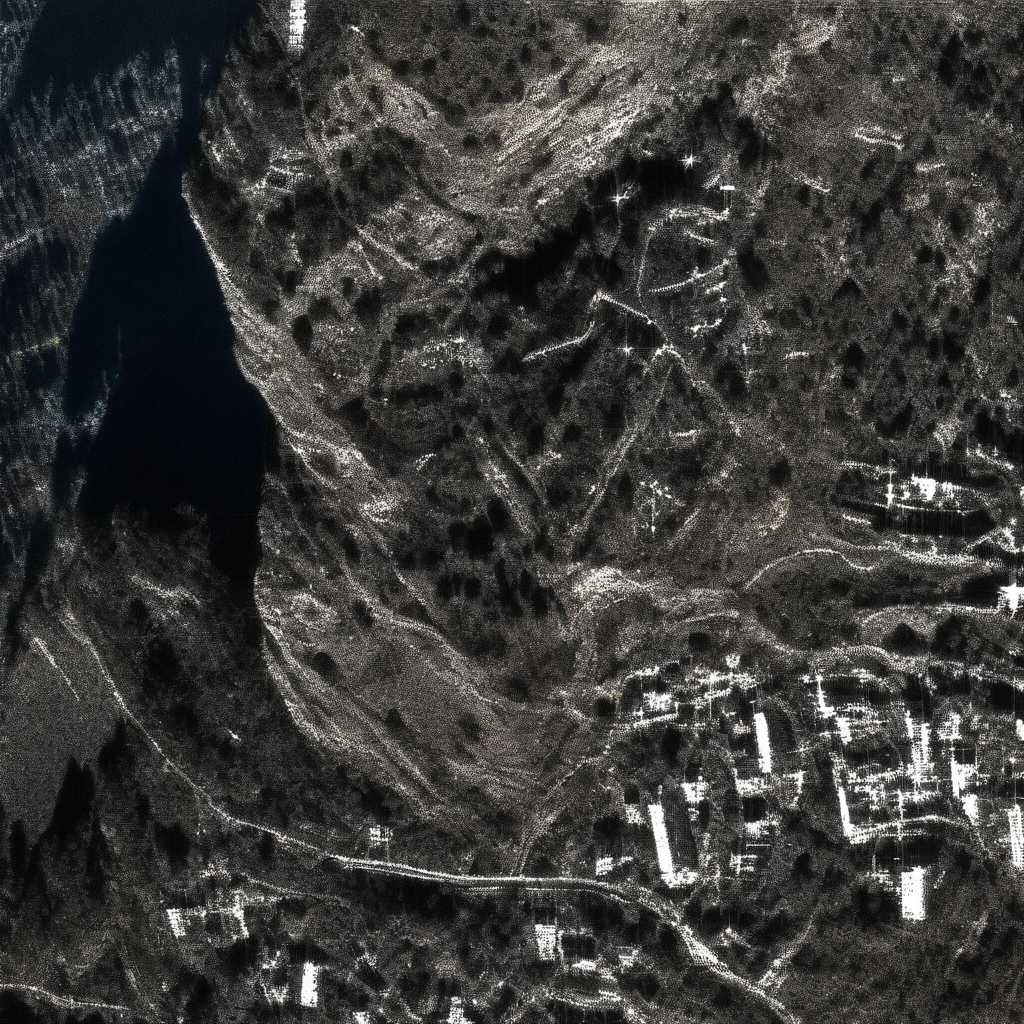} \\

    \end{tabular}
    \end{adjustbox}
    \caption{Refining with VAE vs <SAR> Embedding Learning: Generated images (1024x1024px at 40cm) per category for 9 different models (same seed training and evaluation)}
    \label{fig:subplot_models_vae}
\end{figure*}

In Figures~\ref{fig:subplot_models_full} and~\ref{fig:subplot_models_vae}, we generate images for each model with the same evaluation seed at epoch 8. Furthermore, Stable Diffusion XL behaves deterministically under fixed random seeds and identical training configurations, enabling controlled comparative experiments. We also use the same training seed to ensure that only the configuration changes across experiments. We use several prompts to generate our evaluation images as follows:

\begin{itemize}
    \item \textbf{Airport:} "A satellite view of an expansive airport with multiple runways, parked aircraft, terminal buildings, parking areas, and surrounding roads."
    
    \item \textbf{Seacoast:} "A satellite view of a coastal area with a structured marina housing numerous boats, adjacent to a town with organized roadways, and bordered by a sandy beach."
    
    \item \textbf{Forest:} "A satellite view of a landscape divided into two contrasting areas: a dense forest with a uniform canopy and a barren, plowed field with linear patterns. A winding road cuts through the terrain, connecting the two regions."
    
    \item \textbf{City:} "A satellite view of a dense urban area with a mix of residential and commercial buildings, winding roads, patches of greenery, and a few large parking lots."
    
    \item \textbf{Field:} "A satellite view of a vast agricultural landscape with meticulously organized rectangular fields, a winding canal, and a few isolated structures."
    
    \item \textbf{Mountains/Relief:} "A satellite view of a juxtaposition of rugged mountainous terrain with patches of greenery, and a densely populated urban area with structured roadways, buildings, and swimming pools."
\end{itemize}

\section{Best model images generation}
\label{appendix:heart-rose-2-images}
\setcounter{equation}{0}  
\renewcommand{\theequation}{C.\arabic{equation}}  
\setcounter{figure}{0}  
\renewcommand{\thefigure}{C.\arabic{figure}}  
The images shown in Figure~\ref{fig:heart_rose_grid} were generated using our best-performing model, \textit{heart-rose-2}. The corresponding prompts used for each image are listed below.

(a) "A SAR image of a vast landscape with a rectangular structure in an airport surrounded by roads and patches of vegetation."\\
(b) "A SAR image of a coastal town with rooftops, a winding road, a sandy beach, boats on the water, and a rocky outcrop."\\
(c) "A SAR image of a port with a boat in the water and a forest nearby."\\
(d) "A SAR image of a verdant landscape divided into geometrically patterned fields, a forested area, and a cluster of isolated buildings."\\
(e) "A SAR image of a vast landscape dominated by meticulously organized agricultural fields, intersected by winding roads, and anchored by a sizable building complex surrounded by vegetation."\\
(f) "A SAR image of a juxtaposition of organized residential areas with pools and vegetation, surrounded by meticulously arranged agricultural fields."\\
(g) "A SAR image of a dense forest, an agricultural field with linear patterns, a paved road intersecting the field."\\
(h) "A SAR image of a landscape dominated by agricultural fields, a road, a cluster of buildings, and a solar panel array."\\
(i) "A SAR image of a hilly terrain with patches of vegetation, a winding road, and scattered structures, possibly residential or agricultural buildings."\\
(j) "A SAR image of a landscape divided into two areas of dense forest with a uniform canopy. A winding road cuts through the terrain, connecting the two regions."\\
(k) "A SAR image of a river with a bridge, a town with buildings, roads, and green spaces, and a facility with circular structures."\\
(l) "A SAR image of a mountainous terrain with a mix of dense forested areas and barren patches. There are visible erosion patterns, possibly from water flow."\\
(m) "A SAR image of a coastal area with a structured marina housing numerous boats, adjacent to a town with organized roadways, and a large body of water extending to the horizon."\\
(n) "A SAR image of a residential area with organized streets, houses with varying roof designs, patches of greenery, swimming pools, and a few larger structures that could be commercial or community buildings."\\
(o) "A SAR image of a large water body with circular patterns, a curving road, a cluster of buildings, and a marina with boats."




\newpage
\bibliography{cas-refs}

\begin{thebibliography}{38}
\providecommand{\natexlab}[1]{#1}
\providecommand{\url}[1]{\texttt{#1}}
\expandafter\ifx\csname urlstyle\endcsname\relax
  \providecommand{\doi}[1]{doi: #1}\else
  \providecommand{\doi}{doi: \begingroup \urlstyle{rm}\Url}\fi

\bibitem[Agrawal and Banerjee(2025)]{Agrawal_2025}
K.~Agrawal and R.~Banerjee.
\newblock Synthetic art generation and deepfake detection: A study on jamini roy inspired dataset.
\newblock \emph{TechRxiv}, Mar. 2025.
\newblock \doi{10.36227/techrxiv.174119231.19482547/v1}.
\newblock URL \url{http://dx.doi.org/10.36227/techrxiv.174119231.19482547/v1}.

\bibitem[Auer et~al.(2016)Auer, Bamler, and Reinartz]{7730757}
S.~Auer, R.~Bamler, and P.~Reinartz.
\newblock Raysar - 3d sar simulator: Now open source.
\newblock In \emph{2016 IEEE International Geoscience and Remote Sensing Symposium (IGARSS)}, pages 6730--6733, 2016.
\newblock \doi{10.1109/IGARSS.2016.7730757}.

\bibitem[Baqu{\'e} et~al.(2019)Baqu{\'e}, Dreuillet, and Oriot]{Baqu2019SethiR}
R.~Baqu{\'e}, P.~Dreuillet, and H.~M. Oriot.
\newblock Sethi : Review of 10 years of development and experimentation of the remote sensing platform.
\newblock \emph{2019 International Radar Conference (RADAR)}, 2019.

\bibitem[COCHIN et~al.(2008)COCHIN, POULIGUEN, DELAHAYE, HELLARD, GOSSELIN, and AUBINEAU]{5757200}
C.~COCHIN, P.~POULIGUEN, B.~DELAHAYE, D.~l. HELLARD, P.~GOSSELIN, and F.~AUBINEAU.
\newblock Mocem - an 'all in one' tool to simulate sar image.
\newblock In \emph{7th European Conference on Synthetic Aperture Radar}, pages 1--4, 2008.

\bibitem[Dai et~al.(2025)Dai, Lu, and Li]{dai2025diffusionbasedsyntheticdatageneration}
W.~Dai, L.~Lu, and Z.~Li.
\newblock Diffusion-based synthetic data generation for visible-infrared person re-identification, 2025.
\newblock URL \url{https://arxiv.org/abs/2503.12472}.

\bibitem[Debuysère et~al.(2024)Debuysère, Trouvé, Letheule, Colin, and Lévêque]{debuysere2024synthesizing}
S.~Debuysère, N.~Trouvé, N.~Letheule, E.~Colin, and O.~Lévêque.
\newblock Synthesizing sar images with generative ai: Expanding to large-scale imagery, October 2024.
\newblock \url{https://hal.science/hal-04786104}.

\bibitem[Debuysère et~al.(2025)Debuysère, Trouvé, Letheule, Lévêque, and Colin]{debuysère2025spacebornairbornsarimage}
S.~Debuysère, N.~Trouvé, N.~Letheule, O.~Lévêque, and E.~Colin.
\newblock From spaceborn to airborn: Sar image synthesis using foundation models for multi-scale adaptation, 2025.
\newblock URL \url{https://arxiv.org/abs/2505.03844}.

\bibitem[Dettmers et~al.(2023)Dettmers, Pagnoni, Holtzman, and Zettlemoyer]{dettmers2023qloraefficientfinetuningquantized}
T.~Dettmers, A.~Pagnoni, A.~Holtzman, and L.~Zettlemoyer.
\newblock Qlora: Efficient finetuning of quantized llms, 2023.
\newblock URL \url{https://arxiv.org/abs/2305.14314}.

\bibitem[Gal et~al.(2022)Gal, Alaluf, Atzmon, Patashnik, Bermano, Chechik, and Cohen-Or]{gal2022imageworthwordpersonalizing}
R.~Gal, Y.~Alaluf, Y.~Atzmon, O.~Patashnik, A.~H. Bermano, G.~Chechik, and D.~Cohen-Or.
\newblock An image is worth one word: Personalizing text-to-image generation using textual inversion, 2022.
\newblock URL \url{https://arxiv.org/abs/2208.01618}.

\bibitem[Gao et~al.(2024)Gao, Wu, Wen, Xu, and Chen]{isprs-annals-X-1-2024-83-2024}
D.~Gao, X.~Wu, Z.~Wen, Y.~Xu, and Z.~Chen.
\newblock Few-shot sar vehicle target augmentation based on generative adversarial networks.
\newblock \emph{ISPRS Annals of the Photogrammetry, Remote Sensing and Spatial Information Sciences}, X-1-2024:\penalty0 83--90, 2024.
\newblock \doi{10.5194/isprs-annals-X-1-2024-83-2024}.
\newblock URL \url{https://isprs-annals.copernicus.org/articles/X-1-2024/83/2024/}.

\bibitem[Haralick et~al.(1973)Haralick, Shanmugam, and Dinstein]{4309314}
R.~M. Haralick, K.~Shanmugam, and I.~Dinstein.
\newblock Textural features for image classification.
\newblock \emph{IEEE Transactions on Systems, Man, and Cybernetics}, SMC-3\penalty0 (6):\penalty0 610--621, 1973.
\newblock \doi{10.1109/TSMC.1973.4309314}.

\bibitem[Hong et~al.(2024)Hong, Wang, Ding, Yu, Lv, Wang, Cheng, Huang, Ji, Xue, Zhao, Yang, Gu, Zhang, Feng, Yin, Wang, Qi, Song, Zhang, Liu, Xu, Li, Dong, and Tang]{hong2024cogvlm2visuallanguagemodels}
W.~Hong, W.~Wang, M.~Ding, W.~Yu, Q.~Lv, Y.~Wang, Y.~Cheng, S.~Huang, J.~Ji, Z.~Xue, L.~Zhao, Z.~Yang, X.~Gu, X.~Zhang, G.~Feng, D.~Yin, Z.~Wang, J.~Qi, X.~Song, P.~Zhang, D.~Liu, B.~Xu, J.~Li, Y.~Dong, and J.~Tang.
\newblock Cogvlm2: Visual language models for image and video understanding, 2024.
\newblock URL \url{https://arxiv.org/abs/2408.16500}.

\bibitem[Hu et~al.(2021)Hu, Shen, Wallis, Allen-Zhu, Li, Wang, Wang, and Chen]{hu2021loralowrankadaptationlarge}
E.~J. Hu, Y.~Shen, P.~Wallis, Z.~Allen-Zhu, Y.~Li, S.~Wang, L.~Wang, and W.~Chen.
\newblock Lora: Low-rank adaptation of large language models, 2021.
\newblock URL \url{https://arxiv.org/abs/2106.09685}.

\bibitem[Jia et~al.(2022)Jia, Tang, Chen, Cardie, Belongie, Hariharan, and Lim]{jia2022visualprompttuning}
M.~Jia, L.~Tang, B.-C. Chen, C.~Cardie, S.~Belongie, B.~Hariharan, and S.-N. Lim.
\newblock Visual prompt tuning, 2022.
\newblock URL \url{https://arxiv.org/abs/2203.12119}.

\bibitem[Khanna et~al.(2024)Khanna, Liu, Zhou, Meng, Rombach, Burke, Lobell, and Ermon]{khanna2024diffusionsatgenerativefoundationmodel}
S.~Khanna, P.~Liu, L.~Zhou, C.~Meng, R.~Rombach, M.~Burke, D.~Lobell, and S.~Ermon.
\newblock Diffusionsat: A generative foundation model for satellite imagery, 2024.
\newblock URL \url{https://arxiv.org/abs/2312.03606}.

\bibitem[Kirillov et~al.(2023)Kirillov, Mintun, Ravi, Mao, Rolland, Gustafson, Xiao, Whitehead, Berg, Lo, Dollár, and Girshick]{kirillov2023segment}
A.~Kirillov, E.~Mintun, N.~Ravi, H.~Mao, C.~Rolland, L.~Gustafson, T.~Xiao, S.~Whitehead, A.~C. Berg, W.-Y. Lo, P.~Dollár, and R.~Girshick.
\newblock Segment anything, 2023.
\newblock URL \url{https://arxiv.org/abs/2304.02643}.

\bibitem[Labs(2024)]{flux2024}
B.~F. Labs.
\newblock Flux.
\newblock \url{https://github.com/black-forest-labs/flux}, 2024.

\bibitem[Liu et~al.(2025)Liu, Chen, Zhao, Zou, and Shi]{liu2025text2earthunlockingtextdrivenremote}
C.~Liu, K.~Chen, R.~Zhao, Z.~Zou, and Z.~Shi.
\newblock Text2earth: Unlocking text-driven remote sensing image generation with a global-scale dataset and a foundation model, 2025.
\newblock URL \url{https://arxiv.org/abs/2501.00895}.

\bibitem[{Liu} et~al.(2018){Liu}, {Zhao}, {Liu}, {Dong}, {Liu}, and {Hui}]{2018SPIE10752E..05L}
W.~{Liu}, Y.~{Zhao}, M.~{Liu}, L.~{Dong}, X.~{Liu}, and M.~{Hui}.
\newblock {Generating simulated SAR images using Generative Adversarial Network}.
\newblock In A.~G. {Tescher}, editor, \emph{Applications of Digital Image Processing XLI}, volume 10752 of \emph{Society of Photo-Optical Instrumentation Engineers (SPIE) Conference Series}, page 1075205, Sept. 2018.
\newblock \doi{10.1117/12.2320024}.

\bibitem[Ma et~al.(2025)Ma, Xiao, Dong, Wang, Wang, and Pan]{ma2025sarchatbench2mmultitaskvisionlanguagebenchmark}
Z.~Ma, X.~Xiao, S.~Dong, P.~Wang, H.~Wang, and Q.~Pan.
\newblock Sarchat-bench-2m: A multi-task vision-language benchmark for sar image interpretation, 2025.
\newblock URL \url{https://arxiv.org/abs/2502.08168}.

\bibitem[Pang et~al.(2024)Pang, Cao, Tang, Xu, Bai, Zhou, and Meng]{pang2024hsigenefoundationmodelhyperspectral}
L.~Pang, X.~Cao, D.~Tang, S.~Xu, X.~Bai, F.~Zhou, and D.~Meng.
\newblock Hsigene: A foundation model for hyperspectral image generation, 2024.
\newblock URL \url{https://arxiv.org/abs/2409.12470}.

\bibitem[Podell et~al.(2023)Podell, English, Lacey, Blattmann, Dockhorn, Müller, Penna, and Rombach]{podell2023sdxlimprovinglatentdiffusion}
D.~Podell, Z.~English, K.~Lacey, A.~Blattmann, T.~Dockhorn, J.~Müller, J.~Penna, and R.~Rombach.
\newblock Sdxl: Improving latent diffusion models for high-resolution image synthesis, 2023.
\newblock URL \url{https://arxiv.org/abs/2307.01952}.

\bibitem[Rombach et~al.(2022)Rombach, Blattmann, Lorenz, Esser, and Ommer]{rombach2022highresolutionimagesynthesislatent}
R.~Rombach, A.~Blattmann, D.~Lorenz, P.~Esser, and B.~Ommer.
\newblock High-resolution image synthesis with latent diffusion models, 2022.
\newblock URL \url{https://arxiv.org/abs/2112.10752}.

\bibitem[Ruiz et~al.(2023)Ruiz, Li, Jampani, Pritch, Rubinstein, and Aberman]{ruiz2023dreamboothfinetuningtexttoimage}
N.~Ruiz, Y.~Li, V.~Jampani, Y.~Pritch, M.~Rubinstein, and K.~Aberman.
\newblock Dreambooth: Fine tuning text-to-image diffusion models for subject-driven generation, 2023.
\newblock URL \url{https://arxiv.org/abs/2208.12242}.

\bibitem[Saharia et~al.(2022)Saharia, Chan, Saxena, Li, Whang, Denton, Ghasemipour, Ayan, Mahdavi, Lopes, Salimans, Ho, Fleet, and Norouzi]{saharia2022photorealistictexttoimagediffusionmodels}
C.~Saharia, W.~Chan, S.~Saxena, L.~Li, J.~Whang, E.~Denton, S.~K.~S. Ghasemipour, B.~K. Ayan, S.~S. Mahdavi, R.~G. Lopes, T.~Salimans, J.~Ho, D.~J. Fleet, and M.~Norouzi.
\newblock Photorealistic text-to-image diffusion models with deep language understanding, 2022.
\newblock URL \url{https://arxiv.org/abs/2205.11487}.

\bibitem[Tang et~al.(2024)Tang, Cao, Hou, Jiang, Liu, and Meng]{tang2024crsdiffcontrollableremotesensing}
D.~Tang, X.~Cao, X.~Hou, Z.~Jiang, J.~Liu, and D.~Meng.
\newblock Crs-diff: Controllable remote sensing image generation with diffusion model, 2024.
\newblock URL \url{https://arxiv.org/abs/2403.11614}.

\bibitem[Team(2025)]{chameleonteam2025chameleonmixedmodalearlyfusionfoundation}
C.~Team.
\newblock Chameleon: Mixed-modal early-fusion foundation models, 2025.
\newblock URL \url{https://arxiv.org/abs/2405.09818}.

\bibitem[Trouve et~al.(2024)Trouve, Letheule, Leveque, Rami, and Colin]{trouve2024synthesis}
N.~Trouve, N.~Letheule, O.~Leveque, I.~Rami, and E.~Colin.
\newblock Sar image synthesis using text conditioned pre-trained generative ai models.
\newblock In \emph{Proceedings of EUSAR 2024; 15th European Conference on Synthetic Aperture Radar}, Munich, Germany, 2024. VDE, VDE,ITG.

\bibitem[Wang et~al.(2024)Wang, Fan, Bao, Jiang, and Song]{wang2024borabidimensionalweightdecomposedlowrank}
Q.~Wang, Y.~Fan, J.~Bao, H.~Jiang, and Y.~Song.
\newblock Bora: Bi-dimensional weight-decomposed low-rank adaptation, 2024.
\newblock URL \url{https://arxiv.org/abs/2412.06441}.

\bibitem[Woollard et~al.(2022)Woollard, Blacknell, Griffiths, and Ritchie]{rs14112561}
M.~Woollard, D.~Blacknell, H.~Griffiths, and M.~A. Ritchie.
\newblock Sarcastic v2.0—high-performance sar simulation for next-generation atr systems.
\newblock \emph{Remote Sensing}, 14\penalty0 (11), 2022.
\newblock ISSN 2072-4292.
\newblock \doi{10.3390/rs14112561}.
\newblock URL \url{https://www.mdpi.com/2072-4292/14/11/2561}.

\bibitem[Ye et~al.(2023)Ye, Zhang, Liu, Han, and Yang]{ye2023ipadaptertextcompatibleimage}
H.~Ye, J.~Zhang, S.~Liu, X.~Han, and W.~Yang.
\newblock Ip-adapter: Text compatible image prompt adapter for text-to-image diffusion models, 2023.
\newblock URL \url{https://arxiv.org/abs/2308.06721}.

\bibitem[Yu et~al.(2022)Yu, Wang, Vasudevan, Yeung, Seyedhosseini, and Wu]{yu2022cocacontrastivecaptionersimagetext}
J.~Yu, Z.~Wang, V.~Vasudevan, L.~Yeung, M.~Seyedhosseini, and Y.~Wu.
\newblock Coca: Contrastive captioners are image-text foundation models, 2022.
\newblock URL \url{https://arxiv.org/abs/2205.01917}.

\bibitem[Yu et~al.(2023)Yu, Shi, Pasunuru, Muller, Golovneva, Wang, Babu, Tang, Karrer, Sheynin, Ross, Polyak, Howes, Sharma, Xu, Tamoyan, Ashual, Singer, Li, Zhang, James, Ghosh, Taigman, Fazel-Zarandi, Celikyilmaz, Zettlemoyer, and Aghajanyan]{yu2023scalingautoregressivemultimodalmodels}
L.~Yu, B.~Shi, R.~Pasunuru, B.~Muller, O.~Golovneva, T.~Wang, A.~Babu, B.~Tang, B.~Karrer, S.~Sheynin, C.~Ross, A.~Polyak, R.~Howes, V.~Sharma, P.~Xu, H.~Tamoyan, O.~Ashual, U.~Singer, S.-W. Li, S.~Zhang, R.~James, G.~Ghosh, Y.~Taigman, M.~Fazel-Zarandi, A.~Celikyilmaz, L.~Zettlemoyer, and A.~Aghajanyan.
\newblock Scaling autoregressive multi-modal models: Pretraining and instruction tuning, 2023.
\newblock URL \url{https://arxiv.org/abs/2309.02591}.

\bibitem[Yu et~al.(2024)Yu, Liu, Liu, Shi, and Zou]{yu2024metaearthgenerativefoundationmodel}
Z.~Yu, C.~Liu, L.~Liu, Z.~Shi, and Z.~Zou.
\newblock Metaearth: A generative foundation model for global-scale remote sensing image generation, 2024.
\newblock URL \url{https://arxiv.org/abs/2405.13570}.

\bibitem[Zhang et~al.(2023)Zhang, Rao, and Agrawala]{zhang2023addingconditionalcontroltexttoimage}
L.~Zhang, A.~Rao, and M.~Agrawala.
\newblock Adding conditional control to text-to-image diffusion models, 2023.
\newblock URL \url{https://arxiv.org/abs/2302.05543}.

\bibitem[Zhang et~al.(2024)Zhang, Zhao, Guo, and Yin]{Zhang_2024}
Z.~Zhang, T.~Zhao, Y.~Guo, and J.~Yin.
\newblock Rs5m and georsclip: A large-scale vision- language dataset and a large vision-language model for remote sensing.
\newblock \emph{IEEE Transactions on Geoscience and Remote Sensing}, 62:\penalty0 1–23, 2024.
\newblock ISSN 1558-0644.
\newblock \doi{10.1109/tgrs.2024.3449154}.
\newblock URL \url{http://dx.doi.org/10.1109/TGRS.2024.3449154}.

\bibitem[Zhou et~al.(2022)Zhou, Yang, Loy, and Liu]{Zhou_2022}
K.~Zhou, J.~Yang, C.~C. Loy, and Z.~Liu.
\newblock Learning to prompt for vision-language models.
\newblock \emph{International Journal of Computer Vision}, 130\penalty0 (9):\penalty0 2337–2348, July 2022.
\newblock ISSN 1573-1405.
\newblock \doi{10.1007/s11263-022-01653-1}.
\newblock URL \url{http://dx.doi.org/10.1007/s11263-022-01653-1}.

\bibitem[Zou et~al.(2020)Zou, Zhang, Wang, Wu, and Gu]{s20226673}
L.~Zou, H.~Zhang, C.~Wang, F.~Wu, and F.~Gu.
\newblock Mw-acgan: Generating multiscale high-resolution sar images for ship detection.
\newblock \emph{Sensors}, 20\penalty0 (22), 2020.
\newblock ISSN 1424-8220.
\newblock \doi{10.3390/s20226673}.
\newblock URL \url{https://www.mdpi.com/1424-8220/20/22/6673}.

\end{thebibliography}



\end{document}